\documentclass[10pt]{report}

\usepackage[table]{xcolor}
\usepackage{amsmath}
\usepackage{hyperref} 
\usepackage{xspace}
\usepackage{booktabs}
\usepackage{subcaption} 
\usepackage{verbatim}
\usepackage[inline]{enumitem}

\usepackage{cwpuzzle}

\hypersetup{colorlinks,linkcolor={red!50!black},citecolor={blue!50!black},urlcolor={blue!80!black}}
\usepackage[toc,page]{appendix}
\usepackage[top=1.2in, bottom=1.22in, left=1.2in, right=1.2in]{geometry}

\usepackage[tikz]{bclogo}
\usetikzlibrary{shapes,calc,positioning,automata,arrows,trees}
\tikzset{
  dirtree/.style={
    grow via three points={one child at (0.8,-0.7) and two children at (0.8,-0.7) and (0.8,-1.45)}, 
    edge from parent path={($(\tikzparentnode\tikzparentanchor)+(.4cm,0cm)$) |- (\tikzchildnode\tikzchildanchor)}, growth parent anchor=west, parent anchor=south west},
}
\newcommand\bcpen{\includegraphics[width=15pt]{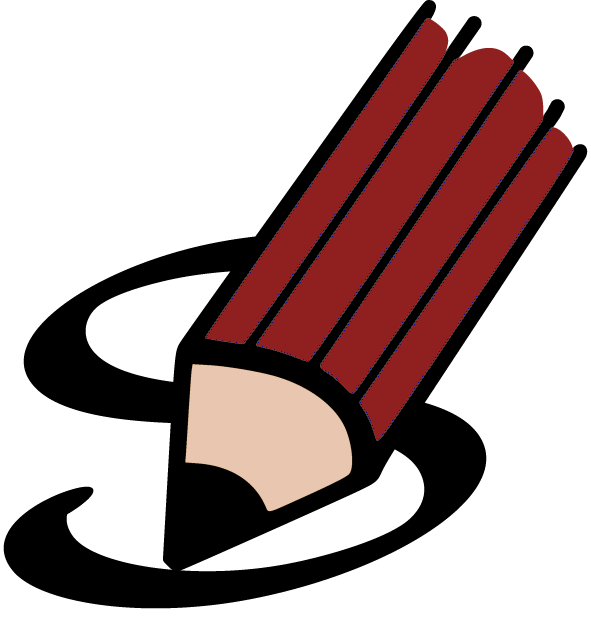}} 
\newcommand\bcdico{\includegraphics[width=15pt]{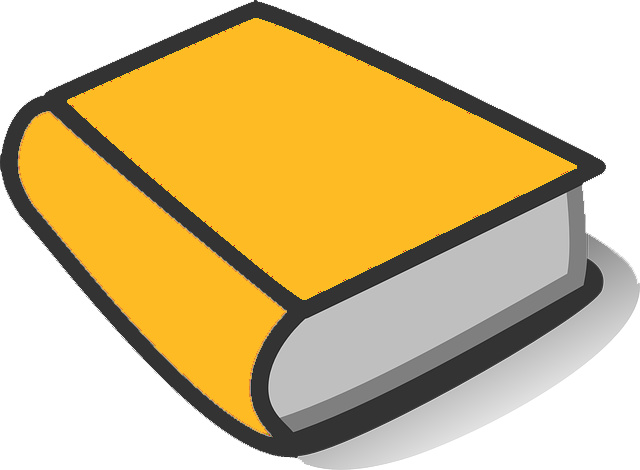}} 

\definecolor{v3lgray}{gray}{0.98}
\definecolor{v2lgray}{gray}{0.85}
\definecolor{vlgray}{gray}{0.92}
\definecolor{mygray}{rgb}{0.92,0.98,0.92}
\definecolor{bgray}{rgb}{0.8,0.8,0.8}
\definecolor{dgray}{rgb}{0.4,0.4,0.4}
\definecolor{mgray}{rgb}{0.55,0.55,0.55}
\definecolor{dblue}{RGB}{0,0,99}
\definecolor{dred}{RGB}{120,6,54}
\definecolor{dgreen}{RGB}{47,135,7}
\definecolor{d2green}{RGB}{47,85,7}
\definecolor{dviolet}{RGB}{102,0,153}
\definecolor{mblue}{RGB}{0,0,180}
\definecolor{m2blue}{RGB}{0,0,220}
\definecolor{colorse}{RGB}{255,248,220}
\definecolor{colorsy}{HTML}{F2F2F2}
\definecolor{colorex}{RGB}{255,248,220} 

\definecolor{grey}{rgb}{0.75,0.75,0.75}
\definecolor{dgray}{rgb}{0.4,0.4,0.4}
\definecolor{dorange}{HTML}{F74A05}
\definecolor{darktangerine}{rgb}{1.0, 0.66, 0.07}
\definecolor{deepchampagne}{rgb}{0.98, 0.84, 0.65}
\definecolor{cornsilk}{rgb}{1.0, 0.97, 0.86}
\definecolor{brickred}{rgb}{0.8, 0.25, 0.33}
\definecolor{officegreen}{rgb}{0.0, 0.5, 0.0}
\definecolor{NavyBlue}{HTML}{2984F2}
\definecolor{background}{HTML}{EEEEEE}
\definecolor{delim}{RGB}{20,105,176}
\colorlet{numb}{magenta!60!black}

\def\N{\mathbb{N}}

\def\ace{ACE\xspace}

\def\mzinc{MiniZinc\xspace}
\def\x3{{\rm XCSP$^3$}\xspace}
\def\jv3{{\rm JvCSP$^3$}\xspace}
\def\p3{{\rm PyCSP$^3$}\xspace}

\newcommand{\xml}[1]{{\tt <#1>}} 
\newcommand{\gb}[1]{{\tt #1}} 
\newcommand{\gbc}[1]{\textcolor{dblue}{{\mathit #1}}} 
\newcommand{\nn}[1]{{\tt #1}} 
\newcommand{\nm}[1]{\mathit{#1}} 
\newcommand{\va}[1]{{\boldsymbol #1}} 
\newcommand{\ns}[1]{{\mathcal #1}}  

\newtheorem{definition}{Definition}
\newtheorem{remark}{Remark}

\newcounter{cntPy}
\newcounter{cntSe}

\usepackage{listingsutf8}
\newcommand*{\com}[1]{\hfill \textcolor{dgray}{// #1}} 

\lstdefinelanguage{json}{
    basicstyle=\ttfamily\footnotesize,
    showstringspaces=false,
    breaklines=true,
    literate=
     *{0}{{{\color{numb}0}}}{1}
      {1}{{{\color{numb}1}}}{1}
      {2}{{{\color{numb}2}}}{1}
      {3}{{{\color{numb}3}}}{1}
      {4}{{{\color{numb}4}}}{1}
      {5}{{{\color{numb}5}}}{1}
      {6}{{{\color{numb}6}}}{1}
      {7}{{{\color{numb}7}}}{1}
      {8}{{{\color{numb}8}}}{1}
      {9}{{{\color{numb}9}}}{1}
      {[}{{{\color{delim}{[}}}}{1}
      {]}{{{\color{delim}{]}}}}{1},
}
\lstnewenvironment{json}{\lstset{
    inputencoding=utf8/latin1,
    language=json,
    framesep=5pt,
    xleftmargin=20pt,xrightmargin=2pt,
  }
}{}

\lstnewenvironment{python}{\lstset{
    inputencoding=utf8/latin1,
    language=python,
    ndkeywords={size,dom,matrix,strict,occurrences},
    ndkeywordstyle=\color{mblue}, 
    keywords=[2]{data,satisfy,minimize,maximize,solve},
    keywordstyle=[2]\color{dred}, 
    deletekeywords={def},
    keywords=[3]{def,If,Then,Else,Match, Cases}, 
    keywordstyle=[3]\color{dorange},
    commentstyle=\color{officegreen}, 
    showstringspaces=false,
    captionpos=t,
    basicstyle=\ttfamily\footnotesize,
    framesep=5pt,
    xleftmargin=5pt,xrightmargin=1pt,
    backgroundcolor=\color{colorex}, 
    escapechar=@
  }
}{}

\lstnewenvironment{xcsp}{\lstset{
    keywords={instance,variables,literals,var,domain,constraints,objectives,annotations,annotation,array,extension,intension, tuples, supports,conflicts,args,quantification,fuzzy,relation,interval,point,agents,agent,allDifferent,allDifferentExcept,count,ordered,vars,ctrs,comm,stages,decision, stochastic,required,possible,nodes,edges,arcs,minimize,maximize,linear,operator,value,allEqual,among,atLeast,atMost,exactly,limit,cumulative,circuit,regular,mdd,hamming,channel,element,values,transitions,origins,durations,indexes,permutation,coeffs,nValues,clause,cube,sort,lex,precedence,cardPath,slide,slidingAmong,slidingSum,sequence,gsc,seqBin,binPacking,maximum,minimum, hardTemplate,softTemplate,capacities,loads,intervals,costMatrix,knapsack,allDisjoint,overlap,exists,forall,aggregate,dbd,output,min,max,varHeuristic,valHeuristic,lastConflict,vals, AC, FC, prepro, search, filtering, branching, restarts,group,width, widths, stretch,lengths,block,start,final,balance,path,tree, root, succs, range, image, allDistant, roots, capacity, bins, sizes, heights, except, function, instantiation,matrix, sumCosts, increasing, index, list,lists,set,mset,terminal,rules,grammar,mapping,cards, ranges, rowCards, colCards, indexes, not, allIntersecting, costMeasure, accept, limits, relation, sum, comparison, operand, condition, conditions, nCircuits, nPaths, nTrees, rowOccurs, colOccurs, occurs, noOverlap,cost, total, number, cardinality, scope, static, equal, row, smart, ifThen, form, adhoc, note
    },
    basewidth  = {.6em,0.6em},
    basicstyle=\ttfamily\footnotesize,
    keywordstyle=\color{mblue}\bfseries,
    identifierstyle=\color{black},
    sensitive=false,
    comment=[l]{//},
    morecomment=[s]{<!--}{-->},
    commentstyle=\color{dred}\ttfamily,
    stringstyle=\color{dgreen}\ttfamily,
    morestring=[b]',
    morestring=[b]",
    escapechar=@,
    showstringspaces=false,
    xleftmargin=1pt,xrightmargin=1pt,
    breaklines=true,
    backgroundcolor=\color{v3lgray}, 
    inputencoding=utf8/latin9,texcl
  }
}{}

\lstdefinelanguage{semantics}{
  keywords={identifier}, basewidth  = {.5em,0.5em}, escapechar=@, xleftmargin=1pt, xrightmargin=1pt,
  breaklines=true, basicstyle=\ttfamily\linespread{1.20}\small, backgroundcolor=\color{colorse}, inputencoding=utf8/latin9, texcl, mathescape
}
\lstnewenvironment{semantics}{\lstset{language=semantics}}{}

\usepackage[skins,breakable,xparse]{tcolorbox}
\newcommand{\core}[1]{ 
  \medskip \begin{tcolorbox}[
    enhanced,breakable,
    boxsep=0pt,top=4pt,bottom=0pt,left=2mm,right=1mm,
    toprule=0.1mm,leftrule=0.1mm,rightrule=0.25mm,bottomrule=0.25mm,shadow={0.2mm}{-0.2mm}{0mm}{dgray},
    overlay unbroken and first={\node (logo) at ([xshift=4mm,yshift=-5mm]frame.north west) {#1}; },
    colframe=dgray,titlerule=-0.2mm,toptitle=3mm,coltitle=dred, fonttitle=\bfseries,
    lines before break=6, pad at break*=10pt
    }
    
\newenvironment{boxse}
 {\stepcounter{cntSe} \core{\bcdico} , colback=colorse, title style={color=colorse}, coltitle=black, title=~ ~ \, Semantics \thecntSe]}
 {\end{tcolorbox}} 

\newenvironment{boxpy}
 {\stepcounter{cntPy} \core{\bcpen} , colback=colorex, title style={color=colorex}, title=~ ~ \,PyCSP$^3$ Model \thecntPy]}
 {\end{tcolorbox}}

 \newenvironment{boxpyno}
 {\core{\bcpen} , colback=colorex, title style={color=colorex}, coltitle = colorex, title = ~ ~ Code]}
 {\end{tcolorbox}} 

\newenvironment{command}
  {\quote\small\verbatim}
  {\endverbatim\endquote}

\usepackage{enumitem}

\setlist{topsep=4pt,itemsep=1pt}

\usepackage{chessboard}
\usepackage[LSBC4,T1]{fontenc}
\setboardfontcolors{blackfieldmask=black!15}

\title{\textcolor{dred}{\p3 \\ Modeling Combinatorial Constrained Problems in Python}} 

\author{Christophe Lecoutre and Nicolas Szczepanski\\
University of Artois \\ CRIL CNRS, UMR 8188 \\ France \\ ~ \\
\{lecoutre,szczepanski\}@cril.fr
}

\date{Version 2.6 (i) -- September 2, 2026 \\~ \\\href{https://www.pycsp.org}{www.pycsp.org}} 

\usepackage{makeidx}
\makeindex

\begin{document}
\maketitle

\begin{abstract}
This document is a complete guide about \p3, a Python library that allows us to write models of combinatorial constrained problems in a declarative manner.
Currently, with \p3, you can write models of constraint satisfaction and optimization problems.
More specifically, you can build CSP (Constraint Satisfaction Problem) and COP (Constraint Optimization Problem) models.
Importantly, there is a complete separation between the modeling and solving phases: you write a model, you compile it (while providing some data) in order to generate an \x3 instance (file), and you solve that problem instance by means of a constraint solver.
You can also directly pilot the solving procedure in \p3, possibly conducting an incremental solving strategy.
In this document, you will find all that you need to know about \p3, with more than 60 illustrative models.
\end{abstract}

In a nutshell, the main ingredients of the complete tool chain we propose for handling combinatorial constrained problems are:
\begin{itemize}
\item \p3: a Python library for modeling constrained problems (described in this document) 
\item \x3: an intermediate format for representing problem instances while preserving the structure of models \cite{xcsp3}
\end{itemize}

\begin{figure}[h]
\begin{center}

\begin{tikzpicture}[scale=1, every node/.style={scale=1}]
\tikzstyle{sn}=[draw,rounded corners,minimum height=10mm,fill=blue!30,text width=5cm,text centered]
\node (user) at (4.95,2.4) {\includegraphics[scale=0.25]{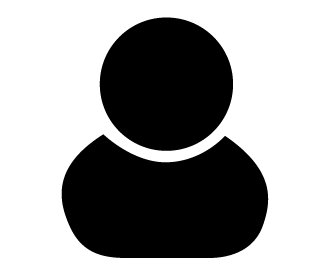}};
\node[draw=none] (root) at (5,1.5) {};
\node[draw=none] (l) at (0,-1.25) {};
\node[draw=none] (r) at (10.5,-1.25) {};
\node[draw=none] (l2) at (10,-3.75) {};
\node[draw=none] (r2) at (10.5,-3.75) {};
\node[draw,minimum height=7mm,fill=grey!30,text width=4cm,text centered] (model) at (2.5,0.5) {\textcolor{dgreen}{Model} \p3  \\{\scriptsize (Python 3)}}; 
\node[draw,minimum height=7mm,fill=grey!30,text width=4cm,text centered] (data) at (7.5,0.5) {\textcolor{dgreen}{Data} \\{\scriptsize (JSON)}}; 

\node[sn] (a) at (5,-1) {Compiler}; 
\node[draw,minimum height=7mm,fill=grey!30,text width=4cm,text centered] (b) at (5,-2.7) {\x3 \textcolor{dgreen}{Instance} \\{\scriptsize (XML)}}; 
\node[sn,text width=1.5cm] (s1) at (0,-4.8) {ACE}; 
\node[sn,text width=1.5cm] (s2) at (2,-4.8) {Choco}; 
\node[sn,text width=1.5cm] (s3) at (4,-4.8) {Picat};
\node[sn,text width=1.5cm] (s4) at (6,-4.8) {CoSoCo}; 
\node[sn,text width=1.5cm] (s5) at (8,-4.8) {...};
\node[sn,text width=1.5cm] (s6) at (10,-4.8) {CP-SAT};

\draw[->,>=latex] (root) -- (model);
\draw[->,>=latex] (root) -- (data);
\draw[->,>=latex] (model) -- (a);
\draw[->,>=latex] (data) -- (a);
\draw[->,>=latex] (a) -- (b);

\draw[->,>=latex] (b) -- (s1);
\draw[->,>=latex] (b) -- (s2);
\draw[->,>=latex] (b) -- (s3);
\draw[->,>=latex] (b) -- (s4);
\draw[->,>=latex] (b) -- (s5);
\draw[->,>=latex] (b) -- (s6);
\node (computer) at (4.98,-6.3) {\includegraphics[scale=0.02]{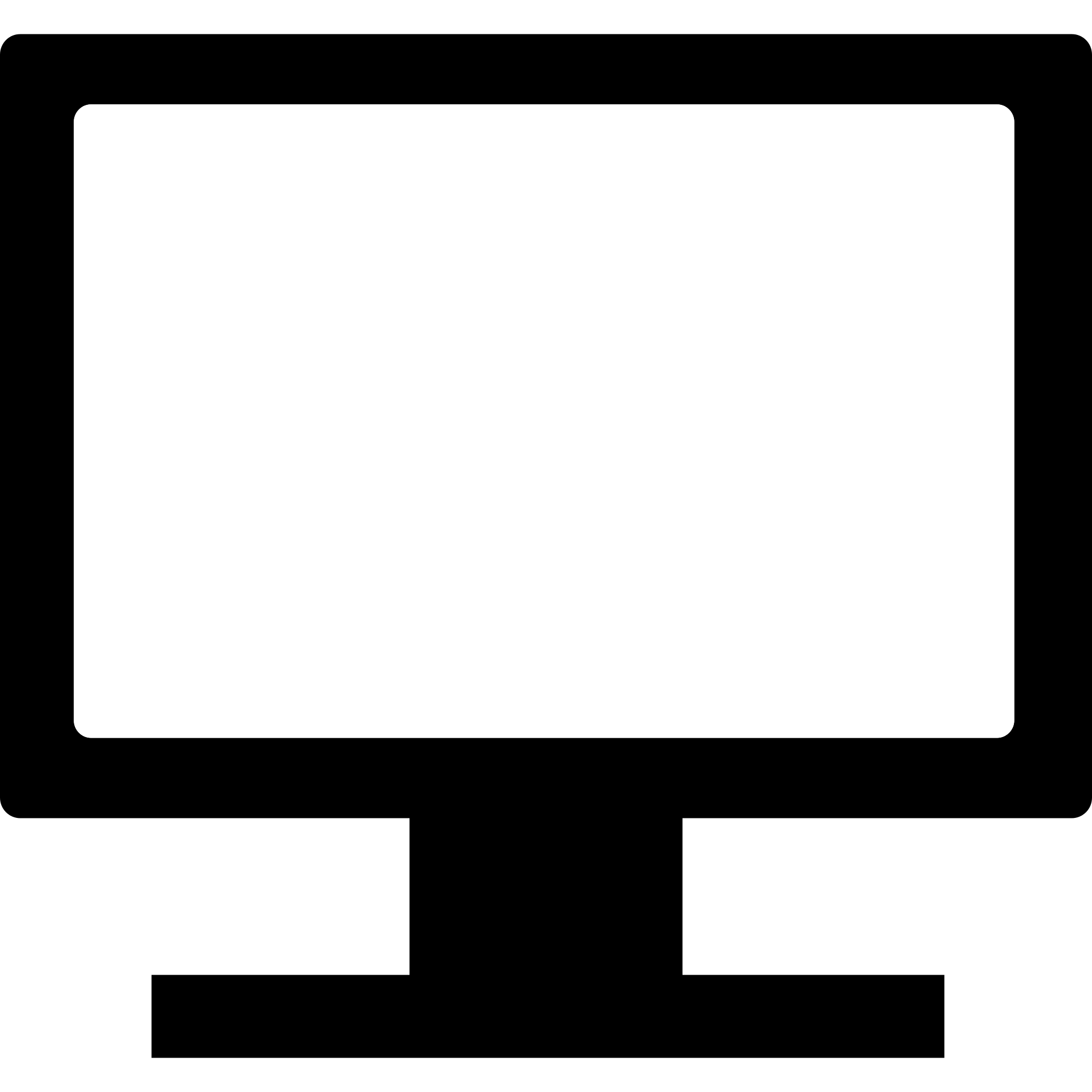}};
\end{tikzpicture}
\end{center}
\caption{Complete process for modeling and solving combinatorial constrained problems.\label{fig:mcsp3}}
\end{figure}

As shown in Figure \ref{fig:mcsp3}, the user who wishes to solve a combinatorial constrained problem has to:
\begin{enumerate}
\item write a model using the Python library \p3 (i.e., write a Python file) 
\item provide a data file (in JSON format) for a specific problem instance to be solved
\item compile both files (model and data) so as to generate an \x3 instance (file)
\item solve the \x3 file (problem instance under format \x3) by using a constraint solver as, e.g., ACE \cite{ace}, Choco \cite{choco}, Picat \cite{ZKF_picat} or CP-SAT \cite{cpsat}.
\end{enumerate}

\medskip
\noindent This approach has many advantages:
\begin{itemize}
\item Python, JSON, and XML are robust mainstream technologies
\item Using JSON for data allows for a unified notation, easy to read for both humans and computers
\item using Python for modeling allows the user to avoid learning again a new programming language
\item Using a coarse-grained XML structure makes it possible to have compact and readable problem instances.
  Note that using JSON instead of XML for representing instances would have been possible but has some drawbacks, as explained in an appendix of \x3 Specifications \cite{xcsp3}.
\end{itemize}

\medskip
\p3 is inspired by both \jv3 \cite{L_jvcsp3} and Numberjack \cite{HOO_numberjack}, and like CPpy \cite{G_cppy}, \p3 can be seen as a Python-embedded CP (Constraint Programming) modeling language.
Currently, \p3 is focused on \x3-core \cite{xcsp3core}, which allows us to use integer variables (with finite domains) and popular constraints.

In this document, you will find more than 60 illustrative models.
Besides, additional ones can be found in \cite{compet22,compet23,compet24,compet25,compet26}, and in our website \href{https://www.pycsp.org}{www.pycsp.org}.

\bigskip\bigskip
\paragraph{Using the Compiler}

As we shall see in this document, for generating an \x3 file from a \p3 model, you have to execute:

\begin{command}
python <model_file> [options] 
\end{command}

with:
\begin{itemize}
\item $<$model\_file$>$: a Python file to be executed, describing a model in \p3
\item $[$options$]$: possible options to be used when compiling
\end{itemize}




\bigskip\bigskip\bigskip
\paragraph{Licence.} 
\p3 is licensed under the \href{https://en.wikipedia.org/wiki/MIT_License}{MIT License}

\paragraph{Code.}
\p3 code is available
\begin{itemize}
\item on Github: \href{https://github.com/xcsp3team/pycsp3}{https://github.com/xcsp3team/pycsp3}
\item as a PyPi package: \href{https://pypi.org/project/pycsp3}{https://pypi.org/project/pycsp3}
\end{itemize}


\tableofcontents

\chapter{Illustrative Models in \p3} \label{sec:illustrative}

{\bf Warning.} In this chapter, we gently introduce \p3 by means of various problems that illustrate the main ingredients of the library.
We also usually show the result of compiling \p3 models into \x3, although that part can be totally ignored.

\section{Single Problems}

We propose to start discovering \p3 with some very simple problems.
We call them {\em single} problems because they are unique (meaning that we do not need to provide any external data when compiling them).

\subsection{A Simple Riddle}\label{sec:riddle}\index{Problems!Riddle}

Remember that when you were young, you were used to playing riddles, some of them having a mathematical background, as for example:
\begin{quote}
{\em Which sequence of four successive integer numbers sum up to 14?}
\end{quote}

\begin{figure}[h]
  \begin{center}
  \includegraphics[scale=0.3]{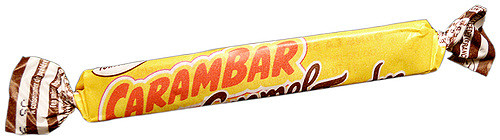}   
\end{center}
  \caption{Famous Riddles in Carambar Candies. \tiny{(image from \href{https://www.flickr.com/photos/bloggyboulga/759966039}{www.flickr.com})} \label{fig:carambar}}  
\end{figure}

If you were already familiar with Mathematics, maybe you were able to formalize this riddle by:
\begin{itemize}
\item introducing four integer variables:
\begin{itemize}
  \item $x_1 \in \N$, $x_2 \in \N$, $x_3 \in \N$, $x_4 \in \N$
\end{itemize}
\item introducing the following mathematical equations (constraints):
\begin{itemize}
\item $x_1+1 = x_2$
\item $x_2+1 = x_3$
\item  $x_3+1=x_4$
\item $x_1 + x_2 + x_3 + x_4 = 14$
\end{itemize}
\end{itemize}

This is a CSP (Constraint Satisfaction Problem) instance, involving four integer variables, three binary constraints (i.e., constraints involving exactly two distinct variables) and one quaternary constraint (i.e., constraint involving exactly four distinct variables).

\medskip
After a rough analysis, we can decide to set 0 as lower bound and 14 as upper bound for the values that can be assigned to the integer variables because, by using that interval of values, we are absolutely certain of not losing any solutions while avoiding reasoning with an infinite set of values.
We then obtain the following \p3 model in a file called `Riddle.py':

\begin{boxpy}\begin{python}
@\imp@

x1 = Var(range(15))
x2 = Var(range(15))
x3 = Var(range(15))
x4 = Var(range(15))

satisfy(
  x1 + 1 == x2,
  x2 + 1 == x3,
  x3 + 1 == x4,
  x1 + x2 + x3 + x4 == 14
)
\end{python}\end{boxpy}

In this Python file, after the first import statement, we declare stand-alone variables by using the \p3 function \nn{Var()}.
Here, we declare four variables called x1, x2, x3, and x4, each one with the set of integers $\{0,1,\dots,14\}$ as domain, which is specified by simply calling the Python function \nn{range()}.

\begin{remark}
In \p3, which is currently targeted to \x3-core, we can only define integer and symbolic variables with finite domains, i.e., variables with a finite set of integers or symbols (strings). 
\end{remark}

To define the domain of a variable, we can simply list values, or use \nn{range()}. For example:

\begin{python}
w = Var(range(15))
x = Var(0, 1)
y = Var(0, 2, 4, 6, 8)
z = Var("a", "b", "c")
\end{python}

declares four variables corresponding to:
\begin{itemize}
\item $w \in \{0, 1, \dots, 14\}$
\item $x \in \{0, 1\}$
\item $y \in \{0, 2, 4, 6, 8\}$
\item $z \in \{a, b,c\}$ 
\end{itemize}

Values can be directly listed as above, or given in a set (and even possibly in a list, although not shown here) as follows:

\begin{python}
w = Var(set(range(15)))
x = Var({0, 1})
y = Var({0, 2, 4, 6, 8})
z = Var({"a", "b", "c"})
\end{python}

It is also possible to name the parameter \nn{dom} when defining the domain:

\begin{python}
w = Var(dom=range(15))
x = Var(dom={0, 1})
y = Var(dom={0, 2, 4, 6, 8})
z = Var(dom={"a", "b", "c"})
\end{python}

\bigskip
Finally, it is of course possible to use generators and comprehension lists/sets. For example, for $y$, we can write:
\begin{python}
y = Var(i for i in range(10) if i % 2 == 0)
\end{python}
or equivalently: 
\begin{python}
y = Var({i for i in range(10) if i % 2 == 0})
\end{python}
or still equivalently:
\begin{python}
y = Var(dom={i for i in range(10) if i % 2 == 0})
\end{python}

\medskip
Now, let us turn to constraints.
When constraints must be imposed on variables, we say that these constraints must be satisfied.
Then, to impose (post) them, we call the \p3 function \nn{satisfy()}, with each constraint passed as a parameter (and so, with commas used as a separator between constraints).
In our example, we have posted four constraints to be satisfied.
These constraints are given in \gb{intension}, by using classical arithmetic, relational and logical operators.
Note that for forcing equality, we need to use `==' in Python (the operator `=' used for assignment cannot be used, because, technically, it cannot be redefined).
In Table \ref{tab:opsexas}, you can find a few other examples of \gb{intension} constraints, while in Tables \ref{tab:ops} and \ref{tab:opsfuncs}, you can find the available operators and functions in \p3.

\begin{table}[b]
  \begin{center}  \begin{tabular}{cp{1cm}c}
  \toprule
  Expressions & & Observations \\
  \midrule
  $x + y < 10$ & & equivalent to $10 > x + y$ \\
  $x*2 - 10*y + 5 == 100$ & & we need to use `==' in Python \\
  $\mathtt{abs}(z[0] - z[1]) >= 2$ & & equivalent to $\mathtt{dist}(z[0], z[1]) >= 2$ \\
  $(x == y)\, |\, (y == 0)$ & & parentheses are required\\
  $\mathtt{either(x == y, y == 0)}$ & & equivalent to  $(x == y)\, |\, (y == 0)$\\
  $\mathtt{disjunction}(x < 2, y < 4, x > y)$ & & equivalent to $(x < 2) \,|\, (y < 4) \,|\, (x > y)$ \\
  $\mathtt{both}(x < w, y > 0)$ & & equivalent to $(x < w) \,\&\, (y > 0)$ \\
  $\mathtt{iff}(x > 0, y > 0)$ & & equivalent to $(x > 0) == (y > 0)$ \\
  $(x == 0)$ \^{} $(y ==1)$ & & use of the logical xor operator \\
  $\mathtt{ift}(x == 0, 5, 10)$ & & the value is 5 if $x$ is 0 else 10 \\ 
  \bottomrule
    \end{tabular}
    \caption{A few examples of expressions denoting \gb{intension} constraints.\label{tab:opsexas}}
  \end{center}
\end{table}

\begin{table}
  \begin{center}
\begin{tabular}{p{0.5cm}l|l}
  \multicolumn{3}{l}{\textcolor{dred}{Arithmetic Operators}} \\
  \midrule
  &  $+$ & addition \\
&  $-$ & subtraction \\
&  $*$ & multiplication \\
&  // & integer division \\
&  \% & remainder \\
&  ** & power \\
  \multicolumn{3}{c}{ } \\
  \multicolumn{3}{l}{\textcolor{dred}{Relational Operators}} \\
  \midrule
  &  $<$ & Less than  \\
&  $<=$ & Less than or equal \\
&  $>=$ & Greater than or equal \\
&  $>$ & Greater than \\
&  $!=$ & Different from \\
  &  $==$ & Equal to \\
   \multicolumn{3}{c}{ } \\
  \multicolumn{3}{l}{\textcolor{dred}{Set Operators}} \\
  \midrule
  &  \nn{in}  & membership \\
  &  \nn{not\, in} & non membership \\
  \multicolumn{3}{c}{ } \\
  \multicolumn{3}{l}{\textcolor{dred}{Logical Operators}} \\
  \midrule
  &  $\sim$ & logical not \\
&  $|$ & logical or \\
  &  \& & logical and \\
    &  \^{} & logical xor \\
\end{tabular}
\caption{Operators that can be used to build expressions (predicates) of \gb{intension} constraints.
  Integer values $0$ and $1$ are respectively equivalent to Boolean values $\nn{False}$ and $\nn{True}$.
  Note that we use the operator $==$ for testing equality and the operators $|$, \& and \^{} for logically combining (sub-)expressions.
  When specifying constraints, we can't use the Python operators \nn{=}, \nn{and}, \nn{or} and \nn{not} (because, technically, they cannot be redefined in Python). \label{tab:ops}}
  \end{center}
\end{table}

\begin{table}
  \begin{center}  \begin{tabular}{p{0.5cm}l|l}
  \multicolumn{3}{l}{\textcolor{dred}{Functions}} \\
  \midrule
  & \nn{abs}() & absolute value of the argument \\
  & \nn{min}() & minimum value of 2 or more arguments \\
  & \nn{max}() & maximum value of 2 or more arguments \\
  & \nn{dist}() & distance between the 2 arguments \\
  & \nn{both}() & conjunction of 2 arguments \\
  & \nn{either}() & disjunction of 2 arguments \\
  & \nn{conjunction}() & conjunction of 2 or more arguments \\
  & \nn{disjunction}() & disjunction of 2 or more arguments \\
  & \nn{imply}() & implication between 2 arguments \\
  & \nn{iff}() & equivalence between 2 or more arguments \\
  & \nn{ift}() & ift(b,u,v) returns u if b is true, v otherwise \\
\end{tabular}
    \caption{Functions that can be used to build expressions (predicates) of \gb{intension} constraints.\label{tab:opsfuncs}}
  \end{center}
\end{table}

\bigskip
Once you have a \p3 model, you can compile it in order to get an \x3 file that can be solved by a constraint solver.
The command is as follows:

\begin{command}
python Riddle.py
\end{command}

The content of the generated \x3 file is: 

\begin{xcsp}
<instance format="XCSP3" type="CSP">
  <variables>
    <var id="x1"> 0..14 </var>
    <var id="x2"> 0..14 </var>
    <var id="x3"> 0..14 </var>
    <var id="x4"> 0..14 </var>
  </variables>
  <constraints>
    <intension> eq(add(x1,1),x2) </intension>
    <intension> eq(add(x2,1),x3) </intension>
    <intension> eq(add(x3,1),x4) </intension>
    <intension> eq(add(x1,x2,x3,x4),14) </intension>
  </constraints>
</instance>
\end{xcsp}

To display the \x3 instance in the standard output (stdout) of the operating system (instead of generating an \x3 file), you can use the option \verb!-display! as follows:

\begin{command}
python Riddle.py -display
\end{command}

Remember that in this first chapter, \x3 files are given to properly understand what is represented by models (and how models are compiled), but if you think that it does not make things clearer for you, you can safely decide to ignore them.
As a user working with the \p3 library and some constraint solvers, you may never need to look at these intermediate \x3 files (although, by experience, it may be helpful in identifying some mistakes in models and some bugs in solvers).

\bigskip
The variables in our model have been declared independently, but it is possible to declare them in a one-dimensional array.
This gives a new \p3 model (version) in a file called `Riddle2.py':

\begin{boxpy}\begin{python}
@\imp@

# x[i] is the ith integer of the sequence
x = VarArray(size=4, dom=range(15))

satisfy(
  x[0] + 1 == x[1],
  x[1] + 1 == x[2],
  x[2] + 1 == x[3],
  x[0] + x[1] + x[2] + x[3] == 14
)

\end{python}\end{boxpy}
and the \x3 file obtained after executing:

\begin{command}
python Riddle2.py
\end{command}

is:
\begin{xcsp}
<instance format="XCSP3" type="CSP">
  <variables>
    <array id="x" note="x[i] is the ith integer of the sequence" size="[4]">
      0..14
    </array>
  </variables>
  <constraints>
    <intension> eq(add(x[0],1),x[1]) </intension>
    <intension> eq(add(x[1],1),x[2]) </intension>
    <intension> eq(add(x[2],1),x[3]) </intension>
    <intension> eq(add(x[0],x[1],x[2],x[3]),14) </intension>
  </constraints>
</instance>
\end{xcsp}

Here, we declare a one-dimensional array of variables: its name (id) is $x$, its size (length) is 4, and each of its variables has $\{0,1,\dots,14\}$ as domain.
Note that we use $x[i]$ for referring to the ($i+1$)th variable of the array (since indexing starts at 0) and that any comment put in the line preceding the declaration of a variable (or variable array) is automatically inserted in the \x3 file.
The \p3 function for declaring an array of variables is \nn{VarArray}() that requires two named parameters \nn{size} and \nn{dom}. 
For declaring a one-dimensional array of variables, the value of \nn{size} must be an integer (or a list containing only one integer), for declaring a two-dimensional array of variables, the value of \nn{size} must be a list containing exactly two integers, and so on.

In some situations, you may want to declare variables in an array with different domains.
For a one-dimensional array, you can give the name of a function that accepts an integer $i$ and returns the domain to be associated with the variable at index $i$ in the array.
For a two-dimensional array, you can give the name of a function that accepts a pair of integers $(i,j)$ and returns the domain to be associated with the variable at indexes $i, j$ in the array.
And so on.
For example, suppose that we have analytically deduced that the first two variables of the array $x$ must be assigned a value strictly less than 6 and the last two variables of the array $x$ must be assigned a value strictly less than 9.
We can write:

\begin{boxpy}\begin{python}
@\imp@

def domain_x(i):
    return range(6) if i < 2 else range(9)  

# x[i] is the ith integer of the sequence
x = VarArray(size=4, dom=domain_x)

satisfy(
  x[0] + 1 == x[1],
  x[1] + 1 == x[2],
  x[2] + 1 == x[3],
  x[0] + x[1] + x[2] + x[3] == 14
)
\end{python}\end{boxpy}

\smallskip
With this new model version, the \x3 file obtained after compilation is:
\begin{xcsp}
<instance format="XCSP3" type="CSP">
  <variables>
     <array id="x" note="x[i] is the ith integer of the sequence" size="[4]">
      <domain for="x[0] x[1]"> 0..5 </domain>
      <domain for="x[2] x[3]"> 0..8 </domain>
    </array>
  </variables>
  <constraints>
    <intension> eq(add(x[0],1),x[1]) </intension>
    <intension> eq(add(x[1],1),x[2]) </intension>
    <intension> eq(add(x[2],1),x[3]) </intension>
    <intension> eq(add(x[0],x[1],x[2],x[3]),14) </intension>
  </constraints>
</instance>
\end{xcsp}

Instead of calling named functions, we can use lambda functions.
This gives: 

\begin{boxpy}\begin{python}
@\imp@

# x[i] is the ith integer of the sequence
x = VarArray(size=4, dom=lambda i: range(6) if i < 2 else range(9))

...  # the rest of the code is similar to the previous model
\end{python}\end{boxpy}

Let us keep analyzing the code of our model.
Because the three binary constraints are similar, one may wonder if we couldn't post these constraints together (in a list).
This is indeed possible by using a comprehension list:

\begin{boxpy}\begin{python}
@\imp@

# x[i] is the ith integer of the sequence
x = VarArray(size=4, dom=range(15))

satisfy(
  # successive integers
  [x[i] + 1 == x[i + 1] for i in range(3)],

  # numbers sum up to 14
  x[0] + x[1] + x[2] + x[3] == 14
)
\end{python}\end{boxpy}

and the \x3 file obtained after compilation is:

\begin{xcsp}
<instance format="XCSP3" type="CSP">
  <variables>
    <array id="x" note="x[i] is the ith integer of the sequence" size="[4]">
      0..14
    </array>
  </variables>
  <constraints>
    <group note="successive integers">
      <intension> eq(add(%0,%1),%2) </intension>
      <args> x[0] 1 x[1] </args>
      <args> x[1] 1 x[2] </args>
      <args> x[2] 1 x[3] </args>
    </group>
    <intension note="numbers sum up to 14">
       eq(add(x[0],x[1],x[2],x[3]),14)
    </intension>
  </constraints>
</instance>
\end{xcsp}

Because of the presence of the comprehension list, we obtain a group of constraints in \x3: basically, we have a constraint template with several parameters identified by \%,
and one ``concrete'' constraint per element \xml{args} providing the effective arguments.
For more information about groups in \x3, see Chapter 10 in \href{https://www.xcsp.org/format3.pdf}{\x3 Specifications}.
Of course, you can use the classical control structures of Python. So, an alternative way of writing the model is:
\begin{boxpy}\begin{python}
@\imp@

# x[i] is the ith integer of the sequence
x = VarArray(size=4, dom=range(15))

for i in range(3):
   satisfy(
     x[i] + 1 == x[i + 1]
   )

satisfy(
  x[0] + x[1] + x[2] + x[3] == 14
)
\end{python}\end{boxpy}

\medskip
Finally, it seems more appropriate to represent the last constraint as a \gb{sum} constraint.
We can then call the \p3 function \nn{Sum}(), which is different from the Python function \nn{sum()}, that builds an object that can be compared, for example, with a value.
This gives:

\begin{boxpy}\begin{python}
@\imp@

# x[i] is the ith integer of the sequence
x = VarArray(size=4, dom=range(15))

satisfy(
  # successive integers
  [x[i] + 1 == x[i + 1] for i in range(3)],

  # numbers sum up to 14
  Sum(x) == 14
)
\end{python}\end{boxpy}

\medskip
and the \x3 file obtained after compilation is:
\begin{xcsp}
<instance format="XCSP3" type="CSP">
  <variables>
    <array id="x" note="x[i] is the ith integer of the sequence" size="[4]">
      0..14
    </array>
  </variables>
  <constraints>
    <group note="successive integers">
      <intension> eq(add(%0,%1),%2) </intension>
      <args> x[0] 1 x[1] </args>
      <args> x[1] 1 x[2] </args>
      <args> x[2] 1 x[3] </args>
    </group>
    <sum note="numbers sum up to 14">
      <list> x[] </list>
      <condition> (eq,14) </condition>
    </sum>
  </constraints>
</instance>
\end{xcsp}

\subsection{Traveling the World}\index{Problems!World Traveling}

Once upon a time, there were three friends called Xavier, Yannick and Zachary, who wanted to travel the world.
However, in their time and country, they were obliged to do their military service.
So, each friend  had to decide if he travels after or before his due military service. 
Xavier and Yannick wanted to travel together.
Xavier and Zachary also wanted to travel together.
However, because Yannick and Zachary didn't always get along very well, they preferred not traveling together.
Can the three friends be satisfied?

\begin{figure}[h!]
\begin{center}
  \includegraphics[scale=0.18]{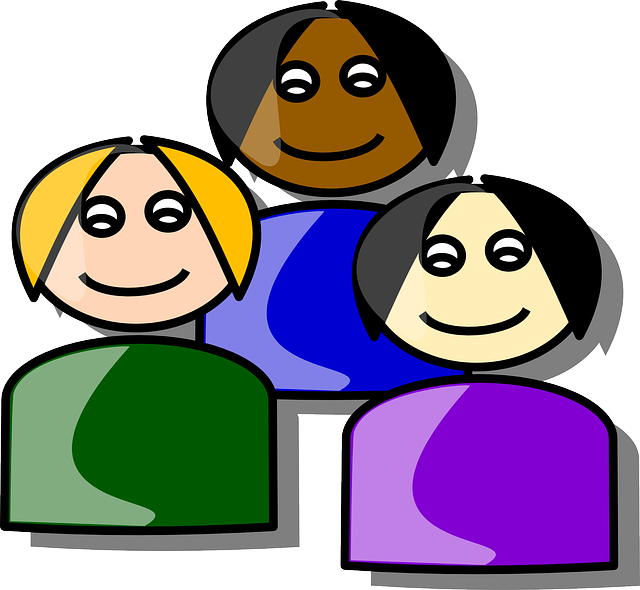}
\end{center}
\caption{Three friends who want to travel the world. \tiny{(image from \href{https://www.maxpixel.net/Three-People-Group-Women-Diversity-Ethnic-23733}{maxpixel.net})} \label{fig:friends}}
\end{figure}

\bigskip
The answer is `no': the three friends cannot make decisions that satisfy all of them.
Certainly, you can deduce this, but imagine that to be quite sure, you want to check it with the help of a constraint solver after having written the model.
For the model, first, we just have to declare three variables $x$, $y$, and $z$ denoting the decisions made by the three friends Xavier, Yannick and Zachary.
For each variable, two values are possible: $a$ (after the military service) and $b$ (before the military service).
Concerning the constraints, we have to enumerate the combinations of values that satisfy each pair of friends.
We obtain a constraint network, which can be drawn 
under the form of a compatibility graph. 
Figure \ref{fig:small} presents the compatibility graph of the small constraint network $P$ depicted above:
\begin{itemize}
\item the set of variables of $P$ is $\{x,y,z\}$, each variable having $\{a,b\}$ as domain;
\item the set of constraints of $P$ is $\{(x,y) \in \{(a,a),(b,b)\}, (x,z) \in \{(a,a),(b,b)\},$ $ (y,z) \in \{(a,b),(b,a)\}\}$.
\end{itemize}

\begin{figure}[h!]
  \begin{center}
    \begin{tikzpicture}[node distance=1cm, >=stealth, auto] 
      \tikzstyle{node}=[state,minimum size=6mm,fill=white,draw,font=\sffamily\normalsize]
      \tikzstyle{arete}=[thick,-,>=stealth]
      \def\xbase{0}
      \def\ybase{2.5}
      
      \node[node] (xa) at (\xbase+0.5,\ybase+0.5) {$a$};
      \node[node] (xb) at (\xbase,\ybase) {$b$};
      \draw (\xbase-0.4,\ybase+0.8) node {$x$};
      
      \node[node] (ya) at (2,1.85) {$a$};
      \node[node] (yb) at (2,1.15) {$b$};
      \draw (2.8,1.5) node {$y$};
      
      \node[node] (za) at (0.5,0) {$a$};
      \node[node] (zb) at (0,0.5) {$b$};
      \draw (-0.4,-0.3) node {$z$};
      
      \draw[rotate around={45:(0.25,2.75)}] (0.25,2.75) ellipse (1cm and 0.5cm); 
      \draw (2,1.5) ellipse (0.5cm and 1cm); 
      \draw[rotate around={-45:(0.25,0.25)}] (0.25,0.25) ellipse (1cm and 0.5cm); 
      
      \draw (xa) -- (ya) ;
      \draw (xb) -- (yb) ;  
      \draw (xa) -- (za) ;
      \draw (xb) -- (zb) ;
      \draw (ya) -- (zb) ;
      \draw (yb) -- (za) ;
    \end{tikzpicture}
  \end{center}
  \caption{The compatibility graph of a small constraint network.\label{fig:small}}
\end{figure}


Here, the constraints directly indicate what is authorized; we call such constraints \gb{extension} constraints (or table constraints). 
For example, we know that we can satisfy the binary constraint involving the variables $x$ and $y$ by assigning both variables with either value $a$ or value $b$. 
The interested reader can observe that the constraint network is arc-consistent (AC) but not path-inverse consistent (PIC).
But don't worry! It doesn't matter here if you do not know anything about these properties. 

\medskip
The \p3 model for our problem, in a file called `WorldTraveling.py', is:

\begin{boxpy}\begin{python}
@\imp@

a, b = "a", "b"  # two symbols (after, before)

x = Var(a,b)
y = Var(a,b)
z = Var(a,b)

satisfy(
  (x,y) in {(a,a), (b,b)},
  (x,z) in {(a,a), (b,b)},
  (y,z) in {(a,b), (b,a)}
)
\end{python}\end{boxpy}

For compiling it, we execute:

\begin{command}
python WorldTraveling.py
\end{command}

and the \x3 file obtained after compilation is:

\begin{xcsp}
<instance format="XCSP3" type="CSP">
  <variables>
    <var id="x" type="symbolic"> a b </var>
    <var id="y" type="symbolic"> a b </var>
    <var id="z" type="symbolic"> a b </var>
  </variables>
  <constraints>
    <extension>
      <list> x y </list>
      <supports> (a,a)(b,b) </supports>
    </extension>
    <extension>
      <list> x z </list>
      <supports> (a,a)(b,b) </supports>
    </extension>
    <extension>
      <list> y z </list>
      <supports> (a,b)(b,a) </supports>
    </extension>
  </constraints>
</instance>
\end{xcsp}

Here, we declare three stand-alone symbolic variables (note how the domain of each of them is simply composed of the two symbols "a" and "b").
And we declare three binary \gb{extension} constraints.
In \p3, we simply use the operator \nn{in} to represent such constraints: a tuple of variables representing the scope of the constraint is given at the left of the operator and a set of tuples of values is given at the right of the operator.
This is basically what we write in mathematical form.
Note that we use \nn{in} when the constraint enumerates the allowed tuples (called supports), as in our example, and \nn{not in} when the constraint enumerates the forbidden tuples (called conflicts).  

Now, suppose that instead of declaring symbolic variables, you prefer to declare integer variables.
By replacing "a" by 0 and "b" by 1, you can write:

\begin{boxpy}\begin{python}
@\imp@

x = Var(0,1)
y = Var(0,1)
z = Var(0,1)

satisfy(
  (x,y) in {(0,0), (1,1)},
  (x,z) in {(0,0), (1,1)},
  (y,z) in {(0,1), (1,0)}
)
\end{python}\end{boxpy}

which, when compiled, gives:

\begin{xcsp}
<instance format="XCSP3" type="CSP">
  <variables>
    <var id="x"> 0 1 </var>
    <var id="y"> 0 1 </var>
    <var id="z"> 0 1 </var>
  </variables>
  <constraints>
    <extension>
      <list> x y </list>
      <supports> (0,0)(1,1) </supports>
    </extension>
    <extension>
      <list> x z </list>
      <supports> (0,0)(1,1) </supports>
    </extension>
    <extension>
      <list> y z </list>
      <supports> (0,1)(1,0) </supports>
    </extension>
  </constraints>
</instance>
\end{xcsp}

Note that the scope of an \gb{extension} constraint is expected to be given under the form of a tuple, but can be given under the form of a list too.
Similarly, the table of an \gb{extension} constraint is expected to be given under the form of a set, but can be given under the form of a list too.
This means that, for example, it is possible to write:

\begin{python}
  [x,y] in [(0,0), (1,1)]
\end{python}

but personally, we prefer to stay closer to pure mathematical forms (but for efficiency reasons, we may use lists for huge tables). 

\section{Academic Problems}

Contrary to single problems, {\em academic} problems require the introduction of some elementary pieces of data from the user: a fixed number of integers (and/or strings).

\subsection{Queens Problem}\label{sec:queens}\index{Problems!Queens}

The problem is stated as follows: can we put 8 queens on a chessboard such that no two queens attack each other?
Two queens attack each other iff they belong to the same row, the same column or the same diagonal.
An illustration is given by Figure \ref{fig:queens}.

By considering boards of various size, the problem can be generalized as follows: can we put $n$ queens on a board of size $n \times n$ such that no two queens attack each other?
Contrary to previously introduced single problems, we have to deal here with a family of problem instances, each of them characterized by a specific value of $n$.
We can try to solve the 8-queens instance, the 10-queens instance, and even the 1000-queens instance.

\begin{figure}[h]
  \centering
    \begin{subfigure}[t]{0.5\textwidth}
        \centering
        \setchessboard{boardfontencoding=LSBC4,showmover=false}
        \def\mylist{}
        \setchessboard{setpieces=\mylist}
        \chessboard
        \caption{Puzzle}
    \end{subfigure}%
    ~ 
    \begin{subfigure}[t]{0.5\textwidth}
        \centering
        \setchessboard{boardfontencoding=LSBC4,showmover=false}
        \def\mylist{Qa3, Qb6, Qc4, Qd2, Qe8, Qf5, Qg7, Qh1}
        \setchessboard{setpieces=\mylist}
        \chessboard
        \caption{Solution}
    \end{subfigure}
    \caption{Putting 8 queens on a chessboard \label{fig:queens}}
\end{figure}

For such problems, we have to separate the description of the model from the description of the data.
In other words, we have to write a model with some kind of parameters.
In \p3, what you have to do is:
\begin{enumerate}
\item clearly identify the parameters of the problem (names and structures)
\item use these parameters in your model by means of the predefined \p3 variable called \nn{data} 
\item specify effective values of these parameters when you compile to \x3   
\end{enumerate}

In our case, we have only one integer parameter called $n$.
If we associate a variable $q_i$ with the $(i+1)$th row of the board, then we can simply post the following \gb{intension} constraints:
\begin{quote}
  $q_i \neq q_j \land |q_i - q_j| \neq j - i, \forall i, j : 0 \leq i < j < n$
\end{quote}
Indeed, this way, we have the guarantee that queens are on different columns (since $q_i \neq q_j$) and on different diagonals (since the column distance $|q_i - q_j|$ is different from the row distance $|i -j| = j -i$).

\medskip
This can be translated into a \p3 model in a file `Queens.py':

\begin{boxpy}\begin{python}
@\imp@

n = data

# q[i] is the column of the ith queen (at row i)
q = VarArray(size=n, dom=range(n))

for i in range(n):
  for j in range(i+1, n):
    satisfy(
      (q[i] != q[j]) & (abs(q[i] - q[j]) != j - i) 
    )
\end{python}\end{boxpy}

Note how the parameter $n$ is given by the value of the predefined \p3 variable \nn{data}.
This is because there is only one parameter here; later, we shall see that for more than one parameter, \nn{data} is given under the form of a tuple. 
In our model, there is a constraint for any pair $(i,j)$ such that $0 \leq i < j < n$.
Note that when expressions are logically combined (here, with the operator '\&'), we need to put them between parentheses.

For iterating over pairs of indexes, we can use the (slightly extended) function \nn{combinations} from package \nn{itertools}, as follows:

\begin{boxpy}\begin{python}
@\imp@

n = data

# q[i] is the column of the ith queen (at row i)
q = VarArray(size=n, dom=range(n))

for i, j in combinations(n, 2):
  satisfy(
    (q[i] != q[j]) & (abs(q[i] - q[j]) != j - i) 
  )
\end{python}\end{boxpy}

We can also use a comprehension list (actually a generator, since brackets are omitted here although we could have inserted them), as follows:

\begin{boxpy}\begin{python}
@\imp@

n = data

# q[i] is the column of the ith queen (at row i)
q = VarArray(size=n, dom=range(n))

satisfy(
  (q[i] != q[j]) & (abs(q[i] - q[j]) != j - i) for i, j in combinations(n, 2) 
)
\end{python}\end{boxpy}

You may find annoying, or rather unclear, to use the symbol '\&' for applying a logical conjunction (logical and) between the two parts of the expression.
There are at least three equivalent alternatives.
First, you can post constraints (parts) separately, by posting two groups (lists):

\begin{boxpy}\begin{python}
@\imp@

n = data

# q[i] is the column of the ith queen (at row i)
q = VarArray(size=n, dom=range(n))

satisfy(
  [q[i] != q[j] for i, j in combinations(n, 2)],

  [abs(q[i] - q[j]) != j - i for i, j in combinations(n, 2)] 
)
\end{python}\end{boxpy}

Second, you can put the two constraints in a tuple (or list) while iterating over the combinations:

\begin{boxpy}\begin{python}
@\imp@
    
n = data

# q[i] is the column of the ith queen (at row i)
q = VarArray(size=n, dom=range(n))

satisfy(
  (
    q[i] != q[j],
    abs(q[i] - q[j]) != j - i
  ) for i, j in combinations(n, 2)
)
\end{python}\end{boxpy}

Third, you can call the function \nn{both()}:

\begin{boxpy}\begin{python}
@\imp@
    
n = data

# q[i] is the column of the ith queen (at row i)
q = VarArray(size=n, dom=range(n))

satisfy(
  both(
    q[i] != q[j],
    abs(q[i] - q[j]) != j - i
  ) for i, j in combinations(n, 2)
)
\end{python}\end{boxpy}

Note that you do not get exactly the same result (i.e., \x3 file) when compiling: while some alternatives (models) force a simple tree expression (constraint), some others generate separate constraints.
In general, this may have an impact on solver efficiency, but there are no general rules for anticipating the right choice (so as to get the most efficient form for the solving process). 

\medskip
Now, the question is: how can we solve a specific instance?
The answer is: just compile the model while indicating with the option \verb!-data! either the value for $n$ or the name of a JSON file containing an object with a unique field $n$.
In the former case, this gives for $n=4$: 

\begin{command}
python Queens.py -data=4
\end{command}

and the \x3 file obtained after compilation is:

\begin{xcsp}
<instance format="XCSP3" type="CSP">
  <variables>
    <array id="q" note="q[i] is the column of the ith queen (at row i)" size="[4]">
      0..3
    </array>
  </variables>
  <constraints>
    <group>
      <intension> and(ne(%0,%1),ne(abs(sub(%0,%1)),%2)) </intension>
      <args> q[0] q[1] 1 </args>
      <args> q[0] q[2] 2 </args>
      <args> q[0] q[3] 3 </args>
      <args> q[1] q[2] 1 </args>
      <args> q[1] q[3] 2 </args>
      <args> q[2] q[3] 1 </args>
    </group>
  </constraints>
</instance>
\end{xcsp}

In the latter case, just build a file `queens-4.json' whose content is:
\begin{json}
{
  "n": 4
}
\end{json}

and execute:

\begin{command}
python Queens.py -data=queens-4.json
\end{command}

In our situation where only one integer is needed (and more generally, for any academic problem), it is a little bit of overkill to use JSON files. 

\medskip
Remember that once you have an \x3 file, you can run any solver that recognizes this format: ACE, Choco, Picat, $\dots$ 
\medskip

\medskip
At this point, suppose that you have been told that it could be a good idea to post \gb{allDifferent} constraints; remember that an \gb{allDifferent} constraint imposes that all involved variables (or expressions) must take different values. 
It is known (you can try to make the mathematical proof) that it suffices to post three constraints as in the following model:

\begin{boxpy}\begin{python}
@\imp@

n = data

# q[i] is the column of the ith queen (at row i)
q = VarArray(size=n, dom=range(n))

satisfy(
  # all queens are put on different columns
  AllDifferent(q),

  # no two queens on the same upward diagonal
  AllDifferent(q[i] + i for i in range(n)),
  
  # no two queens on the same downward diagonal
  AllDifferent(q[i] - i for i in range(n))
)
\end{python}\end{boxpy}

\medskip
After compilation, we obtain:

\begin{xcsp}
<instance format="XCSP3" type="CSP">
  <variables>
    <array id="q" note="q[i] is the column of the ith queen (at row i)" size="[4]">
      0..3
    </array>
  </variables>
  <constraints>
    <allDifferent note="all queens are put on different columns">
      q[]
    </allDifferent>
    <allDifferent note="no two queens on the same upward diagonal">
      add(q[0],0) add(q[1],1) add(q[2],2) add(q[3],3)
    </allDifferent>
    <allDifferent note="no two queens on the same downward diagonal">
      sub(q[0],0) sub(q[1],1) sub(q[2],2) sub(q[3],3)
    </allDifferent>
  </constraints>
</instance>
\end{xcsp}

\begin{remark}
In \p3, most of the global constraints are posted by calling a function whose first letter is uppercase, as for example \nn{AllDifferent}(), \nn{Sum}(), and \nn{Cardinality}(). 
\end{remark}

\medskip
Maybe, you think that it is annoying to have several files for various model variants (as a side remark, have you observed how many frameworks generate hundreds and even thousands of files; this is crazy!).
In fact, you can put different model variants in the same file by using the \p3 function \nn{variant}() that accepts a string as parameter (or nothing).
When you compile, you can then indicate the name of the variant.
Putting the two variants seen earlier in the same file `Queens.py' gives:

\begin{boxpy}\begin{python}
@\imp@

n = data

# q[i] is the column of the ith queen (at row i)
q = VarArray(size=n, dom=range(n))

if not variant():
   satisfy(
     # all queens are put on different columns
     AllDifferent(q),

     # no two queens on the same upward diagonal
     AllDifferent(q[i] + i for i in range(n)),
     
     # no two queens on the same downward diagonal
     AllDifferent(q[i] - i for i in range(n))
   )

elif variant("bin"):
   satisfy(
     both(
       q[i] != q[j],
       abs(q[i] - q[j]) != j - i
     ) for i, j in combinations(n, 2)
   )
\end{python}\end{boxpy}

To compile the main model (variant), just type:

\begin{command}
python Queens.py -data=4
\end{command}

To compile the model variant "bin", just type:

\begin{command}
python Queens.py -data=4 -variant=bin 
\end{command}

\subsection{Board Coloration} \label{sec:boardColoration}\index{Problems!Board Coloration}

The (chess)board coloration problem is to color all squares of a board composed of $n$ rows and $m$ columns such that the four corners of any rectangle in the board must not be assigned the same color.
Importantly, we want to minimize the number of used colors.

\begin{figure}[h]
\begin{center}
  \includegraphics[scale=0.4]{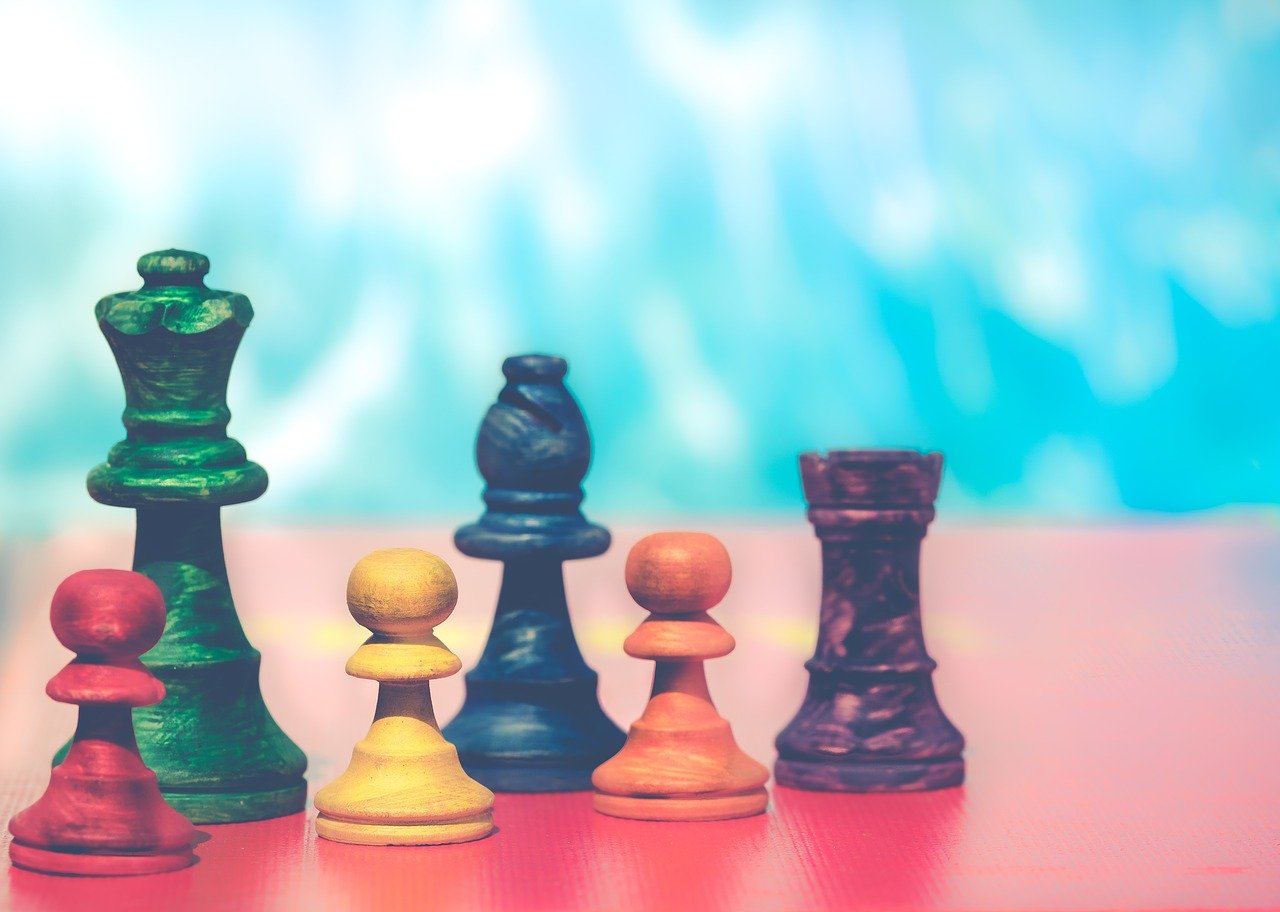} 
\end{center}
\caption{Coloring Boards. \tiny{(image by Ylanite Koppens on \href{https://pixabay.com/fr/photos/pions-figures-d-\%C3\%A9checs-color\%C3\%A9-3467512}{Pixabay})} \label{fig:board}}
\end{figure}

This time, we then need two integer parameters $n$ and $m$.
These values will be given by the predefined \p3 variable \nn{data} that is expected to be a tuple (if data are correctly given at compile time, of course). 
After a very rough analysis, we can decide to use $n \times m$ as an upper bound of the number of used colors. 
This gives a \p3 model in a file `BoardColoration.py':

\begin{boxpy}\begin{python}
@\imp@

n, m = data

# x[i][j] is the color at row i and column j
x = VarArray(size=[n, m], dom=range(n * m))

satisfy(
  # at least 2 corners of different colors for any rectangle inside the board
  NValues(x[i1][j1], x[i1][j2], x[i2][j1], x[i2][j2]) > 1
    for i1, i2 in combinations(n, 2) for j1, j2 in combinations(m, 2) 
)

minimize(
  # minimizing the greatest used color index (and so, the number of colors)
  Maximum(x)
)
\end{python}\end{boxpy}

The user is expected to give two integer values, automatically put in \nn{data} under the form of a tuple.
This is why we have the possibility of using tuple unpacking in our model.
Of course, this is equivalent to writing:

\begin{python}
n, m = data[0], data[1]
\end{python}

Here, we declare a two-dimensional array of variables: its name is $x$, its size is $n \times m$ and each of its variables has $\{0, 1, \dots, n \times m-1\}$ as domain.
We then need to post several \gb{notAllEqual} constraints.
Actually, this constraint is a special case of the \gb{nValues} constraint: we want the number of different values taken by some variables (the scope of the constraint) being strictly greater than 1.
This is given in the model by an expression involving the \p3 function \nn{NValues}().

\medskip
Finally, the objective function corresponds to the minimization of the maximum value taken by any variable in the two-dimensional array $x$.
Because domains are all similar, this is indeed equivalent to minimizing the number of used colors.
For an optimization problem, you can call either the \p3 function \nn{minimize}() or the \p3 function \nn{maximize}().
You can use different kinds of parameters:
\begin{itemize}
\item a stand-alone variable
\item a general arithmetic expression, like in \nn{u * 3 + v} where $u$ and $v$ are two variables 
\item a sum over a list (array) of variables by using the function \nn{Sum}(), like in \nn{Sum(x)}
\item a dot product, like in \nn{[u, v, w] * [2, 4, 3]} where $u$, $v$ and $w$ are three variables
\item a minimum by using the function \nn{Minimum}(), like in \nn{Minimum(x)}
\item a maximum by using the function \nn{Maximum}(), like in \nn{Maximum(x)}
\item a number of different values by using the function \nn{NValues}(), like in \nn{NValues(x)}
\end{itemize}

As we shall see later, it is even possible to build still more general (arithmetic) expressions involving functions \nn{Sum}(), \nn{Minimum}(), etc.

\medskip
To solve a specific instance, as usually, we have first to compile the model while indicating with the option \verb!-data! either the values for $n$ and $m$ (between brackets) or the name of a JSON file containing an object with two integer fields. 
In the former case, this gives for $n=3$ and $m=4$: 

\begin{command}
python BoardColoration.py -data=[3,4]
\end{command}

With some operating systems (shells), you may need to escape brackets, which gives:

\begin{command}
python BoardColoration.py -data=\[3,4\]
\end{command}

The \x3 file obtained after compilation is:

\begin{xcsp}
<instance format="XCSP3" type="COP">
  <variables>
    <array id="x" size="[3][4]" note="x[i][j] is the color at row i and col j">
      0..11
    </array>
  </variables>
  <constraints>
    <group note="at least 2 corners of different colors for any rectangle">
      <nValues>
        <list> %... </list>
        <condition> (gt,1) </condition>
      </nValues>
      <args> x[0][0] x[0][1] x[1][0] x[1][1] </args>
      <args> x[0][0] x[0][2] x[1][0] x[1][2] </args>
      ...  // ellipsis
      <args> x[1][1] x[1][2] x[2][2] x[2][3] </args>
    </group>
  </constraints>
  <objectives>
    <minimize type="maximum"> x[][] </minimize>
  </objectives>
</instance>
\end{xcsp}

Of course, because tuple unpacking is used for data in our model, the order is important: the first value is for $n$ and the second one for $m$.
If ever we use a JSON file for the data, it is also important to have $n$ before $m$:

\begin{json}
{
  "n": 3,
  "m": 4
}
\end{json}


However, you can relax this requirement by avoiding tuple unpacking for data, and instead write in the model something like:

\begin{python}
n, m = data.n, data.m
\end{python}

It means that \nn{data} is now expected to be a named tuple (and not simply a classical tuple).
To benefit from named tuples, you have to either indicate names when specifying data, as for example, in:

\begin{command}
python BoardColoration.py -data=[m=4,n=3]
\end{command}

or use a JSON file (whatever the order of the fields of the root object in the file).

\medskip
This being said, we prefer personally to use tuple unpacking for data because it is more concise.

\medskip
As a matter of fact, this problem has many symmetries.
It is known that we can break variable symmetries by posting a lexicographic constraint between any two successive rows and any two successive columns.
For posting lexicographic constraints, we can use the \p3 functions \nn{LexIncreasing}() and \nn{LexDecreasing}().
Besides, we can use two optional named parameters \nn{strict} and \nn{matrix} whose default values are \nn{False}.
When \nn{matrix} is set to \nn{True}, it means that the constraint must be applied on each row and each column of the specified two-dimensional array.
On the other hand, it is relevant to tag this constraint because it clearly informs us that it is inserted for breaking symmetries: tagging is made possible by putting in a comment line an expression of the form \nn{tag()}, with a token (or a sequence of tokens separated by a white-space) between parentheses. 
The model is now:

\begin{boxpy}\begin{python}
@\imp@

n, m = data

# x[i][j] is the color at row i and column j
x = VarArray(size=[n, m], dom=range(n * m))

satisfy(
  # at least 2 corners of different colors for any rectangle inside the board
  [NValues(x[i1][j1], x[i1][j2], x[i2][j1], x[i2][j2]) > 1
     for i1, i2 in combinations(n, 2)  for j1, j2 in combinations(m, 2)], 

  # tag(symmetry-breaking)
  LexIncreasing(x, matrix=True)
)

minimize(
  # minimizing the greatest used color index (and so, the number of colors)
  Maximum(x)
)
\end{python}\end{boxpy}

\medskip
After compilation, we have the following additional element in the generated \x3 file:

\begin{xcsp}
    <lex class="symmetry-breaking">
      <matrix> x[][] </matrix>
      <operator> le </operator>
    </lex>
\end{xcsp}

\medskip
Note the presence of the attribute \nn{class} that results from the insertion of the expression \nn{tag()}.
Easily, a solver can now solve this instance with or without symmetry breaking.
Indeed, at time of parsing, it is quite easy to discard XML elements with a specified tag (class): this is currently made possible with the available parsers in Java and C++ for \x3.
The interest is that we have only one file, which can be used for testing different model variations.

\subsection{Magic Sequence} \label{sec:magicSequence} \index{Problems!Magic Sequence}

A magic sequence of order $n$ is a sequence of integers $x_0,\dots,x_{n-1}$ between 0 and $n-1$, such that each value $i \in 0..n-1$ occurs exactly $x_i$ times in the sequence.
For example,

\begin{command}
  6 2 1 0 0 0 1 0 0 0
\end{command}

is a magic sequence of order 10 since 0 occurs 6 times, 1 occurs twice, $\dots$ and 9 occurs 0 times.

\medskip
One can mathematically prove that every solution respects:
\begin{quote}
  $x_0 + x_1 + x_2 + x_3 + \dots + x_{n-1} = n$
\end{quote}
and
\begin{quote}
  $-x_0 + 0x_1 + x_2 + 2x_3 + \dots + (n-2)x_{n-1} = 0$
\end{quote}

So, it may be a good idea to post these additional constraints for improving the filtering process of the search space while making it clear that they are redundant (i.e., not modifying the set of solutions) by using an appropriate tag.
This gives a \p3 model in a file `MagicSequence.py':

\begin{boxpy}\begin{python}
@\imp@

n = data

# x[i] is the ith value of the sequence
x = VarArray(size=n, dom=range(n))

satisfy(
  # each value i occurs exactly x[i] times in the sequence
  Cardinality(x, occurrences={i: x[i] for i in range(n)}),

  # tag(redundant)
  [
    Sum(x) == n,
    Sum((i - 1) * x[i] for i in range(n)) == 0
  ]
)
\end{python}\end{boxpy}

On the one hand, the \gb{cardinality} constraint is exactly what we need here.
Here, the \p3 function \nn{Cardinality}() we use simply states that each value $i$ in $0..n-1$ must occur exactly $x[i]$ times; a required named parameter called \nn{occurrences} is given as value a Python dictionary for storing that information\footnote{It is also possible to write simply \texttt{occurrences=x} that automatically builds a dictionary as follows: \\ \texttt{occurrences=\{i:\,x[i] for i in range(len(x))\}}}.
On the other hand, we have put together the two additional constraints in a list, permitting to tag these two constraints with the token ``redundant''.

Now, if we execute:

\begin{command}
python MagicSequence.py -data=6
\end{command}

we obtain the following \x3 instance:

\begin{xcsp}
<instance format="XCSP3" type="CSP">
  <variables>
    <array id="x" note="x[i] is the ith value of the sequence" size="[6]">
      0..5
    </array>
  </variables>
  <constraints>
    <cardinality note="each value i occurs exactly x[i] times in the sequence">
      <list> x[] </list>
      <values> 0 1 2 3 4 5 </values>
      <occurs> x[] </occurs>
    </cardinality>
    <block class="redundant">
      <sum>
        <list> x[] </list>
        <condition> (eq,6) </condition>
      </sum>
      <sum>
        <list> x[] </list>
        <coeffs> -1 0 1 2 3 4 </coeffs>
        <condition> (eq,0) </condition>
      </sum>
    </block>
  </constraints>
</instance>
\end{xcsp}

\subsection{Golomb Ruler}\label{sec:golomb} \index{Problems!Golomb Ruler}

This problem (and its variants) is said to have many practical applications including sensor placements for x-ray crystallography and radio astronomy. 
A Golomb ruler is defined as a set of $n$ integers $0 = a_1 < a_2 < ... < a_n$ such that the $n \times (n-1)/2$ differences $a_j - a_i$, $1 \leq i < j \leq n$, are distinct. 
Such a ruler is said to contain $n$ marks (or ticks) and to be of length $a_n$. 
The objective is to find optimal rulers (i.e., rulers of minimum length). 
An optimal ruler for $n=4$ is illustrated below:


\begin{figure}[h]
\begin{center}
  \includegraphics[scale=0.13]{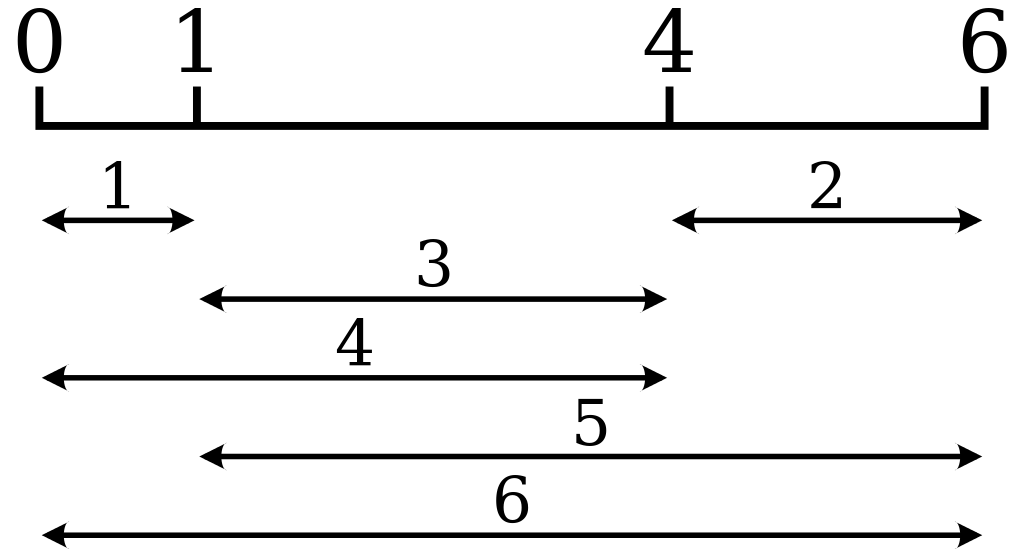}
\end{center}
\caption{An Optimal Golomb Ruler with 4 Ticks. \tiny{(image from \href{https://commons.wikimedia.org/wiki/File:Golomb_Ruler-4.svg}{commons.wikimedia.org})} \label{fig:golomb}}
\end{figure}

Dimitromanolakis has computed relatively short Golomb rulers
and thus showed with computer aid that the optimal ruler for $n \leq 65,000$ has length less than $n^2$.

A simple model involves a single constraint \gb{allDifferent}:

\begin{boxpy}\begin{python}
@\imp@

n = data

# x[i] is the position of the ith tick
x = VarArray(size=n, dom=range(n * n))

satisfy(
  # all distances are different
  AllDifferent(abs(x[i] - x[j]) for i, j in combinations(n, 2))
)

minimize(
  # minimizing the position of the rightmost tick
  Maximum(x)
) 
\end{python}\end{boxpy}

Another model variant involves auxiliary variables and ternary constraints.
This variant shows how we can handle holes (``undefined'' variables) in variable arrays. 
This variant is:

\begin{boxpy}\begin{python}
@\imp@

n = data

def domain_y(i, j):
   return range(1, n * n) if i < j else None

# x[i] is the position of the ith tick
x = VarArray(size=n, dom=range(n * n))

# y[i][j] is the distance between x[i] and x[j] for i strictly less than j
y = VarArray(size=[n, n], dom=domain_y) 

satisfy(
  # all distances are different
  AllDifferent(y),

  # linking variables from both arrays
  [x[j] == x[i] + y[i][j] for i, j in combinations(n, 2)]
)

minimize(
  # minimizing the position of the rightmost tick
  Maximum(x)
)
\end{python}\end{boxpy}

Here, we declare a two-dimensional array of variables, called $y$, even if only the part in this array above the main diagonal really contains variables.
This is handled by the auxiliary function \nn{domain\_y}() that returns an actual domain for a pair $(i,j)$ when $i < j$, and \nn{None} otherwise.
This way, we can simply post a constraint \gb{allDifferent} by specifying the array $y$ (even if $y$ contains some ``undefined'' cells/variables).

Of course, it is possible to use a lambda function when defining domains.
Concerning symmetry breaking, we can decide to force $x[0]$ to be equal to 0, and to impose a strict increasing order on variables of $x$.
When we want the values of a sequence of variables to be in increasing or decreasing order, we can call the \p3 functions \nn{Increasing}() or \nn{Decreasing}(); the named parameter \nn{strict} can be used to indicate that the order must be strict. 
We obtain now:

\begin{boxpy}\begin{python}
@\imp@

n = data

# x[i] is the position of the ith tick
x = VarArray(size=n, dom=range(n * n))

# y[i][j] is the distance between x[i] and x[j] for i strictly less than j
y = VarArray(size=[n, n], dom=lambda i, j: range(1, n * n) if i < j else None) 

satisfy(
  # all distances are different
  AllDifferent(y),

  # linking variables from both arrays
  [x[j] == x[i] + y[i][j] for i, j in combinations(n, 2)],
  
  # tag(symmetry-breaking)
  [
    x[0] == 0,
    Increasing(x, strict=True)
  ]
)

minimize(
  # minimizing the position of the rightmost tick
  Maximum(x)
)
\end{python}\end{boxpy}

For $n=4$, we obtain: 




\begin{xcsp}
<instance format="XCSP3" type="COP">
  <variables>
    <array id="x" note="x[i] is the position of the ith tick" size="[4]">
      0..16
    </array>
    <array id="y" note="y[i][j] is the distance between x[i] and x[j] for i strictly less than j" size="[4][4]">
      1..16
    </array>
  </variables>
  <constraints>
    <allDifferent note="all distances are different">
      y[0][1..3] y[1][2..3] y[2][3]
    </allDifferent>
    <group note="linking variables from both arrays">
      <intension> eq(%0,add(%1,%2)) </intension>
      <args> x[1] x[0] y[0][1] </args>
      <args> x[2] x[0] y[0][2] </args>
      <args> x[3] x[0] y[0][3] </args>
      <args> x[2] x[1] y[1][2] </args>
      <args> x[3] x[1] y[1][3] </args>
      <args> x[3] x[2] y[2][3] </args>
    </group>
    <block class="symmetry-breaking">
      <intension> eq(x[0],0) </intension>
      <ordered>
        <list> x[] </list>
        <operator> lt </operator>
      </ordered>
    </block>
  </constraints>
  <objectives>
    <minimize note="minimizing the position of the rightmost tick" type="maximum">
      x[]
    </minimize>
  </objectives>
</instance>
\end{xcsp}


Technically, the undefined variables of the array $y$ in the \p3 model are not identified as such in the \x3 instance (see the element \xml{array} for $y$).
However, although not explicitly identified as undefined, they can be discarded by solvers because they are involved nowhere (neither in the constraints nor in the objective); see how the constraint \xml{allDifferent} only involves the variables in the upper half of the two-dimensional array $y$.

\section{Structured Problems}

Some problems need more than elementary data, that is to say, more than a few elementary pieces of data such as integers. 
In this document, we call them {\em structured} problems. 

\subsection{Sudoku}\label{sec:sudoku} \index{Problems!Sudoku}

\newcounter{row}
\newcounter{col}

\newcommand\setrow[9]{
  \setcounter{col}{1}
  \foreach \n in {#1, #2, #3, #4, #5, #6, #7, #8, #9} {
    \edef\x{\value{col} - 0.5}
    \edef\y{9.5 - \value{row}}
    \node[anchor=center] at (\x, \y) {\n};
    \stepcounter{col}
  }
  \stepcounter{row}
}

This well-known problem is stated as follows: fill in a grid using digits ranging from 1 to 9 such that:
\begin{itemize}
\item all digits occur on each row
\item all digits occur on each column
  \item all digits occur in each $3 \times 3$ block (starting at a position multiple of 3) 
\end{itemize}
An illustration is given by Figure \ref{fig:sudoku}.

\begin{figure}[h]
  \centering
\begin{tikzpicture}[scale=.5]
  \begin{scope}
    \draw (0, 0) grid (9, 9);
    \draw[very thick, scale=3] (0, 0) grid (3, 3);
    \setcounter{row}{1}
    \setrow { }{2}{ }  {5}{ }{1}  { }{9}{ }
    \setrow {8}{ }{ }  {2}{ }{3}  { }{ }{6}
    \setrow { }{3}{ }  { }{6}{ }  { }{7}{ }
    \setrow { }{ }{1}  { }{ }{ }  {6}{ }{ }
    \setrow {5}{4}{ }  { }{ }{ }  { }{1}{9}
    \setrow { }{ }{2}  { }{ }{ }  {7}{ }{ }   
    \setrow { }{9}{ }  { }{3}{ }  { }{8}{ }
    \setrow {2}{ }{ }  {8}{ }{4}  { }{ }{7}
    \setrow { }{1}{ }  {9}{ }{7}  { }{6}{ }
    \node[anchor=center] at (4.5, -0.5) {Puzzle};
  \end{scope}

  \begin{scope}[xshift=12cm]
    \draw (0, 0) grid (9, 9);
    \draw[very thick, scale=3] (0, 0) grid (3, 3);
    \setcounter{row}{1}
    \setrow { }{2}{ }  {5}{ }{1}  { }{9}{ }
    \setrow {8}{ }{ }  {2}{ }{3}  { }{ }{6}
    \setrow { }{3}{ }  { }{6}{ }  { }{7}{ }
    \setrow { }{ }{1}  { }{ }{ }  {6}{ }{ }
    \setrow {5}{4}{ }  { }{ }{ }  { }{1}{9}
    \setrow { }{ }{2}  { }{ }{ }  {7}{ }{ }
    \setrow { }{9}{ }  { }{3}{ }  { }{8}{ }
    \setrow {2}{ }{ }  {8}{ }{4}  { }{ }{7}
    \setrow { }{1}{ }  {9}{ }{7}  { }{6}{ }
    \node[anchor=center] at (4.5, -0.5) {Solution};
    \begin{scope}[blue, font=\sffamily\slshape]
      \setcounter{row}{1}
      \setrow {4}{ }{6}  { }{7}{ }  {3}{ }{8}
      \setrow { }{5}{7}  { }{9}{ }  {1}{4}{ }
      \setrow {1}{ }{9}  {4}{ }{8}  {2}{ }{5}
      \setrow {9}{7}{ }  {3}{8}{5}  { }{2}{4}
      \setrow { }{ }{3}  {7}{2}{6}  {8}{ }{ }
      \setrow {6}{8}{ }  {1}{4}{9}  { }{5}{3}
      \setrow {7}{ }{4}  {6}{ }{2}  {5}{ }{1}
      \setrow { }{6}{5}  { }{1}{ }  {9}{3}{ }
      \setrow {3}{ }{8}  { }{5}{ }  {4}{ }{2}
    \end{scope}
  \end{scope}
\end{tikzpicture}
\caption{Solving a Sudoku Grid\label{fig:sudoku}  \tiny{(example from \href{https://www.texample.net/tikz/examples/sudoku/}{/www.texample.net/tikz})}}
\end{figure}

Because there are several clues, and because their number cannot be anticipated, we need a parameter \texttt{clues} that represents a two-dimensional array of integer values.
When $\texttt{clues}[i][j]$ is 0, it means that the cell is empty, whereas when it contains a digit between 1 and 9, it means that it represents a fixed value (clue).
A \p3 model is given by the following file `Sudoku.py':

\begin{boxpy}\begin{python}
@\imp@

clues = data  # if not 0, clues[i][j] is a value imposed at row i and col j

# x[i][j] is the value at row i and col j
x = VarArray(size=[9, 9], dom=range(1, 10))

satisfy(
  # imposing distinct values on each row and each column
  AllDifferent(x, matrix=True),

  # imposing distinct values on each block  tag(blocks)
  [AllDifferent(x[i:i + 3, j:j + 3]) for i in [0, 3, 6] for j in [0, 3, 6]],
  
  # imposing clues  tag(clues)
  [
    x[i][j] == clues[i][j]
      for i in range(9) for j in range(9) if clues and clues[i][j] > 0
  ]
)
\end{python}\end{boxpy}

First, note how the named parameter \nn{matrix} is used to ensure that all digits are different on each row and each column of the two-dimensional array $x$; this is the matrix version of \gb{allDifferent}.
Second, note how the notation $x[i:i + 3, j:j + 3]$ extracts a list of variables corresponding to a block of size $3 \times 3$ in $x$.
This is similar to notations used in package NumPy. 
Finally, each clue is naturally imposed under the form of a unary \gb{intension} constraint.

Suppose now that we have a file `grid.json' containing:

\begin{json}
{
  "clues": [
    [0, 4, 0, 0, 0, 0, 0, 0, 0],
    [5, 3, 9, 0, 0, 1, 0, 6, 0],
    [0, 0, 1, 0, 0, 2, 0, 5, 0],
    [4, 0, 7, 2, 0, 9, 0, 0, 6],
    [0, 0, 6, 0, 0, 0, 5, 0, 0],
    [8, 0, 0, 6, 0, 3, 1, 0, 7],
    [0, 8, 0, 7, 0, 0, 2, 0, 0],
    [0, 6, 0, 3, 0, 0, 4, 1, 8],
    [0, 0, 0, 0, 0, 0, 0, 7, 0]
  ]
}
\end{json}

then, we can execute:

\begin{command}
python Sudoku.py -data=grid.json
\end{command}

and we obtain the following \x3 instance (simplified here as not all clues are shown):
\begin{xcsp}
<instance format="XCSP3" type="CSP">
  <variables>
    <array id="x" note="x[i][j] is the value at row i and col j" size="[9][9]">
      1..9
    </array>
  </variables>
  <constraints>
    <allDifferent note="imposing distinct values on each row and each column">
      <matrix> x[][] </matrix>
    </allDifferent>
    <group note="imposing distinct values on each block" class="blocks">
      <allDifferent> %... </allDifferent>
      <args> x[0..2][0..2] </args>
      <args> x[0..2][3..5] </args>
      <args> x[0..2][6..8] </args>
      <args> x[3..5][0..2] </args>
      <args> x[3..5][3..5] </args>
      <args> x[3..5][6..8] </args>
      <args> x[6..8][0..2] </args>
      <args> x[6..8][3..5] </args>
      <args> x[6..8][6..8] </args>
    </group>
    <instantiation note="imposing clues" class="clues">
      <list> x[0][1] x[8][7] </list>   // only two of them inserted here for conciseness
      <values> 4 7 </values>
    </instantiation>
  </constraints>
</instance>
\end{xcsp}

Once again, we have used tags.
This way, it will be easy at parsing time to discard blocks or clues, if wished. 
Suppose now that we want to generate an instance without any clue.
Of course, we can build a grid only containing the value 0, but this is a little bit tedious. 
Actually, you just need to use a JSON file like this:

\begin{json}
{
  "clues": null
}
\end{json}

An alternative is simply to execute:

\begin{command}
python Sudoku.py -data=None
\end{command}

or

\begin{command}
python Sudoku.py -data=null
\end{command}

or even
\begin{command}
python Sudoku.py 
\end{command}

For these three last commands, the value \nn{None} is set to the predefined \p3 variable \nn{data}.

\subsection{Warehouse Location} \label{sec:warehouse} \index{Problems!Warehouse Location}

\begin{figure}[h!]
\begin{center}
  \includegraphics[scale=0.42]{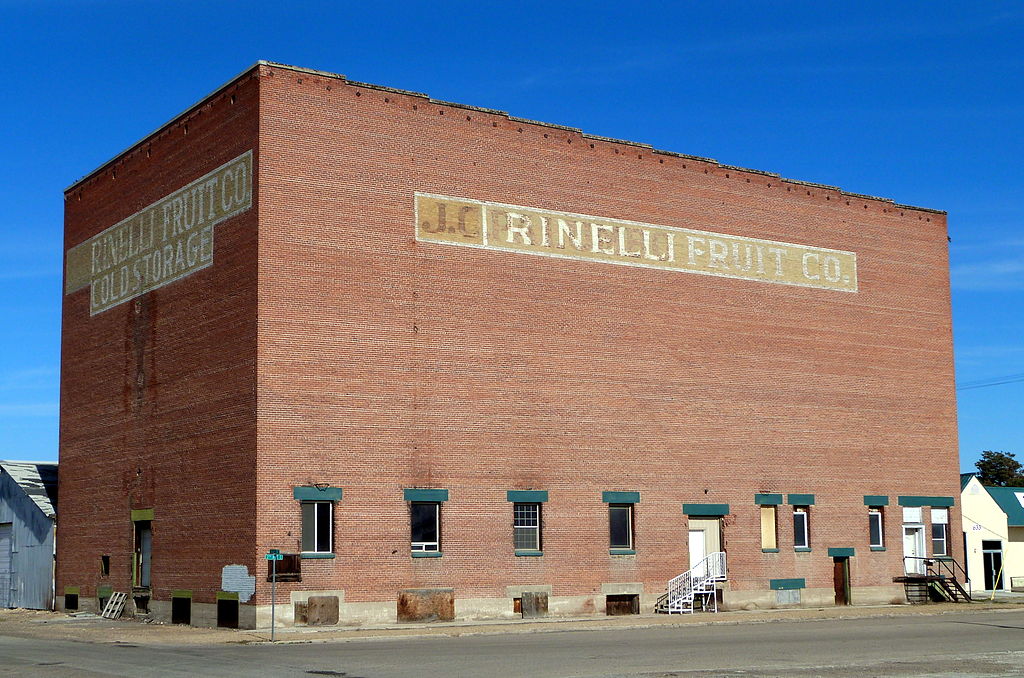}
\end{center}
\caption{Palumbo Fruit Company Warehouse. \tiny{(image from \href{https://commons.wikimedia.org/wiki/File:Palumbo_Fruit_Company_Warehouse_-_Payette_Idaho.jpg}{/commons.wikimedia.org})} \label{fig:warehouse}}
\end{figure}


In the Warehouse Location Problem (WLP), a company considers opening warehouses at some candidate locations in order to supply its existing stores.
Each possible warehouse has the same maintenance cost, and a capacity designating the maximum number of stores that it can supply.
Each store must be supplied by exactly one open warehouse.
The supply cost to a store depends on the warehouse.
The objective is to determine which warehouses to open, and which of these warehouses should supply the various stores, such that the sum of the maintenance and supply costs is minimized.
See \href{https://www.csplib.org/Problems/prob034/}{CSPLib--Problem 034} for more information.
An example of data is the file `warehouse.json' containing:

\begin{json}
{
  "fixedCost": 30,
  "warehouseCapacities": [1, 4, 2, 1, 3],
  "storeSupplyCosts": [
    [100, 24, 11, 25, 30], [28, 27, 82, 83, 74],
    [74, 97, 71, 96, 70], [2, 55, 73, 69, 61],
    [46, 96, 59, 83, 4], [42, 22, 29, 67, 59],
    [1, 5, 73, 59, 56], [10, 73, 13, 43, 96],
    [93, 35, 63, 85, 46], [47, 65, 55, 71, 95]
  ]
}
\end{json}

A \p3 model of this problem is given by the following file `Warehouse.py':

\begin{boxpy}\begin{python}
@\imp@

wcost, capacities, costs = data  # wcost is the fixed cost when opening a warehouse
nWarehouses, nStores = len(capacities), len(costs)

# w[i] is the warehouse supplying the ith store
w = VarArray(size=nStores, dom=range(nWarehouses))

# c[i] is the cost of supplying the ith store
c = VarArray(size=nStores, dom=lambda i: costs[i])

# o[j] is 1 if the jth warehouse is open
o = VarArray(size=nWarehouses, dom={0, 1})

satisfy(
  # capacities of warehouses must not be exceeded
  [Count(w, value=j) <= capacities[j] for j in range(nWarehouses)],

  # the warehouse supplier of the ith store must be open
  [o[w[i]] == 1 for i in range(nStores)],
  
  # computing the cost of supplying the ith store
  [costs[i][w[i]] == c[i] for i in range(nStores)]
)

minimize(
  # minimizing the overall cost
  Sum(c) + Sum(o) * wcost
)
\end{python}\end{boxpy}

Concerning data, the root object in the JSON file is expected to have three fields.
We then expect to get a named tuple of size 3 that can be unpacked. 
An alternative is to write something like:

\begin{python}
wcost = data.fixedCost  # for each open warehouse
capacities = data.warehouseCapacities
costs = data.storeSupplyCosts
nWarehouses, nStores = len(capacities), len(costs)
\end{python}

In our model, we associate a specific domain with each variable of the array $c$ by means of a lambda function.
Note that it is possible to give a list, \nn{costs[i]}, instead of a set, \nn{set(costs[i])}, as the list will be automatically converted to a set.
For dealing with warehouse capacities, we use the \gb{count} constraint by calling the \p3 function \nn{Count()}: the number of variables in a given list (here, $w$) that take the value specified by the named parameter \nn{value} must be less than a constant.
For linking stores with warehouses, we use the \gb{element} constraint: the variable in the array $o$ at index $w[i]$ must be 1 because this variable denotes the warehouse supplying the ith store, and it must be open.
Note that the index is not a constant but a variable of our model.
Similarly, we use the \gb{element} constraint for computing the actual costs; this time the array contains values (and not variables) and the target to reach is given by a variable. 
Finally, the objective function corresponds to minimizing two partial sums.
After executing:
\begin{command}
python Warehouse.py -data=warehouse.json
\end{command}

we obtain the following \x3 instance (some parts are omitted; see the presence of ellipsis):

\begin{xcsp}
<instance format="XCSP3" type="COP">
  <variables>
    <array id="w" note="w[i] is the warehouse supplying the ith store" size="[10]">
      0..4
    </array>
    <array id="c" note="c[i] is the cost of supplying the ith store" size="[10]">
      <domain for="c[0]"> 11 24 25 30 100 </domain>
      <domain for="c[1]"> 27 28 74 82 83 </domain>
      ...  // ellipsis
    </array>
    <array id="o" note="o[j] is 1 if the jth warehouse is open" size="[5]">
      0 1
    </array>
  </variables>
  <constraints>
    <block note="capacities of warehouses must not be exceeded">
      <count>
        <list> w[] </list>
        <values> 0 </values>
        <condition> (le,1) </condition>
      </count>
      ... // ellipsis
    </block>
    <group note="the warehouse supplier of the ith store must be open">
      <element>
        <list> o[] </list>
        <index> %0 </index>
        <value> 1 </value>
      </element>
      <args> w[0] </args>
      <args> w[1] </args>
      ...  // ellipsis
    </group>
    <block note="computing the cost of supplying the ith store">
      <element>
        <list> 100 24 11 25 30 </list>
        <index> w[0] </index>
        <value> c[0] </value>
      </element>
      ... // ellipsis
    </block>
  </constraints>
  <objectives>
    <minimize note="minimizing the overall cost" type="sum">
      <list> c[] o[] </list>
      <coeffs> 1 1 1 1 1 1 1 1 1 1 30 30 30 30 30 </coeffs>
    </minimize>
  </objectives>
</instance>
\end{xcsp}

In the model above, we have introduced three arrays of variables, allowing us to write a rather simple objective.
However, a more compact model is possible because one can write more complex forms of objectives.
This gives:

\begin{boxpy}\begin{python}
@\imp@

wcost, capacities, costs = data  # wcost is the fixed cost when opening a warehouse
nWarehouses, nStores = len(capacities), len(costs)

# w[i] is the warehouse supplying the ith store
w = VarArray(size=nStores, dom=range(nWarehouses))

satisfy(
  # capacities of warehouses must not be exceeded
  Count(w, value=j) <= capacities[j] for j in range(nWarehouses)
)

minimize(
  # minimizing the overall cost
  Sum(costs[i][w[i]] for i in range(nStores)) + NValues(w) * wcost
)
\end{python}\end{boxpy}

When compiling, in order to remain in the perimeter of \x3-core (see Chapter \ref{ch:logical}), some auxiliary variables may be introduced.
Here, this is the case for WLP, and the reader is invited to observe that the result of the compilation (i.e., \x3 files) for both model variants (depicted above) is rather similar. 

\subsection{Black Hole (Solitaire)} \label{sec:blackhole} \index{Problems!Blackhole}

\begin{figure}[h!]
\begin{center}
  \includegraphics[scale=1.3]{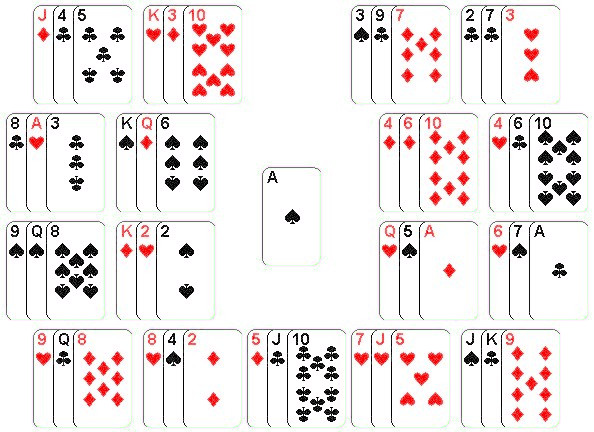}  
\end{center}
\caption{A Deal of Black Hole Solitaire. \label{fig:solitaire}} 
\end{figure}

From WikiPedia: ``Black Hole is a solitaire card game. Invented by David Parlett, this game's objective is to compress the entire deck into one foundation.
The cards are dealt to a board in piles of three.
The leftover card, dealt first or last, is placed as a single foundation called the Black Hole.
This card usually is the Ace of Spades.
Only the top cards of each pile in the tableau are available for play and in order for a card to be placed in the Black Hole, it must be a rank higher or lower than the top card on the Black Hole. This is the only allowable move in the entire game.
The game ends if there are no more top cards that can be moved to the Black Hole. The game is won if all of the cards end up in the Black Hole.''
An illustration is given by Figure \ref{fig:solitaire}.

We may want to play with various sizes of piles and various number of cards per suit.
An example of data is given by the file `blackhole-4.json' containing:

\begin{json}
{
  "nCardsPerSuit": 4,
  "piles": [[1 ,4 ,13] ,[15 ,9 ,6] ,[14 ,2 ,12] ,[7 ,8 ,5] ,[11 ,10 ,3]]
}
\end{json}

A \p3 model of this problem is given by the following file `Blackhole.py':

\begin{boxpy}\begin{python}
@\imp@

m, piles = data
nCards = 4 * m

# x[i] is the value j of the card at position i of the stack
x = VarArray(size=nCards, dom=range(nCards))

# y[j] is the position i of the card whose value is j
y = VarArray(size=nCards, dom=range(nCards))

T = {(i, j) for i in range(nCards) for j in range(nCards)
              if i % m == (j + 1) % m or j % m == (i + 1) % m}

satisfy(
  # linking variables of x and y
  Channel(x, y),

  # the Ace of Spades is initially put on the stack
  y[0] == 0,
  
  # cards must be played in the order of the piles
  [Increasing([y[j] for j in pile], strict=True) for pile in piles],
  
  # each new card put on the stack must be at a higher or lower rank 
  [(x[i], x[i + 1]) in T for i in range(nCards - 1)]
)
\end{python}\end{boxpy}

Note how the \gb{channel} constraint is used to make a channeling between the two arrays $x$ and $y$ (we have $x[i] = j \Leftrightarrow y[j]=i$),
how the value of the first variable of $y$ is imposed by a unary \gb{intension} constraint, how we guarantee to take cards from each pile in a strict increasing order with \gb{increasing} constraints
and how \gb{extension} constraints are posted after having precomputed a table $T$. 

Because the same table constraint is imposed on successive pairs of variables, we can use the meta-constraint \gb{slide}, introduced in Section \ref{sec:slide}. 
It suffices to replace the last argument of \nn{satisfy}() with:

\begin{python}
   Slide((x[i], x[i + 1]) in T for i in range(nCards - 1))
\end{python}

With this meta-constraint \gb{slide}, after executing:

\begin{command}
python Blackhole.py -data=blackhole.json
\end{command}

we obtain the following \x3 instance:

\begin{xcsp}
<instance format="XCSP3" type="CSP">
  <variables>
    <array id="x" note="x[i] is the value j of the card at position i of the stack" size="[16]">
      0..15
    </array>
    <array id="y" note="y[j] is the position i of the card whose value is j" size="[16]">
      0..15
    </array>
  </variables>
  <constraints>
    <channel note="linking variables of x and y">
      <list> x[] </list>
      <list> y[] </list>
    </channel>
    <intension note="the Ace of Spades is initially put on the stack">
      eq(y[0],0)
    </intension>
    <group note="cards must be played in the order of the piles">
      <ordered>
        <list> %0 %1 %2 </list>
        <operator> lt </operator>
      </ordered>
      <args> y[1] y[4] y[13] </args>
      <args> y[15] y[9] y[6] </args>
      <args> y[14] y[2] y[12] </args>
      <args> y[7..8] y[5] </args>
      <args> y[11] y[10] y[3] </args>
    </group>
    <slide note="each new card put on the stack must be at a higher or lower rank">
      <list> x[] </list>
      <extension>
        <list> %0 %1 </list>
        <supports> (0,1)(0,3)(0,5)(0,7)(0,9)(0,11)(0,13)(0,15)(1,0)(1,2)(1,4)(1,6)(1,8)(1,10)(1,12)(1,14)(2,1)(2,3)(2,5)(2,7)(2,9)(2,11)(2,13)(2,15)(3,0)(3,2)(3,4)(3,6)(3,8)(3,10)(3,12)(3,14)(4,1)(4,3)(4,5)(4,7)(4,9)(4,11)(4,13)(4,15)(5,0)(5,2)(5,4)(5,6)(5,8)(5,10)(5,12)(5,14)(6,1)(6,3)(6,5)(6,7)(6,9)(6,11)(6,13)(6,15)(7,0)(7,2)(7,4)(7,6)(7,8)(7,10)(7,12)(7,14)(8,1)(8,3)(8,5)(8,7)(8,9)(8,11)(8,13)(8,15)(9,0)(9,2)(9,4)(9,6)(9,8)(9,10)(9,12)(9,14)(10,1)(10,3)(10,5)(10,7)(10,9)(10,11)(10,13)(10,15)(11,0)(11,2)(11,4)(11,6)(11,8)(11,10)(11,12)(11,14)(12,1)(12,3)(12,5)(12,7)(12,9)(12,11)(12,13)(12,15)(13,0)(13,2)(13,4)(13,6)(13,8)(13,10)(13,12)(13,14)(14,1)(14,3)(14,5)(14,7)(14,9)(14,11)(14,13)(14,15)(15,0)(15,2)(15,4)(15,6)(15,8)(15,10)(15,12)(15,14) </supports>
      </extension>
    </slide>
  </constraints>
  </instance>
\end{xcsp}

Here, the main interest of using \gb{slide} is that the generated \x3 file is made more compact (while emphasizing the sliding structure).
However, in our illustration, because the sliding form is not circular and because two successive constraints only share one variable,
any solver reasoning individually with the sliding constraints will reach the same efficiency (i.e., will reach the same level of filtering of the search space) as reasoning with the meta-constraint.

If you are worried about using the \p3 function \nn{Slide}() in the model, you can leave the model as it was given initially, and in case you are however interested in the more compact sliding form, you can use the option \nn{-recognizeSlides} as in the following command:

\begin{command}
python Blackhole.py -data=blackhole-4.json -recognizeSlides 
\end{command}



\subsection{Rack Configuration}\label{sec:rack} \index{Problems!Rack Configuration}

\begin{figure}[h]
\begin{center}
  \includegraphics[scale=0.35]{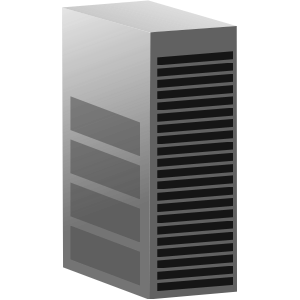}
\end{center}
\caption{A Rack. \tiny{(image from \href{https://freesvg.org/vector-image-of-racks}{freesvg.org})} \label{fig:rack}}
\end{figure}

The rack configuration problem consists of plugging a set of electronic cards into racks with electronic connectors.
Each card plugged into a rack uses a connector.
In order to plug a card into a rack, the rack must be of a rack model.
Each card is characterized by the power it requires.
Each rack model is characterized by the maximal power it can supply, its size (number of connectors), and its price.
The problem is to decide how many of the available racks are actually needed such that: 
\begin{itemize}
\item every card is plugged into one rack
\item the total power demand and the number of connectors required by the cards do not exceed that available for a rack
\item the total price is minimized.
\end{itemize}
See \href{https://www.csplib.org/Problems/prob031/}{CSPLib--Problem 031} for more information.

An example of data is given by the file `rack.json' containing:

\begin{json}
{
  "nRacks": 10,
  "models": [[150, 8, 150], [200, 16, 200]],
  "cardTypes": [[20, 20], [40, 8], [50, 4], [75, 2]]
}
\end{json}

A \p3 model for this problem is given by the following file `Rack.py':

\begin{boxpy}\begin{python}
@\imp@

nRacks, models, cardTypes = data
models.append([0, 0, 0])  # we add first a dummy model (0,0,0)
powers, sizes, costs = zip(*models)
cardPowers, cardDemands = zip(*cardTypes)
nModels, nTypes = len(models), len(cardTypes)

# m[i] is the model used for the ith rack
m = VarArray(size=nRacks, dom=range(nModels))

# p[i] is the power of the model used for the ith rack
p = VarArray(size=nRacks, dom=powers)

# s[i] is the size (number of connectors) of the model used for the ith rack
s = VarArray(size=nRacks, dom=sizes)

# c[i] is the cost (price) of the model used for the ith rack
c = VarArray(size=nRacks, dom=costs)

# nc[i][j] is the number of cards of type j put in the ith rack
nc = VarArray(size=[nRacks, nTypes],
              dom=lambda i, j: range(min(max(sizes), cardDemands[j]) + 1))

T = {(i, powers[i], sizes[i], costs[i]) for i in range(nModels)}
              
satisfy(
  # linking rack models with powers, sizes and costs
  [(m[i], p[i], s[i], c[i]) in T for i in range(nRacks)],

  # connector-capacity constraints
  [Sum(nc[i]) <= s[i] for i in range(nRacks)],
  
  # power-capacity constraints
  [nc[i] * cardPowers <= p[i] for i in range(nRacks)],
  
  # demand constraints
  [Sum(nc[:, j]) == cardDemands[j] for j in range(nTypes)],
  
  # tag(symmetry-breaking)
  [
    Decreasing(m),

    If(
      m[0] == m[1],
      Then=nc[0][0] >= nc[1][0]
    )
  ]
)

minimize(
  # minimizing the total cost being paid for all racks
  Sum(c)
)
\end{python}\end{boxpy}

From data, we build first some auxiliary lists that are useful for easily writing our model.
Note that using the Python function \nn{zip()} is simpler and compacter than writing for example:

\begin{python}
cardPowers, cardDemands = [row[0] for row in cardTypes], [row[1] for row in cardTypes]
\end{python}

After declaring five arrays of variables, a quaternary table constraint is first posted.
See how it is easy to link variables of 4 arrays with a simple table.
Then, three lists of \gb{sum} constraints are posted.
In the second list, we use a dot product, and in the third list, we use the notation $nc[:, j]$ to extract the jth column of the array $nc$, as in NumPy. 
For breaking symmetries, we use a complex expression based on the control structure \nn{If ... Then} that will be introduced later.
Note that the first letter is capitalized ('if' versus 'If') and the condition (test) involves a variable from the model (it is forbidden to use a classical test as with the classical 'if' of Python).

As usual, for generating an \x3 instance, we just need to execute: 

\begin{command}
python Rack.py -data=rack.json
\end{command}

One drawback with the previous model is that it is difficult to understand the role of each piece of data, when looking independently at the JSON file.
One remedy is then to choose a clearer structure as in this file `rack2.json':

\begin{json}
{
  "nRacks": 10,
  "rackModels": [
    {"power": 150, "nConnectors": 8, "price": 150},
    {"power": 200, "nConnectors": 16, "price": 200}
  ],
  "cardTypes": [
    {"power": 20, "demand": 20},
    {"power": 40, "demand": 8},
    {"power": 50, "demand": 4},
    {"power": 75, "demand": 2}
  ]
}
\end{json}

In \p3, it is quite easy to change the representation (structure) of data.
It suffices to update the way the predefined \p3 variable \nn{data} is used in the model.
In our case, with this new representation, we only need to replace:

\begin{python}
models.append([0, 0, 0])  # we add first a dummy model (0,0,0)
\end{python}

with:

\begin{python}
models.append(models[0].__class__(0, 0, 0))  # we add first a dummy model (0,0,0) 
\end{python}

Again we add a dummy rack model to those defined in the JSON file.
To do that, and in order to avoid breaking the homogeneity of the data, we get the class of the used named tuples to build and add a new one.
As any JSON object is automatically converted to a named tuple, we still have the possibility to use the function \nn{zip()} in our model.

\chapter{Data, Variables and Objectives}\label{ch:data}

In this chapter, we give some additional details and illustrations about data, variables and objectives, although many examples can already be found in the other chapters. 

\section{Specifying Data}\label{sec:specdata}

In this section, we describe the following options:
\begin{itemize}
\item \verb!-data!
\item  \verb!-parser! (can also be written \verb!-dataparser!)
\item \verb!-export! (can also be written \verb!-dataexport!)
\item \verb!-format! (can also be written \verb!-dataformat!)
\item \verb!-output!
\end{itemize}

\bigskip
\noindent Except for ``single'' problems, each problem usually represents a large (often, infinite) family of cases, called instances, that one may want to solve.
All these instances are uniquely identified by some specific data.

First, recall that the command to be run for generating an \x3 instance (file), given a model and some data is:

\begin{command}
python <model_file> -data=<data_values>
\end{command}

where \verb!<model_file>! (is a Python file that) represents a \p3 model, and  \verb!<data_values>! represents some specific data.
In our context, an {\em elementary} value is a value of one of these built-in data types: integer (int), real (float), string (str) and boolean (bool).
Specific data can be given as:
\begin{enumerate}
\item a single elementary value, as in \verb!-data=5!
\item a list of elementary values, between square (or round) brackets\footnote{According to the operating system, one might need to escape brackets.} and with comma used as a separator, 
as in \verb!-data=[9,0,0,3,9]!
\item a list of named elementary values, between square (or round) brackets and with comma used as a separator, as in \verb!-data=[v=9,b=0,r=0,k=3,l=9]!
\item a JSON file (possibly, given by a URL), as in \verb!-data=Bibd-9-3-9.json!
\item a text file (i.e., a non-JSON file in any arbitrary format) while providing with the option \nn{-parser} some Python code to load it, as in \verb!-data=puzzle.txt -parser=ParserPuzzle.py!  
\end{enumerate}


Then, {\bf data can be directly used in \p3 models by means of a predefined variable called \nn{data}}.
The value of the predefined \p3 variable \nn{data} is set as follows:
\begin{enumerate}
\item if the option \verb!-data! is not specified, or if it is specified as \verb!-data=null! or  \verb!-data=None!, then the value of \nn{data} is \nn{None}. See, for example, Section \ref{sec:sudoku}.
\item if a single elementary value is given (possibly, between brackets), then the value of \nn{data} is directly this value. See, for example, Section \ref{sec:golomb}.
\item if a JSON file containing a root object with only one field is given, then the value of \nn{data} is directly this value. See, for example, Section \ref{sec:sudoku}.
\item if a list of (at least two) elementary values is given, then the value of \nn{data} is a tuple containing those values in sequence. See, for example, Section \ref{sec:boardColoration}.
\item if a list of (at least two) named elementary values is given, then the value of \nn{data} is a named tuple. See, for example, Section \ref{sec:boardColoration}.
\item if a JSON file containing a root object with at least two fields is given, then the value of \nn{data} is a named tuple. Actually, any encountered JSON object in the file is (recursively) converted into a named tuple. See, for example, Section \ref{sec:warehouse} and Section \ref{sec:rack}. 
\end{enumerate}

\bigskip
Although various cases have already been illustrated in Chapter \ref{sec:illustrative}, we introduce below a few additional examples.

\paragraph{All-Interval Series.} \index{Problems!All-Interval Series} Given the twelve standard pitch-classes (c, c\#, d, $\dots$), represented by numbers $0,1,\dots,11$, find a series in which each pitch-class occurs exactly once and in which the musical intervals between neighboring notes cover the full set of intervals from the minor second (1 semitone) to the major seventh (11 semitones). 
That is, for each of the intervals, there is a pair of neighboring pitch-classes in the series, between which this interval appears. 

\begin{figure}[h]
\begin{center}
  \includegraphics[scale=0.18]{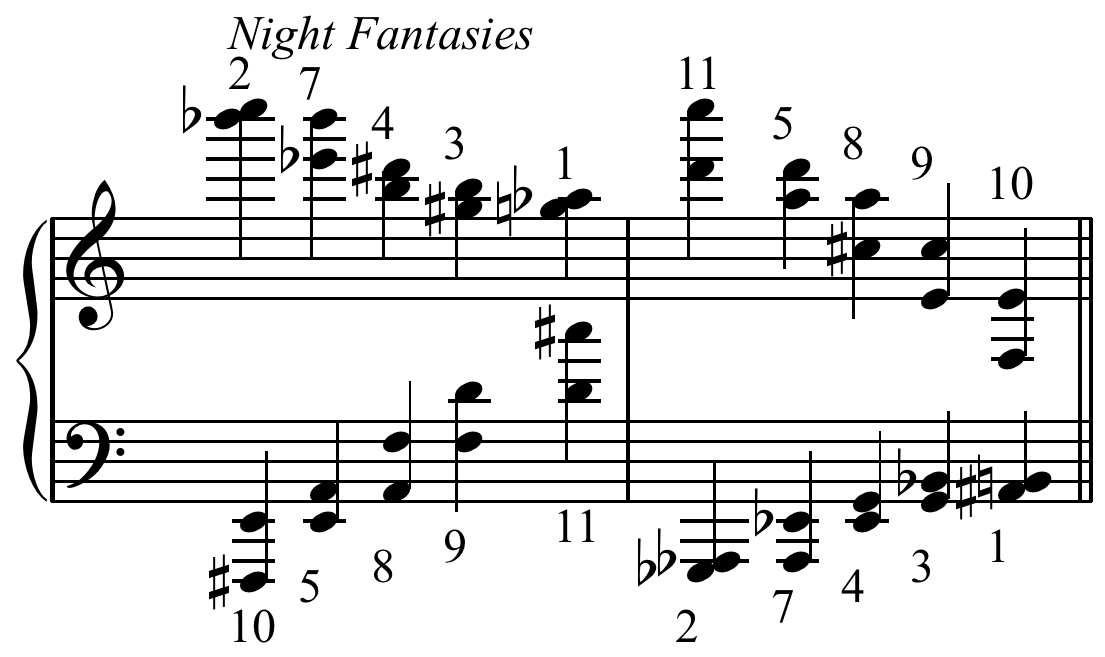} 
\end{center}
\caption{Elliott Carter often bases his all-interval sets on the list generated by Bauer-Mendelberg and Ferentz and uses them as a "tonic" sonority \tiny{(image from \href{https://commons.wikimedia.org/wiki/File:Carter_all-interval_sets.png}{commons.wikimedia.org})}}
\end{figure}

The problem of finding such a series can be easily formulated as an instance of a more general arithmetic problem.
Given a positive integer $n$, find a sequence $x = \langle x_0, x_1, \dots, x_{n-1} \rangle$, such that:
\begin{enumerate}
\item $x$ is a permutation of $\{0,1,...,n-1\}$; 
\item the interval sequence $y = \langle |x_1-x_0|, |x_2-x_1|, ... |x_{n-1}-x_{n-2}| \rangle$ is a permutation of $\{1,2,...,n-1\}$. 
\end{enumerate}
A sequence satisfying these conditions is called an all-interval series of order $n$; the problem of finding such a series is the all-interval series problem of order $n$. 
For example, for $n=8$, a solution is: 
\begin{quote}
\begin{verbatim}
1 7 0 5 4 2 6 3
\end{verbatim}
\end{quote}

A \p3 model of this problem is given by the following file `AllInterval.py':

\begin{boxpy}\begin{python}
@\imp@

n = data

# x[i] is the ith note of the series
x = VarArray(size=n, dom=range(n))

satisfy(
  # notes must occur once, and so form a permutation
  AllDifferent(x),

  # intervals between neighbouring notes must form a permutation
  AllDifferent(abs(x[i] - x[i + 1]) for i in range(n - 1)),
)
\end{python}\end{boxpy}

Here, the required data is a single integer value.
So, to generate the \x3 instance of \nn{AllInterval} for order $12$, we just execute:

\begin{command}
python AllInterval.py -data=12
\end{command}

\paragraph{Balanced Incomplete Block Designs.}\label{par:bibd} \index{Problems!Balanced Incomplete Block Designs} From \href{https://www.csplib.org/Problems/prob028}{CSPLib}:
``Balanced Incomplete Block Design (BIBD) generation is a standard combinatorial problem from design theory, originally used in the design of statistical experiments but has since found other applications such as cryptography.
It is a special case of Block Design, which also includes Latin Square problems.
BIBD generation is described in most standard textbooks on combinatorics.
A BIBD is defined as an arrangement of $v$ distinct objects into $b$ blocks such that each block contains exactly $k$ distinct objects, each object occurs in exactly $r$ different blocks, and every two distinct objects occur together in exactly $\lambda$ blocks.
Another way of defining a BIBD is in terms of its incidence matrix, which is a $v$ by $b$ binary matrix with exactly $r$ ones per row, $k$ ones per column, and with a scalar product of $\lambda$ between any pair of distinct rows.
A BIBD is therefore specified by its parameters $(v,b,r,k,\lambda)$.''

\begin{figure}[h]
\begin{center}
  \includegraphics[scale=0.6]{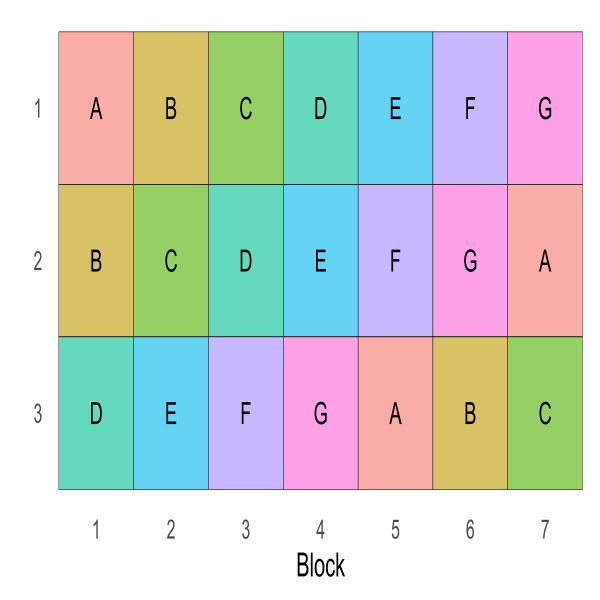}
\end{center}
\caption{7 treatments in 7 blocks of size 3 \tiny{(image from \href{https://www.maths.usyd.edu.au/u/UG/SM/STAT3022/r/current/Lecture/lecture29\_2020JC.html}{Incomplete Block Design Lesson at Univ. Sydney})} \label{fig:bibd}}
\end{figure}

An example of a solution for $(7,7,3,3,1)$ is:
\begin{center}
0 1 1 0 0 1 0 \\
1 0 1 0 1 0 0 \\
0 0 1 1 0 0 1 \\
1 1 0 0 0 0 1 \\
0 0 0 0 1 1 1 \\
1 0 0 1 0 1 0 \\
0 1 0 1 1 0 0 \\
\end{center}

Hence, we need five integers $v$, $b$, $r$, $k$, $l$ (for $\lambda$) for specifying a unique instance; possibly, $b$ and $r$ can be set to 0, so that these values are automatically computed according to a classical BIBD template.
A \p3 model of this problem is given by the following file `BIBD.py': 

\begin{boxpy}\begin{python}
@\imp@

v, b, r, k, l = data 
b = (l * v * (v - 1)) // (k * (k - 1)) if b == 0 else b  
r = (l * (v - 1)) // (k - 1) if r == 0 else r  

# x[i][j] is the value of the matrix at row i and column j
x = VarArray(size=[v, b], dom={0, 1})

satisfy(
  # constraints on rows
  [Sum(row) == r for row in x],

  # constraints on columns
  [Sum(col) == k for col in columns(x)],
  
  # scalar constraints with respect to lambda
  [row1 * row2 == l for row1, row2 in combinations(x, 2)]
)
\end{python}\end{boxpy}

To generate an \x3 instance (file), we can for example execute:

\begin{command}
python BIBD.py -data=[9,0,0,3,9]
\end{command}

As mentioned earlier, with some command interpreters (shells), you may have to escape the characters '[' and ']', which gives:

\begin{command}
python BIBD.py -data=\[9,0,0,3,9\]
\end{command}

You can also use round brackets instead of square brackets: 

\begin{command}
python BIBD.py -data=(9,0,0,3,9)
\end{command}

If it causes some problem with the command interpreter (shell), you have to escape the characters '(' and ')', which gives:

\begin{command}
python BIBD.py -data=\(9,0,0,3,9\)
\end{command}

Unless specified otherwise with the option \verb!-output!, the filename of the generated \x3 instance is `BIBD-9-0-0-3-9.xml'.
This means that if we execute:
\begin{command}
python BIBD.py -data=[9,0,0,3,9] -output=My-Bibd
\end{command}
the generated filename is 'My-Bibd.xml' (if not present as a suffix, `.xml' is automatically added).
It is also possible to indicate the path to the output file. If we execute:
\begin{command}
python BIBD.py -data=[9,0,0,3,9] -output=test/My-Bibd
\end{command}
the file 'My-Bibd.xml' is generated in the directory 'test'.
If we just indicate the name of directory:
\begin{command}
python BIBD.py -data=[9,0,0,3,9] -output=test
\end{command}
the file `BIBD-9-0-0-3-9.xml' is generated in the directory 'test'.


\medskip
Suppose that you would prefer to have a JSON file for storing these data values.
You can execute:

\begin{command}
python BIBD.py -data=[9,0,0,3,9] -datexport
\end{command}

You then obtain the following JSON file `BIBD-9-0-0-3-9.json'

\begin{json}
{
  "v":9,
  "b":0,
  "r":0,
  "k":3,
  "l":9
}
\end{json}

And now, to generate the same \x3 instance (file) as above, you can execute:

\begin{command}
python BIBD.py -data=BIBD-9-0-0-3-9.json
\end{command}

\begin{remark}
  At the Windows command line, different escape characters may be needed (for example, depending whether you use Windows Powershell or not).
However, note that you can always run a command from a batch script file (or use a JSON file). 
\end{remark}

\paragraph{Filenames with Formatted Data.}
As shown above, when data are given under the form of elementary values on the command line, they are integrated in the filename of the generated instance.
However, sometimes, it may be interesting to format a little bit such filenames.
This is possible by using the format \verb!-format! (or \verb!-dataformat!).
The principle is that the string passed to this option will serve to apply formatting to the values in  \verb!-data!.
For example,

\begin{command}
python BIBD.py -data=[9,0,0,3,9] -format={:02d}-{:01d}-{:01d}-{:02d}-{:02d}
\end{command}
will generate an \x3 file with filename `BIBD-09-0-0-03-09.xml'.
If the same pattern must be applied to all pieces of data, we can write:

\begin{command}
python BIBD.py -data=[9,0,0,3,9] -format={:02d}
\end{command}
so as to obtain an \x3 file with filename `BIBD-09-00-00-03-09.xml'.
A more direct way of formatting some input is to insert leading zeros.
For example,

\begin{command}
python BIBD.py -data=[09,0,0,03,009] 
\end{command}
will generate an \x3 file with filename `BIBD-09-0-0-03-009.xml'.

\paragraph{Balanced Academic Curriculum Problem (BACP).} \index{Problems!Balanced Academic Curriculum (BACP)} From \href{https://www.csplib.org/Problems/prob030}{CSPLib}:
``The goal of BACP is to design a balanced academic curriculum by assigning periods to courses in a way that the academic load of each period is balanced, i.e., as similar as possible.
  An academic curriculum is defined by a set of courses and a set of prerequisite relationships among them.
  Courses must be assigned within a maximum number of academic periods.
  Each course is associated to a number of credits or units that represent the academic effort required to successfully follow it.

\begin{center}
\includegraphics[scale=1.8]{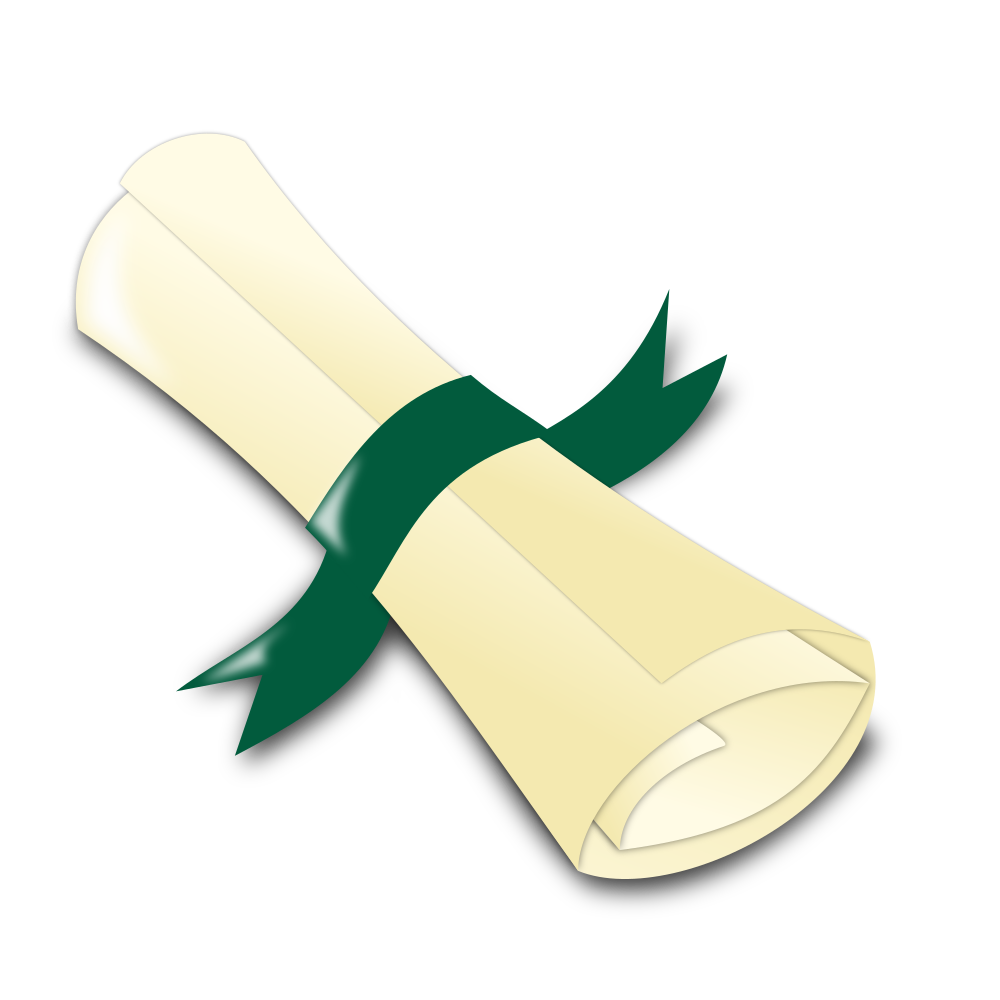}  
\end{center} 

  The curriculum must obey the following regulations:
  \begin{itemize}
    \item minimum academic load: a minimum number of academic credits per period is required to consider a student as full time
    \item maximum academic load: a maximum number of academic credits per period is allowed in order to avoid overload
    \item minimum number of courses: a minimum number of courses per period is required to consider a student as full time
    \item maximum number of courses: a maximum number of courses per period is allowed in order to avoid overload
  \end{itemize}
The goal is to assign a period to every course in a way that the minimum and maximum academic load for each period, the minimum and maximum number of courses for each period, and the prerequisite relationships are satisfied. An optimal balanced curriculum minimizes the maximum academic load for all periods.''

\bigskip
When analyzing this problem, we identify its parameters as being the number of periods (an integer), the minimum and the maximum number of credits (two integers), the minimum and the maximum number of courses (two integers),
the credits for each course (a one-dimensional array of integers) and the prerequisites (a two-dimensional array of integers, with each row indicating a prerequisite). 
An example of data is given by the following JSON file `example.json':

\begin{json}
{
  "nPeriods": 4,
  "minCredits": 2,
  "maxCredits": 5,
  "minCourses": 2,
  "maxCourses": 3,
  "credits": [2,3,1,3,2,3,3,2,1],
  "prerequisites": [[2,0],[4,1],[5,2],[6,4]]
}
\end{json}

\bigskip
A \p3 model of this problem is given by the following file `BACP.py':

\begin{boxpy}\begin{python}
@\imp@

nPeriods, minCredits, maxCredits, minCourses, maxCourses, credits, prereq = data
nCourses = len(credits)

# s[c] is the period (schedule) for course c
s = VarArray(size=nCourses, dom=range(nPeriods))

# co[p] is the number of courses at period p
co = VarArray(size=nPeriods, dom=range(minCourses, maxCourses + 1))

# cr[p] is the number of credits at period p
cr = VarArray(size=nPeriods, dom=range(minCredits, maxCredits + 1))

# cp[c][p] is 0 if the course c is not planned at period p,
#             the number of credits for c otherwise
cp = VarArray(size=[nCourses, nPeriods], dom=lambda c, p: {0, credits[c]})

def table(c):
   return {(0,) * p + (credits[c],) + (0,) * (nPeriods - p - 1) + (p,)
     for p in range(nPeriods)}

satisfy(
  # channeling between arrays cp and s
  [(*cp[c], s[c]) in table(c) for c in range(nCourses)],

  # counting the number of courses in each period
  [Count(s, value=p) == co[p] for p in range(nPeriods)],
  
  # counting the number of credits in each period
  [Sum(cp[:, p]) == cr[p] for p in range(nPeriods)],
  
  # handling prerequisites
  [s[c1] < s[c2] for (c1, c2) in prereq]
)

minimize(
  # minimizing the maximum number of credits in periods
  Maximum(cr)
)
\end{python}\end{boxpy}

The command to execute for compiling is then:

\begin{command}
python BACP.py -data=example.json
\end{command}

Because tuple unpacking is used, it is important to note that the fields of the root object in the JSON file must be given in this exact order.
If it is not the case, as for example:

\begin{json}
{
  "nPeriods": 4,
  "prequisites": [[2,0],[4,1],[5,2],[6,4]],
  "minCredits": 2,
  "maxCredits": 5,
  "credits": [2,3,1,3,2,3,3,2,1],
  "minCourses": 2,
  "maxCourses": 3
}
\end{json}

there will be a problem when unpacking data.
If you wish a safer model (because, for example, you have no guarantee about the way the data are generated), you must specifically refer to the fields of the named tuple instead: 

\begin{python}
@\imp@

nPeriods = data.nPeriods
minCredits, maxCredits = data.minCredits, data.maxCredits
minCourses, maxCourses = data.minCourses, data.maxCourses
credits, prereq = data.credits, data.prerequisites
nCourses = len(credits)
\end{python}

Now, let us suppose that you would like to use the data from this MiniZinc file `data.mzn':

\begin{lstlisting}[xleftmargin=20pt, basicstyle=\ttfamily\footnotesize, framesep=5pt]
include "curriculum.mzn.model";
n_courses = 9;
n_periods = 4;
load_per_period_lb = 2;
load_per_period_ub = 5;
courses_per_period_lb = 2;
courses_per_period_ub = 3;
course_load = [2, 3, 1, 3, 2, 3, 3, 2,1, ];
constraint prerequisite(2, 0);
constraint prerequisite(4, 1);
constraint prerequisite(5, 2);
constraint prerequisite(6, 4);
\end{lstlisting}

We need to write a piece of code in Python for building the variable \nn{data} that will be used in our model.
After importing everything (*) from \nn{pycsp3.problems.data.parsing}, we can use some \p3 functions such as \nn{next\_line}(), \nn{number\_in}(), \nn{remaining\_lines}(),$\dots$
Here, we also use the classical function \nn{split}() of module \nn{re} to parse information concerning prerequisites. 
Note that you have to add relevant fields to the predefined dictionary\footnote{At this stage, \nn{data} is a dictionary. Later, it will be automatically converted to a named tuple.}  \nn{data}, as in the following file `BACP\_ParserZ.py':

\begin{python}
from pycsp3.problems.@data@.parsing import *

nCourses = number_in(next_line())
data["nPeriods"] = number_in(next_line())
data["minCredits"] = number_in(next_line())
data["maxCredits"] = number_in(next_line())
data["minCourses"] = number_in(next_line())
data["maxCourses"] = number_in(next_line())
data["credits"] = numbers_in(next_line())
data["prerequisites"] = [[int(v) - 1
    for v in re.split(r'constraint prerequisite\(|,|\);', line) if len(v) > 0]
      for line in remaining_lines(skip_curr=True)]
\end{python}

To generate the \x3 instance (file), you have to execute:

\begin{command}
python BACP.py -data=data.mzn -parser=BACP_ParserZ.py
\end{command}

If you want the same data put in a JSON file, execute:

\begin{command}
python BACP.py -data=data.mzn -parser=BACP_ParserZ.py -dataexport
\end{command}

You obtain a file called `BACP-data.json' equivalent to the one introduced earlier.
If you want to specify the name of the output JSON file, give it as a value to the option \nn{-dataexport}, as e.g., in:

\begin{command}
python BACP.py -data=data.mzn -parser=BACP_ParserZ.py -dataexport=instance0
\end{command}

The generated JSON file is then called `instance0.json'. 

\paragraph{Special Rules when Loading JSON Files. \label{par:special}}
The rules that are used when loading a JSON file in order to set the value of the \p3 predefined variable \nn{data} are as follows.  
\begin{enumerate}
\item For any field $f$ of the root object in the JSON file, we obtain a field \nn{f} in the generated named tuple \nn{data} such that:
\begin{itemize}
\item if \nn{f} is a JSON list (or recursively, a list of lists) containing only integers, the type of \nn{data.f} is `pycsp3.tools.curser.ListInt' instead of `list'; `ListInt' being a subclass of `list'.
  The main interest is that \nn{data.f} can be directly used as a vector for the global constraint \gb{element}.
  See Mario Problem, page \pageref {pb:mario}, for an illustration.
\item if \nn{f} is an object, \nn{data.f} is a named tuple with the same fields as \nn{f}.
  See Rack Configuration Problem in Section \ref{sec:rack} for an illustration.
\end{itemize}
\item The rules above apply recursively.
\end{enumerate}

\paragraph{Special Rule when Building Arrays of Variables.}
When we define a list (array) $x$ of variables with \nn{VarArray}(), the type of $x$ is `pycsp3.tools.curser.ListVar' instead of `list'. The main interest is that $x$ can be directly used as a vector for the global constraint \gb{element}.

\paragraph{Special Values \nn{null} and \nn{None}.}

When the value \nn{null} occurs in a JSON file, it becomes \nn{None} in \p3 after loading the data file.
An illustration is given at the end of Section \ref{sec:sudoku}.

\paragraph{Loading Several JSON Files.}

It is possible to load data from several JSON files.
It suffices to indicate a list of JSON filenames between brackets.
For example, let `file1.json' be:
\begin{json}
{
  "a": 4,
  "b": 12
}
\end{json}

let `file2.json' be:
\begin{json}
{
  "c": 10,
  "d": 1
}
\end{json}

and let `Test.py' be:
\begin{python}
@\imp@

a, b, c, d = data

print(a, b, c, d)

...
\end{python}

then, by executing: 

\begin{command}
python Test.py -data=[file1.json,file2.json]
\end{command}

we obtain the expected values in the four Python variables, because the order of fields
is guaranteed (as if the two JSON files had been concatenated); behind the scenes, an OrderedDict is used, and the method `update()' is called.

\paragraph{Combining JSON Files and Named Elementary Values.}

It may be useful to load data from JSON files, while updating some (named) elementary values.
It means that we can indicate between brackets JSON filenames as well as named elementary values.
The rule is simple: any field of the variable \nn{data} is given as value the last statement concerning it when loading.

For example, the command:
\begin{command}
python Test.py -data=[file1.json,file2.json,c=5]
\end{command}
defines the variable \nn{data} from the two JSON files, except that the variable \nn{c} is set to 5.

\medskip
However, the command:
\begin{command}
python Test.py -data=[c=5,file1.json,file2.json]
\end{command}
is not appropriate because the value of \nn{c} will be overridden when considering `file2.json'.

\medskip
Just remember that named elementary values must be given after JSON files.

\paragraph{Loading Several Text Files.}

It is also possible to load data from several text (non-JSON) files.
It suffices to indicate a list of filenames between brackets, which then will be concatenated just before soliciting an appropriate parser.
For example, let `file1.txt' be:
\begin{lstlisting}[xleftmargin=20pt, basicstyle=\ttfamily\footnotesize, framesep=5pt]
5
2 4 12 3 8  
\end{lstlisting}

let `file2.txt' be:
\begin{lstlisting}[xleftmargin=20pt, basicstyle=\ttfamily\footnotesize, framesep=5pt]
3 3
0 1 1
1 0 1
0 0 1
\end{lstlisting}

then, at the time the file `Test2\_Parser.py' is executed after typing:

\begin{command}
python Test2.py -data=[file1.txt,file2.txt] -parser=Test2_Parser.py
\end{command}

we can read a sequence of text lines as if a single file was initially given with content:
\begin{lstlisting}[xleftmargin=20pt, basicstyle=\ttfamily\footnotesize, framesep=5pt]
5
2 4 12 3 8  
3 3
0 1 1
1 0 1
0 0 1
\end{lstlisting}


It is even possible to add arbitrary lines to the intermediate concatenated file.
For example, 
\begin{command}
python Test2.py -data=[file1.txt,file2.txt,10] -parser=Test2_Parser.py
\end{command}
adds a last line containing the value 10.
Because whitespace is not tolerated, one may need to surround additional lines with quotes (or double quotes).
For example, at time `Test2\_Parser.py' is executed after typing:
\begin{command}
python Test2.py -data=[file1.txt,file2.txt,10,"3 5",partial] -parser=Test2_Parser.py
\end{command}
the sequence of text lines is as follows:
\begin{lstlisting}[xleftmargin=20pt, basicstyle=\ttfamily\footnotesize, framesep=5pt]
5
2 4 12 3 8  
3 3
0 1 1
1 0 1
0 0 1
10
3 5
partial
\end{lstlisting}

\paragraph{Default Data.}
Except for single problems, data must be specified by the user in order to generate specific problem instances.
If data are not specified, an error is raised.
However, when writing the model, it is always possible to indicate some default data, notably by using the behaviour of the Python operator \nn{or}.
For setting a JSON file as being the default data file, we must call the function \nn{default\_data()}. 
Handling default data is illustrated with BIBD and BACP problems.

For BIBD, If we replace:
\begin{python}
v, b, r, k, l = data 
\end{python}
by
\begin{python}
v, b, r, k, l = data or (9,0,0,3,9)
\end{python}
then, we can generate the default instance with:
\begin{command}
python BIBD.py
\end{command}

For BACP, if we replace:

\begin{python}
nPeriods, minCredits, maxCredits, minCourses, maxCourses, credits, prereq = data
\end{python}
by
\begin{python}
nPeriods, minCredits, maxCredits, minCourses, maxCourses, credits, \
    prereq = data or default_data(example.json)
\end{python}
then, we can generate the default instance with:
\begin{command}
python BACP.py
\end{command}

\paragraph{Loading a JSON Data File.}

If for some reasons, it is convenient to load some data independently of the option \verb!-data!, one can call the function \nn{load\_json\_data()}.
This function accepts a parameter that is the filename of a JSON file (possibly given by a URL), and returns a named tuple containing loaded data.

\section{Declaring Variables}\label{sec:variables}

\subsection{Stand-alone Variables}

Stand-alone variables can be declared by means of the \p3 function \nn{Var()}.
To define the domain of a variable, we can simply list values, or use \nn{range()}. For example:

\begin{python}
w = Var(range(15))
x = Var(0, 1)
y = Var(0, 2, 4, 6, 8)
z = Var("a", "b", "c")
\end{python}

declares four variables corresponding to:
\begin{itemize}
\item $w \in \{0, 1, \dots, 14\}$
\item $x \in \{0, 1\}$
\item $y \in \{0, 2, 4, 6, 8\}$
\item $z \in \{a, b, c \}$
\end{itemize}

Values can be directly listed as above, or given in a set as follows:

\begin{python}
w = Var(set(range(15))) 
x = Var({0, 1})
y = Var({0, 2, 4, 6, 8})
z = Var({"a", "b", "c"})
\end{python}

It is also possible to name the parameter \nn{dom} when defining the domain:

\begin{python}
w = Var(dom=range(15))   # or equivalently, w = Var(dom=set(range(15)))
x = Var(dom={0, 1})
y = Var(dom={0, 2, 4, 6, 8})
z = Var(dom={"a", "b", "c"})
\end{python}

Finally, it is of course possible to use generators and comprehension sets. For example, for $y$, we can write:
\begin{python}
y = Var(i for i in range(10) if i % 2 == 0)
\end{python}
or equivalently: 
\begin{python}
y = Var({i for i in range(10) if i % 2 == 0})
\end{python}
or still equivalently:
\begin{python}
y = Var(dom={i for i in range(10) if i % 2 == 0})
\end{python}

\begin{remark}
In \p3, which is currently targeted to \x3-core, we can only define integer and symbolic variables with finite domains, i.e., variables with a finite set of integers or symbols (strings). 
\end{remark}

\subsection{Arrays of Variables} \label{sec:varArray}

The \p3 function for declaring an array of variables is \nn{VarArray}() that requires two named parameters \nn{size} and \nn{dom}. 
For declaring a one-dimensional array of variables, the value of \nn{size} must be an integer (or a list containing only one integer), for declaring a two-dimensional array of variables, the value of \nn{size} must be a list containing exactly two integers, and so on.
Since Version 2.6, one can use ranges instead of integers when specifying the value of \nn{size}, but each range must start at 0, and each range step must be equal to 1 (each range is seen, here, as equivalent to its length, which could be specified instead).
The named parameter \nn{dom} indicates the domain of each variable in the array.

The signature of the function \nn{VarArray}() is:

\begin{python}
  def VarArray(*, size, dom): 
\end{python}

An illustration is given by:

\begin{python}
x = VarArray(size=10, dom={0, 1})
y = VarArray(size=[5, 20], dom=range(10))
z = VarArray(size=[4, 3, 4], dom={1, 5, 10, 20})
\end{python}

We have:
\begin{itemize}
\item $x$, a one-dimensional array of 10 variables with domain $\{0,1\}$
\item $y$, a two-dimensional array of $5 \times 20$ variables with domain $\{0,1,\dots, 9\}$
\item $z$, a three-dimensional array of $4 \times 3 \times 4$ variables with domain $\{1, 5, 10, 20\}$
\end{itemize}

Indexing starts at 0. For example, $x[2]$ is the third variable of $x$, and $y[1]$ is the second row of $y$.
Technically, variable arrays are objects that are instances of \nn{ListVar}, a subclass of \nn{list}; additional functionalities of such objects are useful, for example, when posting the \gb{element} constraint.  

In some situations, you may want to declare variables in an array with different domains.
For a one-dimensional array, you can give the name of a function that accepts an integer $i$ and returns the domain to be associated with the variable at index $i$ in the array.
For a two-dimensional array, you can give the name of a function that accepts a pair of integers $(i,j)$ and returns the domain to be associated with the variable at indexes $i, j$ in the array.
And so on.

For example, suppose that the domain of all variables of the first column of $y$ is \nn{range(5)} instead of \nn{range(10)}.
We can write:

\begin{python}
def domain_y(i,j):
   return range(5) if j == 0 else range(10)

y = VarArray(size=[5, 20], dom=domain_y)
\end{python}
  
We can also use a lambda function:

\begin{python}
y = VarArray(size=[5, 20], dom=lambda i,j: range(5) if j == 0 else range(10))
\end{python}

Sometimes, not all variables in an array are relevant.
For example, you may only want to use the variables in the lower part of a two-dimensional array (matrix).
In that case, the value \nn{None} must be used.
An illustration is given below:

\paragraph{Golomb Ruler.} This problem was introduced in Section \ref{sec:golomb}.
Here is a snippet of the \p3 model:  

\begin{python}
# y[i][j] is the distance between x[i] and x[j] for i strictly less than j
y = VarArray(size=[n, n], dom=lambda i, j: range(1, n * n) if i < j else None)
\end{python}

In the array $y$, the lower part (below the main downward diagonal) only contains \nn{None}. For example, $y[1][0]$ is equal to \nn{None}.
This is taken into consideration when the \x3 file is generated by compilation.

\bigskip
Sometimes, one may want to be able to refer to variables in arrays in an individual manner.
It suffices to use facilities offered by Python, as shown in the following model.

\paragraph{Allergy.} \index{Problems!Allergy}
Four friends (two women named Debra and Janet, and two men named Hugh and Rick) found that each of them is allergic to something different:
 eggs, mold, nuts and ragweed.
We would like to match each one's surname (Baxter, Lemon, Malone and Fleet) with his or her allergy.
We know that:
\begin{itemize}
\item Rick isn't allergic to mold
\item Baxter is allergic to eggs
\item Hugh isn't surnamed Lemon or Fleet
\item Debra is allergic to ragweed
\item Janet (who isn't Lemon) isn't allergic to eggs or mold
\end{itemize}

\begin{figure}[h]
\begin{center}
  \includegraphics[scale=0.35]{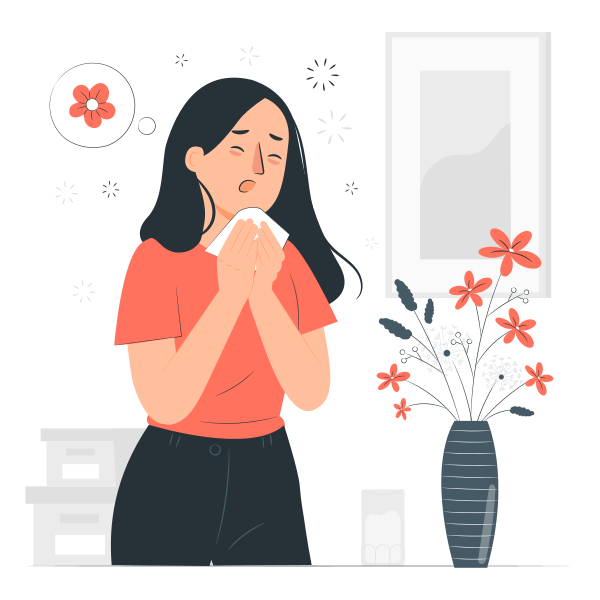}
\end{center}
\caption{Who is allergic? \tiny{(image from \href{https://fr.freepik.com/vecteurs-libre/illustration-concept-allergie-au-pollen_15577658.htm\#query=allergy&position=4&from_view=search&track=sph&uuid=f6b0bc8d-50c1-4865-84dd-c8e769acfb26}{storyset on Freepik})} \label{fig:allergy}}
\end{figure}

\bigskip
A \p3 model of this problem is given by the following file `Allergy.py':

\begin{boxpy}\begin{python}
@\imp@

Debra, Janet, Hugh, Rick = friends = ["Debra", "Janet", "Hugh", "Rick"]

# foods[i] is the friend allergic to the ith food
eggs, mold, nuts, ragweed = foods = VarArray(size=4, dom=friends)

# surnames[i] is the friend with the ith surname
baxter, lemon, malone, fleet = surnames = VarArray(size=4, dom=friends)

satisfy(
  AllDifferent(foods),
  AllDifferent(surnames),

  mold != Rick,
  eggs == baxter,
  lemon != Hugh,
  fleet != Hugh,
  ragweed == Debra,
  lemon != Janet,
  eggs != Janet,
  mold != Janet
)
\end{python}\end{boxpy}

Note how we define an array of variables, and unpack its elements.
This way, we can reason with either the array or individual variables.
Any comment put in the line preceding the declaration of a variable (or variable array) is automatically inserted in the \x3 file,
except for cases where individual variables and arrays are declared on the same line, as in the model above.

\subsection{Naming Variables and Arrays of Variables}\label{sec:naming}

Since Version 2.1, when declaring a stand-alone variable, one can set the name (id) with the parameter \nn{id} .
Then, to designate the variable, you just have to call the function \nn{var()} with the specified name.
Here is an example of model:

\begin{boxpy}\begin{python}
@\imp@

x = Var(range(10))
y = Var(dom=range(5), id="yy_12")
Var(dom=range(10), id="z")

d = dict()
a = 1
d[0] = Var(0, 1, id="d_0")
d[a] = Var(0, 1, id="d_1")

satisfy(
  x >= 3,
  var("x") <= 6,
  y > 2,
  var("yy_12") < 4,
  var("z") != 4,
  d[0] + d[1] != 0,
  var("d_0") + var("d_1") != 2
)
\end{python}\end{boxpy}

which, when compiled, gives:

\begin{xcsp}
<instance format="XCSP3" type="CSP">
  <variables>
    <var id="x"> 0..9 </var>
    <var id="yy_12"> 0..4 </var>
    <var id="z"> 0..9 </var>
    <var id="d_0"> 0 1 </var>
    <var id="d_1"> 0 1 </var>
  </variables>
  <constraints>
    <intension> ge(x,3) </intension>
    <intension> le(x,6) </intension>
    <intension> gt(yy_12,2) </intension>
    <intension> lt(yy_12,4) </intension>
    <intension> ne(z,4) </intension>
    <intension> ne(add(d_0,d_1),0) </intension>
    <intension> ne(add(d_0,d_1),2) </intension>
  </constraints>
</instance>
\end{xcsp}

Similarly, one can use the parameter \nn{id} when declaring arrays of variables, and call the function \nn{var()} to get access to arrays.
Here is an example of model:

\begin{boxpy}\begin{python}
@\imp@

x = VarArray(size=3, dom={0, 1})
y = VarArray(size=3, dom={0, 1}, id="yy")
VarArray(size=3, dom={0, 1}, id="zz")

d = dict()
a = 1
d[0] = VarArray(size=3, dom={0, 1}, id="d0")
d[a] = VarArray(size=3, dom={0, 1}, id="d_a")

satisfy(
  Sum(x) == 1,
  Sum(y) > 0,
  Sum(var("yy")) < 2,
  Sum(var("zz")) < 2,
  Sum(d[0] + d[a]) > 0,
  Sum(var("d0") + var("d_a")) < 2,
  var("d_a")[1] == 1
)
\end{python}\end{boxpy}

which, when compiled, gives:

\begin{xcsp}
<instance format="XCSP3" type="CSP">
  <variables>
    <array id="x" size="[3]"> 0 1 </array>
    <array id="yy" size="[3]"> 0 1 </array>
    <array id="zz" size="[3]"> 0 1 </array>
    <array id="d0" size="[3]"> 0 1 </array>
    <array id="d_a" size="[3]"> 0 1 </array>
  </variables>
  <constraints>
    <sum>
      <list> x[] </list>
      <condition> (eq,1) </condition>
    </sum>
    <sum>
      <list> yy[] </list>
      <condition> (gt,0) </condition>
    </sum>
    <sum>
      <list> yy[] </list>
      <condition> (lt,2) </condition>
    </sum>
    <sum>
      <list> zz[] </list>
      <condition> (lt,2) </condition>
    </sum>
    <sum>
      <list> d0[] d_a[] </list>
      <condition> (gt,0) </condition>
    </sum>
    <sum>
      <list> d0[] d_a[] </list>
      <condition> (lt,2) </condition>
    </sum>
    <intension> eq(d_a[1],1) </intension>
  </constraints>
</instance>
\end{xcsp}

\paragraph{Ghoulomb.} \index{Problems!Ghoulomb} From \href{https://github.com/MiniZinc/mzn-challenge}{Minizinc Challenge 2013}:
This is a variation of the classic Golomb ruler problem, proposed for the 2010 and 2013 Minizinc challenges:
\begin{itemize}
\item three Golomb rulers are constructed, but only the second one has to be minimized,
\item the constraint \gb{cumulative} is used instead of the constraint \gb{allDifferent},
\item instead of a resource with capacity 1 and tasks that use 1 capacity unit, the capacity is set to use more than half of the possible maximum capacity.
\end{itemize}

\begin{figure}[h]
\begin{center}
  \includegraphics[scale=0.5]{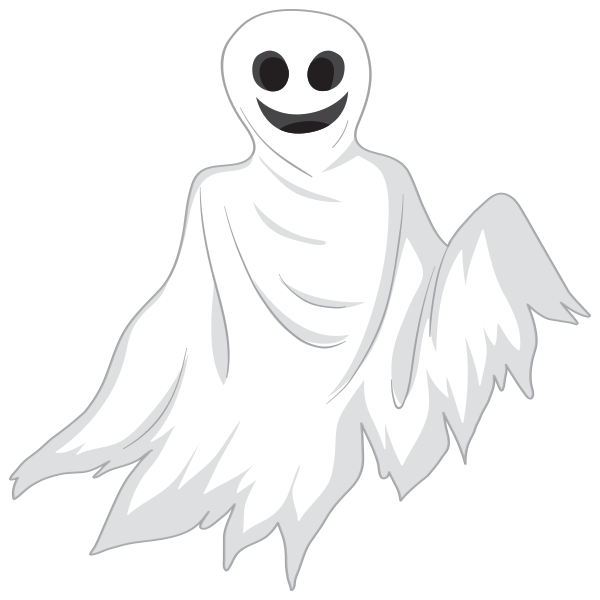}
\end{center}
\caption{Ghoulomb: based on the classic Golomb ruler problem, with a few twists to make it more evil (from Minizinc Challenge 2013). \tiny{(image from \href{https://www.freepik.com/free-vector/white-scary-ghost-isolated_21196381.htm}{brgfx on Freepik})} \label{fig:ghoul}}
\end{figure}


\bigskip
A \p3 model (which can be seen as the close translation of the one submitted to the 2010/2013 Minizinc challenges) of this problem is given by the following file `Ghoulomb.py':

\begin{boxpy}\begin{python}
@\imp@

m1, m2, m3 = data
k = 0  

def ruler(m):
   global k  # we need a global variable for the comments (see 'k') 
   k += 1

   # x'k'[i] is the location of the ith mark in ruler 'k'
   x = VarArray(size=m, dom=range(m * m + 1), id="x" + str(k))

   # d'k'[k] is the distance between the kth pair of marks of the ruler 'k'
   d = VarArray(size=(m * (m - 1)) // 2, dom=range(m * m + 1), id="d" + str(k))

   satisfy(
     # ensuring constraints for the ruler 'k'
     [
       [d[k] == x[j] - x[i] for k, (i, j) in enumerate(combinations(m, 2))],

       x[0] == 0,

       Increasing(x, strict=True),

       Cumulative(
         origins=d,
         lengths=1,
         heights=11
       ) <= 15,

       d[0] < d[-1]
     ]
   )

ruler(m1)
ruler(m2)
ruler(m3)

minimize(
  var("x2")[-1]
)
\end{python}\end{boxpy}

Note how we can define ``local'' arrays of variables \nn{x} and \nn{d} in Function \nn{ruler()}, while ensuring that the id of these arrays are different.
Note also how we can insert the value of the variable k in comments (by using simple quotes).
By executing:

\begin{command}
python Ghoulomb.py -data=[3,10,20]
\end{command}

we obtain an XCSP3 file where one can check, for example, that arrays \nn{x1}, \nn{x2} and \nn{x3} have been correctly defined:

\begin{xcsp}
<array id="x1" note="x1[i] is the location of the ith mark in ruler 1" size="[3]">
   0..9
</array>
<array id="x2" note="x2[i] is the location of the ith mark in ruler 2" size="[10]">
   0..100
</array>
<array id="x3" note="x3[i] is the location of the ith mark in ruler 3" size="[20]">
   0..400
</array>
\end{xcsp}

\subsection{Arrays of Multiple Variables (Arrays with Several Fields)}\label{sec:multiple}

Since version 2.5, it has been possible to declare arrays of variables involving several fields.
Such arrays are said to be arrays of {\em multiple} variables. 
This can be useful for writing certain models more compactly and clearly.

The \p3 function for declaring arrays of multiple variables is \nn{VarArrayMultiple}(), which requires two named parameters \nn{size} and \nn{fields}. 
The named parameter \nn{fields} must be a dictionary that gives the domain (value) of each field name (key).
The interest of arrays of multiple variables is that one can handle all fields together (in which case, we have a tuple of variables) or any field sperately (in which case, we have a single variable). 

The signature of the function \nn{VarArrayMultiple}() is:

\begin{python}
  def VarArrayMultiple(*, size, fields): 
\end{python}

For example, the statement:
\begin{python}
c = VarArrayMultiple(size=10, fields={"x": range(100), "y": range(200)})
\end{python}
introduces $10*2$ variables.
We have the possibility of writing $c[2].x$ (equivalent to $c[2][0]$), and $c[2]$ represents the pair $(c[2].x,c[2].y)$.

\paragraph{Portal Game.} \index{Problems!Portal} From \href{https://github.com/MiniZinc/mzn-challenge}{Minizinc Challenge 2024}:
The game is played on a 2D grid. Players can move up, down, left, or right.
They can also shoot a portal in any direction.
The portal travels in a straight line until it hits a wall.
Then, the player can shoot a second portal.
If the player moves onto a portal, they will be teleported to the other portal.
Shooting a third portal will make the first portal disappear.
Players can only shoot one portal at a time.
They can only have two portals on the board at a time.
They can only move one square at a time.

\begin{figure}[h!]
\begin{center}
  \includegraphics[scale=0.3]{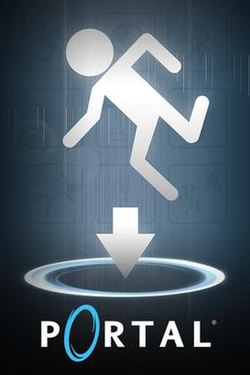}
\end{center}
\caption{Portal Game. \tiny{(image from \href{https://en.wikipedia.org/wiki/Portal}{wikipedia})} \label{fig:portal}}
\end{figure} 

As an example of data, we have a JSON file containing:

\begin{json}
{
  "inputBoard": [
    "XXXXXXXXXX",
    "XN       X",
    "XN      XX",
    "XN O     X",
    "XNNNN    X",
    "X O!N    X",
    "XO      OX",
    "X      ^ X",
    "X        X",
    "XXXXXXXXXX"
  ],
  "maxTime": 10
}
\end{json}

A \p3 model (which can be seen as the close translation of the one submitted to the 2024 Minizinc challenges) of this problem is given by the following file `Portal.py':

\begin{boxpy}\begin{python}
@\imp@
    
board, k = data or load_json_data("rand-10-9-010.json")

n, m = len(board), len(board[0])

PLAYER_NORTH, PLAYER_SOUTH, PLAYER_EAST, PLAYER_WEST, WALL, WALL_NO_PORTAL,
  EMPTY, PIT, GOAL = cells = ('^', 'v', '>', '<', 'X', 'N', ' ', 'O', '!')
NORTH, SOUTH, EAST, WEST = headings = range(4)
LEFT, RIGHT, MOVE, SHOOT = actions = range(4)
nHeadings, nActions = len(headings), len(actions)

ABSENT = -1
init = ... 
goal = ... 
init_heading = cells.index(board[init[0]][init[1]])
accessible = ...
next_heading = ...

Tw = ...
Td = ...

# pos[i] is the position of the player at the ith step
pos = VarArrayMultiple(size=k, fields={"x": range(n), "y": range(n)})

# h[i] is the heading of the player at the ith step
h = VarArray(size=k, dom=range(nHeadings))

# a[i] is the action of the player at the ith step
a = VarArray(size=k - 1, dom=range(nActions))

# p1[i] is the position of the first portal at the ith step
p1 = VarArrayMultiple(size=k, fields={"x": range(-1, n), "y": range(-1, m)})

# p2[i] is the position of the second portal at the ith step
p2 = VarArrayMultiple(size=k, fields={"x": range(-1, n), "y": range(-1, m)})

# wall[i] is the position of the first meetable wall at the ith step
wall = VarArrayMultiple(size=k - 1, fields={"x": range(-1, n), "y": range(-1, m)})

# delt[i] is the position of the player if moved at the ith step
delt = VarArrayMultiple(size=k - 1, fields={"x": range(-1,n+1), "y": range(-1,m+1)})

# z is the number of actions to reach the goal
z = Var(dom=range(k + 1))

# acc[i] is 1 if the next cell is accessible when moving at the ith step
acc = VarArray(size=k - 1, dom={0, 1})

satisfy(
   # some initializations
   [
      pos[0] == init,
      pos[-1] == goal,

      h[0] == init_heading,

      p1[0] == ABSENT,
      p2[0] == ABSENT
   ],

   # avoiding pits
   [pos[t] in {(i, j) for i in range(n) for j in range(m) if board[i][j] != PIT}
      for t in range(k)],

   # ensuring that portals are different
   [
      If(
         p1[t].x != ABSENT,  # if at least one portal
         Then=p1[t] != p2[t]
      ) for t in range(k - 1)
   ],

   # computing next headings
   [h[t + 1] == next_heading[h[t]][a[t]] for t in range(k - 1)],

   # computing next reachable walls
   [(pos[t], h[t], wall[t]) in Tw for t in range(k - 1)],

   # computing next positions
   [(pos[t], h[t], delt[t]) in Td for t in range(k - 1)],

   # computing accessibility if moving
   [acc[t] == accessible[delt[t]] for t in range(k - 1)],

   # managing action 'move'
   [
      If(
         a[t] == MOVE,
         Then=If(
            p2[t].x != ABSENT,  # if two portals
            Then=If(
               delt[t] == p1[t],
               Then=pos[t + 1] == p2[t],
               Else=If(
                  delt[t] == p2[t],
                  Then=pos[t + 1] == p1[t],
                  Else=[acc[t], pos[t + 1] == delt[t]]
               )
            ),
            Else=[acc[t], pos[t + 1] == delt[t]]
         ),
         Else=pos[t + 1] == pos[t]
      ) for t in range(k - 1)
   ],

   # managing action 'shoot'
   [
      If(
         a[t] == SHOOT, wall[t].x != ABSENT,
         Then=[p1[t + 1] == wall[t], p2[t + 1] == p1[t]],
         Else=[p1[t + 1] == p1[t], p2[t + 1] == p2[t]]
      ) for t in range(k - 1)
   ],

   # computing z
   pos[z] == goal,

   # tag(symmetry-breaking)
   [
      a[z - 1] == MOVE,
      [If(t >= z, Then=[pos[t] == goal, a[t] == LEFT]) for t in range(k - 1)],
      [a[t:t + 2] not in {(LEFT, RIGHT), (RIGHT, LEFT)} for t in range(k - 2)],
      [a[t:t + 3] not in {(RIGHT, RIGHT, RIGHT)} for t in range(k - 3)]  
   ]
)

minimize(
   # minimizing the number of steps
   z + 1
)
\end{python}\end{boxpy}

See \href{https://github.com/xcsp3team/PyCSP3-models/tree/main/recreational/Portal}{https://github.com/xcsp3team/PyCSP3-models/tree/main/recreational/Portal} for the ellipses in the model.
And note the arrays of multiple variables, which simplify the process of posting constraints in many places.

Also, in order to be able to compile (see Page \pageref{sec:indexElt}), note that one must use the option \verb!-force_element_index! when compiling the model, or to write \verb!options.force_element_index = True! at the beginning of the model file.
Another possibility is to write \verb!((aux := Var()) == z - 1, a[aux] == MOVE)! instead of \verb!a[z - 1] == MOVE!.


\section{Specifying Objectives} \label{sec:objectives}

For specifying an objective to optimize, you must call one of the two following functions:

\begin{python}
def minimize(term):
\end{python}

\begin{python}
def maximize(term):
\end{python}

The argument \nn{term} can be:
\begin{itemize}
\item a variable, as in \verb!minimize(v)!
\item an expression, as in \verb!minimize(v + w * w)!
\item a sum, as in \verb!minimize(Sum(x))!
\item a dot product, as in \verb!minimize([u,v,w] * [3, 2, 5])!
\item a generator, as in \verb!minimize(Sum((x[i] > 1) * c[i] for i in range(n)))! 
\item a minimum, as in \verb!minimize(Minimum(x))!
\item a maximum, as in \verb!minimize(Maximum(x))!
\item a number of distinct values, as in \verb!minimize(NValues(x))!
\item $\dots$
\end{itemize}

An illustration is given by the three different variants of the following problem.

\paragraph{RLFAP.} \index{Problems!Radio Link Frequency Assignment} From Cabon et al. \cite{CGLSW_radio}:
``When radio communication links are assigned the same or closely related frequencies, there is a potential for interference.
Consider a radio communication network, defined by a set of radio links.
The Radio Link Frequency Assignment Problem (RLFAP) \cite{CGLSW_radio} is to assign, from limited spectral resources, a frequency to each of these links in such a way that all the links may operate together without noticeable interference.
Moreover, the assignment has to comply with certain regulations and physical constraints of the transmitters.
Among all such assignments, one will naturally prefer those which make good use of the available spectrum, trying to save the spectral resources for a later extension of the network.

\begin{center}
\begin{tabular}{ccc}
 & \includegraphics[scale=0.25]{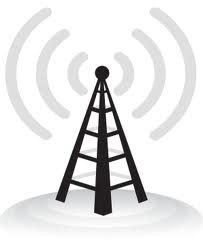}\\
\includegraphics[scale=0.25]{figures/radar.jpeg} & & \includegraphics[scale=0.25]{figures/radar.jpeg} \\
\end{tabular}
\end{center}

{\em Formal Definition:} we are given a set $X$ of unidirectional radio links.
For each link $i \in X$, a frequency $f_i$ has to be chosen from a finite set $D_i$ of frequencies available for the transmitter which yield unary constraints of type:
\begin{equation}\label{eq:dom}
f_i \in D_i
\end{equation}

Depending on the type of the problem (bulk or updating problem), some links may already have a pre-assigned frequency which define unary constraints of the type
\begin{equation}\label{eq:unary}
f_i = p_i
\end{equation}

Binary constraints are defined on pairs of links $\{i,k\}$. These constraints may be either of type:
\begin{equation}\label{eq:gt}
|f_i - f_j| > d_{ij}
\end{equation}
or of type:
\begin{equation}\label{eq:eq}
|f_i - f_j| = d_{ij}
\end{equation}

Depending on the instance considered, some of the constraints may actually be soft constraints which may be violated at some cost.
A mobility cost $m$ is defied for changing pre-assigned values, defined by constraints of type \ref{eq:unary} and an interference cost $c$ is defined for violation of soft constraints of type \ref{eq:gt}.
Constraints of type \ref{eq:dom} and \ref{eq:eq} are always hard.
The complete set of constraints $C$ is therefore partitioned in a set $H$ of hard constraints and a set $S$ of soft constraints.
Several variants can be defined:
\begin{enumerate}
\item Minimum span (SPAN): if all the constraints in $C$ can be satisfied together, one can try to minimize the largest frequency used in the assignment.
\item Minimum cardinality (CARD): if all the constraints in $C$ can be satisfied together, one can try to minimize the number of different frequencies used in the assignment.
\item Maximum Feasibility (MAX): if all the constraints in $C$ cannot be satisfied simultaneously, one should try to find an assignment that satisfies all constraints in $H$ and that minimizes the sum of all the violation costs (interference cost and mobility cost) for constraints in $S$.''
\end{enumerate}

As an illustration of data specifying an instance of this problem, we have:
\begin{json}
{
  "domains": [
    [16, 30, 44, 58, 72, 86, 100, 114, 128, 142, 156, 254, 268, ...],
    [30, 58, 86, 114, 142, 268, 296, 324, 352, 380, 414, 442, 470, ...],
    ...
  ],
  "vars": [
    {"domain": 0, "value": null, "mobility": null },
    {"domain": 1, "value": 58, "mobility": 0 },
    ...
  ],
  "ctrs":[
    {"x": 13, "y": 14, "operator": ">", "limit": 238, "weight": 0 },
    {"x": 13, "y": 16, "operator": "=", "limit": 186, "weight": 1 },
    ...
  ],
  "mobilityCosts": [0, 0, 0, 0, 0],
  "interferenceCosts": [0, 1000, 100, 10, 1]  
}
\end{json}

The fields \nn{mobility} and \nn{weight} are indexes for getting the actual cost in the two arrays \nn{mobilityCosts} and \nn{interferenceCosts}.
For more details, we refer the reader to \cite{CGLSW_radio}.

\bigskip
A \p3 model of this problem is given by the following file `RLFAP.py':

\begin{boxpy}\begin{python}
@\imp@

domains, variables, constraints, mobilityCosts, interferenceCosts = data
n = len(variables)

# f[i] is the frequency of the ith radio link
f = VarArray(size=n, dom=lambda i: domains[variables[i].domain])

satisfy(
  # managing pre-assigned frequencies
  [f[i] == v for i, (_, v, mob) in enumerate(variables)
     if v and not (variant("max") and mob)],

  # hard constraints on radio-links
  [expr(op, abs(f[i] - f[j]), k) for (i, j, op, k, wgt) in constraints
     if not (variant("max") and wgt)]
)

if variant("span"):
   minimize(
     # minimizing the largest frequency
     Maximum(f)
   )
elif variant("card"):
   minimize(
     # minimizing the number of used frequencies
     NValues(f)
   )
elif variant("max"):
   minimize(
     # minimizing the sum of violation costs
     Sum(ift(f[i] == v, 0, mobilityCosts[mob])
        for i, (_, v, mob) in enumerate(variables) if v and mob)
     + Sum(ift(expr(op, abs(f[i] - f[j]), k), 0, interferenceCosts[wgt])
        for (i, j, op, k, wgt) in constraints if wgt)
   )
\end{python}\end{boxpy}

Constraints of types \ref{eq:unary} and \ref{eq:gt} are considered to be hard when the variant is not ``max'' or the index (for mobility/interference cost) is not 0.
Note that we use the \p3 function \nn{expr()} to post the binary constraint on pairs of links; the first parameter is a string denoting an operator that can be chosen among "$<$", "$<=$", "$>=$", "$>$", "=", "==", "!=", "lt", "le", "ge", "gt", "eq", "ne", $\dots$
In our context, the code
\begin{python}
expr(op, abs(f[i] - f[j]), k)
\end{python}

is equivalent to:
\begin{python}
abs(f[i] - f[j]) == k if op == "=" else abs(f[i] - f[j]) > k
\end{python}

Concerning the objective, we have three kinds of minimization.
Note how we can combine several partial computations (here, sums), when dealing with variant ``max''.
Remember that the \p3 ternary function \nn{ift()} (if-then-else) returns either the second parameter or the third parameter according to the fact the first parameter evaluates to \nn{True} or \nn{False}.



\chapter{Twenty Five Popular Constraints}\label{ch:twenty}

In this chapter, we introduce twenty five popular constraints, those from \x3-core that are recognized by many constraint solvers.
Figure \ref{fig:intCtrs} shows their classification.
We also show, at the end of this chapter how one can manage arbitrary constraints, which have very specific semantics (not corresponding to classical ones, as defined in \x3 format).

\begin{figure}[p]
\resizebox{!}{0.97\textheight}{
\begin{tikzpicture}[dirtree,every node/.style={draw=black,thick,anchor=west}] 
\tikzstyle{selected}=[fill=colorex]
\tikzstyle{optional}=[fill=gray!12]
\node {{\bf Constraints over Integer Variables}}
    child { node [selected] {Generic Constraints}
      child { node {Constraint \gb{intension}}}
      child { node {Constraint \gb{extension}}}
    }		
    child [missing] {}
    child [missing] {}	
    child { node [selected] {Language-based Constraints}
      child { node {Constraint \gb{regular}}}
      child { node {Constraint \gb{mdd}}}
    }		
    child [missing] {}	
    child [missing] {}	
    child { node [selected] {Comparison-based Constraints}
      child { node {Constraints \gb{allDifferent}, \gb{allDifferentList}, \gb{allEqual}}}
      child { node {Constraints \gb{increasing}, \gb{decreasing}}}
      child { node {Constraints \gb{lexIncreasing}, \gb{lexDecreasing}}}
      child { node {Constraint \gb{precedence}}}
    }
    child [missing] {}
    child [missing] {}
    child [missing] {}
    child [missing] {}		
    child { node [selected] {Counting Constraints}
      child { node {Constraint \gb{sum}}}
      child { node {Constraint \gb{count}}}
      child { node {Constraint \gb{nValues}}}
      child { node {Constraint \gb{cardinality}}}
    } 
    child [missing] {}	
    child [missing] {}
    child [missing] {}	
    child [missing] {}
    child { node [selected] {Connection Constraints}
      child { node {Constraints \gb{maximum} and \gb{maximumArg}}}
      child { node {Constraints \gb{minimum} and \gb{minimumArg}}}
      child { node {Constraint \gb{element} and \gb{channel}}}
    } 
    child [missing] {}	
    child [missing] {}
    child [missing] {}
    child { node [selected] {Packing and Scheduling Constraints}
      child { node {Constraint \gb{noOverlap}}} 
      child { node {Constraint \gb{cumulative}}}
      child { node {Constraints \gb{binPacking} and \gb{knapsack}}}
    } 
    child [missing] {}
    child [missing] {}
    child [missing] {}	
    child { node [selected] {Other Constraints}
      child { node {Constraint \gb{circuit}}}
      child { node {Meta-Constraint \gb{slide}}}
    } 
    child [missing] {}	
    child [missing] {}	
    child [missing] {} 
    ;
\end{tikzpicture}
}
\caption{Popular constraints over integer variables.\label{fig:intCtrs}} 
\end{figure}

\paragraph{Semantics.}
Concerning the semantics of constraints, here are a few important remarks:

\begin{itemize}
\item when presenting the semantics, we distinguish between a variable $x$ and its assigned value $\va{x}$ (note the bold face on the symbol $x$).
\item in many constraints, quite often, we need to introduce numerical conditions (comparisons) composed of an operator $\odot$ in $\{<,\leq,>,\geq,=, \neq, \in, \notin\}$ and a right-hand side operand $k$ that can be a value (constant), a variable of the model, an interval or a set; the left-hand side being indirectly defined by the constraint.
  The numerical condition is a kind of terminal operation to be applied after the constraint has ``performed some computation''.
  In Python, the operator $\odot$ is from $\{<, <=, >, >=, ==, !\!\!=, \mathtt{in}, \mathtt{not\; in}\}$ and an interval is given by a \nn{range} object.
  A few examples of constraints involving numerical conditions are:
  
  \begin{tabular}{l}
    $\mathtt{Sum(x) > 10}$, \\
    $\mathtt{Count(x, value=1)\; in\; range(10)}$, \\
    $\mathtt{NValues(x)\; in\; \{2, 4, 6\}}$, \\
    $\mathtt{Minimum(x) == y}$
  \end{tabular}

  Of course, we can also write $\mathtt{10 < Sum(x)}$ and $\mathtt{y == Minimum(x)}$, but for simplicity of the presentation, we shall always assume that numerical conditions are on the right side.
For the semantics of a numerical condition $(\odot,k)$, and depending on the form of $k$ (a value, a variable, an interval or a set), we shall indiscriminately use $\va{k}$ to denote the value of the constant $k$, the value of the variable $k$, the interval $l..u$ represented by $k$, or the set $\{a_1,\ldots,a_p\}$ represented by $k$.
\end{itemize}

\medskip
\noindent {\bf Important.} To add constraints to a model, one has to call the \p3 function \nn{satisfy()} while passing as parameter(s):
\begin{itemize}
\item a stand-alone constraint
\item a list of constraints
\item a generator of constraints
\item a sequence of (lists of) constraints (with commas used as a separator between constraints)
\end{itemize}

We say that constraints are posted (to the model), and every call to  \nn{satisfy()} is said to be a {\em posting operation}.

\section{Constraint \gb{intension}}

An \gb{intension} constraint corresponds to a Boolean expression, which is usually called predicate.
For example, the constraint $x+y = z$ corresponds to an equation, which is an expression evaluated to $\nm{false}$ or $\nm{true}$ according to the values assigned to the variables $x$, $y$ and $z$.
However, note that for equality, we need to use `==' in Python (the operator `=' used for assignment cannot be redefined), and so, the previous constraint must be written $x+y == z$ in \p3.
To build predicates, classical arithmetic, relational and logical operators (and functions) are available; they are presented in Table \ref{tab:ops} and Table \ref{tab:opsfuncs}.
In Table \ref{tab:opsexas}, you can find a few examples of \gb{intension} constraints.
Note that the integer values $0$ and $1$ are respectively equivalent to the Boolean values $\nm{false}$ and $\nm{true}$
This allows us to combine Boolean expressions with arithmetic operators (for example, addition) without requiring any type conversions.
For example, it is valid to write $(x < 5) + (y < z) == 1$ for stating that exactly one of the Boolean expressions $x<5$ and $y<z$ must be true, although it may be possible (and/or relevant) to write it differently.

Below, $P$ denotes a predicate expression with $r$ formal parameters (not shown here, for simplicity), $X=\langle x_0,x_1,\ldots,x_{r-1} \rangle$ denotes a sequence of $r$ variables, the scope of the constraint, and $P(\va{x}_0,\va{x}_1,\ldots,\va{x}_{r-1})$ denotes the value (0/false or 1/true) returned by $P$ for a specific instantiation of the variables of $X$. 

\begin{boxse}
\begin{semantics}
  $\gb{intension}(X,P)$, with $X=\langle x_0,x_1,\ldots,x_{r-1} \rangle$ and $P$ a predicate iff 
  $P(\va{x}_0,\va{x}_1,\ldots,\va{x}_{r-1}) = true\; (1)$  @\com{recall that 1 is equivalent to true}@
\end{semantics}
\end{boxse}

\begin{remark}\label{rem:divneg}
Building constraining expressions that involve integer division (with operator // or \%) where either operand can be negative is {\bf strongly discouraged}.
In case of such a situation, the rule is ``rounding towards 0'' (as in C or Java).
Do note that language designers had to choose if their language will round towards zero, negative infinity, or positive infinity when doing integer division. 
\end{remark}

\paragraph{Zebra Puzzle.} \index{Problems!Zebra} The Zebra puzzle (sometimes referred to as Einstein's puzzle) is defined as follows.
There are five houses in a row, numbered from left to right.
Each of the five houses is painted a different color, and has one inhabitant.
The inhabitants are all of different nationalities, own different pets, drink different beverages and have different jobs.

\begin{figure}[h]
\begin{center}
  \includegraphics[scale=1.5]{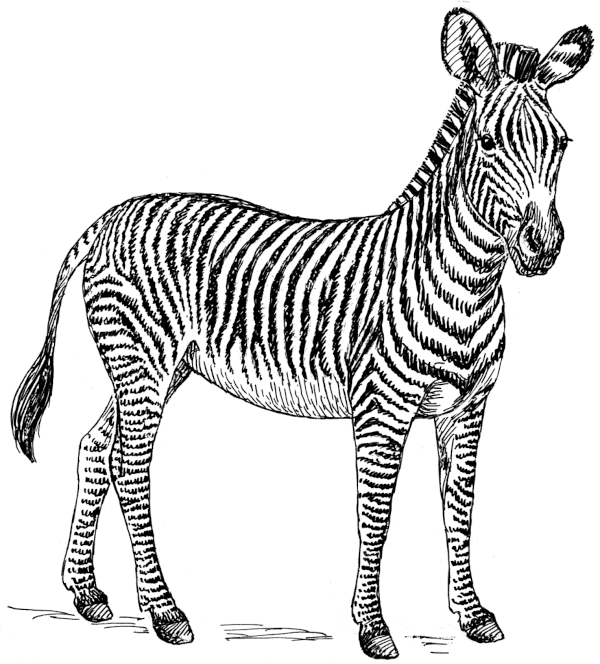} 
\end{center}
\caption{In which house lives the zebra? \tiny{(image from \href{https://commons.wikimedia.org/wiki/File:Zebra_(PSF).png}{/commons.wikimedia.org})} \label{fig:zebra}}
\end{figure}

We know that:
\begin{itemize}
 \item colors are yellow, green, red, white, and blue
 \item nations of inhabitants are italy, spain, japan, england, and norway
 \item pets are cat, zebra, bear, snails, and horse
 \item drinks are milk, water, tea, coffee, and juice
 \item jobs are painter, sculptor, diplomat, pianist, and doctor
 \item The painter owns the horse
 \item The diplomat drinks coffee
 \item The one who drinks milk lives in the white house
 \item The Spaniard is a painter
 \item The Englishman lives in the red house
 \item The snails are owned by the sculptor
 \item The green house is on the left of the red one
 \item The Norwegian lives on the right of the blue house
 \item The doctor drinks milk
 \item The diplomat is Japanese
 \item The Norwegian owns the zebra
 \item The green house is next to the white one
 \item The horse is owned by the neighbor of the diplomat
 \item The Italian either lives in the red, white or green house
\end{itemize}

\bigskip
A \p3 model of this problem is given by the following file `Zebra.py':

\begin{boxpy}\begin{python}
@\imp@

houses = range(5)  # each house has a number from 0 (left) to 4 (right)

# colors[i] is the house of the ith color
yellow, green, red, white, blue = colors = VarArray(size=5, dom=houses)

# nations[i] is the house of the inhabitant with the ith nationality
italy, spain, japan, england, norway = nations = VarArray(size=5, dom=houses)

# jobs[i] is the house of the inhabitant with the ith job
painter, sculptor, diplomat, pianist, doctor = jobs = VarArray(size=5, dom=houses)

# pets[i] is the house of the inhabitant with the ith pet
cat, zebra, bear, snails, horse = pets = VarArray(size=5, dom=houses)

# drinks[i] is the house of the inhabitant with the ith preferred drink
milk, water, tea, coffee, juice = drinks = VarArray(size=5, dom=houses)

satisfy(
  AllDifferent(colors),
  AllDifferent(nations),
  AllDifferent(jobs),
  AllDifferent(pets),
  AllDifferent(drinks),
  
  painter == horse,
  diplomat == coffee,
  white == milk,
  spain == painter,
  england == red,
  snails == sculptor,
  green + 1 == red,
  blue + 1 == norway,
  doctor == milk,
  japan == diplomat,
  norway == zebra,
  abs(green - white) == 1,
  horse in {diplomat - 1, diplomat + 1},
  italy in {red, white, green}
)    
\end{python}\end{boxpy}

In this model, there are many equations. We also use the operator \nn{in} for expressing a choice between several values.
Note how we define arrays of variables and unpack them so as to simplify the task of posting constraints.
For example, \nn{colors} is an array of 5 variables, the first one \nn{colors[0]} being given \nn{yellow} as alias, the second one \nn{colors[1]} being given \nn{green} as alias, and so on.

\bigskip
{\bf Important.} Note that we use the operators $|$, \& and \^{} for logically combining (sub-)expressions. We can't use the Python operators \nn{and}, \nn{or} and \nn{not} (because they cannot be redefined).
For example, instead of writing:
\begin{python}
horse in {diplomat - 1, diplomat + 1}
\end{python}
one could have written:
\begin{python}
(horse == diplomat - 1) | (horse == diplomat + 1)
\end{python}
However, if instead of $|$, we ever use \nn{or}:
\begin{python}
(horse == diplomat - 1) or (horse == diplomat + 1)   @\textcolor{dred}{\#  ERROR: 'or' cannot be used}@
\end{python}
we have a problem: only the first part of the disjunction is generated in \x3 (because of the short-circuit evaluation of \nn{or} by Python). 
Also, be careful about parentheses.
If ever you write:
\begin{python}
horse == diplomat - 1 | horse == diplomat + 1  @\textcolor{dred}{\#  ERROR: not what you certainly mean}@
\end{python}
this is equivalent to:
\begin{python}
horse == (diplomat - 1 | horse) == diplomat + 1
\end{python}
which is not what we wish (besides, in \p3, we cannot build expressions for \gb{intension} constraints with chaining comparison). 
Finally, when two terms must be logically combined with $|$, it is possible to use the function \nn{either()} instead, as in:
\begin{python}
either(horse == diplomat - 1, horse == diplomat + 1)
\end{python}

\section{Constraint \gb{extension}}\label{sec:extension}

An \gb{extension} constraint is often referred to as a \gb{table} constraint. 
It is defined by enumerating in a set the tuples of values that are allowed (tuples are called supports) or forbidden (tuples are called conflicts) for a sequence of variables.
A positive table constraint is then defined by a scope (a sequence or tuple of variables) $\mathtt{\langle scope \rangle}$ and a table (a set of tuples of values) $\mathtt{\langle table \rangle}$  as follows:
\begin{quote}
  $\mathtt{\langle scope \rangle\; \in\; \langle table \rangle}$
\end{quote}
When the table constraint is negative (i.e., enumerates forbidden tuples), we have:
\begin{quote}
  $\mathtt{\langle scope \rangle\; \notin\; \langle table \rangle}$
\end{quote}

With $X$ denoting a scope (sequence or tuple of variables), and $S$ and $C$ denoting sets of supports and conflicts, we have the following semantics for non-unary positive table constraints:

\begin{boxse}
\begin{semantics}
$\gb{extension}(X,S)$, with $X=\langle x_0,x_1,\ldots,x_{r-1} \rangle$ and $S$ a set of supports, iff 
  $\langle \va{x}_0,\va{x}_1,\ldots,\va{x}_{r-1}  \rangle \in S$ 

$\gbc{Prerequisite}: \forall \tau \in S, |\tau|=|X| \geq 2$
\end{semantics}
\end{boxse}

and this one for non-unary negative table constraints:

\begin{boxse}
\begin{semantics}
$\gb{extension}(X,C)$, with $X=\langle x_0,x_1,\ldots,x_{r-1} \rangle$ and $C$ a set of conflicts, iff
  $\langle \va{x}_0,\va{x}_1,\ldots,\va{x}_{r-1} \rangle \notin C$ 

$\gbc{Prerequisite}: \forall \tau \in C, |\tau|=|X| \geq 2$
\end{semantics}
\end{boxse}

In \p3, we can directly write table constraints in mathematical forms, by using tuples, sets and the operators \nn{in} and \nn{not in}.
The scope is given by a tuple of variables on the left of the constraining expression and the table is given by a set of tuples of values on the right of the constraining expression.
Although not recommended (except for huge tables), it is possible to write scopes and tables under the form of lists.
Note that for posting \gb{extension} constraints, you can also use the function \nn{Table}() if you prefer (since Version 2.4), as illustrated later with the problem TTPPV.

\paragraph{Traffic Lights.} \index{Problems!Traffic Lights}

From \href{https://www.csplib.org/Problems/prob016}{CSPLib}:
``Consider a four way traffic junction with eight traffic lights.
Four of the traffic lights are for the vehicles and can be represented by the variables $v1$ to $v4$ with domains $\{r,ry,g,y\}$ (for red, red-yellow, green and yellow).
The other four traffic lights are for the pedestrians and can be represented by the variables $p1$ to $p4$ with domains $\{r,g\}$.
 The constraints on these variables can be modeled by quaternary constraints on $(v_i, p_i, v_j, p_j)$ for $1 \leq i \leq 4, j=(1+i)\, \mathtt{mod}\, 4$ which allow just the tuples $\{(r,r,g,g), (ry,r,y,r), (g,g,r,r), (y,r,ry,r)\}$.''

\begin{figure}[bh!]
\begin{center}
  \includegraphics[scale=0.2]{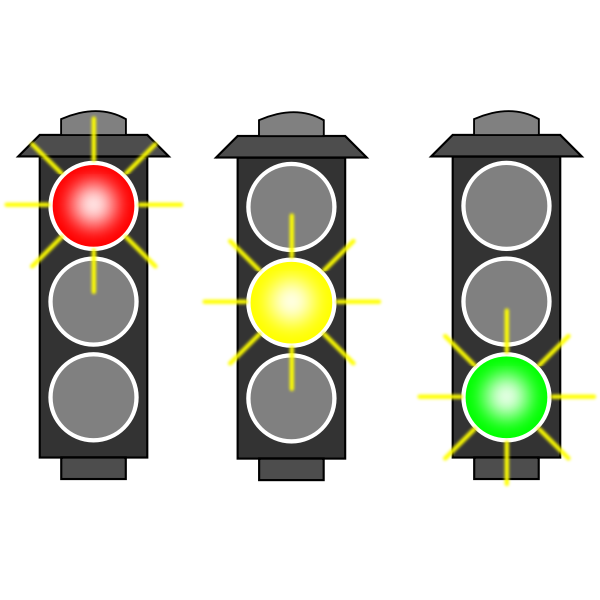} 
\end{center}
\caption{How to adjust traffic lights? \tiny{(image from \href{https://freesvg.org/traffic-lights-selection-vector-image}{freesvg.org})} \label{fig:trafficlight}}
\end{figure}

\bigskip
A \p3 model of this (single) problem is given by the following file `TrafficLights.py':

\begin{boxpy}\begin{python}
@\imp@

R, RY, G, Y = "red", "red-yellow", "green", "yellow"

T = {(R, R, G, G), (RY, R, Y, R), (G, G, R, R), (Y, R, RY, R)}

# v[i] is the color for the ith vehicle traffic light
v = VarArray(size=4, dom={R, RY, G, Y})

# p[i] is the color for the ith pedestrian traffic light
p = VarArray(size=4, dom={R, G})

satisfy(
  # ensuring the coherence of traffic lights
  (v[i], p[i], v[i + 1], p[i + 1]) in T for i in range(4)
)
\end{python}\end{boxpy}

Note how we naturally build a set of tuples (with symbolic values, here).
Four quaternary table constraints are posted in this model.  

Certainly, the attentive reader may wonder why some indexes are not out of range.
Indeed, when $i$ is set to 3, $i+1$ is equal to 4, which is out of the range of the possible indexes for $v$ and $p$.
However, in \p3, you can benefit from an auto-adjustment of array indexing (while a warning message is displayed): when an index $i$ is greater than or equal to the length of a list $t$, it is automatically transformed into i modulo the length of $t$.
More specifically, this is valid for lists of type 'ListVar' and 'ListInt', meaning arrays of variables declared in the model and arrays of integers coming from specified data.
For that reason, the statement:
\begin{python}
   (v[i], p[i], v[i + 1], p[i + 1]) in T for i in range(4)
\end{python}
is equivalent to:
\begin{python}
   (v[i], p[i], v[(i + 1) % 4], p[(i + 1) % 4]) in T for i in range(4)
\end{python}
Note that you can prevent such indexing auto-adjustment with the option '-dontadjustindexing' (when compiling).

\paragraph{Traveling Tournament with Predefined Venues.} \index{Problems!Traveling Tournament with Pred. Venues} \label{sec:ttpv}  From \href{https://www.csplib.org/Problems/prob068}{CSPLib}:
``The Traveling Tournament Problem with Predefined Venues (TTPPV) was introduced in \cite{MUR_traveling} and consists of finding an optimal compact single round robin schedule for a sport event.
Given a set of $n$ teams, each team has to play against every other team exactly once.
In each round, a team plays either at home or away, however no team can play more than two (or three) consecutive times at home or away.
The sum of the traveling distance of each team has to be minimized.
The particularity of this problem resides on the venue of each game that is predefined, i.e. if team $a$ plays against $b$ it is already known whether the game is going to be held at $a$'s home or at $b$'s home.
The original instances assume symmetric circular distances: for $i \leq j, d_{i,j}=d_{j,i}=\min(j-i,i-j+n)$.''

\begin{figure}[h]
\begin{center}
  \includegraphics[scale=0.25]{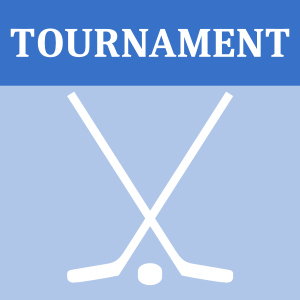} 
\end{center}
\caption{Traveling Tournament \tiny{(image from \href{https://freesvg.org/vector-graphics-of-hockey-tournament-icon}{freesvg.org})} \label{fig:tournament}}
\end{figure}

An example of data is given by the following JSON file:

\begin{json}
{
  "nTeams": 8,
  "predefinedVenues": [
    [0,1,1,0,0,0,0,1],
    [0,0,0,1,0,1,0,1],
    ...  
  ]
}
\end{json}

\medskip
A \p3 model of this problem is given by the following file `TTTPV.py':

\begin{boxpy}\begin{python}
@\imp@

nTeams, pv = data
nRounds = nTeams - 1

def cdist(i, j):  # circular distance between i and j
   return min(abs(i - j), nTeams - abs(i - j))

def T1(i):  # table used for first and last games
   # when playing at home (whatever the opponent, travel distance is 0)
   return {(1, ANY, 0)} | {(0, j, cdist(i, j)) for j in range(nTeams) if j != i}

def T2(i):  # table used for other games 
   return ({(1, 1, ANY, ANY, 0)} |
      {(0, 1, j, ANY, cdist(j, i)) for j in range(nTeams) if j != i} |
      {(1, 0, ANY, j, cdist(i, j)) for j in range(nTeams) if j != i} |
      {(0, 0, j, k, cdist(j, k)) for j in range(nTeams) for k in range(nTeams)
        if different_values(i, j, k)})

def automaton():
   q, q01, q02, q11, q12 = "q", "q01", "q02", "q11", "q12"
   t = [(q, 0, q01), (q, 1, q11), (q01, 0, q02), (q01, 1, q11), (q11, 0, q01),
        (q11, 1, q12), (q02, 1, q11), (q12, 0, q01)]
   return Automaton(start=q, transitions=t, final={q01, q02, q11, q12})

# o[i][k] is the opponent (team) of the ith team  at the kth round
o = VarArray(size=[nTeams, nRounds], dom=range(nTeams))

# h[i][k] is 1 iff the ith team plays at home at the kth round
h = VarArray(size=[nTeams, nRounds], dom={0, 1})

# t[i][k] is the traveled distance by the ith team at the kth round.
# An additional round is considered for returning at home.
t = VarArray(size=[nTeams, nRounds + 1], dom=range(nTeams // 2 + 1))

satisfy(
  # a team cannot play against itself
  [o[i][k] != i for i in range(nTeams) for k in range(nRounds)],

  # ensuring predefined venues 
  [pv[i][o[i][k]] == h[i][k] for i in range(nTeams) for k in range(nRounds)],
  
  # ensuring symmetry of games: if team i plays against j, then j plays against i
  [o[:, k][o[i][k]] == i for i in range(nTeams) for k in range(nRounds)],
  
  # each team plays once against all other teams
  [AllDifferent(row) for row in o],
  
  # at most 2 consecutive games at home, or consecutive games away
  [h[i] in automaton() for i in range(nTeams)],
  
  # handling traveling for the first game
  [(h[i][0], o[i][0], t[i][0]) in T1(i) for i in range(nTeams)],
  
  # handling traveling for the last game
  [(h[i][-1], o[i][-1], t[i][-1]) in T1(i) for i in range(nTeams)],
  
  # handling traveling for two successive games
  [(h[i][k], h[i][k + 1], o[i][k], o[i][k + 1], t[i][k + 1]) in T2(i)
     for i in range(nTeams) for k in range(nRounds - 1)]
)

minimize(
  # minimizing summed up traveled distance   
  Sum(t)
)
\end{python}\end{boxpy}

Two functions, called \nn{T1}() and \nn{T2}(), are introduced here to build {\em short} tables, i.e., tables that contain the special symbol '*', denoted in \p3 by the constant ANY.
When the symbol '*' is present, it means that any value from the domain of the corresponding variable can be present at its position.
For more information about short tables, see e.g., \cite{JN_extending,VLS_extending}.
Remember that the symbol $|$ can be used in Python to perform the union of two sets, and that we use the notation $o[:, k]$ to extract the kth column of the array $o$, as in NumPy.
Some \gb{regular} constraints (based on automata) are also posted, but we shall discuss them in the next section. 

Since Version 2.4, one can alternatively use the function \nn{Table}() for posting \gb{extension} constraints.
Depending on the context, it may render the code clearer or not (it is also a matter of taste).
For example, the two last groups of extensional constraints from the model above can be written:

\begin{python}
satisfy(
  ...
  
  # handling traveling for the last game
  [
    Table(
      scope=(h[i][-1], o[i][-1], t[i][-1]),
      supports=T1(i)
    ) for i in range(nTeams)
  ],
  
  # handling traveling for two successive games
  [
    Table(
      scope=(h[i][k], h[i][k + 1], o[i][k], o[i][k + 1], t[i][k + 1]),
      supports=T2(i)
    ) for i in range(nTeams) for k in range(nRounds - 1)
  ]

\end{python}

\paragraph{Subgraph Isomorphism Problem.} \index{Problems!Subgraph Isomorphism} An instance of the {\em subgraph isomorphism problem} is defined by a pattern graph $G_p=(V_p,E_p)$ and a target graph $G_t=(V_t,E_t)$: the  objective is to determine whether $G_p$ is isomorphic to some subgraph(s) in $G_t$.
Finding a solution to such a problem instance means then finding a {\em subisomorphism function}, that is an injective mapping $f : V_p \rightarrow V_t$ such that all edges of $G_p$ are preserved: $\forall (v,v') \in E_p, (f(v_p),f(v'_p)) \in E_t$. 
Here, we refer to the partial, and not the induced subgraph isomorphism problem.

\begin{figure}[h]
  \centering
  \begin{subfigure}[t]{0.48\textwidth}
    \centering
    \begin{tikzpicture}[scale=1]
      \tikzstyle{node}=[circle,fill=white,draw,font=\sffamily\normalsize\bfseries]
      \tikzstyle{arete}=[thick,-,>=stealth]
      \node[node] (1) at (1.5,1.5) {$1$};
      \node[node] (2) at (0,0) {$2$};
      \node[node] (3) at (3,0) {$3$};
      \node[node] (4) at (1.5,0.5) {$4$};
      \draw[arete] (1) -- (2) ;
      \draw[arete] (1) -- (3) ;
      \draw[arete] (1) -- (4) ;
      \draw[arete] (2) -- (3) ;
      \draw[arete] (2) -- (4) ;
      \draw[arete] (3) -- (4) ;
    \end{tikzpicture}
    \caption{Pattern Graph}
  \end{subfigure}
  \begin{subfigure}[t]{0.48\textwidth}
    \centering
    \begin{tikzpicture}[scale=1]
      \tikzstyle{node}=[circle,fill=white,draw,font=\sffamily\normalsize\bfseries]
      \tikzstyle{arete}=[thick,-,>=stealth]      
      \node[node] (a) at (0,1.5) {$a$};
      \node[node] (b) at (0,0) {$b$};
      \node[node] (c) at (3,0) {$c$};
      \node[node] (d) at (3,1.5) {$d$};
      \node[node] (e) at (1.5,0.75) {$e$};      
      \draw[arete] (a) -- (b) ;
      \draw[arete] (a) -- (d) ;
      \draw[arete] (a) -- (e) ;
      \draw[arete] (b) -- (c) ;
      \draw[arete] (b) -- (e) ;
      \draw[arete] (c) -- (d) ;
      \draw[arete] (c) -- (e) ;
      \draw[arete] (d) -- (e) ;
    \end{tikzpicture}
    \caption{Target Graph}
  \end{subfigure}
  \caption{An Instance of the Subgraph Isomorphism Problem \label{fig:subgraphIsomorphism}}
\end{figure}

An example of data is given by the following JSON file:

\begin{json}
{
  "nPatternNodes": 180,
  "nTargetNodes": 200,
  "patternEdges":[[0,1], [0,3], [0,17], ... ], 
  "targetEdges":[[0,34], [0,65], [0,129], ...]
}
\end{json}

A \p3 model of this problem is given by the following file `Subisomorphism.py':

\begin{boxpy}\begin{python} 
@\imp@

n, m, p_edges, t_edges = data

# useful auxiliary structures
T = {(i, j) for i, j in t_edges} | {(j, i) for i, j in t_edges}
p_loops =  [i for (i, j) in p_edges if i == j]
t_loops =  [i for (i, j) in t_edges if i == j]
p_degrees = [len([edge for edge in p_edges if i in edge]) for i in range(n)]
t_degrees = [len([edge for edge in t_edges if i in edge]) for i in range(m)]
conflicts = [{j for j in range(m) if t_degrees[j] < p_degrees[i]} for i in range(n)]

# x[i] is the target node to which the ith pattern node is mapped
x = VarArray(size=n, dom=range(m))

satisfy(
  # ensuring injectivity
  AllDifferent(x),

  # preserving edges
  [(x[i], x[j]) in T for (i, j) in p_edges],
  
  # being careful of self-loops
  [x[i] in t_loops for i in p_loops],
  
  # tag(redundant)
  [x[i] not in C for i, C in enumerate(conflicts)]
)
\end{python}\end{boxpy}

In this model, some binary \gb{extension} constraints are posted for preserving edges, and some unary \gb{extension} constraints are posted for handling self-loops as well as for reducing domains by reasoning from node degrees.
Note that for a unary \gb{extension} constraint, we use the form: $x$ \nn{in} $S$ (and  $x$ \nn{not in} $S$) where $x$ is a variable of the model and $S$ a set of values. 
For a negative table constraint, if ever the length of the table is 0, no constraint is posted.

\section{Constraint \gb{regular}} \label{sec:regular}

\begin{definition}[DFA]
A {\em deterministic finite automaton} (DFA) is a 5-tuple $(Q,\Sigma,\delta,q_0,F)$ where $Q$ is a finite set of states, $\Sigma$ is a finite set of symbols called the alphabet, $\delta:Q \times \Sigma \rightarrow Q$  is a transition function,  $q_0 \in Q$ is the initial state, and $F \subseteq Q$ is the set of final states.
\end{definition}

Given an input string (a finite sequence of symbols taken from the alphabet $\Sigma$), the automaton starts in the initial state $q_0$, and for each symbol in sequence of the string, applies the transition function to update the current state.  
If the last state reached is a final state then the input string is accepted by the automaton.  
The set of strings that the automaton $A$ accepts constitutes a language, denoted by $L(A)$, which is technically a regular language.
When the automaton is non-deterministic, we can find two transitions $(q_i,a,q_j)$ and $(q_i,a,q_k)$ such that $q_j \neq q_k$.

A \gb{regular} constraint \cite{CB_constraints,P_regular} ensures that the sequence of values assigned to the variables of its scope must belong to a given regular language (i.e., forms a word that can be recognized by a deterministic, or non-deterministic, finite automaton).
For such constraints, a DFA is then used to  determine whether or not a given tuple is accepted. 
This can be an attractive approach when constraint relations can be naturally represented by regular expressions in a known regular language.  
For example, in rostering problems, regular expressions can represent valid patterns of activities.
The semantics is:

\begin{boxse}
\begin{semantics}
$\gb{regular}(X,A)$, with $X=\langle x_0,x_1,\ldots,x_{r-1} \rangle$ and $A$ a finite automaton, iff 
  $\va{x}_0\va{x}_1\ldots\va{x}_{r-1} \in L(A)$ 
\end{semantics}
\end{boxse}

In \p3, we can directly write \gb{regular} constraints in mathematical forms, by using tuples, automata and the operator \nn{in}. 
The scope of a constraint is given by a tuple of variables on the left of the constraining expression and an automaton is given on the right of the constraining expression.
Automata in \p3 are objects of Class \nn{Automaton} that are built by calling the following constructor:

\begin{python}
  def __init__(self, *, start, transitions, final):
\end{python}

Three named parameters are required:
\begin{itemize}
\item \nn{start} is the name of the initial state (a string)
\item \nn{transitions} is a set (or list) of 3-tuples
\item \nn{final} is the set (or list) of the names of final states (strings)
\end{itemize}

Note that the set of states and the alphabet can be inferred from \nn{transitions}.

\begin{center}
\begin{tikzpicture}[node distance=1cm, >=stealth, auto] 
 \tikzstyle{snode}=[state,minimum size=6mm]
 \node[snode, initial]            (a)                       {$a$};
 \node[snode]                     (b)[right=of a]          {$b$};
 \node[snode]                     (c)[right=of b]          {$c$};
 \node[snode]                     (d)[right=of c]          {$d$};
 \node[snode, accepting]          (e)[right=of d]          {$e$};
 \path[->] 
 (a)  edge[loop above]  node{0}  (a)
 (a)  edge node{1}   (b)
 (b)  edge node{1}   (c)
 (c)  edge node{0}   (d)
 (d)  edge[loop above]  node{0}  ()
 (d)  edge node{1}   (e)
 (e)  edge[loop above]  node{0}  ();
\end{tikzpicture}
\end{center}

As an example, the constraint defined on scope $\langle x_1,x_2,\ldots,x_7 \rangle$ from the simple automation depicted above  is given in \p3 by:

\begin{python}
a, b, c, d, e = "a", "b", "c", "d", "e"
t = {(a,0,a), (a,1,b), (b,1,c), (c,0,d), (d,0,d), (d,1,e), (e,0,e)}
A = Automaton(start=a, transitions=t, final=e)

satisfy(
  (x1, x2, x3, x4, x5, x6, x7) in A,

  ...
)
\end{python}

This gives, after compiling to \x3:

\begin{xcsp}  
<regular>
   <list> x1 x2 x3 x4 x5 x6 x7 </list>
   <transitions> 
     (a,0,a)(a,1,b)(b,1,c)(c,0,d)(d,0,d)(d,1,e)(e,0,e) 
   </transitions>
   <start> a </start>
   <final> e </final>
</regular>
 \end{xcsp}

\paragraph{Traveling Tournament with Predefined Venues.} This problem was introduced in Section \ref{sec:ttpv}.
Here is a snippet of the \p3 model:  

\begin{python}
def automaton():
   q, q01, q02, q11, q12 = "q", "q01", "q02", "q11", "q12"
   t = [(q, 0, q01), (q, 1, q11), (q01, 0, q02), (q01, 1, q11), (q11, 0, q01),
        (q11, 1, q12), (q02, 1, q11), (q12, 0, q01)]
   return Automaton(start=q, transitions=t, final={q01, q02, q11, q12})

A = automaton()
   
satisfy(
  # at most 2 consecutive games at home, or consecutive games away
  [h[i] in A for i in range(nTeams)]
)
\end{python}

Since Version 2.4, one can alternatively use the function \nn{Regular}() for posting \gb{regular} constraints.
Depending on the context, it may render the code clearer or not (it is also a matter of taste).
For example, the group of constraints above can be written:

\begin{python}
def automaton():
   q, q01, q02, q11, q12 = "q", "q01", "q02", "q11", "q12"
   t = [(q, 0, q01), (q, 1, q11), (q01, 0, q02), (q01, 1, q11), (q11, 0, q01),
        (q11, 1, q12), (q02, 1, q11), (q12, 0, q01)]
   return Automaton(start=q, transitions=t, final={q01, q02, q11, q12})

A = automaton()
   
satisfy(
  # at most 2 consecutive games at home, or consecutive games away
  [
    Regular(
      scope=h[i],
      automaton=A
    ) for i in range(nTeams)
  ]
)
\end{python}

\section{Constraint \gb{mdd}}

The constraint \gb{mdd} \cite{CY_maintaining,CY_maintainingr,CY_mdd,PR_GAC4} ensures that the sequence of values assigned to the variables it involves follows a path going from the root of the described MDD (Multi-valued Decision Diagram) to the unique terminal node.
Because the graph is directed, acyclic, with only one root node and only one terminal node, we just need to introduce the set of transitions.

Below, $L(M)$ denotes the language recognized by an MDD $M$.

\begin{boxse}
\begin{semantics}
$\gb{mdd}(X,M)$, with  $X=\langle x_0,x_1,\ldots,x_{r-1} \rangle$ and $M$ an MDD, iff 
  $\va{x}_0\va{x}_1\ldots\va{x}_{r-1} \in L(M)$ 
\end{semantics}
\end{boxse}

In \p3, we can directly write \gb{mdd} constraints in mathematical forms, by using tuples, MDDs and the operator \nn{in}. 
The scope of a constraint is given by a tuple of variables on the left of the constraining expression and an MDD is given on the right of the constraining expression.
MDDs in \p3 are objects of Class \nn{MDD} that are built by calling the following constructor:

\begin{python}
   def __init__(self, transitions):
\end{python}

The named parameter \nn{transitions} is required: this is a list (not a set) of 3-tuples.
As said above, the root and terminal nodes (and the full set of states) can be inferred from \nn{transitions}, if the MDD is well constructed.

\begin{center}
\begin{tikzpicture}[>=stealth]
  \tikzstyle{snode}=[draw,circle,minimum size=6mm]
  \tikzstyle{sedge}=[draw,->,>=latex]
  \tikzstyle{slabel}=[midway,scale=1.2]
  \node[snode] (0) at (0,4.5) {$r$};
  \node[snode] (1) at (-1.5,3)  {$n_1$};
  \node[snode] (2) at (0,3) {$n_2$};
  \node[snode] (3) at (1.5,3) {$n_3$};
  \node[snode] (4) at (-0.75,1.5)  {$n_4$};
  \node[snode] (5) at (0.75,1.5) {$n_5$};
  \node[snode] (6) at (0,0) {$t$};
  \node[minimum size=6mm] (x1) at (3.2,3.75)  {$u$};
  \node[minimum size=6mm] (x2) at (3.2,2.25) {$v$};
  \node[minimum size=6mm] (x3) at (3.2,0.75) {$w$};

  \draw[sedge] (0) -- node[slabel,left]{0} (1);
  \path[sedge] (0) edge node[slabel,right]{$1$} (2);
  \path[sedge] (0) edge node[slabel,right]{$2$} (3);
  \path[sedge] (1) edge node[slabel,left]{$2$} (4);
  \path[sedge] (2) edge node[slabel,right]{$2$} (4);
  \path[sedge] (3) edge node[slabel,right]{$0$} (5);
  \path[sedge] (4) edge node[slabel,left]{$0$} (6);
  \path[sedge] (5) edge node[slabel,right]{$0$} (6);
\end{tikzpicture}
\end{center}

As an example, the constraint of scope $\langle u,v,w \rangle$ is defined from the simple MDD depicted above (with root node $r$ and terminal node $t$) as:

\begin{python}
r, n1, n2, n3, n4, n5, t = "r", "n1", "n2", "n3", "n4", "n5", "t" 
transitions = [(r,0,n1), (r,1,n2), (r,2,n3), (n1,2,n4),
               (n2,2,n4), (n3,0,n5), (n4,0,t), (n5,0,t)]
M = MDD(transitions)

satisfy(
  (u, v, w) in M,

  ...
)
\end{python}

\paragraph{Word Design for DNA Computing on Surfaces.} \index{Problems!Word Design} From \href{https://www.csplib.org/Problems/prob033}{CSPLib}:
``The problem is to find as large as possible a set $S$ of strings (words) of length 8 over the alphabet $W = \{ A,C,G,T \}$ with the following properties:
\begin{itemize}
\item each word in $S$ has 4 symbols from $\{ C,G \}$
\item each pair of distinct words in $S$ differ in at least 4 positions
\item each pair of words $x$ and $y$ in S (where $x$ and $y$ may be identical) are such that $x^R$ and $y^C$ differ in at least 4 positions.
  Here, $(x_1,\dots,x_8 )^R = (x_8,\dots,x_1)$ is the reverse of $x=(x_1,\dots,x_8)$ and $(x_1,\dots,x_8)^C$ is the Watson-Crick complement of $x=(x_1,\dots,x_8)$, i.e. the word where each A is replaced by a T and vice versa, and each C is replaced by a G and vice versa.
\end{itemize}

This problem has its roots in Bioinformatics and Coding Theory.''

\begin{figure}[h!]
\begin{center}
  \includegraphics[scale=0.03]{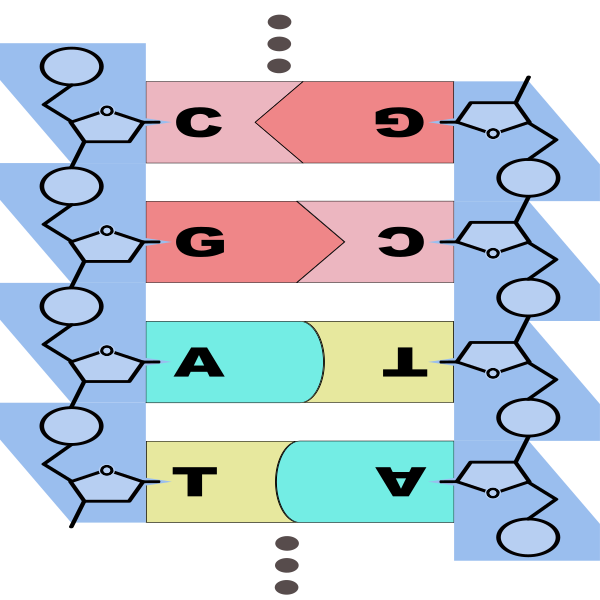}
\end{center}
\caption{Forming words from the four bases found in DNA: Adenine, Cytosine, Guanine and Thymine.  \tiny{(image from \href{https://commons.wikimedia.org/wiki/File:Genetik-DNA-4-Basen-einfach.svg}{commons.wikimedia.org})} \label{fig:dna}}
\end{figure}

A first precomputed JSON file, called `words.json', indicates the possible words (each word has 4 symbols from $\{1,2\} = \{C,G \}$) and is such that its reverse and Watson-Crick complement differ in at least 4 positions):
\begin{json}
{
  "words": [
    [0,0,0,0,1,1,1,1],
    [0,0,0,0,1,1,1,2],
    [0,0,0,0,1,1,2,1],
    [0,0,0,0,1,1,2,2],
    ...
  ]
}
\end{json}

A second precomputed JSON file, called `mdd.json', indicates the transitions of an MDD that can be used to enforce the restrictions on pairs of words:
\begin{json}
{
  "transitions":[
    ["root",0,"n3"],
    ["root",1,"n286276"],
    ["root",2,"n430777"],
    ...
  ]
}
\end{json}

A \p3 model of this problem is given by the following file `WordDesign2.py':

\begin{boxpy}\begin{python}
@\imp@

words, transitions, n = data  
M = MDD(transitions)

# x[i][k] is the kth letter (0-A, 1-C, 2-G, 3-T) of the ith word
x = VarArray(size=[n, 8], dom=range(4))

satisfy(
  # each word must be well-formed
  [x[i] in words for i in range(n)],

  # ordering words  tag(symmetry-breaking)
  LexIncreasing(x, strict=True),
  
  # ensuring the validity of any pair of words
  [x[i] + x[j] in M for i, j in combinations(n, 2)]
)
\end{python}\end{boxpy} 

This model involves 1 array of variables and 3 types of constraints: \gb{Extension}, \gb{LexIncreasing} and \gb{MDD}.
For generating an \x3 instance (file), you can execute for example:
\begin{command}
python WordDesign2.py -data=[words.json,mdd.json,n=15]
\end{command}
Note how we can append a specific parameter to the data coming from two JSON files.

\section{Constraint \gb{allDifferent}}

The constraint \gb{allDifferent}, see \cite{R_filtering,H_alldiff,GMN_generalized}, ensures that the variables in a specified list $X$ must all take different values.
A variant, called \gb{allDifferentExcept} in the literature \cite{BCR_global,C_dulmage}, enforces variables to take distinct values, except those that are assigned to some specified values (often, the single value 0).
This is the role of the set $E$ below.

\begin{boxse}
\begin{semantics}
$\gb{allDifferent}(X,E)$, with $X=\langle x_0,x_1,\ldots \rangle$, iff 
  $\forall  (i,j) : 0 \leq i < j < |X|, \va{x}_i \neq \va{x}_j \lor \va{x}_i \in E \lor \va{x}_j \in E$
$\gb{allDifferent}(X)$ iff $\gb{allDifferent}(X,\emptyset)$ 
\end{semantics}
\end{boxse}

In \p3, to post a constraint \gb{allDifferent}, we must call the function \nn{AllDifferent}() whose signature is:

\begin{python}
  def AllDifferent(term, *others, excepting=None, @matrix@=None):
\end{python}

The two parameters \texttt{term} and \texttt{others} are positional, and allow us to pass the terms either in sequence (individually) or under the form of a list.
The optional named parameter \nn{excepting} indicates the value (or the set of values) that must be ignored, and the optional named parameter \nn{matrix} indicates if a constraint \gb{allDifferent} must be imposed on both rows and columns of a two-dimensional list (matrix). 
More accurately, the terms can be given as:
\begin{itemize}
\item a list of variables, as in \nn{AllDifferent(x)} 
\item a sequence of individual variables, as in \nn{AllDifferent(u, v, w)}
\item a generator of variables, as in \nn{AllDifferent(x[i] for i in range(n) if i\%2 > 0)}
\item a sequence of individual expressions, as in \nn{AllDifferent(x[1] + 1, x[2] + 2, x[3] + 3)}
\item a generator of expressions, as in \nn{AllDifferent(x[i] + i for i in range(n))}
\end{itemize}

Below, we introduce some additional models involving the \gb{allDifferent} constraint.

\paragraph{Send-More-Money.} \index{Problems!Send-More-Money} \label{sec:sendmore}  From \href{https://en.wikipedia.org/wiki/Verbal_arithmetic}{Wikipedia}:
Cryptarithmetic is a type of mathematical game consisting of a mathematical equation among unknown numbers, whose digits are represented by letters.
The goal is to identify the value of each letter.
The classic example, published in the July 1924 issue of Strand Magazine by Henry Dudeney is: 

\begin{center}
\begin{minipage}[b]{0.3\linewidth}
\begin{verbatim}
     S E N D
+    M O R E
=  M O N E Y
\end{verbatim}
\end{minipage}
\end{center}

A \p3 model for this specific example is given by:

\begin{boxpy}\begin{python} 
@\imp@

# letters[i] is the digit of the ith letter involved in the equation
s, e, n, d, m, o, r, y = letters = VarArray(size=8, dom=range(10))

satisfy(
  # letters are given different values
  AllDifferent(letters),

  # words cannot start with 0
  [s > 0, m > 0],
  
  # respecting the mathematical equation
     [s, e, n, d] * [1000, 100, 10, 1]
  +  [m, o, r, e] * [1000, 100, 10, 1]
  == [m, o, n, e, y] * [10000, 1000, 100, 10, 1]
)
\end{python}\end{boxpy}

\medskip
It is important to note that not only variables but also general expressions can be involved in the \gb{allDifferent} constraint, as shown in Section \ref{sec:queens} and the following model.

\paragraph{Costas Arrays.} \index{Problems!Costas Arrays} \label{sec:costas}
From \href{https://www.csplib.org/Problems/prob076/}{CSPLib}:
``A costas array is a pattern of $n$ marks on an $n \times n$ grid, one mark per row and one per column, in which the $n \times (n - 1)/2$ (displacement) vectors between the marks are all-different.
Such patterns are important as they provide a template for generating radar and sonar signals with ideal ambiguity functions.''

\begin{figure}[h!] 
\begin{center}
  \includegraphics[scale=1]{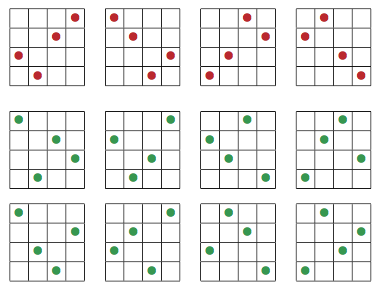}
\end{center}
\caption{The 12 Costas arrays of order 4. \tiny{(image from \href{https://commons.wikimedia.org/wiki/File:CostasArray44.png}{commons.wikimedia.org})} \label{fig:costas}}
\end{figure}

\bigskip
A \p3 model of this problem is given by the following file `CostasArray.py': 

\begin{boxpy}\begin{python}
@\imp@

n = data

# x[i] is the row where is put the ith mark (on the ith column)
x = VarArray(size=n, dom=range(n))

satisfy(
  # all marks are on different rows (and columns)
  AllDifferent(x),

  # all displacement vectors between the marks must be different
  [AllDifferent(x[i] - x[i + d] for i in range(n - d)) for d in range(1, n - 1)]
)
\end{python}\end{boxpy}

Now, assuming that $x$ is a two-dimensional list (array) of variables, the matrix variant of \gb{allDifferent} is imposed on $x$ by:
\nn{AllDifferent(x, matrix=True)}.
If $x = [ [u_1, u_2, u_3, u_4], [v_1, v_2, v_3, v_4], [w_1, w_2, w_3, w_4]]$, then the posted constraint is equivalent to having posted:
\begin{itemize}
\item \nn{AllDifferent($u_1,u_2,u_3,u_4$)}
\item \nn{AllDifferent($v_1,v_2,v_3,v_4$)}
\item \nn{AllDifferent($w_1,w_2,w_3,w_4$)}
\item \nn{AllDifferent($u_1,v_1,w_1$)} 
\item \nn{AllDifferent($u_2,v_2,w_2$)}
\item \nn{AllDifferent($u_3,v_3,w_3$)}
\item \nn{AllDifferent($u_4,v_4,w_4$)}
\end{itemize}

The matrix variant of \gb{allDifferent} was introduced in Section \ref{sec:sudoku}.
Here is another illustration.

\paragraph{Futoshiki.} \index{Problems!Futoshiki} \label{sec:futoshiki}  From \href{https://en.wikipedia.org/wiki/Futoshiki}{Wikipedia}:
``Futoshiki is a logic puzzle game from Japan, which was developed by Tamaki Seto in 2001.
The puzzle is played on a square grid, and the objective is to place the numbers such that each row and column contains only one of each digit.
Some digits may be given at the start, and inequality constraints are initially specified between some of the squares, such that one must be higher or lower than its neighbor.''

\begin{figure}[h]
  \centering
    \begin{subfigure}[t]{0.5\textwidth}
        \centering
        \includegraphics[scale=0.22]{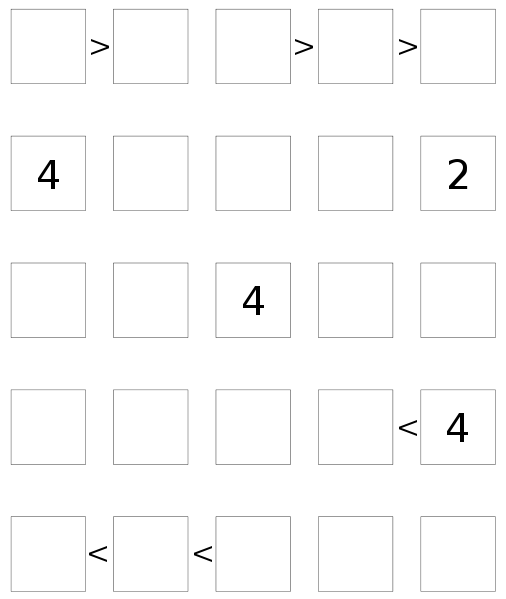}
        \caption{Puzzle}
    \end{subfigure}%
    ~ 
    \begin{subfigure}[t]{0.5\textwidth}
        \centering
        \includegraphics[scale=0.22]{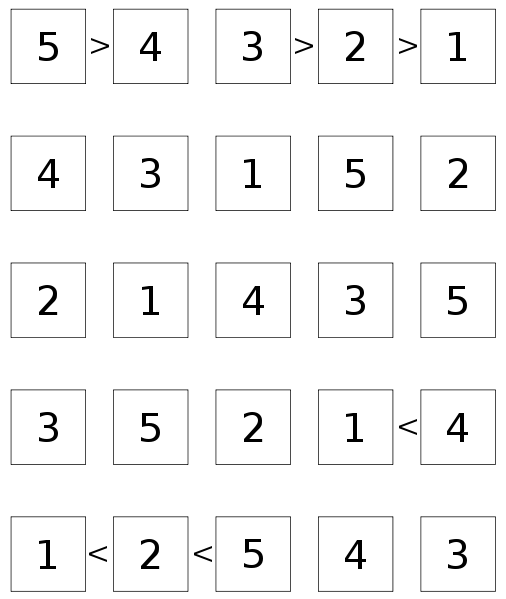}
        \caption{Solution}
    \end{subfigure}
    \caption{Solving a Futoshiki Puzzle. \tiny{(images from \href{https://commons.wikimedia.org/wiki/File:Futoshiki1.svg}{commons.wikimedia.org})} \label{fig:futoshiki}} 
\end{figure}

An example of data is given by the following JSON file:

\begin{json}
{
  "size": 3,
  "nbHints": [{"row":0, "col":0, "number":2}],
  "opHints": [
    {"row":0, "col":1, "lessThan":true, "horizontal":true},
    {"row":2, "col":0, "lessThan":true, "horizontal":true}
  ]
}
\end{json}

A \p3 model of this problem is given by the following file `Futoshiki.py': 

\begin{boxpy}\begin{python}
@\imp@

n, nbHints, opHints = data  # n is the order of the grid

# x[i][j] is the number put at row i and column j
x = VarArray(size=[n, n], dom=range(1, n + 1))

satisfy(
  # different values on each row and each column
  AllDifferent(x, matrix=True),

  # respecting number hints
  [x[i][j] == k for (i, j, k) in nbHints],
  
  # respecting operator hints
  [y < z if lt else y > z
     for (y, z, lt) in [(x[i][j], x[i][j + 1] if hr else x[i + 1][j], lt)
        for (i, j, lt, hr) in opHints]]
)
\end{python}\end{boxpy}

Because objects from the JSON file are automatically converted to named tuples, note how we can use tuple unpacking when iterating over lists of such objects.

\medskip
Here is now an illustration concerning the ``except'' variant of \gb{allDifferent}.

\paragraph{Progressive Party.} This problem will be introduced in Section \ref{sec:progressive}. Here is a snippet of the \p3 model:

\begin{python}
# s[b][p] is the scheduled (visited) boat by the crew of boat b at period p
s = VarArray(size=[nBoats, nPeriods], dom=range(nBoats))

satisfy(
  ...

  # a guest crew cannot revisit a host
  [AllDifferent(s[b], excepting=b) for b in range(nBoats)],

  ... 
)
\end{python}

Because the crew can stay several periods on his boat, while visiting different boats on other periods, we need  \gb{allDifferent} with the named parameter \nn{excepting}.

\section{Constraint \gb{allDifferentList}}

The constraint \gb{allDifferentList} admits as parameters two (or more) lists of integer variables, and ensures that the tuple of values taken by variables of the first list is different from the tuple of values taken by variables of the second list.
If more than two lists are given, all tuples must be different. 
A variant enforces tuples to take distinct values, except those that are assigned to some specified tuples (often, the single tuple containing only 0). 


\begin{boxse}
\begin{semantics}
$\gb{allDifferentList}(\ns{X},E)$, with $\ns{X}=\langle X_1, X_2,\ldots \rangle$, $E$ the set of discarded tuples, iff 
  $\forall (i,j) : 1 \leq i < j \leq |\ns{X}|, \va{X}_i \neq \va{X}_j \lor \va{X}_i \in E \lor \va{X}_j \in E$ 
$\gb{allDifferentList}(\ns{X})$ iff $\gb{allDifferentList}(\ns{X},\emptyset)$ 

$\gbc{Prerequisite}: |\ns{X}| \geq 2 \land \forall i : 1 \leq i < |\ns{X}|, |X_i| = |X_{i+1}| \geq 2 \land \forall \tau \in E, |\tau|=|X_1|$
\end{semantics}
\end{boxse}

In \p3, to post a constraint \gb{allDifferentList}, we must call the function \nn{AllDifferentList}() whose signature is:

\begin{python}
  def AllDifferentList(term, *others, excepting=None):
\end{python}

The two parameters \texttt{term} and \texttt{others} are positional, and allow us to pass the terms either in sequence (individually) or under the form of a matrix.
The optional named parameter \nn{excepting} indicates the tuple (or the set of tuples) that must be ignored.

\paragraph{Crossword Generation.} \index{Problems!Crossword Generation}
``Given a grid with imposed black cells (spots) and a dictionary, the problem is to fulfill the grid with the words contained in the dictionary.''
An illustration is given by Figure \ref{fig:cross}.

\begin{figure}[h]
  \centering
  \begin{subfigure}[t]{0.35\textwidth}
    \centering
    \begin{Puzzle}{6}{6}
      |A  |L  |O  |H  |A  |*  |.
      |X  |*  |R  |I  |C  |H  |.
      |I  |C  |E  |*  |H  |A  |.
      |O  |R  |*  |W  |E  |I  |.
      |M  |A  |M  |A  |*  |F  |.
      |*  |G  |E  |N  |O  |A  |.
    \end{Puzzle}
    \caption{Crossword Grid}
  \end{subfigure}%
  ~
  \begin{subfigure}[t]{0.35\textwidth}
    \centering
    \PuzzleSolution
    \begin{Puzzle}{6}{6}
      |A  |L  |O  |H  |A  |*  |.
      |X  |*  |R  |I  |C  |H  |.
      |I  |C  |E  |*  |H  |A  |.
      |O  |R  |*  |W  |E  |I  |.
      |M  |A  |M  |A  |*  |F  |.
      |*  |G  |E  |N  |O  |A  |.
    \end{Puzzle}
        \caption{Solution}
    \end{subfigure}
    \caption{Making a Crossword Puzzle. \label{fig:cross}} 
\end{figure}

An example of data is given by the following JSON file `grid-ogd.json':
\begin{json}
{
  "spots": [
    [0,0,0,0,0,1],
    [0,1,0,0,0,0],
    [0,0,0,1,0,0],
    [0,0,1,0,0,0],
    [0,0,0,0,1,0],
    [1,0,0,0,0,0]],
  "dictFileName": "ogd"
}
\end{json}

The grid is specified by the field \nn{spots} of the root object in the JSON file; when present, the value 1 means the presence of a spot (black cell).
The name of the dictionary to be used is also given (it is clearly unreasonable to include the content of the dictionary in the JSON file if we expect to generate several instances from the same dictionary).

A \p3 model of this problem is given by the following file `Crossword.py': 

\begin{boxpy}\begin{python} 
@\imp@

spots, dict_name = data
words = dict()  # we load/build the dictionary of words
for line in open(dict_name):
  code = alphabet_positions(line.strip().lower())
  words.setdefault(len(code), []).append(code)

def find_holes(tab, transposed):
  def build_hole(row, col, size, horizontal):
    sl = slice(col, col + size)
    return Hole(row, sl, size) if horizontal else Hole(sl, row, size)

  Hole = namedtuple("Hole", "i j r")  # i and j are indexes (one being a slice)
  p, q = len(tab), len(tab[0])
  t = []
  for i in range(p):
    start = -1
    for j in range(q):
      if tab[i][j] == 1:
        if start != -1 and j - start >= 2:
          t.append(build_hole(i, start, j - start, not transposed))
        start = -1
      elif start == -1:
        start = j
      elif j == q - 1 and q - start >= 2:
        t.append(build_hole(i, start, q - start, not transposed))
  return t

holes = find_holes(spots, False) + find_holes(columns(spots), True)
arities = sorted(set(arity for (_, _, arity) in holes))
n, m, nHoles = len(spots), len(spots[0]), len(holes)

#  x[i][j] is the letter, number from 0 to 25, at row i and column j (when no spot)
x = VarArray(size=[n, m], dom=lambda i, j: range(26) if spots[i][j] == 0 else None)

satisfy(
  # fill the grid with words
  [x[i, j] in words[r] for (i, j, r) in holes],

  # tag(distinct-words)
  [
    AllDifferentList(
      x[i, j] for (i, j, r) in holes if r == arity
    ) for arity in arities
  ]
)
\end{python}\end{boxpy}

One can then execute: 
\begin{command}
python Crossword.py -data=grid-ogd.json
\end{command}

If one wants to use another dictionary, as e.g., the dictionary (file) `words', one can execute:
\begin{command}
python Crossword.py -data=[grid-ogd.json,dictFileName='words']
\end{command}

Finally, one can find irrelevant the fact of having both the grid and the dictionary specified in the JSON file.
One may prefer to have a JSON file `grid.json' depicting the grid:

\begin{json}
{
  "spots": [
    [0,0,0,0,0,1],
    [0,1,0,0,0,0],
    [0,0,0,1,0,0],
    [0,0,1,0,0,0],
    [0,0,0,0,1,0],
    [1,0,0,0,0,0]]
}
\end{json}

and execute:
\begin{command}
python Crossword.py -data=[grid.json,dictFileName='ogd']
\end{command}
or
\begin{command}
python Crossword.py -data=[grid.json,dictFileName='words']
\end{command}

\section{Constraint \gb{allEqual}}

The constraint \gb{allEqual} ensures that all involved variables take the same value.

\begin{boxse}
\begin{semantics}
$\gb{allEqual}(X)$, with $X=\langle x_0,x_1,\ldots \rangle$, iff 
  $\forall (i, j) : 0 \leq i < j < |X|, \va{x}_i = \va{x}_j$
\end{semantics}
\end{boxse}

In Python, we can call the function \nn{AllEqual}() with a list of variables as parameter.

\medskip
\paragraph{Domino.} \index{Problems!Domino} As an illustration, let us consider the problem Domino that was introduced in \cite{ZY_AC3.1} to emphasize the sub-optimality of a generic constraint propagation algorithm (called AC3).
Each instance, characterized by two integers $n$ and $d$, is binary and corresponds to an undirected constraint graph with a cycle. 
More precisely, $n$ denotes the number of variables, each with $\{0,\dots,d-1\}$ as domain, and there exist:
\begin{itemize}
\item $n-1$ equality constraints: $x_i = x_{i+1}, \forall i \in \{0,\dots,n-2\}$
\item a trigger constraint: $(x_0 + 1 = x_{n-1}) \vee (x_0 = x_{n-1} = d-1)$
\end{itemize}

\begin{figure}[h]
\begin{center}
  \includegraphics[scale=0.15]{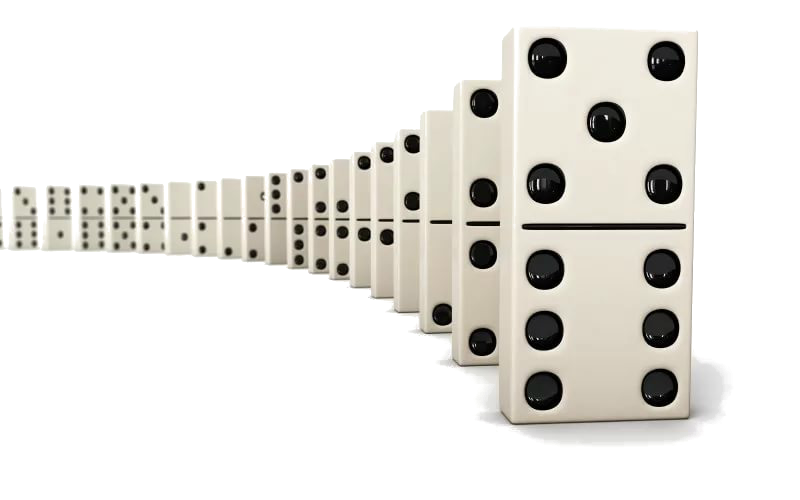} 
\end{center}
\caption{Filtering as a Domino (cascade) effect.  \tiny{(image from \href{https://pngimg.com/download/54995}{pngimg.com})} \label{fig:domino}}
\end{figure}

Those who are interested in the way domains of variables can be filtered (i.e., reduced) in this problem will observe a kind of Domino (cascade) effect \cite{ZY_AC3.1,L_constraint}.
A \p3 model of this problem is given by the following file `Domino.py':

\begin{boxpy}\begin{python}
@\imp@

n, d = data

# x[i] is the value of the ith domino
x = VarArray(size=n, dom=range(d))

satisfy(
  AllEqual(x),

  either(
    x[0] + 1 == x[-1],
    both(x[0] == x[-1], x[0] == d - 1)
  )
)
\end{python}\end{boxpy}

Of course, it is possible to replace the constraint \gb{allEqual} by:

\begin{python}
  [x[i] == x[i + 1] for i in range(n - 1)],
\end{python}

The constraint \gb{allEqual} is mainly introduced for its ease of use.

\section{Constraints \gb{increasing} and \gb{decreasing}}

The constraint \gb{ordered} ensures that the variables of a specified list of variables $X$ are ordered in sequence, according to a specified relational operator $\odot \in \{<,\leq,\geq,>\}$.
An optional list of integers or variables $L$ indicates the minimum distance between any two successive variables of $X$.

\begin{boxse}
\begin{semantics}
$\gb{ordered}(X,L,\odot)$, with $X=\langle x_0,x_1,\ldots \rangle$, $L=\langle l_0,l_1,\ldots \rangle$ and $\odot \in \{<,\leq,\geq,>\}$, iff 
  $\forall i : 0 \leq i < |X|-1, \va{x}_i + l_i \odot \va{x}_{i+1}$
$\gb{ordered}(X,\odot)$, with $X=\langle x_0,x_1,\ldots \rangle$ and $\odot \in \{<,\leq,\geq,>\}$, iff 
  $\forall i : 0 \leq i < |X|-1, \va{x}_i \odot \va{x}_{i+1}$

$\gbc{Prerequisite}: |X| = |L| + 1$
\end{semantics}
\end{boxse}

In \p3, to post a constraint \gb{ordered}, we must call either the function \nn{Increasing}() or the function \nn{Decreasing}(),  whose signatures are:

\begin{python}
  def Increasing(term, *others, strict=False, lengths=None):
  def Decreasing(term, *others, strict=False, lengths=None):
\end{python}

The two parameters \texttt{term} and \texttt{others} are positional, and allow us to pass the variables either in sequence (individually) or under the form of a list.
The optional named parameter \nn{strict} indicates if the relation must be strict or not, and the optional named parameter \nn{lengths} is for specifying minimum distances.  
In other words, assuming that $x = [u, v, w]$ is a simple list of variables, ordering variables of $x$ can be imposed by: 
\begin{itemize}
\item \nn{Increasing(x, strict=True)}\\ensuring $u < v < w$
\item \nn{Increasing(x)} \\ensuring $u \leq v \leq w$
\item \nn{Decreasing(x)} \\ensuring $u \geq v \geq w$
\item \nn{Decreasing(x, strict=True)}\\ensuring $u > v > w$
\end{itemize}

The constraints \gb{increasing} and \gb{decreasing} are mainly an ease of use, as it is possible to post equivalent \gb{intension} constraints.
For example,  \nn{Increasing(x, strict=True)} can be equivalently written as:
\begin{python}
  [x[i] < x[i + 1] for i in range(len(x) - 1)]
\end{python}

\paragraph{Steiner Triple Systems.} \index{Problems!Steiner Triple Systems} From \href{https://www.csplib.org/Problems/prob044}{CSPLib}:
``The ternary Steiner problem of order $n$ consists of finding a set of $n \times (n - 1)/6$ triples of distinct integer elements in $\{1,2, \dots,n\}$ such that any two triples have at most one common element.
It is a hypergraph problem coming from combinatorial mathematics where $n$ modulo $6$ has to be equal to $1$ or $3$.
One possible solution for $n=7$ is $\{\{1, 2, 3\}, \{1, 4, 5\}, \{1, 6, 7\}, \{2, 4, 6\}, \{2, 5, 7\}, \{3, 4, 7\}, \{3, 5, 6\}\}$.
This is a particular case of the more general Steiner system.''

A \p3 model of this problem is given by the following file `Steiner3.py':

\begin{boxpy}\begin{python}
@\imp@

n = data
nTriples = (n * (n - 1)) // 6
T = {(i1, i2, i3, j1, j2, j3)
       for (i1, i2, i3, j1, j2, j3) in product(range(1, n + 1), repeat=6)
         if different_values(i1, i2, i3) and different_values(j1, j2, j3)
            and len({i for i in {i1, i2, i3} if i in {j1, j2, j3}}) <= 1}

# x[i] is the ith triple of value
x = VarArray(size=[nTriples, 3], dom=range(1, n + 1))

satisfy(
  # each triple must be formed of strictly increasing integers
  [Increasing(triple, strict=True) for triple in x],

  # each pair of triples must share at most one value
  [(triple1 + triple2) in T for (triple1, triple2) in combinations(x, 2)]
)
\end{python}\end{boxpy}

\begin{figure}[h!]
  \begin{center}
\begin{tikzpicture}
  \draw (30:1)  -- (210:2)
        (150:1) -- (330:2)
        (270:1) -- (90:2)
        (90:2)  -- (210:2) -- (330:2) -- cycle
        (0:0)   circle (1);
  \fill (0:0)   circle(3pt)
        (30:1)  circle(3pt)
        (90:2)  circle(3pt)
        (150:1) circle(3pt)
        (210:2) circle(3pt)
        (270:1) circle(3pt)
        (330:2) circle(3pt);
\end{tikzpicture}
  \end{center}
\caption{The Fano plane is a Steiner triple system. The triples (blocks) correspond to the 7 lines, each containing 3 points. Every pair of points belongs to a unique line. \label{fig:steiner}}
\end{figure}

\section{Constraints \gb{lexIncreasing} and \gb{lexDecreasing}} \label{sec:lex}

The constraint \gb{ordered} can be naturally lifted to lists, by considering the lexicographic order.
Because this constraint is very popular, it is called \gb{lex}, instead of \gb{ordered} over lists of integer variables.
The constraint \gb{lex}, see \cite{CB_revisiting,FHKMW_global}, ensures that the tuple formed by the values assigned to the variables of a first specified list $X_1$ is related to the tuple formed by the values assigned to the variables of a second specified list $X_2$ with respect to a specified lexicographic order operator $\odot \in  \{<_{lex}, \leq_{lex}, \geq_{lex}, >_{lex}\}$.
If more than two lists of variables are specified, the entire sequence of tuples must be ordered; this captures then \gb{lexChain} \cite{CB_arc}.


\begin{boxse}
\begin{semantics}
$\gb{lex}(\ns{X},\odot)$, with $\ns{X}=\langle X_0,X_1,\ldots \rangle$ and $\odot \in \{<_{lex}, \leq_{lex}, \geq_{lex}, >_{lex}\}$, iff 
  $\forall i : 0 \leq i < |\ns{X}|-1, \va{X}_i \odot \va{X}_{i+1}$ 

$\gbc{Prerequisite}: |\ns{X}| \geq 2 \land \forall i : 0 \leq i < |\ns{X}|-1, |X_i| = |X_{i+1}| \geq 2$
\end{semantics}
\end{boxse}

In \p3, to post a constraint \gb{lex}, we must call either the function \nn{LexIncreasing}() or the function \nn{LexDecreasing}(),  whose signatures are:

\begin{python}
def LexIncreasing(term, *others, strict=False, matrix=False):
def LexDecreasing(term, *others, strict=False, matrix=False):
\end{python}

The two parameters \texttt{term} and \texttt{others} are positional, and allow us to pass the lists either in sequence (individually) or under the form of a two-dimensional list.
The optional named parameter \nn{strict} indicates if the relation must be strict or not, and the optional named parameter \nn{matrix} indicates if a lexicographic order must be imposed on both rows and columns of a two-dimensional list (matrix). 
In other words, assuming that $x$, $y$ and $z$ are simple lists of variables, ordering lexicographically $x$, $y$ and $z$ can be imposed by:
\begin{itemize}
\item \nn{LexIncreasing(x y, z, strict=True)}\\ ensuring $x <_{lex} y <_{lex} z$
\item \nn{LexIncreasing(x, y, z)} \\ ensuring $x \leq_{lex} y \leq_{lex} z$
\item \nn{LexDecreasing(x, y, z)} \\ ensuring $x \geq_{lex} y \geq_{lex} z$
\item \nn{LexDecreasing(x, y, z, strict=True)} \\ ensuring $x >_{lex} y >_{lex} z$
\end{itemize}

Now, assuming that $x$ is a two-dimensional list of variables, the matrix variant of \gb{lex} with $\leq_{lex}$ (for example) as operator is imposed on $x$ by:
\nn{LexIncreasing(x, matrix=True)}.
If $x = [ [p, q, r], [u, v, w]]$, then the posted constraint is equivalent to having posted:
\begin{itemize}
\item $(p,q,r) \leq_{lex} (u,v,w)$ 
\item $(p,u) \leq_{lex} (q,v) \leq_{lex} (r,w)$
\end{itemize}

Since Version 2.3, it is possible to use the Python operators '<', '<=', '>' and '>=' for posting lexicographic constraints involving exactly two lists of variables.
If $x$ and $y$ are two lists of variables, then we can write:
\begin{python}
  satisfy(
    x <= y
  )
\end{python}
instead of:
\begin{python}
  satisfy(
    LexIncreasing(x,y)
  )
\end{python}

\paragraph{Social Golfers.} \index{Problems!Social Golfers}
``The coordinator of a local golf club has come to you with the following problem.
In their club, there are 32 social golfers, each of whom play golf once a week, and always in groups of 4.
They would like you to come up with a schedule of play for these golfers, to last as many weeks as possible, such that no golfer plays in the same group as any other golfer on more than one occasion. 
The problem can easily be generalized to that of scheduling $G$ groups of $K$ golfers over at most $W$ weeks, such that no golfer plays in the same group as any other golfer twice (i.e. maximum socialisation is achieved).
For the original problem, the values of $G$ and $K$ are respectively 8 and 4.''
See \href{https://www.csplib.org/Problems/prob010}{CSPLib}.

\begin{figure}[h!]
\begin{center}
  \includegraphics[scale=0.9]{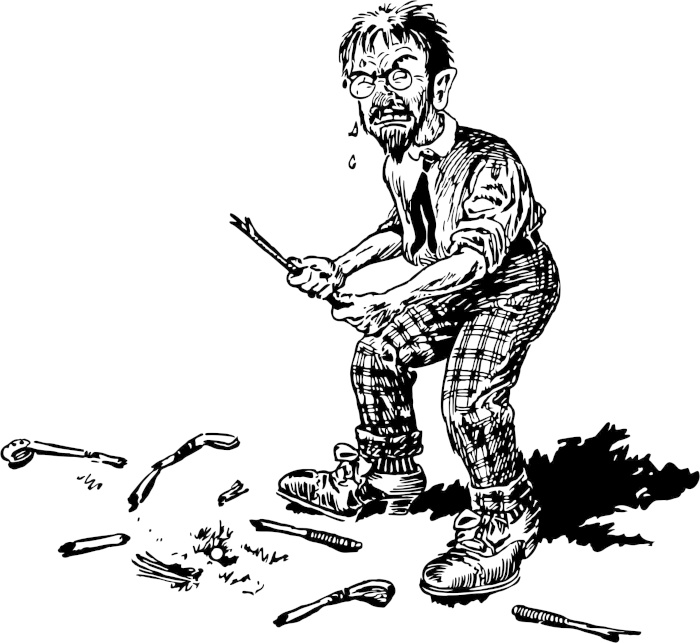}
\end{center}
\caption{A golfer who apparently needs socialization. \tiny{(image from \href{https://www.publicdomainpictures.net/pictures/140000/velka/angry-golf-man.jpg}{www.publicdomainpictures.net})} \label{fig:socialGolfers}}
\end{figure}

A \p3 model of this problem is given by the following file `SocialGolfers.py':

\begin{boxpy}\begin{python}
@\imp@

nGroups, @size@, nWeeks = data
nPlayers = nGroups * @size@

# g[w][p] is the group admitting on week w the player p
g = VarArray(size=[nWeeks, nPlayers], dom=range(nGroups))

satisfy(
  # ensuring that two players don't meet more than once
  [
    If(
      g[w1][p1] == g[w1][p2],
      Then=g[w2][p1] != g[w2][p2]
    ) for w1, w2 in combinations(nWeeks, 2) for p1, p2 in combinations(nPlayers, 2)
  ],

  # respecting the size of the groups
  [
    Cardinality(
      within=g[w],
      occurrences={i: @size@ for i in range(nGroups)}
    ) for w in range(nWeeks)
  ],

  # tag(symmetry-breaking)
  LexIncreasing(g, matrix=True)
)
\end{python}\end{boxpy}

We have the guarantee of keeping at least one solution if the instance is satisfiable, when the matrix \gb{lex} constraint is posted.

\section{Constraint \gb{precedence}}\label{sec:precedence}

The constraint \gb{precedence}, see \cite{LL_precedence,W_precedence}, ensures that if a variable $x$ of a specified list $X$ is assigned the $i+1th$ value of a specified list $V$ of values, then another variable of $X$, that precedes $x$, is assigned the $ith$ value of $V$.
In general, this constraint is useful for breaking value symmetries.
For the semantics, $V^{\nm{cv}}$ means \texttt{covered}=\nn{true}.

\begin{boxse}
\begin{semantics}
$\gb{precedence}(X,V)$, with $X=\langle x_1,x_2,\ldots \rangle$ and $V=\langle v_1,v_2,\ldots \rangle$ iff  
 $\forall i : 1 \leq i < |V|, v_{i+1} \in \{\va{x}_i : 1 \leq i \leq |X|\} \Rightarrow  v_{i} \in \{\va{x}_i : 1 \leq i \leq |X|\}$
 $\forall i : 1 \leq i < |V| \land v_{i+1} \in \{\va{x}_i : 1 \leq i \leq |X|\}$, 
   $\min\{j : 1 \leq j \leq |X| \land \va{x}_j = v_i\} < \min\{j :  1 \leq j \leq |X| \land \va{x}_j = v_{i+1}\}$
$\gb{precedence}(X,V^{\nm{cv}})$ iff  $\gb{precedence}(X,V) \wedge v_{|V|} \in \{\va{x}_i : 1 \leq i \leq |X|\}$
\end{semantics}
\end{boxse}

In \p3, to post a constraint \gb{precedence}, we must call the function \nn{Precedence}() whose signature is:

\begin{python}
  def Precedence(scope, *, values=None, covered=False)  
\end{python}

Only the list (scope) is required.
When absent, the list of values is assumed to be the ordered set of values collected over the domains of all variables in the scope.
The parameter \texttt{covered} is optional: when \nn{true}, each value of the specified list must be assigned by at least one variable in the scope of the constraint.

\paragraph{Community Detection.} \index{Problems!Community Detection}
The problem of constrained community detection is described with many details in \cite{GBS_community}.
The problem is to partition the set of nodes of a graph (the parts forming so-called communities) while seeking maximum modularity (as defined by a matrix).
Among possible constraints related to some background knowledge, one can impose that some pairs of nodes must be assigned to the same or different communities.

\begin{figure}[h!]
\begin{center}
  \includegraphics[scale=0.25]{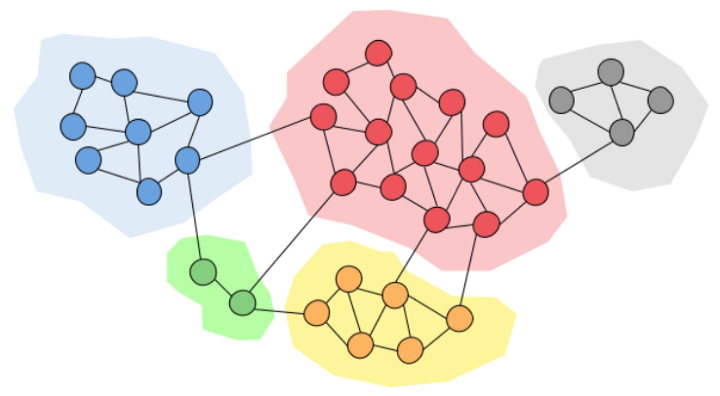}
\end{center}
\caption{Forming Communities. \tiny{(image by Thamindu Dilshan Jayawickrama)} \label{fig:community}}
\end{figure}

An example of data is given by the following JSON file `comm1.json':

\begin{json}
{
  "graph": [
    [0,0,1,0,0,1],
    [0,1,0,0,0,0],
    [0,0,0,1,0,0],
    [0,0,1,0,0,0],
    [0,0,1,0,1,0],
    [1,0,1,0,1,1]],
  "together": [[0,5], [2,3], [4,5]],
  "separate": [[1,3], [3, 5]],
  "maxCommunities": 3
}
\end{json}

The graph is given by its adjacency matrix, and nodes that must be put together or in separate communities are indicated by lists.
The maximum number of communities is also indicated.

A \p3 model (similar to the one proposed for the 2021 Minizinc challenge)  of this problem is given by the following file `CommunityDetection.py':

\begin{boxpy}\begin{python}
@\imp@

graph, together, separate, m = data  # m is the maximum number of communities
n = len(graph)  # number of nodes

def modularity_matrix():
   degrees = [sum(graph[i]) for i in range(n)]  # node degrees
   sum_degrees = sum(degrees)  # multiplier used to avoid fractions
   return [[sum_degrees * graph[i][j] - degrees[i] * degrees[j] for j in range(n)]
     for i in range(n)]

W = modularity_matrix()

# x[i] is the community of the ith node
x = VarArray(size=n, dom=range(m))

satisfy(
  # considering nodes that must belong to the same community
  [x[i] == x[j] for i, j in together],

  # considering nodes that must not belong to the same community
  [x[i] != x[j] for i, j in separate],
  
  # tag(symmetry-breaking)
  Precedence(x)
)

maximize(
  Sum((x[i] == x[j]) * W[i][j] for i, j in combinations(n, 2) if W[i][j] != 0)
)
\end{python}\end{boxpy}

As mentioned in \cite{GBS_community}, the constraint \gb{precedence}, called \gb{value\_precede\_chain} in Minizinc, can be very useful:
``This constraint enforces a unique community numbering for any particular partition. It can be viewed as a lexicographic ordering constraint on the assignment of vertices to communities.''
It prevents the search from exploring k! symmetric equivalent solutions obtained by permuting the community numbers.
Note that \verb!Precedence(x)! is equivalent to writing \verb!Precedence(x, values=range(m))!.

\medskip
Since \x3 Specifications 3.1, \gb{precedence} belongs to \x3-core.

\section{Constraint \gb{sum}}\label{sec:sum}

The constraint \gb{sum} is one of the most important constraints. 
This constraint may involve (integer or variable) coefficients, and is subject to a numerical condition $(\odot, k)$.
For example, a form of \gb{sum}, sometimes called \gb{subset-sum} or \gb{knapsack} \cite{T_dynamic,PQ_counting} involves the operator $\mathtt{in}$, and ensures that the computed sum belongs to a specified interval.
Below, we introduce the semantics while considering a main list $X$ of variables and a list $C$ of coefficients:

\begin{boxse}
\begin{semantics}
$\gb{sum}(X,C,(\odot,k))$, with $X=\langle x_0,x_1,\ldots \rangle$, and $C=\langle c_0,c_1,\ldots\rangle$, iff
  $(\sum_{i=0}^{|X|-1} \va{c}_i \times \va{x}_i) \odot \va{k}$ 

$\gbc{Prerequisite}: |X| = |C| \geq 2$
\end{semantics}
\end{boxse}

In \p3, to post a constraint \gb{sum}, we must call the function \nn{Sum}() whose signature is:

\begin{python}
  def Sum(term, *others):
\end{python}

The two parameters \texttt{term} and \texttt{others} are positional, and allow us to pass the terms either in sequence (individually) or under the form of a list.
More accurately, the terms can be given as:
\begin{itemize}
\item a list of variables, as in \nn{Sum(x)} 
\item a sequence of individual variables, as in \nn{Sum(u, v, w)}
\item a generator of variables, as in \nn{Sum(x[i] for i in range(n) if i\%2 > 0)}
\item a generator of variables, with coefficients, as in \nn{Sum(x[i] * costs[i] for i in range(n))}
\item a generator of expressions, as in \nn{Sum(x[i] > 0 for i in range(n))}
\item a generator of expressions, with coefficients, as in \nn{Sum((x[i] + y[i]) * costs[i] for i in range(n))}
\end{itemize}

Note that arguments are flattened, meaning that variables (and expressions) are collected from arguments to form a simple list even if multi-dimensional structures (lists) are involved, and while discarding any occurrence of the value \nn{None}.
For example, flattening \nn{[ [u, v], [None, w]]} gives \nn{[u, v, w]}.

The object obtained when calling \nn{Sum}() must be restricted by a condition (typically, defined by a relational operator and a limit).

\paragraph{Magic Sequence.} This problem was introduced in Section \ref{sec:magicSequence}.
Here is a snippet of the \p3 model:   

\begin{python}
satisfy(
  ...

  # tag(redundant)
  [
    Sum(x) == n,

    Sum((i - 1) * x[i] for i in range(n)) == 0
  ]
)
\end{python}

The first \gb{sum} constraint involves a simple list $x$ of variables whereas the second one involves terms that are products of variables and coefficients. 

\bigskip
Importantly, it is possible to combine several objects \nn{Sum} with operators $+$ and $-$ (and to compare them, which is equivalent to a subtraction).
This is illustrated below, with a general model for crypto-arithmetic puzzles (in Section \ref{sec:sendmore}, we introduced a specific model dedicated to `send+more=money').

\paragraph{Crypto Puzzle.} \index{Problems!Crypto Puzzle}
In crypto-arithmetic problems, digits (values between 0 and 9) are represented by letters.
Different letters stand for different digits, and different occurrences of the same letter denote the same digit.
The problem is then represented as an arithmetic operation between words.
The task is to find out which letter stands for which digit, so that the result of the given arithmetic operation is true.

For example,

\begin{center}
\color{dred}{
\begin{minipage}[b]{0.1\linewidth}
~ ~ 
\end{minipage}
\begin{minipage}[b]{0.2\linewidth}\centering
\begin{verbatim}
     N O
+    N O 
=  Y E S  
\end{verbatim}
\end{minipage}
\hspace{0.2cm}
\begin{minipage}[b]{0.3\linewidth}\centering
\begin{verbatim}
     C R O S S
+    R O A D S
=  D A N G E R 
\end{verbatim}
\end{minipage}
\hspace{0.2cm}
\begin{minipage}[b]{0.3\linewidth}\centering
\begin{verbatim}
   D O N A L D
+  G E R A L D
=  R O B E R T
\end{verbatim}
\end{minipage}
}
\end{center}

A \p3 model of this problem is given by the following file `CryptoPuzzle.py':

\begin{boxpy}\begin{python}
@\imp@

word1, word2, word3 = words = [w.lower() for w in data]
letters = set(alphabet_positions(word1 + word2 + word3))
n = len(word1)
assert len(word2) == n and len(word3) in {n, n + 1}

# x[i] is the value assigned to the ith letter (if present) of the alphabet
x = VarArray(size=26, dom=lambda i: range(10) if i in letters else None)

# auxiliary lists of variables associated with the three words
x1, x2, x3 = [[x[i] for i in reversed(alphabet_positions(word))] for word in words]

satisfy(
  # all letters must be assigned different values
  AllDifferent(x),

  # the most significant letter of each word cannot be equal to 0
  [x1[-1] != 0, x2[-1] != 0, x3[-1] != 0],
  
  # ensuring the crypto-arithmetic sum
  Sum((x1[i] + x2[i]) * 10 ** i for i in range(n))
  == Sum(x3[i] * 10 ** i for i in range(len(x3)))
)
\end{python}\end{boxpy}

The \p3 function \nn{alphabet\_positions}() returns a tuple composed with the position in the alphabet of all letters of a specified string.
For example, \nn{alphabet\_positions("about")} returns \nn{(0, 1, 14, 20, 19)}.
Note how two objects \nn{Sum} are involved.
Of course the crypto-arithmetic sum could also have been written as:

\begin{python}
   Sum((x1[i] + x2[i]) * 10 ** i for i in range(n))
     - Sum(x3[i] * 10 ** i for i in range(len(x3))) == 0
\end{python}

To properly understand the way the constraint \gb{sum} is constructed, note that executing:

\begin{command}
python CryptoPuzzle.py -data=[SEND,MORE,MONEY]
\end{command}

yields the following \x3 file:

\begin{xcsp}
<instance format="XCSP3" type="CSP">
  <variables>
    <array id="x" note="x[i] is the value assigned to the ith letter (if present) of the alphabet" size="[26]"> 0..9 </array>
  </variables>
  <constraints>
    <allDifferent note="all letters must be assigned different values">
      x[3..4] x[12..14] x[17..18] x[24]
    </allDifferent>
    <group note="the most significant letter of each word cannot be equal to 0">
      <intension> ne(%0,0) </intension>
      <args> x[18] </args>
      <args> x[12] </args>
      <args> x[12] </args>
    </group>
    <sum note="ensuring the crypto-arithmetic sum">
      <list> add(x[3],x[4]) add(x[13],x[17]) add(x[4],x[14]) add(x[18],x[12])
             x[24] x[4] x[13..14] x[12] </list>
      <coeffs> 1 10 100 1000 -1 -10 -100 -1000 -10000 </coeffs>
      <condition> (eq,0) </condition>
    </sum>
  </constraints>
</instance>
\end{xcsp}

Finally, it is possible to use dot product to build a weighted sum.
It means that it suffices to use the operator $*$ between two lists involving variables, integers or expressions to obtain an object \nn{Sum} as e.g., in
\nn{[u, v, w] * [2, 4, 3]} which represents $u*2 + v*4 + w*3$.
An illustration is given below.

\paragraph{Template Design.} \index{Problems!Template Design} From \href{https://www.csplib.org/Problems/prob002}{CSPLib}:
``This problem arises from a colour printing firm which produces a variety of products from thin board, including cartons for human and animal food and magazine inserts.
Food products, for example, are often marketed as a basic brand with several variations (typically flavours).
Packaging for such variations usually has the same overall design, in particular the same size and shape, but differs in a small proportion of the text displayed and/or in colour.
For instance, two variations of a cat food carton may differ only in that on one is printed 'Chicken Flavour' on a blue background whereas the other has 'Rabbit Flavour' printed on a green background.
A typical order is for a variety of quantities of several design variations.
Because each variation is identical in dimension, we know in advance exactly how many items can be printed on each mother sheet of board, whose dimensions are largely determined by the dimensions of the printing machinery.
Each mother sheet is printed from a template, consisting of a thin aluminium sheet on which the design for several of the variations is etched.
Each design of carton is made from an identically sized and shaped piece of board.
Several cartons can be printed on each mother sheet (in slots), and several different designs can be printed at once, on the same mother sheet.
The problem is to decide, firstly, how many distinct templates to produce, and secondly, which variations, and how many copies of each, to include on each template, in order to 
minimize the amount of waste produced.'' 
More details, and an example, are given on CSPLib.

\begin{figure}[h!]
\begin{center}
  \includegraphics[scale=1]{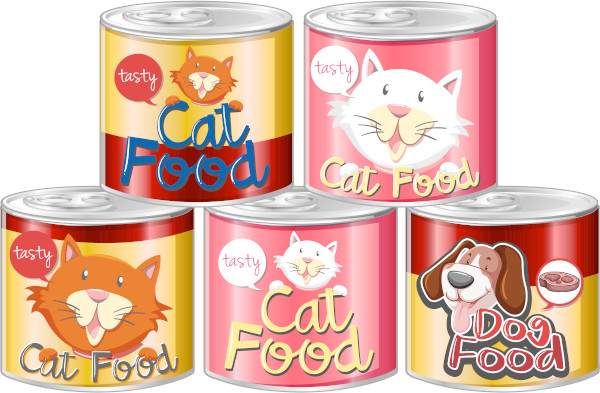}
\end{center}
\caption{Cat Food Cartons. \tiny{(image from \href{https://www.vecteezy.com/vector-art/294802-set-of-canned-pet-food}{www.vecteezy.com})} \label{fig:catFood}}
\end{figure}

An example of data is given by the following JSON file:

\begin{json}
{
  "nSlots": 9,
  "demands": [250, 255, 260, 500, 500, 800, 1100]
}
\end{json}

\bigskip
A \p3 model of this problem is given by the following file `TemplateDesign.py':

\begin{boxpy}\begin{python}
@\imp@
from math import ceil, floor

nSlots, demands = data
nTemplates = nVariations = len(demands)

def variation_interval(v):
    return range(ceil(demands[v] * 0.95), floor(demands[v] * 1.1) + 1)

# d[i][j] is the number of occurrences of the jth variation on the ith template
d = VarArray(size=[nTemplates, nVariations], dom=range(nSlots + 1))

# p[i] is the number of printings of the ith template
p = VarArray(size=nTemplates, dom=range(max(demands) + 1))

satisfy(
  # all slots of all templates are used
  [Sum(d[i]) == nSlots for i in range(nTemplates)],

  # respecting printing bounds for each variation
  [p * d[:, j] in variation_interval(j) for j in range(nVariations)]
)

minimize(
  # minimizing the number of used templates
  Sum(p[i] > 0 for i in range(nTemplates))   
)
\end{python}\end{boxpy}

The two arguments of \nn{satisfy}() correspond to two lists of \gb{sum} constraints; the second list involves dot products, each one built from the array (list) of variables $p$ and the jth column of the two-dimensional array (list) $d$, and imposed to belong to a certain interval.

\bigskip
It is also possible to use the function \gb{Hamming} that represents a constraint \gb{Sum} counting the number of equal terms between two lists of the same size. 

\paragraph{Hamming Vectors.} \index{Problems!Hamming Vectors}
The task is to build $n$ vectors of length $m$ with $d$ possible values, while ensuring that the Hamming distance (i.e., the number of distinct elements) is at least equal to $k$.  
For example, for $n=9$, $m=4$, $d=3$ and $k=3$, one possible solution is:
\begin{verbatim}
  (0, 0, 0, 0)  
  (0, 1, 1, 1)
  (0, 2, 2, 2)
  (1, 0, 1, 2)
  (1, 1, 2, 0)
  (1, 2, 0, 1)
  (2, 0, 2, 1)
  (2, 1, 0, 2)
\end{verbatim}

\begin{figure}[h!]
\begin{center}
  \includegraphics[scale=0.05]{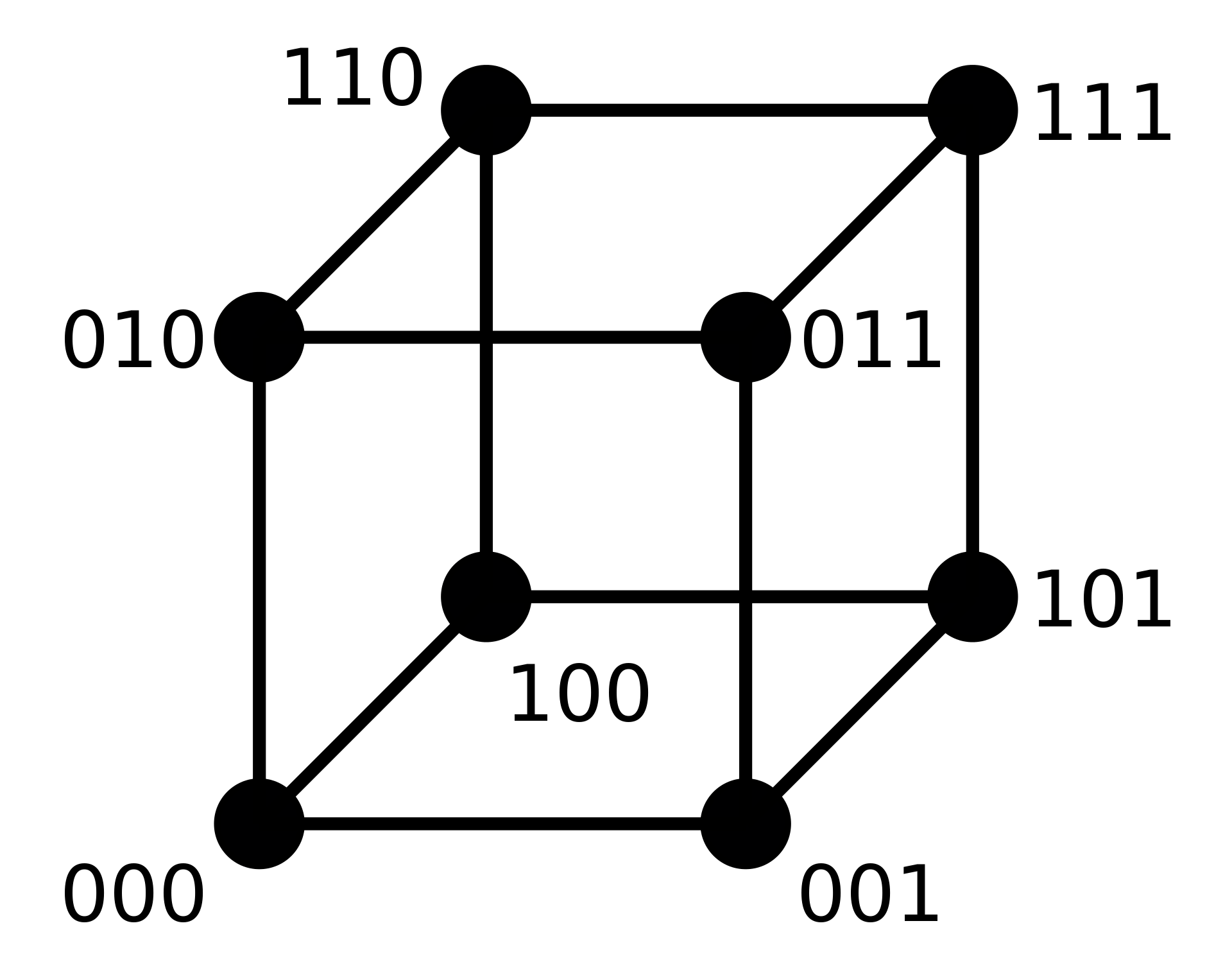}
\end{center}
\caption{3-bit binary cube for finding Hamming distance. \tiny{(image from \href{https://commons.wikimedia.org/wiki/File:Hamming_distance_3_bit_binary.svg}{commons.wikimedia.org})} \label{fig:hamming}}
\end{figure}

A \p3 model of this problem is given by the following file `HammingVectors.py':

\begin{boxpy}\begin{python}
@\imp@

n, m, d, k = data or (9, 4, 3, 3)

# x[i][j] is the jth value of the ith vector
x = VarArray(size=[n, m], dom=range(d))

satisfy(
  # ensuring a Hamming distance of at least 'k' between any two vectors
  [Hamming(row1, row2) >= k for row1, row2 in combinations(x, 2)],

  # tag(symmetry-breaking)
  LexIncreasing(x)
)
\end{python}\end{boxpy}

\section{Constraint \gb{count}}\label{sec:count}

The constraint \gb{count}\footnote{initially introduced in CHIP \cite{BC_chip} and Sicstus \cite{COC_open}}, imposes that the number of variables from a specified list of variables $X$ that take their values from a specified set $V$ respects a numerical condition $(\odot,k)$.
This constraint captures known constraints (usually) called \gb{atLeast}, \gb{atMost}, \gb{exactly} and \gb{among}.
To simplify, we assume for the semantics that $V$ is a set of integer values.

\begin{boxse}
\begin{semantics}
$\gb{count}(X,V,(\odot,k))$, with $X=\langle x_0,x_1,\ldots \rangle$, iff 
  $|\{i : 0 \leq i < |X| \land \va{x}_i \in V\}| \odot \va{k}$ 
\end{semantics}
\end{boxse}

In \p3, to post a constraint \gb{count}, we must call the function \nn{Count}() whose signature is:

\begin{python}
  def Count(term, *others, value=None, values=None):
\end{python}

The two parameters \texttt{term} and \texttt{others} are positional, and allow us to pass the main list of variables $X$ either in sequence (individually) or under the form of a list.
The two named parameters allow us to specify either a single value (unique target for counting) or a set of values.
Exactly one of these two parameters must be different from \nn{None}.
Assuming that $x$ is a list of variables, here are a few examples:

\begin{itemize}
\item \nn{Count(x, values=\{1, 5, 8\}) == k} \\stands for '$k$ variables from $x$ must take their values {\em among} those in $\{1, 5, 8\}$' 
\item \nn{Count(x, value=0) > 1}  \\stands for '{\em at least} 2 variables from $x$ must be assigned to the value 0'  
\item \nn{Count(x, value=1) <= k} \\stands for '{\em at most} $k$ variables from $x$ must be assigned to the value 1'  
\item \nn{Count(x, value=z) == k} \\stands for '{\em exactly} $k$ variables from $x$ must be assigned to the value $z$'  
\end{itemize}

\paragraph{Warehouse Location.} This problem was introduced in Section \ref{sec:warehouse}.
Here is a snippet of the \p3 model:   
\begin{python}
satisfy(
  # capacities of warehouses must not be exceeded
  [Count(w, value=j) <= capacities[j] for j in range(nWarehouses)],  

  ...
)
\end{python}

Each \gb{count} constraint imposes that the number of variables in $w$ that take the value $j$ is at most equal to the capacity of the jth warehouse.

\paragraph{Pizza Voucher Problem.} \index{Problems!Pizza Voucher} From the Intelligent Systems CMPT 417 course at Simon Fraser University.
``The problem arises in the University College Cork student dorms.
There is a large order of pizzas for a party, and many of the students have vouchers for acquiring discounts in purchasing pizzas.
A voucher is a pair of numbers e.g. $(2,4)$, which means if you pay for $2$ pizzas then you can obtain for free up to $4$ pizzas as long as they each cost no more than the cheapest of the $2$ pizzas you paid for.
Similarly a voucher $(3,2)$ means that if you pay for $3$ pizzas you can get up to $2$ pizzas for free as long as they each cost no more than the cheapest of the $3$ pizzas you paid for.
The aim is to obtain all the ordered pizzas for the least possible cost.
Note that not all vouchers need to be used.'' 

\begin{figure}[h!]
\begin{center}
  \includegraphics[scale=0.25]{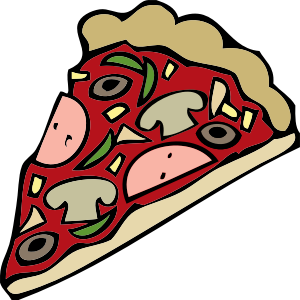}
\end{center}
\caption{A Nice Pizza Slice. \tiny{(image from \href{https://freesvg.org/pizza-slice-vector-image}{freesvg.org})} \label{fig:pizza}}
\end{figure}

An example of data is given by the following JSON file:

\begin{json}
{
  "pizzaPrices": [50, 60, 90, 70, 80, 100, 20, 30, 40, 10],
  "vouchers":[
    {"payPart":1,"freePart":2},
    {"payPart":2,"freePart":3},
    ...
  ]
}
\end{json}

A \p3 model of this problem is given by the following file `PizzaVoucher.py':

\begin{boxpy}\begin{python}
@\imp@

prices, vouchers = data
nPizzas, nVouchers = len(prices), len(vouchers)

# v[i] is the voucher used for the ith pizza. 0 means that no voucher is used.
# A negative (resp., positive) value i means that the ith pizza contributes
# to the pay (resp., free) part of voucher |i|.
v = VarArray(size=nPizzas, dom=range(-nVouchers, nVouchers + 1))

# p[i] is the number of paid pizzas wrt the ith voucher
p = VarArray(size=nVouchers, dom=lambda i: {0, vouchers[i].payPart})

# f[i] is the number of free pizzas wrt the ith voucher
f = VarArray(size=nVouchers, dom=lambda i: range(vouchers[i].freePart + 1))

satisfy(
  # counting paid pizzas
  [Count(v, value=-i - 1) == p[i] for i in range(nVouchers)],

  # counting free pizzas
  [Count(v, value=i + 1) == f[i] for i in range(nVouchers)],
  
  # a voucher, if used, must contribute to have at least one free pizza.
  [(f[i] == 0) == (p[i] != vouchers[i].payPart) for i in range(nVouchers)],
  
  # a free pizza must be cheaper than any pizza paid wrt the used voucher
  [
    If(
      v[i] < 0,
      Then= v[i] != -v[j]
    ) for i in range(nPizzas) for j in range(nPizzas) 
         if i != j and prices[i] < prices[j]
  ]  
)

minimize(
  # minimizing summed up costs of pizzas
  Sum((v[i] <= 0) * prices[i] for i in range(nPizzas))    
)
\end{python}\end{boxpy}
    

It is really frequent to perform counting when modeling, and there are some specific situations where it is possible to use some derived functions:

\begin{itemize}
\item \gb{Exist}: corresponds to a constraint \gb{Count} being satisfied iff at least one term (variable or tree expression) is true (i.e., equal to 1); see an illustration Page \pageref{pb:steelMillSlab} for Steel Mill Slab Problem
\item \gb{NotExist}: corresponds to a constraint \gb{Count} being satisfied iff no term (variable or tree expression) is true (i.e., equal to 1); see War or Peace Problem below
\item \gb{ExactlyOne}: corresponds to a constraint \gb{Count} being satisfied iff exactly one term (variable or tree expression) is true (i.e., equal to 1); see an illustration Page \pageref{pb:amaze} for Amaze Problem
\item \gb{AtLeastOne}: corresponds to a constraint \gb{Count} being satisfied iff at least one term (variable or tree expression) is true (i.e., equal to 1); note that this is an alias for \gb{Exist}
\item \gb{AtMostOne}: corresponds to a constraint \gb{Count} being satisfied iff at most one term (variable or tree expression) is true (i.e., equal to 1)
\item \gb{AllHold}: corresponds to a constraint \gb{Count} being satisfied iff every term (variable or tree expression) is true (i.e., equal to 1)  
\end{itemize}

\paragraph{War or Peace.}\index{Problems!War or Peace}
There are $n$ countries such that:
\begin{itemize}
  \item each pair of two countries is either at war or has a peace treaty,
  \item each pair of two countries that has a common enemy has a peace treaty.
\end{itemize}
What is the minimum number of peace treaties?

\begin{figure}[h!]
\begin{center}
  \includegraphics[scale=0.3]{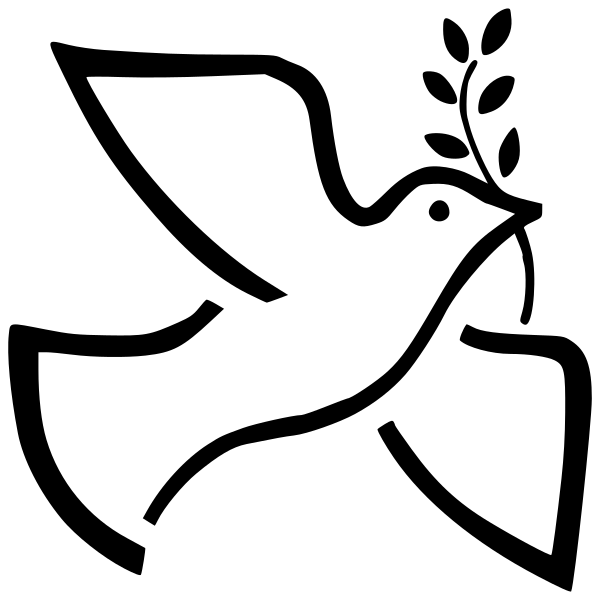}
\end{center}
\caption{Make Peace not War. \tiny{(image from \href{https://freesvg.org/dove-with-branch}{freesvg.org})} \label{fig:dove}}
\end{figure}

\bigskip
A \p3 model of this problem is given by the following file `WarOrPeace.py':

\begin{boxpy}\begin{python}
@\imp@

n = data   # number of countries
WAR, PEACE = 0, 1

# x[i][j] is 1 iff countries i and j have a peace treaty
x = VarArray(size=[n, n], dom=lambda i, j: {WAR, PEACE} if i < j else None)


satisfy(
  If(
    x[i][j] != PEACE,
    Then=NotExist(
      both(
        x[min(i, k)][max(i, k)] == WAR,
        x[min(j, k)][max(j, k)] == WAR
      ) for k in range(n) if different_values(i, j, k)
    )
  ) for i, j in combinations(n, 2)
)

minimize(
  # minimizing the number of peace treaties
  Sum(x)
)
\end{python}\end{boxpy}

\section{Constraint \gb{nValues}}\label{sec:nValues}

The constraint \gb{nValues} \cite{BHHKW_nvalue}, ensures that the number of distinct values taken by the variables of a specified list $X$ respects a numerical condition $(\odot,k)$.
A variant, called \gb{nValuesExcept} \cite{BHHKW_nvalue} discards some specified values of a set $E$ (often, the single value 0).

\begin{boxse}
\begin{semantics}
$\gb{nValues}(X,E,(\odot,k))$, with $X=\langle x_0,x_1,\ldots \rangle$, iff 
  $|\{\va{x}_i : 0 \leq i < |X|\} \setminus E| \odot \va{k}$ 
$\gb{nValues}(X,(\odot,k))$ iff $\gb{nValues}(X,\emptyset,(\odot,k))$
\end{semantics}
\end{boxse}


In \p3, to post a constraint \gb{nValues}, we must call the function \nn{NValues}() whose signature is:

\begin{python}
  def NValues(term, *others, excepting=None): 
\end{python}

The two parameters \texttt{term} and \texttt{others} are positional, and allow us to pass the variables either in sequence (individually) or under the form of a list.
The optional named parameter  \texttt{excepting} allows us to specify a value (integer) or a list of values.
The object obtained when calling \nn{NValues}() must be restricted by a condition (typically, defined by a relational operator and a limit).

\paragraph{Board Coloration.} This problem was introduced in Section \ref{sec:boardColoration}.
The constraint \gb{nValues} was introduced for capturing \gb{notAllEqual}.


\paragraph{RLFAP.} This problem was introduced in Section \ref{sec:objectives}.
The function \nn{NValues}() was used to specify the objective of one variant of the problem. 

\section{Constraint \gb{cardinality}}

The constraint \gb{cardinality}, also called \gb{globalCardinality} or \gb{gcc} in the literature, see \cite{R_gcc, H_integrated}, ensures that the number of occurrences of each value in a specified set $V$, taken by the variables of a specified list $X$, is equal to a specified value (or variable), or belongs to a specified interval (information given by a set $O$).
A Boolean option \texttt{closed}, when set to \nn{true}, means that all variables of $X$ must be assigned a value from $V$.

For simplicity, for the semantics below, we assume that $V$ only contains values and $O$ only contains variables.
Note that $^{\nm{cl}}$ means that \texttt{closed} is \nn{true}.

\begin{boxse}
\begin{semantics}
$\gb{cardinality}(X,V,O)$, with $X=\langle x_0,x_1,\ldots \rangle$, $V=\langle v_0,v_1,\ldots \rangle$, $O=\langle o_0, o_1,\ldots \rangle$,
  iff $\forall j : 0 \leq j < |V|, |\{i : 0 \leq i < |X| \land \va{x}_i =v_j\}| = \va{o}_j$
$\gb{cardinality}^{\nm{cl}}(X,V,O)$ iff $\gb{cardinality}(X,V,O) \land \forall i : 0 \leq i < |X|, \va{x}_i \in V$

$\gbc{Prerequisite}: |X| \geq 2 \land |V| = |O| \geq 1$
\end{semantics}
\end{boxse}

The form of the constraint obtained by only considering variables in the sets $X$, $V$ and $O$ is called \gb{distribute} in \mzinc.
In that case, for the semantics, we must additionally guarantee:
\begin{quote}
$\forall (i,j) : 0 \leq i < j < |V|, \va{v}_i \neq \va{v}_j$.
\end{quote}

In \p3, to post a constraint \gb{cardinality}, we must call the function \nn{Cardinality}() whose signature is:

\begin{python}
  def Cardinality(term, *others, @{occurrences}, closed=False):
\end{python}

The two parameters \texttt{term} and \texttt{others} are positional, and allow us to pass the variables either in sequence (individually) or under the form of a list.
The value of the required named parameter  \texttt{occurrences} must be a dictionary: each entry $(k,v)$ in the dictionary means that the number of occurrences of $k$ is given by $v$.
The optional named parameter\texttt{closed} , when set to \nn{true}, means that all variables specified by the two positional parameters must be assigned a value that corresponds to a key in the dictionary.

\paragraph{Labeled Dice.} \index{Problems!Labeled Dice} From \href{https://jimorlin.wordpress.com/2009/02/17/colored-letters-labeled-dice-a-logic-puzzle/}{Jim Orlin's Blog}:
``There are 13 words as follows: buoy, cave, celt, flub, fork, hemp, judy, junk, limn, quip, swag, visa, wish.
There are 24 different letters that appear in the 13 words.
The question is: can one assign the 24 letters to 4 different cubes so that the four letters of each word appear on different cubes.
There is one letter from each word on each cube.
The puzzle was created by Humphrey Dudley''

A \p3 model of this problem is given by the following file `LabeledDice.py':
\begin{boxpy}\begin{python}
@\imp@

words = ["buoy", "cave", "celt", "flub", "fork", "hemp",
         "judy", "junk", "limn", "quip", "swag", "visa"]

# x[i] is the cube where the ith letter of the alphabet is put
x = VarArray(size=26, dom=lambda i: range(1, 5)
       if i in alphabet_positions("".join(words)) else None)

satisfy(
  # the four letters of each word appears on different cubes
  [AllDifferent(x[i] for i in alphabet_positions(w)) for w in words],

  # each cube is assigned 6 letters
  Cardinality(x, occurrences={i: 6 for i in range(1, 5)})
)
\end{python}\end{boxpy}

The \p3 function \nn{alphabet\_positions}() returns a tuple composed with the position in the alphabet of all letters of a specified string.
For example, \nn{alphabet\_positions("about")} returns \nn{(0, 1, 14, 20, 19)}.
The posted \gb{cardinality} constraint ensures that we have 6 letters per cube (using an index $i$ for cubes, ranging from 1 to 4).

\paragraph{Magic Sequence.} This problem was introduced in Section \ref{sec:magicSequence}.
Here is a snippet of the \p3 model:  

\begin{python}
# x[i] is the ith value of the sequence
x = VarArray(size=n, dom=range(n))

satisfy(
  # each value i occurs exactly x[i] times in the sequence
  Cardinality(x, occurrences={i: x[i] for i in range(n)}),

  ...
)
\end{python}

Here, one can see that variables are used for counting the number of occurrences, and besides, this is a special case where these variables are from the main list (first parameter $x$).

\paragraph{Sports Scheduling.} \index{Problems!Sports Scheduling} From CSPLib:
``The problem is to schedule a tournament of $n$ teams over $n-1$ weeks, with each week divided into $n/2$ periods, and each period divided into two slots indicating the two involved teams (for example, one playing at home, and the other away).
A tournament must satisfy the following three conditions:
\begin{itemize}
\item every team plays every other team.
\item every team plays once a week;
\item every team plays at most twice in the same period over the tournament;
\end{itemize}
''

\begin{figure}[h!]
\begin{center}
  \includegraphics[scale=0.3]{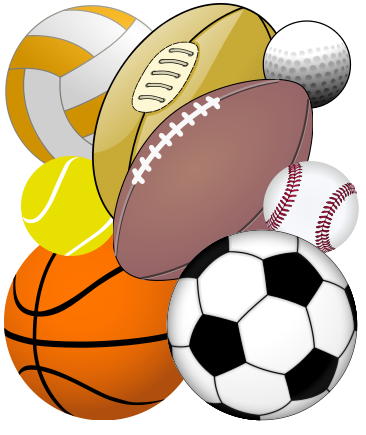}
\end{center}
\caption{Sports Scheduling. \tiny{(image from \href{https://commons.wikimedia.org/wiki/File:Sports_portal_bar_icon.png}{commons.wikimedia.org})} \label{fig:sportsScheduling}}
\end{figure}

A \p3 model of this problem is given by the following file `SportsScheduling.py':
\begin{boxpy}\begin{python}
@\imp@

nTeams = data or 8
nWeeks, nPeriods, nMatches = nTeams - 1, nTeams // 2, (nTeams - 1) * nTeams // 2


def match_number(t1, t2):
    return nMatches - ((nTeams - t1) * (nTeams - t1 - 1)) // 2 + (t2 - t1 - 1)


T = {(t1, t2, match_number(t1, t2)) for t1, t2 in combinations(range(nTeams), 2)}

# m[w][p] is the number of the match at week w and period p
m = VarArray(size=[nWeeks, nPeriods], dom=range(nMatches))

# x[w][p] is the first team for the match at week w and period p
x = VarArray(size=[nWeeks, nPeriods], dom=range(nTeams))

# y[w][p] is the second team for the match at week w and period p
y = VarArray(size=[nWeeks, nPeriods], dom=range(nTeams))

satisfy(
  # all matches are different (no team can play twice against another team)
  AllDifferent(m),

  # linking variables through ternary table constraints
  [(x[w][p], y[w][p], m[w][p]) in T for w in range(nWeeks) for p in range(nPeriods)],
  
  # each week, all teams are different (each team plays each week)
  [AllDifferent(x[w] + y[w]) for w in range(nWeeks)],
  
  # each team plays at most two times in each period
  [
    Cardinality(
      within=x[:,p] + y[:,p],
      occurrences={t: range(1, 3) for t in range(nTeams)}
    ) for p in range(nPeriods)
  ]
)
\end{python}\end{boxpy}

Here, we can see that the interval $1..2$ (given by \nn{range(1,3)}) is used to control the number of occurrences of each team in each period, when posting \gb{cardinality} constraints. 
Note that we could add some symmetry breaking constraints to the model.

\section{Constraint \gb{maximum}}\label{sec:maximum}

The constraint \gb{maximum} ensures that the maximum value among those assigned to the variables of a specified list $X$ respects a numerical condition $(\odot,k)$.

\begin{boxse}
\begin{semantics}
$\gb{maximum}(X,(\odot,k))$, with $X=\langle x_0,x_1,\ldots\rangle$, iff 
  $\max\{\va{x}_i : 0 \leq i < |X|\} \odot \va{k}$
\end{semantics}
\end{boxse}

In \p3, to post a constraint \gb{maximum}, we must call the function \nn{Maximum}() whose signature is:

\begin{python}
  def Maximum(term, *others)
\end{python}

The two parameters \texttt{term} and \texttt{others} are positional, and allow us to pass the variables either in sequence (individually) or under the form of a list.
The object obtained when calling \nn{Maximum}() must be restricted by a condition (typically, defined by a relational operator and a limit).

\paragraph{Open Stacks.} \index{Problems!Open Stacks} From Steven Prestwich:
``A manufacturer has a number of orders from customers to satisfy.
Each order is for a number of different products, and only one product can be made at a time.
Once a customer's order is started a stack is created for that customer.
When all the products that a customer requires have been made the order is sent to the customer, so that the stack is closed.
Because of limited space in the production area, the number of stacks that are simultaneously open should be minimized.''

An example of data is given by the following JSON file:

\begin{json}
{
  "orders": [
    [0,1,1,1,1,1,0,1,1,1,1,1,1,1,1,1,1,0,1,1,1,1,0,1,1,0,1,1,1,1,1,1,1,1,0], 
    [1,1,1,1,1,1,1,1,1,1,1,1,0,1,1,1,1,1,1,1,0,1,1,1,1,1,1,1,1,1,1,1,1,1,1], 
    ...
  ]
}
\end{json}

Each row of \nn{orders} corresponds to a customer order indicating with 0 or 1 if the jth product is needed. 
A \p3 model of this problem is given by the following file `OpenStacks.py':

\begin{boxpy}\begin{python}
@\imp@

orders = data
n, m = len(orders), len(orders[0])  # n orders (customers), m possible products

def table(t):
   return {(ANY, te, 0) for te in range(t)} |
          {(ts, ANY, 0) for ts in range(t + 1, m)} |
          {(ts, te, 1) for ts in range(t + 1) for te in range(t, m)}
    
# p[j] is the period (time) of the jth product
p = VarArray(size=m, dom=range(m))

# s[i] is the starting time of the ith stack
s = VarArray(size=n, dom=range(m))

# e[i] is the ending time of the ith stack
e = VarArray(size=n, dom=range(m))

# o[i][t] is 1 iff the ith stack is open at time t
o = VarArray(size=[n, m], dom={0, 1})

satisfy(
  # all products are scheduled at different times
  AllDifferent(p),

  # computing starting times of stacks
  [Minimum(p[j] for j in range(m) if orders[i][j]) == s[i] for i in range(n)],
  
  # computing ending times of stacks
  [Maximum(p[j] for j in range(m) if orders[i][j]) == e[i] for i in range(n)],
  
  # inferring when stacks are open
  [(s[i], e[i], o[i][t]) in table(t) for i in range(n) for t in range(m)],
)

minimize(
  # minimizing the number of stacks that are simultaneously open
  Maximum(Sum(o[:, t]) for t in range(m))
)
\end{python}\end{boxpy}

Note that each list of variables is given to \nn{Maximum}() under the form of a comprehension list (generator).
The \p3 function \nn{Maximum}() is also used for building the expression to be minimized.

\section{Constraint \gb{maximumArg}}\label{sec:maximumArg}

A form related to \gb{maximum} is the constraint \gb{maximumArg}, sometimes called \gb{arg\_max}, which ensures that the index of a maximum variable (i.e., a variable with a maximal value) in a list respects a numerical condition.
The semantics is:

\begin{boxse}
\begin{semantics}
$\gb{maximum}(X,i)$, with $X=\langle x_0,x_1,\ldots \rangle$, iff 
  $\va{i} \in \{j : 0 \leq j < |X| \land \va{x}_j = \max\{\va{x}_k : 0 \leq k < |X|\}\}$
$\gb{maximumArg}(X,(\odot,k))$ iff $\exists i : \gb{maximum}(X,i) \land i \odot \va{k}$
\end{semantics}
\end{boxse}

In \p3, to post a constraint \gb{maximumArg}, we must call the function \nn{MaximumArg}() whose signature is:

\begin{python}
  def MaximumArg(term, *others, rank=None)
\end{python}

The two parameters \texttt{term} and \texttt{others} are positional, and allow us to pass the variables either in sequence (individually) or under the form of a list.
The optional parameter \texttt{rank} can be \nn{None} or take a value among \nn{TypeRank.FIRST}, \nn{TypeRank.ANY}, \nn{TypeRank.LAST}. 
The object obtained when calling \nn{MaximumArg}() must be restricted by a condition (typically, defined by a relational operator and a limit).

\section{Constraint \gb{minimum}}\label{sec:minimum}

The constraint \gb{minimum} ensures that the minimum value among those assigned to the variables of a specified list $X$ respects a numerical condition $(\odot,k)$.

\begin{boxse}
\begin{semantics}
$\gb{minimum}(X,(\odot,k))$, with $X=\langle x_0,x_1,\ldots\rangle$, iff 
  $\min\{\va{x}_i : 0 \leq i < |X|\} \odot \va{k}$
\end{semantics}
\end{boxse}

In \p3, to post a constraint \gb{minimum}, we must call the function \nn{Minimum}() whose signature is:

\begin{python}
  def Minimum(term, *others)
\end{python}

The two parameters \texttt{term} and \texttt{others} are positional, and allow us to pass the variables either in sequence (individually) or under the form of a list.
The object obtained when calling \nn{Minimum}() must be restricted by a condition (typically, defined by a relational operator and a limit).

\paragraph{Open Stacks.} See the model introduced in the previous section.

\section{Constraint \gb{minimumArg}}\label{sec:minimumArg}

A form related to \gb{minimum} is the constraint \gb{minimumArg}, sometimes called \gb{arg\_min}, which ensures that the index of a minimum variable (i.e., a variable with a minimal value) in a list respects a numerical condition.
The semantics is:

\begin{boxse}
\begin{semantics}
$\gb{minimum}(X,i)$, with $X=\langle x_0,x_1,\ldots \rangle$, iff 
  $\va{i} \in \{j : 0 \leq j < |X| \land \va{x}_j = \min\{\va{x}_k : 0 \leq k < |X|\}\}$
$\gb{minimumArg}(X,(\odot,k))$ iff $\exists i : \gb{minimum}(X,i) \land i \odot \va{k}$
\end{semantics}
\end{boxse}

In \p3, to post a constraint \gb{minimumArg}, we must call the function \nn{MinimumArg}() whose signature is:

\begin{python}
  def MinimumArg(term, *others, rank=None)
\end{python}

The two parameters \texttt{term} and \texttt{others} are positional, and allow us to pass the variables either in sequence (individually) or under the form of a list.
The optional parameter \texttt{rank} can be \nn{None} or take a value among \nn{TypeRank.FIRST}, \nn{TypeRank.ANY}, \nn{TypeRank.LAST}. 
The object obtained when calling \nn{MinimumArg}() must be restricted by a condition (typically, defined by a relational operator and a limit).

\section{Constraint \gb{element}}

The constraint \gb{element} \cite{HC_generality} ensures that the element of a specified list $X$ at a specified index $i$ has a specified value $v$.
The semantics is \nn{X[i] = v}, or equivalently:

\begin{boxse}
\begin{semantics}
$\gb{element}(X,i,v)$, with $X=\langle x_0,x_1,\ldots \rangle$, iff 
  $\va{x}_{\va{i}} = \va{v}$ 
\end{semantics}
\end{boxse}

It is important to note that $i$ must be an integer variable (and not a constant).
In Python, to post an \gb{element} constraint, we use the facilities offered by the language, meaning that we can write expressions involving relational and indexing ([]) operators. 

There are three variants of \gb{element}:
\begin{itemize}
\item variant 1: $X$ is a list of variables, $i$ is an integer variable and $v$ is an integer variable
\item variant 2: $X$ is a list of variables, $i$ is an integer variable and $v$ is an integer (constant)
\item variant 3: $X$ is a list of integers, $i$ is an integer variable and $v$ is an integer variable
\end{itemize}

Although variant 3 can be reformulated as a binary extensional constraint,
it is often used when modeling.

\paragraph{The Sandwich Case.} \index{Problems!Sandwich} From beCool (UCLouvain):
Someone in the university ate Alice's sandwich at the cafeteria.
We want to find out who the culprit is.
The witnesses are unanimous about the following facts:
\begin{enumerate}
\item Three persons were in the cafeteria at the time of the crime: Alice, Bob, and Sascha.
\item The culprit likes Alice.
\item The culprit is taller than Alice.
\item Nobody is taller than himself.
\item If A is taller than B, then B is not taller than A.
\item Bob likes no one that Alice likes.
\item Alice likes everybody except Bob.
\item Sascha likes everyone that Alice likes.
\item Nobody likes everyone.
\end{enumerate}

This is a single problem (no external data is required).
A \p3 model of this problem is given by the following file `Sandwich.py':

\begin{boxpy}\begin{python}
@\imp@

alice, bob, sascha = persons = 0, 1, 2

# culprit is among alice (0), bob (1) and sascha (2)
culprit = Var(persons)

# liking[i][j] is 1 iff the ith guy likes the jth guy
liking = VarArray(size=[3, 3], dom={0, 1})

# taller[i][j] is 1 iff the ith guy is taller than the jth guy
taller = VarArray(size=[3, 3], dom={0, 1})

satisfy(
  # the culprit likes Alice
  liking[culprit][alice] == 1,

  # the culprit is taller than Alice
  taller[culprit][alice] == 1,
  
  # nobody is taller than himself
  [taller[p][p] == 0 for p in persons],
  
  # the ith guy is taller than the jth guy iff the reverse is not true
  [taller[p1][p2] != taller[p2][p1] for p1 in persons for p2 in persons if p1 != p2],
  
  # Bob likes no one that Alice likes
  [If(liking[alice][p], Then=~liking[bob][p]) for p in persons],
  
  # Alice likes everybody except Bob
  [liking[alice][p] == 1 for p in persons if p != bob],
  
  # Sascha likes everyone that Alice likes
  [If(liking[alice][p], Then=liking[sascha][p]) for p in persons],
  
  # nobody likes everyone
  [Count(liking[p], value=0) >= 1 for p in persons]
)
\end{python}\end{boxpy}

Variant 2 of \gb{element} is illustrated by:
\begin{python}
  liking[culprit][alice] == 1,
\end{python}
as it basically encodes ``the variable at index \nn{culprit} in the column 0 (alice) of the 2-dimensional array of variables \nn{liking} must be equal to 1''.

\paragraph{Warehouse Location.} This problem was introduced in Section \ref{sec:warehouse}.
Here is a snippet of the \p3 model:   
\begin{python}
satisfy(
  ...

  # computing the cost of supplying the ith store
  [costs[i][w[i]] == c[i] for i in range(nStores)]
)
\end{python}

Variant 3 of \gb{element} is illustrated by:
\begin{python}
  costs[i][w[i]] == c[i]
\end{python}
as it basically encodes ``the variable at index \nn{w[i]} in the ith row  of the 2-dimensional array of integers \nn{costs} must be equal to c[i]''.

\bigskip
Interestingly, it is also possible to use a variant of \gb{element} on matrices, i.e., by using two indexes given by integer variables.
The semantics is \nn{M[i][j] = v}, or equivalently:
\begin{boxse}
\begin{semantics}
$\gb{element}(\ns{M},\langle i,j \rangle,v)$, with $\ns{M}=[ \langle x_{1,1}, x_{1,2},\ldots,x_{1,m}\rangle, \langle x_{2,1}, x_{2,2},\ldots,x_{2,m}\rangle, \ldots ]$, iff 
  $\va{x}_{\va{i},\va{j}} = \va{v}$ 
\end{semantics}
\end{boxse}

It is important to note that $i$ and $j$ must be two integer variables (and not constants).
In Python, to post an \gb{element} constraint on matrices, we use the facilities offered by the language, meaning that we can write expressions involving relational and indexing ([]) operators. 

There are three variants of \gb{element} on matrices:
\begin{itemize}
\item variant 1: $M$ is a matrix of variables, $i$ and $j$ are integer variables and $v$ is an integer variable
\item variant 2: $M$ is a matrix of variables, $i$ and $j$ are integer variables and $v$ is an integer (constant)
\item variant 3: $M$ is a matrix of integers, $i$ and $j$ are integer variables and $v$ is an integer variable
\end{itemize}

Although variant 3 can be reformulated as a ternary extensional constraint, it is often used when modeling.  

\paragraph{Quasigroup Existence.} \index{Problems!Quasigroup Existence} \label{sec:quasigroup}  From \href{https://www.csplib.org/Problems/prob003/}{CSPLib}:
``A quasigroup of order $n$ is a $n \times n$ multiplication table in which each element occurs once in every row and column (i.e., is a Latin square), while satisfying some specific properties.
Hence, the result $a*b$ of applying the multiplication operator $*$ on $a$ (left operand) and $b$ (right operand) is given by the value in the table at row $a$ and column $b$.
Classical variants of quasigroup existence correspond to taking into account the following properties:
  \begin{itemize}
  \item QG3: quasigroups for which $(a*b)*(b*a)=a$
  \item QG4: quasigroups for which $(b*a)*(a*b)=a$
  \item QG5: quasigroups for which $((b*a)*b)*b=a$
  \item QG6: quasigroups for which $(a*b)*b=a*(a*b)$
  \item QG7: quasigroups for which $(b*a)*b=a*(b*a)$
  \end{itemize}
For each of these problems, we may additionally demand that the quasigroup is idempotent. That is, $a*a=a$ for every element $a$.''

A \p3 model of this problem is given by the following file `Quasigroup.py':

\begin{boxpy}\begin{python}
@\imp@

n = data

# x[i][j] is the value at row i and column j of the quasi-group
x = VarArray(size=[n, n], dom=range(n))

satisfy(
  # ensuring a Latin square
  AllDifferent(x, matrix=True),

  # ensuring idempotence  tag(idempotence)
  [x[i][i] == i for i in range(n)]
)

if variant("v3"):
   satisfy(
     x[x[i][j], x[j][i]] == i for i in range(n) for j in range(n)
   )
elif variant("v4"):
   satisfy(
     x[x[j][i], x[i][j]] == i for i in range(n) for j in range(n)
   )
elif variant("v5"):
   satisfy(
     x[x[x[j][i], j], j] == i for i in range(n) for j in range(n)
   )
elif variant("v6"):
   satisfy(
     x[x[i][j], j] == x[i, x[i][j]] for i in range(n) for j in range(n)
  )
elif variant("v7"):
   satisfy(
     x[x[j][i], j] == x[i, x[j][i]] for i in range(n) for j in range(n)
   )    
\end{python}\end{boxpy}

Variant 2 of \gb{element} on matrices is illustrated by:
\begin{python}
  x[x[i][j], x[j][i]] == i
\end{python}
as it basically encodes ``the variable in the matrix $x$ at row index \nn{x[i][j]} (a variable) and column index \nn{x[j][i]} (a variable) must be equal to the integer $i$''.
Note how we can write complex operations involving several (partial forms of) \gb{element} constraints; when compiling, auxiliary variables may possibly be introduced (the interested reader can look at the generated \x3 files).

\paragraph{Traveling Salesman Problem (TSP).} \index{Problems!Traveling Salesman Problem (TSP)} \label{sec:tsp}  From \href{https://en.wikipedia.org/wiki/Travelling_salesman_problem}{Wikipedia}:
``Given a list of cities and the distances between each pair of cities, what is the shortest possible route that visits each city and returns to the origin city?''

An example of data is given by the following JSON file: 

\begin{json}
{
  "distances": [
    [0, 5, 6, 6, 6],
    [5, 0, 9, 8, 4],
    [6, 9, 0, 1, 7],
    [6, 8, 1, 0, 6],
    [6, 4, 7, 6, 0]
  ]
}
\end{json}

\begin{figure}[h!]
\begin{center}
  \includegraphics[scale=0.26]{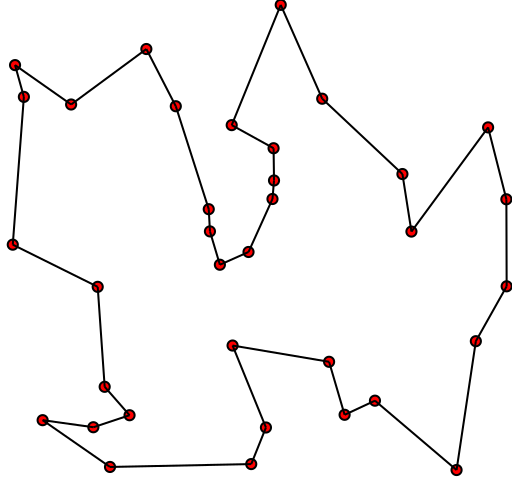}
\end{center}
\caption{A Solution for a TSP instance. \tiny{(image from \href{https://commons.wikimedia.org/wiki/File:GLPK_solution_of_a_travelling_salesman_problem.svg}{commons.wikimedia.org})} \label{fig:tsp}}
\end{figure}

A \p3 model of this problem is given by the following file `TravelingSalesman.py':

\begin{boxpy}\begin{python}
@\imp@

distances = data
nCities = len(distances)

# c[i] is the ith city of the tour
c = VarArray(size=nCities, dom=range(nCities))

# d[i] is the distance between the cities i and i+1 chosen in the tour
d = VarArray(size=nCities, dom=distances)

satisfy(
  # Visiting each city only once
  AllDifferent(c)
)

if not variant():
   satisfy(
     # computing the distance between any two successive cities in the tour
     distances[c[i]][c[i + 1]] == d[i] for i in range(nCities)
   )

elif variant("table"):
   T = {(i, j, distances[i][j]) for i in range(nCities) for j in range(nCities)}

   satisfy(
     # computing the distance between any two successive cities in the tour
     (c[i], c[i + 1], d[i]) in T for i in range(nCities)
   )

minimize(
  # minimizing the traveled distance
  Sum(d)
)
\end{python}\end{boxpy}

First, note that auto-adjustment of array indexing is used here, as \verb!c[i + 1]! is interpreted in \p3 as \verb!c[(i + 1) % nCities]! (while a warning message is displayed).
Variant 3 of \gb{element} on matrices is illustrated by:
\begin{python}
  distances[c[i]][c[(i + 1) % nCities]] == d[i]
\end{python}
as it basically encodes ``the integer in the matrix \nn{distances} at row index \nn{c[i]} (a variable) and column index \nn{c[(i + 1) \% nCities]} (a variable) must be equal to the variable \nn{d[i]}''.
Variant ``table'' shows which ternary table constraints are equivalent to the \gb{element} constraints on matrices (of integers).  
Note that writing \verb!dom=distances! is equivalent (and more compact) to writing \verb!dom={v for row in distances for v in row}!.

\paragraph{Correctly handling indexing of constraints  \gb{element}.} \label{sec:indexElt}

There exist certain situations where the index used (a variable or a constraining expression) when posting a constraint \gb{element} does not exactly correspond to the range of possible indexes of the indexed array.
We illustrate several such representative situations, while introducing the option \verb!-force_element_index!.

\medskip
A first illustrative \p3 model is given by the following file `TestElementIdx1.py':

\begin{boxpy}\begin{python}
@\imp@

x = VarArray(size=10, dom=range(6))

y = Var(dom=range(-1, 10))

if not variant():
   satisfy(
      x[y] == 2  # needs -force_element_index
   )
elif variant("aux1"):
   satisfy(
      (aux := Var()) == y,
      x[aux] == 2
   )
elif variant("aux2"):
   aux = Var()

   satisfy(
      aux == y,
      x[aux] == 2
   )
\end{python}\end{boxpy}

The important point is that the variable $y$ is used as an index of the 1-dimensional variable array $x$.
The problem is that the indexes of $x$ range from 0 to 9 while the domain of $y$ ranges from -1 to 9.
If $y$ is set to -1, the expression $x[y]$ is not well defined.

If we execute:
\begin{verbatim}
    python TestElementIdx1.py
\end{verbatim}
we obtain an error message:
\begin{verbatim}
    ERROR:  Problem with an indexing expression ; maybe use -force_element_index,
    or replace array[index] by array[aux] after posing ((aux := Var()) == index)
\end{verbatim}

If we execute:
\begin{verbatim}
    python TestElementIdx1.py -force_element_index
\end{verbatim}

we obtain:

\begin{xcsp}
<instance format="XCSP3" type="CSP">
  <variables>
    <array id="x" size="[10]"> 0..5 </array>
    <var id="y"> -1..9 </var>
    <array id="aux_gb" note="aux_gb[i] is the ith auxiliary variable having been automatically introduced" size="[1]"> 0..9 </array>
  </variables>
  <constraints>
    <intension> eq(y,aux_gb[0]) </intension>
    <element>
      <list> x[] </list>
      <index> aux_gb[0] </index>
      <value> 2 </value>
    </element>
  </constraints>
</instance>
\end{xcsp}

which is fine.

If we execute:
\begin{verbatim}
    python TestElementIdx1.py -variant=aux1
\end{verbatim}

or 
\begin{verbatim}
    python TestElementIdx1.py -variant=aux2
\end{verbatim}

we obtain the same \x3 instance (and note that we don't need to use \verb!-force_element_index!).
When \verb!Var()! is used, it introduces a variable with domain \verb!-infinity..+infinity!.
However, when this same variable is used as an index of a constraint  \gb{element}, its domain is automatically reduced to an appropriate range.

\medskip
A second illustrative \p3 model is given by the following file `TestElementIdx2.py':

\begin{boxpy}\begin{python}
@\imp@

x = Var(range(4))

start = cp_array([0, 2])

if not variant():
   satisfy(
      If(
         x >= 2,
         Then=start[x - 2] == 2,  # PROBLEM even with -force_element_index
      )
   )
elif variant("aux1"):
   satisfy(
      If(
         x >= 2,
         Then=[(y := Var()) == x - 2, start[y] == 2]
      )
   )
elif variant("aux2"):
   y = Var()  

   satisfy(
      If(
         x >= 2,
         Then=[y == x - 2, start[y] == 2]
      )
   )
\end{python}\end{boxpy}

If we execute:
\begin{verbatim}
    python TestElementIdx2.py
\end{verbatim}
we obtain an error message:
\begin{verbatim}
    ERROR:  Problem with an indexing expression ; maybe use -force_element_index,
    or replace array[index] by array[aux] after posing ((aux := Var()) == index)
\end{verbatim}

If we execute:
\begin{verbatim}
    python TestElementIdx2.py -force_element_index
\end{verbatim}

we obtain:

\begin{xcsp}
<instance format="XCSP3" type="CSP">
  <variables>
    <var id="x"> 0..3 </var>
    <array id="aux_gb" note="aux_gb[i] is the ith auxiliary variable having been automatically introduced" size="[2]">
      <domain for="aux_gb[0]"> 0 1 </domain>
      <domain for="aux_gb[1]"> 0..2 </domain>
    </array>
  </variables>
  <constraints>
    <intension> imp(ge(x,2),eq(aux_gb[1],2)) </intension>
    <intension> eq(sub(x,2),aux_gb[0]) </intension>
    <element>
      <list> 0 2 </list>
      <index> aux_gb[0] </index>
      <value> aux_gb[1] </value>
    </element>
  </constraints>
</instance>
\end{xcsp}

which is, unfortunately, not correct because the constraint $x -2 = \mathtt{aux\_gb}[0]$ should be  $x \geq 2 \Rightarrow x -2 = \mathtt{aux\_gb}[0]$.
This is technically difficult to intercept at compilation time (but we shall attempt to fix that in future versions).
Consequently, one must adopt solutions proposed by variants `aux1' and `aux2' that are safe.

For example, if we execute:
\begin{verbatim}
    python TestElementIdx2.py -variant=aux1
\end{verbatim}

we obtain:

\begin{xcsp}
<instance format="XCSP3" type="CSP">
  <variables>
    <var id="x"> 0..3 </var>
    <array id="aux_gb" note="aux_gb[i] is the ith auxiliary variable having been automatically introduced" size="[2]">
      <domain for="aux_gb[0]"> 0 1 </domain>
      <domain for="aux_gb[1]"> 0..2 </domain>
    </array>
  </variables>
  <constraints>
    <intension> or(lt(x,2),eq(aux_gb[0],sub(x,2))) </intension>
    <intension> or(lt(x,2),eq(aux_gb[1],2)) </intension>
    <element>
      <list> 0 2 </list>
      <index> aux_gb[0] </index>
      <value> aux_gb[1] </value>
    </element>
  </constraints>
</instance>
\end{xcsp}

which is correct.

\medskip
A third illustrative \p3 model is given by the following file `TestElementIdx3.py':

\begin{boxpy}\begin{python}
@\imp@

x = Var(range(4))

start = cp_array([0, 2])

if not variant():
   satisfy(
      If(
         x < 2,
         Then=start[x] == 2,  # PROBLEM even with -force_element_index
      )
   )
elif variant("aux1"):
   satisfy(
      If(
         x < 2,
         Then=[(aux := Var()) == x, start[aux] == 2]
      )
   )
elif variant("aux2"):
   aux = Var()

   satisfy(
      If(
         x < 2,
         Then=[aux == x, start[aux] == 2]
      )
   )
\end{python}\end{boxpy}

Once again, because indexing for a constraint \gb{element} has to be handled in the part \nn{Then} (this is similar for the part \nn{Else}) of a complex structure \nn{If}, one has to safely use an explictly introduced auxiliary variable.

\section{Constraint \gb{channel}}

The first variant of the constraint \gb{channel} is defined on a single list of variables, and ensures that if the $ith$ variable of the list is assigned the value $j$, then the $jth$ variable of the same list must be assigned the value $i$.

\begin{boxse}
\begin{semantics}
$\gb{channel}(X)$, with $X=\langle x_0,x_1,\ldots \rangle$, iff 
  $\forall i : 0 \leq i < |X|, \va{x}_i =j \Rightarrow \va{x}_j=i$
\end{semantics}
\end{boxse}

A second classical variant of \gb{channel}, sometimes called \gb{inverse} or \gb{assignment} in the literature, is defined from two separate lists (of the same size) of variables. 
It ensures that the value assigned to the $ith$ variable of the first list gives the position of the variable of the second list that is assigned to $i$, and vice versa.


\begin{boxse}
\begin{semantics}
$\gb{channel}(X,Y)$, with $X=\langle x_0,x_1,\ldots \rangle$ and $Y=\langle y_0,y_1,\ldots \rangle$, iff 
  $\forall i : 0 \leq i < |X|, \va{x}_i = j \Leftrightarrow \va{y}_j = i$ 

@{\em Prerequisite}@: $2 \leq |X| = |Y|$
\end{semantics}
\end{boxse}

It is also possible to use this form of \gb{channel}, with two lists of different sizes.
The constraint then imposes restrictions on all variables of the first list, but not on all variables of the second list.
The syntax is the same, but the semantics is the following (note that the equivalence has been replaced by an implication):

\begin{boxse}
\begin{semantics}
$\gb{channel}(X,Y)$, with $X=\langle x_0,x_1,\ldots \rangle$ and $Y=\langle y_0,y_1,\ldots \rangle$, iff 
  $\forall i : 0 \leq i < |X|, \va{x}_i = j \Rightarrow \va{y}_j = i$ 

@{\em Prerequisite}@: $2 \leq |X| < |Y|$
\end{semantics}
\end{boxse}

Finally, a third variant of \gb{channel} is obtained by considering a list of 0/1 variables to be channeled with an integer variable.
This third form of constraint \gb{channel} ensures that the only variable of the list that is assigned to 1 is at an index (position) that corresponds to the value assigned to the stand-alone integer variable.

\begin{boxse}
\begin{semantics}
$\gb{channel}(X,v)$, with $X=\{x_0,x_1,\ldots\}$, iff 
  $\forall i : 0 \leq i < |X|, \va{x}_i = 1 \Leftrightarrow \va{v} = i$
  $\exists i : 0 \leq i < |X| \land \va{x}_i = 1$
\end{semantics}
\end{boxse}

In \p3, to post a constraint \gb{channel}, we must call the function \nn{Channel}() whose signature is:

\begin{python}
  def Channel(list1, list2=None, *, start_index1=0, start_index2=0):
\end{python}

For the first variant, in addition to the positional parameter \texttt{list1}, one may use the optional attribute \texttt{start\_index1} that gives the number used for indexing the first variable in this list (0, by default).
For the second variant, two lists must be specified, and optionally the two named parameters can be used.
For the third variant, the positional parameter \texttt{list2} must be a variable (or a list only containing one variable).

\paragraph{Black Hole.} This problem was introduced in Section \ref{sec:blackhole}.
Here is a snippet of the \p3 model:   
\begin{python} 
...

# x[i] is the value j of the card at position i of the stack
x = VarArray(size=nCards, dom=range(nCards))

# y[j] is the position i of the card whose value is j
y = VarArray(size=nCards, dom=range(nCards))

satisfy(
  Channel(x, y),

  ...
)
\end{python}
The constraint \gb{channel} (second variant) links the dual roles of variables from arrays $x$ and $y$.

\paragraph{Progressive Party.} \index{Problems!Progressive Party} \label{sec:progressive} From \href{https://www.csplib.org/Problems/prob013/}{CSPLib}: 
``The problem is to timetable a party at a yacht club.
Certain boats are to be designated hosts, and the crews of the remaining boats in turn visit the host boats for several successive half-hour periods.
The crew of a host boat remains on board to act as hosts while the crew of a guest boat together visits several hosts.
Every boat can only hold a limited number of people at a time (its capacity) and crew sizes are different.
The total number of people aboard a boat, including the host crew and guest crews, must not exceed the capacity.
A guest boat cannot revisit a host and guest crews cannot meet more than once.
The problem facing the rally organizer is that of minimizing the number of host boats.''

\begin{figure}[h!]
\begin{center}
  \includegraphics[scale=0.32]{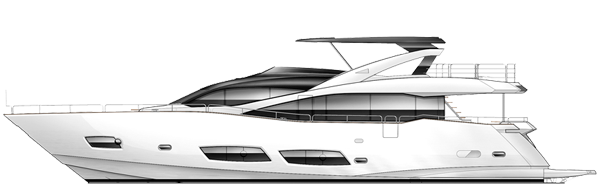}
\end{center}
\caption{Progressive Party at a Yacht Club. \tiny{(image from \href{https://pngimg.com/download/5422}{pngimg.com})} \label{fig:ship}}
\end{figure}

An example of data is given by the following JSON file:
\begin{json}
{
  "nPeriods": 5,
  "boats": [
    {"capacity": 6, "crewSize": 2},
    {"capacity": 8, "crewSize": 2},
    ...
  ]
}
\end{json}

A \p3 model of this problem is given by the following file `ProgressiveParty.py':

\begin{boxpy}\begin{python}
@\imp@

nPeriods, boats = data
nBoats = len(boats)
capacities, crews = zip(*boats)

# h[b] indicates if the boat b is a host boat
h = VarArray(size=nBoats, dom={0, 1})

# s[b][p] is the scheduled (visited) boat by the crew of boat b at period p
s = VarArray(size=[nBoats, nPeriods], dom=range(nBoats))

# g[b1][p][b2] is 1 if s[b1][p] = b2
g = VarArray(size=[nBoats, nPeriods, nBoats], dom={0, 1})

satisfy(
  # identifying host boats (when receiving)
  [h[b] == (s[b][p] == b) for b in range(nBoats) for p in range(nPeriods)],

  # identifying host boats (from visitors)
  [h[s[b][p]] == 1 for b in range(nBoats) for p in range(nPeriods)],

   # channeling variables from arrays s and g
  [Channel(g[b][p], s[b][p]) for b in range(nBoats) for p in range(nPeriods)],
  
  # boat capacities must be respected
  [g[:,p,b]*crews <= capacities[b] for b in range(nBoats) for p in range(nPeriods)],
  
  # a guest crew cannot revisit a host
  [AllDifferent(s[b], excepting=b) for b in range(nBoats)],
  
  # guest crews cannot meet more than once
  [Sum(s[b1][p] == s[b2][p] for p in range(nPeriods)) <= 1
    for b1, b2 in combinations(nBoats, 2)]
}

minimize(
  # minimizing the number of host boats
  Sum(h)
)
\end{python}\end{boxpy}

This is the third variant of \gb{channel} that is used here: \nn{g[b][p]} is an array of 0/1 variables while \nn{s[b][p]} is a stand-alone integer variable.
Below, note how the symbol ':' is used to take a complete slice of a 3-dimensional array of variables, when posting constraints about boat capacities.
Instead, we could have written:

\begin{python}
  [[g[i][p][b] for i in range(nBoats)] * crews <= capacities[b]
    for b in range(nBoats) for p in range(nPeriods)],
\end{python}

Concerning the last list of \gb{sum} constraints, as the Boolean expression \nn{s[b1][p] == s[b2][p]} is considered to return integers, 0 for false and 1 for true, it is possible to perform a summation.

\section{Constraint \gb{noOverlap}} \label{sec:noOverlap}

We start with the one dimensional form of \gb{noOverlap} \cite{H_integrated} that corresponds to \gb{disjunctive} \cite{C_one} and ensures that some objects (e.g., tasks), defined by their origins (e.g., starting times) and lengths (e.g., durations), must not overlap.
The semantics is given by: 

\begin{boxse}
\begin{semantics}
$\gb{noOverlap}(X,L)$, with $X=\langle x_0,x_1,\ldots \rangle$ and $L=\langle l_0,l_1,\ldots \rangle$, iff  
  $\forall (i,j) : 0 \leq i < j < |X|,  \va{x}_{i} + \va{l}_{i} \leq \va{x}_{j} \vee \va{x}_{j} + \va{l}_{j} \leq \va{x}_{i}$

$\gbc{Prerequisite}: |X| = |L| \geq 2$
\end{semantics}
\end{boxse}

In \p3, to post a constraint \gb{noOverlap}, we must call the function \nn{NoOverlap}() whose signature is:

\begin{python}
  def NoOverlap(*, origins, lengths, zero_ignored=False):
\end{python}

Note that all parameters must be named (see '*' at first position), and that the parameter \texttt{zero\_ignored} is optional (value \nn{False} by default).
If ever we are in a situation where there exist some zero-length object(s), then if the  parameter \texttt{zero\_ignored} is set to \nn{False}, it indicates that zero-length objects cannot be packed anywhere (cannot overlap with other objects).
Arguments given to \texttt{origins} and \texttt{lengths} when calling the function \nn{NoOverlap}() are expected to be lists of the same length; \texttt{origins} must be given a list of variables whereas \texttt{lengths} must be given either a list of variables or a list of integers.  

\paragraph{Flow Shop Scheduling.} \index{Problems!Flow Shop Scheduling} From WikiPedia: ``There are n machines and m jobs. Each job contains exactly n operations.
The ith operation of the job must be executed on the ith machine.
No machine can perform more than one operation simultaneously.
For each operation of each job, execution time is specified.
Operations within one job must be performed in the specified order.
The first operation gets executed on the first machine, then (as the first operation is finished) the second operation on the second machine, and so on until the nth operation.
Jobs can be executed in any order, however.
Problem definition implies that this job order is exactly the same for each machine.
The problem is to determine the optimal such arrangement, i.e. the one with the shortest possible total job execution makespan.''

\begin{figure}[h!]
\begin{center}
  \includegraphics[scale=0.25]{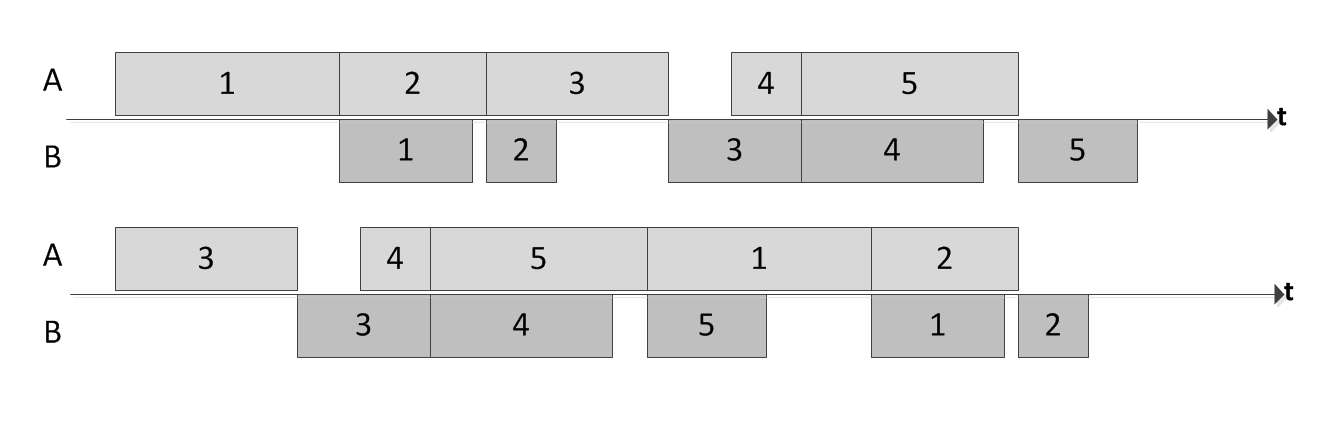}
\end{center}
\caption{Example of (no-wait) flow-shop scheduling with five jobs on two machines A and B. A comparison of total makespan is given for two different job sequences. \tiny{(image from \href{https://commons.wikimedia.org/wiki/File:No-wait_flow_shop_example.png}{commons.wikimedia.org})} \label{fig:flowShop}}
\end{figure}

To specify a problem instance, we just need a two-dimensional array of integers for recording durations, as in the following JSON file:  

\begin{json}
{
  "durations":[
    [26,59,78,88,69],
    [38,62,90,54,30],
    ...
  ]
}
\end{json}

A \p3 model of this problem is given by the following file `FlowShopScheduling.py':

\begin{boxpy}\begin{python}
@\imp@

durations = data  # durations[i][j] is the duration of operation/machine j for job i
horizon = sum(sum(t) for t in durations) + 1
n, m = len(durations), len(durations[0])

# s[i][j] is the start time of the jth operation for the ith job
s = VarArray(size=[n, m], dom=range(horizon))

satisfy(
  # operations must be ordered on each job
  [Increasing(s[i], lengths=durations[i]) for i in range(n)],

  # no overlap on resources
  [
    NoOverlap(
      origins=s[:, j],
      lengths=durations[:, j]
    ) for j in range(m)
  ]
)

minimize(
  # minimizing the makespan
  Maximum(s[i][-1] + durations[i][-1] for i in range(n))
)
\end{python}\end{boxpy}

In this model, for each operation (or equivalently, machine) $j$, we collect the list of variables from the jth column of \nn{s} and the list of integers from the jth column of \nn{durations} when posting a constraint \gb{noOverlap}.
Remember that the notation \nn{[:, j]} stands for the jth column of a two-dimensional array (list).

\medskip
The k-dimensional form of \gb{noOverlap} corresponds to \gb{diffn} \cite{BC_chip} and ensures that, given a set of $n$-dimensional boxes; for any pair of such boxes, there exists at least one dimension where one box is after the other, i.e., the boxes do not overlap.
The semantics is:


\begin{boxse}
\begin{semantics}
$\gb{noOverlap}(\ns{X},\ns{L})$, with $\ns{X}=\langle (x_{1,1},\ldots,x_{1,n}),(x_{2,1},\ldots,x_{2,n}),\ldots \rangle$ and $\ns{L}=\langle (l_{1,1},\ldots,l_{1,n}),(l_{2,1},\ldots,l_{2,n}),\ldots \rangle$, iff  
  $\forall (i,j) : 1 \leq i < j \leq |\ns{X}|,  \exists k \in 1..n :  \va{x}_{i,k} + \va{l}_{i,k} \leq \va{x}_{j,k} \vee \va{x}_{j,k} + \va{l}_{j,k} \leq \va{x}_{i,k}$

$\gbc{Prerequisite}: |\ns{X}| = |\ns{L}| \geq 2$
\end{semantics}
\end{boxse}

In \p3, to post a constraint \gb{noOverlap}, we must call the function \nn{NoOverlap}() whose signature is:

\begin{python}
  def NoOverlap(*, origins, lengths, zero_ignored=False):
\end{python}

Note that all parameters must be named (see '*' at first position), and that the parameter \texttt{zero\_ignored} is optional (value \nn{False} by default).
If ever we are in a situation where there exist some zero-length box(es), then if the  parameter \texttt{zero\_ignored} is set to \nn{False}, it indicates that zero-length boxes cannot be packed anywhere (cannot overlap with other boxes).
Arguments given to \texttt{origins} and \texttt{lengths} when calling the function \nn{NoOverlap}() are expected to be two-dimensional lists of the same length; \texttt{origins} must only involve variables whereas \texttt{lengths} must involve either only variables or only integers.

\paragraph{Rectangle Packing Problem. }\index{Problems!Rectangle Packing}
The rectangle packing problem consists of finding a way of putting a given set of rectangles (boxes) in an enclosing rectangle (container) without overlap.


\begin{figure}[h!]
  \begin{center}
  \begin{tikzpicture}[scale=0.35]
  \draw [thick,color=black] (-0.1,-0.1) rectangle (18.1,10.1) ;
  \draw [thick,color=orange] (0.1,0.1) rectangle (1.9,2.9) ;
  \draw [thick,color=orange] (0.1,3.1) rectangle (1.9,5.9) ;
  \draw [thick,color=orange] (0.1,6.1) rectangle (2.9,9.9) ;

  \draw [thick,color=blue] (2.1,0.1) rectangle (4.9,3.9) ;
  \draw [thick,color=blue] (5.1,0.1) rectangle (7.9,3.9) ;
  \draw [thick,color=blue] (8.1,0.1) rectangle (10.9,3.9) ;

  \draw [thick,color=red] (3.1,4.1) rectangle (8.9,9.9) ;

  \draw [thick,color=yellow] (9.1,4.1) rectangle (17.9,6.9) ;
  \draw [thick,color=yellow] (9.1,7.1) rectangle (17.9,9.9) ;

  \draw [thick,color=olive] (11.1,0.1) rectangle (17.9,1.9) ;
  \draw [thick,color=olive] (11.1,2.1) rectangle (17.9,3.9) ;
\end{tikzpicture}
\end{center}
\caption{Packing Rectangles in a Container. \label{fig:packing}}
\end{figure}

An example of data is given by the following JSON file:

\begin{json}
{
  "container":{"width":112,"height":112},
  "boxes":[
    {"width":2,"height":2},
    {"width":4,"height":4},
    ...
  ]
}
\end{json}

A \p3 model of this problem is given by the following file `RectanglePacking.py':

\begin{boxpy}\begin{python}
@\imp@
    
width, height = data.container
boxes = data.boxes
nBoxes = len(boxes)

# x[i] is the x-coordinate where is put the ith box (rectangle)
x = VarArray(size=nBoxes, dom=range(width))

# y[i] is the y-coordinate where is put the ith box (rectangle)
y = VarArray(size=nBoxes, dom=range(height))

satisfy(
  # unary constraints on x
  [x[i] + boxes[i].width <= width for i in range(nBoxes)],

  # unary constraints on y
  [y[i] + boxes[i].height <= height for i in range(nBoxes)],
  
  # no overlap on boxes
  NoOverlap(
    origins=[(x[i], y[i]) for i in range(nBoxes)],
    lengths=boxes
  ),
  
  # tag(symmetry-breaking)
  [
    x[-1] <= math.floor((width - boxes[-1].width) // 2.0),
    y[-1] <= x[-1]
  ] if width == height else None
)
\end{python}\end{boxpy}

\section{Constraint \gb{cumulative}} \label{sec:cumulative}

The constraint \gb{cumulative} is useful when a resource of limited quantity must be shared for achieving several tasks.
For example, in a scheduling context where several tasks require some specific quantities of a single resource,
the cumulative constraint imposes that a strict limit on the total consumption of the resource is never exceeded at each point of a time line.
The tasks may overlap but their cumulative resource consumption must never exceed the limit.
In Figure \ref{fig:cumulative}, five tasks (some of them overlapping) are scheduled while never exceeding the capacity (5) of the resource.
The interested reader can check that there is no better scheduling scenario, that is to say, a way of scheduling the five tasks in less than 7 time units.

\begin{figure}[h!]
  \begin{center}
\begin{tikzpicture}[scale=0.8]
  \draw [dotted] (0,0) grid (8,6);
  \foreach \y in {1,2,...,6} \draw (-0.2,\y) node[left] {\y};
  \foreach \x in {1,2,...,8} \draw (\x,0) node[below] {\x};
  \draw[>=latex,->] (0,0) -- (8,0) ;
  \draw (4.5,-0.5) node[below]{Time} ;
  \draw [>=latex,->] (0,0) -- (0,6);
  \draw (-1.1,3) node[rotate=90]{Resource Consumption} ;
  \draw[very thick] (0,5) -- (8,5) node[right] {Limit} ; 

  \draw [thick,fill=orange] (0,0) rectangle (3,3) node[midway] {Task 1};
  \draw [thick,fill=orange!40] (0,3) rectangle (2,5) node[midway] {Task 2};
  \draw [thick,fill=orange!80] (3,0) rectangle (7,2) node[midway] {Task 4};
  \draw [thick,fill=orange!60] (3,2) rectangle (5,5) node[midway] {Task 5};
  \draw [thick,fill=orange!20] (5,2) rectangle (7,4) node[midway] {Task 3};

\end{tikzpicture}
\end{center}
\caption{Example of a Limited Cumulative Resource.\label{fig:cumulative}}
\end{figure}

So, the context is to manage a collection of tasks, each one being described by 4 attributes: its starting time \texttt{origin}, its length or duration \texttt{length}, its stopping time \texttt{end} and its resource consumption \texttt{height}.
Usually, the values for \texttt{length} and \texttt{height} are given while the values for \texttt{origin} (and \texttt{end} by deduction) must be computed.

The constraint \gb{cumulative} \cite{AB_extending} enforces that at each point in time, the cumulated height of tasks that overlap that point, respects a numerical condition $(\odot,k)$.
The semantics is given by:

\begin{boxse}
\begin{semantics}
$\gb{cumulative}(X,L,H,(\odot,k))$, with $X=\langle x_0,x_1,\ldots\rangle$, $L=\langle l_0,l_1,\ldots\rangle$, $H=\langle h_0,h_1,\ldots\rangle$, iff 
  $\forall t \in \N, \sum \{\va{h}_i : 0 \leq i < |H| \land \va{x}_i \leq t < \va{x}_i + \va{l}_i\} \odot \va{k}$

$\gbc{Prerequisite}: |X| = |L| = |H| \geq 2$
\end{semantics}
\end{boxse}

If the attributes \texttt{end} are present while reasoning, we have additionally a set $E=\langle e_0,e_1,\ldots\rangle$ such that:
\begin{quote}
$\forall i : 0 \leq i < |X| , \va{x}_i + \va{l}_i = \va{e}_i$
\end{quote}

In \p3, to post a constraint \gb{cumulative}, we must call the function \nn{Cumulative}() whose signature is:

\begin{python}
  def Cumulative(*, origins, lengths, heights, ends=None):
\end{python}

Note that all parameters must be named (see '*' at first position) and the parameter \texttt{ends} is optional (value \nn{None} by default).
Arguments given when calling the function are expected to be lists of the same length.
The object obtained when calling \nn{Cumulative}() must be restricted by a condition (typically, defined by a relational operator and a limit).

\paragraph{RCPSP.} \index{Problems!Resource Constrained Project Scheduling} From CSPLib: ``The Resource-Constrained Project Scheduling Problem is a classical problem in operations research.
A number of activities are to be scheduled.
Each activity has a duration and cannot be interrupted.
There are a set of precedence relations between pairs of activities which state that the second activity must start after the first has finished.
There are a set of renewable resources. Each resource has a maximum capacity and at any given time slot no more than this amount can be in use.
Each activity has a demand (possibly zero) on each resource. 
The problem is usually stated as an optimization problem where the makespan (i.e., the completion time of the last activity) is minimized.'' 
See \href{https://csplib.org/Problems/prob061/}{CSPLib--Problem 061} for more information.

An example of data is given by the following JSON file:

\begin{json}
{
  "horizon":158,
  "resourceCapacities":[12,13,4,12],
  "jobs":[
    {"duration":0, "successors":[1,2,3], "requiredQuantities":[0,0,0,0]},
    {"duration":8, "successors":[5,10,14], "requiredQuantities":[4,0,0,0]},
    ...
  ]
}
\end{json}

A \p3 model of this problem is given by the following file `Rcpsp.py':

\begin{boxpy}\begin{python}
@\imp@

jobs, horizon, capacities, _ = data
durations, successors, quantities = zip(*jobs)  # [job.duration for job in jobs]
nJobs = len(jobs)

# s[i] is the starting time of the ith job
s = VarArray(size=nJobs, dom=lambda i: {0} if i == 0 else range(horizon))

satisfy(
  # precedence constraints
  [s[i] + durations[i] <= s[j] for i in range(nJobs) for j in successors[i]],

  # resource constraints
  [
    Cumulative(
      Task(
        origin=s[i],
        length=durations[i],
        height=quantities[i][k]
      ) for i in range(nJobs) if quantities[i][k] > 0
    ) <= capacity for k, capacity in enumerate(capacities)
  ]
)

minimize(
  s[-1]
)
\end{python}\end{boxpy}

Observe how a \nn{Cumulative} constraint is posted to respect the capacity of each resource.

\section{Constraint \gb{binPacking}} \label{sec:binPacking}

The first form of the constraint \gb{binPacking} \cite{S_constraint,S_solving,CMOS_bin} ensures that a list of items, whose sizes are given, are put in different bins in such a way that the total size of the items in each bin
respects a numerical condition (always the same, because the capacity is assumed to be the same for all bins). 
When the operator ``le'' is used, this corresponds to not exceeding the capacity of each bin. 

\begin{boxse}
\begin{semantics}
$\gb{binPacking}(X,S,(\odot,k))$, with $X=\langle x_1,x_2,\ldots \rangle$ and $S=\langle s_1,s_2,\ldots \rangle$, iff 
  $\forall b \in \{\va{x}_i : 1 \leq i \leq |X|\}, \sum \{s_i : 1 \leq i \leq |S| \land \va{x}_i =b\} \odot \va{k}$

$\gbc{Prerequisite}: |X| = |S| \geq 2$
\end{semantics}
\end{boxse}

The second form of the constraint \gb{binPacking} associates a specific limit (capacity) with each bin.
The limits are given either by integer values or by integer variables.

\begin{boxse}
\begin{semantics}
$\gb{binPacking}(X,S,C)$, with $X=\langle x_1,x_2,\ldots \rangle$, $S=\langle s_1,s_2,\ldots \rangle$ and $C=\langle c_1,c_2,\ldots \rangle$ iff 
  $\forall b \in \{\va{x}_i : 1 \leq i \leq |X|\}$, $\sum \{s_i : 1 \leq i \leq |S| \land \va{x}_i =b\} \leq \va{c_b}$
  
$\gbc{Prerequisite}: |X| = |S| \geq 2$
\end{semantics}
\end{boxse}

The third form of the constraint \gb{binPacking} associates a specific load with each bin.
The loads are given either by integer values or by integer variables.

\begin{boxse}
\begin{semantics}
$\gb{binPacking}(X,S,L)$, with $X=\langle x_1,x_2,\ldots \rangle$, $S=\langle s_1,s_2,\ldots \rangle$ and $L=\langle l_1,l_2,\ldots \rangle$ iff 
  $\forall b \in \{\va{x}_i : 1 \leq i \leq |X|\}$, $\sum \{s_i : 1 \leq i \leq |S| \land \va{x}_i =b\} = \va{l_b}$
  
$\gbc{Prerequisite}: |X| = |S| \geq 2$
\end{semantics}
\end{boxse}

In \p3, to post a constraint \gb{binPacking}, we must call the function \nn{BinPacking}() whose signature is:

\begin{python}
   def BinPacking(term, *others, sizes, limits=None, loads=None):
\end{python}

The two parameters \texttt{term} and \texttt{others} are positional, and allow us to pass the terms either in sequence (individually) or under the form of a list.
The named parameter \texttt{sizes} gives the respective size of the items to be packed.
For the first form of \gb{binPacking}, mentioned above, \texttt{limits} and \texttt{loads} are both \nn{None} and the object obtained when calling \nn{BinPacking}() represents the maximum accumulated size in a bin and must be restricted by a condition (typically, defined by a relational operator and a limit).
For the second form of \gb{binPacking}, \texttt{limits} is specified and no extern condition is present.
For the third form of \gb{binPacking}, \texttt{loads} is specified and no extern condition is present.

\paragraph{Cardinality Constrained Multi-cycle Problem.} \index{Problems!CCMcP}
Studied in \cite{V_kidney}, the Cardinality Constrained Multi-cycle Problem (CCMcP) is a variation of the Kidney Exchange Problem (KEP).
One can consider the CCMcP as an Asymmetric Travelling Salesman Problem (ATSP) with subtours (cycles) allowed (but of limited size $k$).
Each arc has an associated weight.
The arcs with negative weights cannot be part of subtours, and the objective is to maximize the sum of weights occurring along the arcs of the computed subtours (cycles).

An example of data is given by the following JSON file:

\begin{json}
{
  "weights": [
    [0, -1, 4, 2, 5, 6],
    [-1, 0, -1, 8, -1, -1],
    [3, 8, 0, -1, 9, 3],
    [5, -1, -1, 0, 6, 5],
    [4, -1, 9, -1, 0, 2],
    [7, 8, 8, 2, -1, 0]
  ],
  "k": 3
}
\end{json}

A \p3 model (inspired by the one written for the 2019 Minizinc challenge) of this problem is given by the following file `CCMcP.py':

\begin{boxpy}\begin{python}
@\imp@

weights, k = data
nNodes = len(weights)

# x[i] is the successor node of node i (in the cycle where i belongs)
x = VarArray(size=nNodes, dom=range(nNodes))

# y[i] is the cycle (index) where the node i belongs
y = VarArray(size=nNodes, dom=range(nNodes))

satisfy(
  AllDifferent(x),

  # ensuring correct cycles
  [y[i] == y[x[i]] for i in range(nNodes)],
  
  # disabling infeasible arcs
  [x[i] != j for i in range(nNodes) for j in range(nNodes) if weights[i][j] < 0],
  
  # each cycle contains at most k arcs
  BinPacking(y, sizes=1) <= k,
  
  # tag(symmetry-breaking)
  Precedence(y)
)

maximize(
  # maximizing the sum of arc weights of selected cycles
  Sum(weights[i][x[i]] for i in range(nNodes))
)
\end{python}\end{boxpy}

Note that the first form of \gb{binPacking} is used in this model of CCMcP: it ensures that the longest subtour (and, consequently, each subtour) is formed of at most $k$ arcs.
The constraint \gb{precedence} breaks some value symmetries.

\paragraph{Warehouse Location.} This problem was introduced in Section \ref{sec:warehouse}.
Actually, the following group of constraints \gb{count}: 

\begin{python}
  # capacities of warehouses must not be exceeded
  [Count(w, value=j) <= capacities[j] for j in range(nWarehouses)],
\end{python}

can be replaced by a single constraint \gb{binPacking}:
\begin{python}
  # capacities of warehouses must not be exceeded
  BinPacking(w, sizes=1, limits=capacities),
\end{python}

Here, this is the second form of \gb{binPacking}: it ensures that for each warehouse $j$ its capacity is not exceeded.
Note that the parameter \texttt{sizes} is given a unique integer as value (1), implicitly indicating that this is the size to be used for all items.
The interest of making this change (i.e., using \gb{binPacking} on this problem) may depend on the used underlying solvers.

\paragraph{Steel Mill Slab.} \label{pb:steelMillSlab}\index{Problems!Steel Mill Slab} From \href{https://www.csplib.org/Problems/prob038}{CSPLib}:
``Steel is produced by casting molten iron into slabs.
A steel mill can produce a finite number of slab sizes.
An order has two properties, a colour corresponding to the route required through the steel mill, and a weight.
Given a set of orders, the problem is to assign the orders to slabs, the number and size of which are also to be determined, such that the total weight of steel produced is minimised.
This assignment is subject to two further constraints:
\begin{itemize}
\item colour constraints: each slab can contain at most p colours (p is usually 2); 
\item capacity constraints: the total weight of orders assigned to a slab cannot exceed the slab capacity.
\end{itemize}
The colour constraints arise because it is expensive to cut up slabs in order to send them to different parts of the mill.''

\begin{figure}[h!]
\begin{center}
  \includegraphics[scale=0.18]{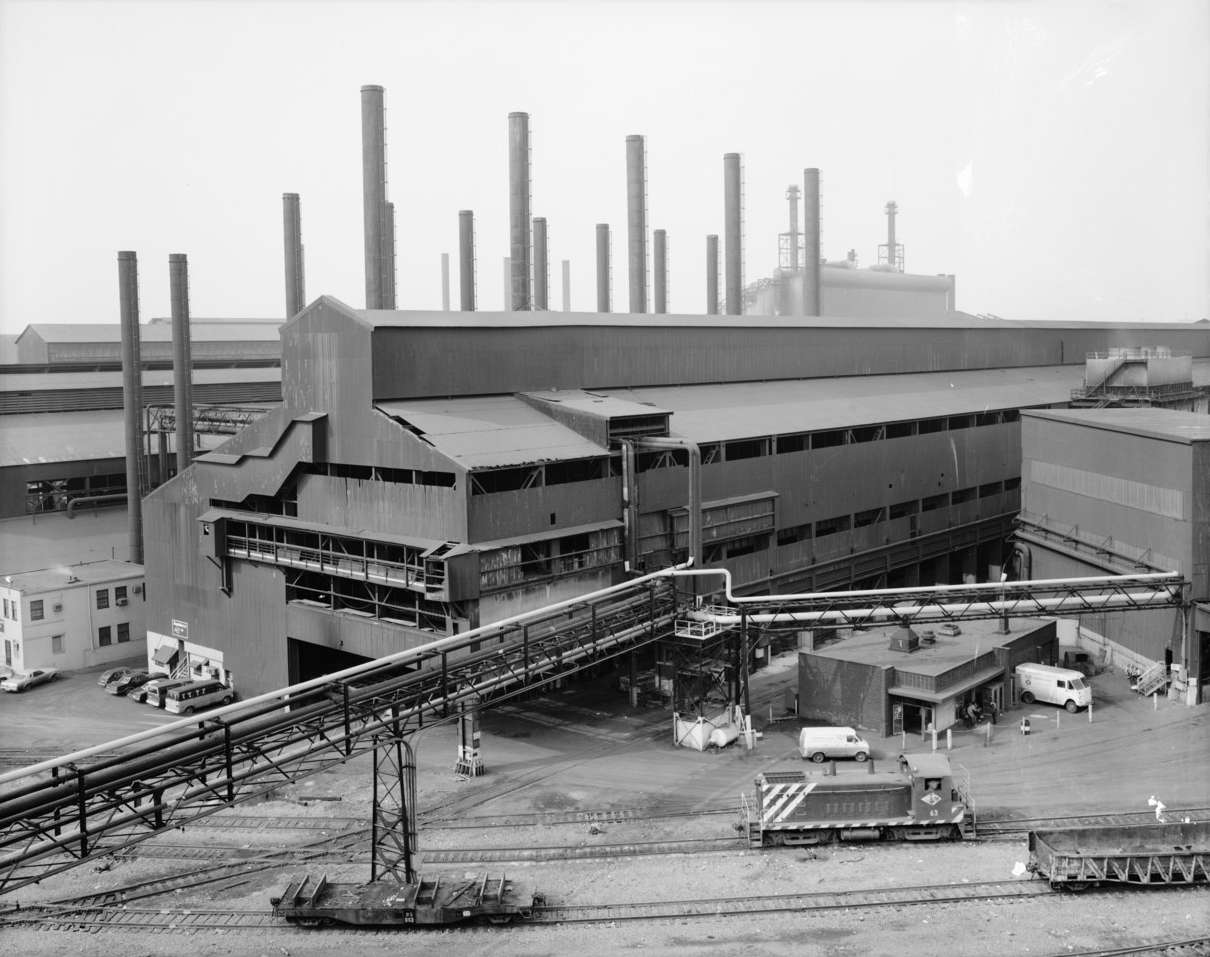}
\end{center}
\caption{Slabbing and Blooming Mills. \tiny{(image from \href{https://commons.wikimedia.org/wiki/Category:Blooming-slabbing_mill}{commons.wikimedia.org})} \label{fig:steelMill}}
\end{figure}

An example of data is given by the following JSON file:

\begin{json}
{
  "slabSizes": [0, 11, 14, ..., 49],
  "orders": [{"color": 0, "size": 4}, {"color": 1, "size": 22},
    {"color": 2, "size": 9}, {"color": 3, "size": 5}, ...]
}
\end{json}

A \p3 model of this problem is given by the following file `SteelMillSlab.py':

\begin{boxpy}\begin{python}
@\imp@

slabSizes, orders = data
colors, sizes = zip(*orders)
nOrders, nColors = len(orders), len(set(colors))
nSlabs, slabSizeLimit = nOrders, max(slabSizes) + 1

# gaps between each required size s and the closest slab with a size greater than s
gaps = cp_array(min(c-s for c in slabSizes if c >= s) for s in range(slabSizeLimit))

# x[k] is the slab for the kth order
x = VarArray(size=nOrders, dom=range(nSlabs))

# y[i] is the size (load) of the ith slab
y = VarArray(size=nSlabs, dom=range(slabSizeLimit))

satisfy(
  # each slab can contain at most 2 colors
  [
    Sum(
      Exist(x[k]==i for k in range(nOrders) if colors[k]==c) for c in range(nColors)
    ) <= 2 for i in range(nSlabs)
  ],

  # computing loads of slabs
  BinPacking(x, sizes=sizes, loads=y),
)

minimize(
  # minimizing total weight of steel produced
  Sum(gaps[y[i]] for i in range(nSlabs))
)
\end{python}\end{boxpy}

The third form of \gb{binPacking} is used here; it allows us to compute loads of all slabs (with just one constraint).
This model is a rare illustration of the explicit need to call the function \texttt{cp\_array}(), in order to use it when posting the objective. 
Also, note that we call \gb{Exist} that corresponds to a constraint \gb{Count}.
Indeed:
\begin{python}
Exist(x[k] == i for k in range(nOrders) if colors[k] == c)
\end{python}
is equivalent to:
\begin{python}
Count(x[k] == i for k in range(nOrders) if colors[k] == c, value=1) >= 1
\end{python}
Finally, some symmetries could be broken by adding some constraints (see for example, the model used for the 2019 Minizinc challenge).

\medskip
Since \x3 Specifications 3.1, \gb{binPacking} belongs to \x3-core (notably because solvers not equipped with a specific propagator can handle that constraint easily by posting $b$ constraints \gb{sum}, one per bin).

\section{Constraint \gb{knapsack}} \label{sec:knapsack}

The constraint \gb{knapsack} \cite{FS_cost,S_approximated,MSS_filtering} ensures that some items are packed in a knapsack with certain weight and profit restrictions.
So, the context is to manage a collection of items, each one being described by 2 attributes: its weight and its profit.
We have to decide how many copies of each item must be selected while respecting a numerical condition $(\odot_w,k_w)$ on accumulated weights and a numerical condition $(\odot_p,k_p)$ on accumulated profits.
The operator of the first condition is expected to be in $\{<,\leq,=\}$ whereas the operator of the second condition is expected to be in $\{>,\geq,=\}$.

The semantics is given by:

\begin{boxse}
\begin{semantics}
  $\gb{knapsack}(X,W,(\odot_w,k_w),P,(\odot_p,k_p))$, with $X=\langle x_1,x_2,\ldots \rangle$, $W=\langle w_1,w_2,\ldots\rangle$, and $P=\langle p_1,p_2,\ldots \rangle$, iff
  $\sum_{i=1}^{|X|} (w_i \times \va{x}_i) \odot_w \va{k_w}$
  $\sum_{i=1}^{|X|} (p_i \times \va{x}_i) \odot_p \va{k_p}$

$\gbc{Prerequisite}: |X| = |W| = |P| \geq 2$
\end{semantics}
\end{boxse}

In \p3, to post a constraint \gb{knapsack}, we must call the function \nn{Knapsack}() whose signature is:

\begin{python}
  def Knapsack(term, *others, weights, wlimit=None, wcondition=None, profits):
\end{python}

The two parameters \texttt{term} and \texttt{others} are positional, and allow us to pass the terms either in sequence (individually) or under the form of a list.
The named parameters \texttt{weights} and \texttt{profits} are obviously required.
The first condition, on weights, is given either by \texttt{wlimit} or \texttt{wcondition}: exactly one of these two parameters must be different from \nn{None}.
The value of \texttt{wlimit} is either an integer value or an integer variable (and the implicit operator is then $\leq$). 
The value of \texttt{wcondition} can be built by calling a function among \nn{lt}(), \nn{le}(), \nn{eq}(), \dots 
The object obtained when calling \nn{Knapsack}() represents the accumulated profit and must be restricted by a condition (typically, defined by a relational operator and a limit).

\paragraph{Optimized Knapsack.\label{pb:knapsack}}\index{Problems!Optimized Knapsack}
We illustrate the constraint \gb{knapsack} with a very simple problem, which is composed of only one constraint \gb{knapsack} together with a summing objective.
The goal is first to ensure that the benefit of selected objects exceeds a given threshold (\texttt{p\_limit}), and then to maximize, if possible, that benefit.   

\begin{figure}[h!]
\begin{center}
  \includegraphics[scale=0.08]{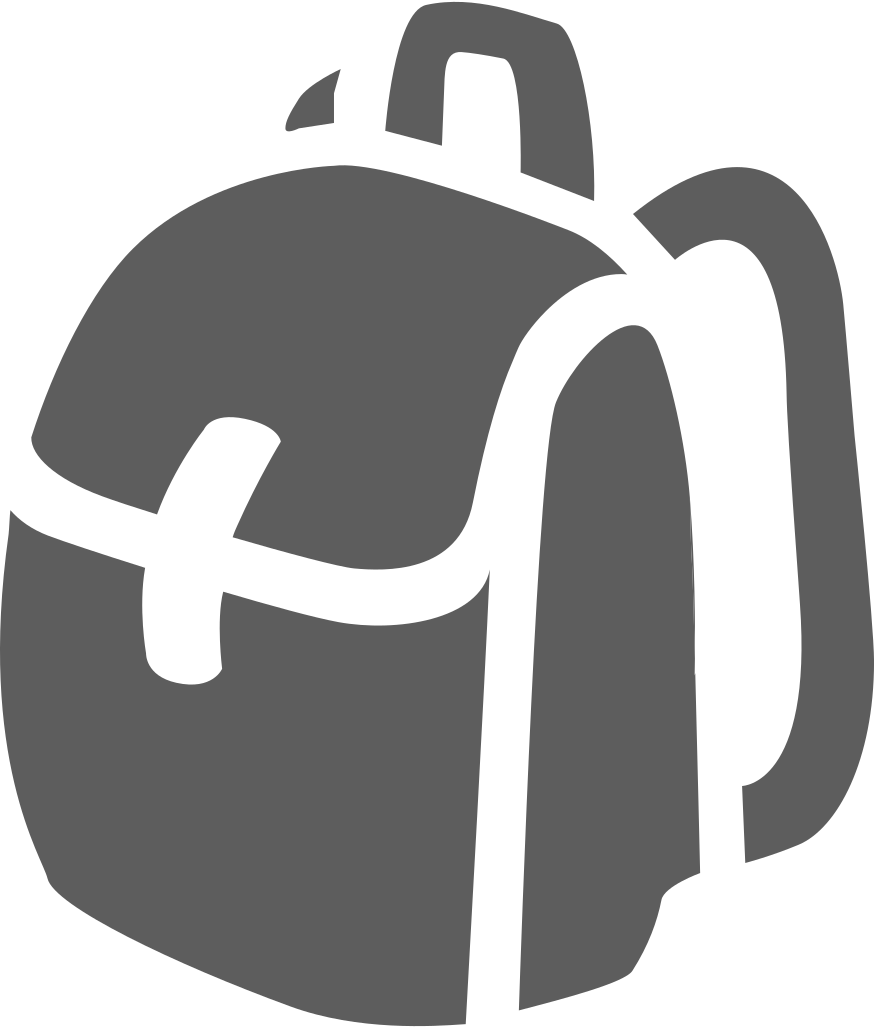}
\end{center}
\caption{Packing Items in a Knapsack. \tiny{(image from \href{https://commons.wikimedia.org/wiki/File:Backpack_(2551)_-_The_Noun_Project.svg}{commons.wikimedia.org})} \label{fig:bag}}
\end{figure}

A \p3 model of this problem is given by the following file `OptKnapsack.py':

\begin{boxpy}\begin{python}
@\imp@

weights = [10, 2, 6, 11, 21, 4, 8, 3, 8, 10]
profits = [20, 4, 3, 9, 13, 2, 3, 4, 7, 8]
w_limit, p_limit = 20, 30

nItems = len(weights)

# x[i] is 1 if the item i is packed
x = VarArray(size=nItems, dom={0, 1})

satisfy(
  Knapsack(x, weights=weights, wlimit=w_limit, profits=profits) >= p_limit
)

maximize(
  x * profits
)
\end{python}\end{boxpy}

Note that it is equivalent to writing: 

\begin{python}
   Knapsack(x, weights=weights, wcondition=le(w_limit), profits=profits) >= p_limit
\end{python}

\medskip
Since \x3 Specifications 3.1, \gb{knapsack} belongs to \x3-core (notably because solvers not equipped with a specific propagator can handle that constraint easily by posting two constraints \gb{sum}).

\section{Constraint \gb{circuit}} \label{sec:circuit}

Sometimes, problems involve graphs that are defined with integer variables (encoding called ``successors variables'').
In that context, graph-based constraints, like \gb{circuit}, involve a main list of variables $x_0, x_1,\ldots$
The assumption is that each pair $(i,\va{x}_i)$ represents an arc (or edge) of the graph to be built; if $\va{x}_i=j$, then it means that the successor of node $i$ is node $j$.
Note that a {\em loop} (also called self-loop) corresponds to a variable $x_i$ such that $\va{x}_i=i$. 

The constraint \gb{circuit} \cite{BC_chip} ensures that the values taken by the variables of the specified list forms a circuit, with the assumption that each pair $(i,\va{x}_i)$ represents an arc.
It is also possible to indicate that the circuit must be of a given size (strictly greater than $1$).
The semantics is given by:

\begin{boxse}
\begin{semantics}
$\gb{circuit}(X)$, with $X=\langle x_0,x_1,\ldots\rangle$, iff  @\com{capture \gb{subcircuit}}@
  $\{(i,\va{x}_i) : 0 \leq i < |X| \land i \neq \va{x}_i \}$ forms a circuit of size $> 1$ 
$\gb{circuit}(X,s)$, with $X=\langle x_0,x_1,\ldots\rangle$, iff  
  $\{(i,\va{x}_i) : 0 \leq i < |X| \land i \neq \va{x}_i \}$ forms a circuit of size $\va{s} > 1$
\end{semantics}
\end{boxse}

In \p3, to post a constraint \gb{circuit}, we must call the function \nn{Circuit}() whose signature is:

\begin{python}
  def Circuit(term, *others, start_index=0, size=None, no_self_looping=False):
\end{python}

The first two parameters \texttt{term} and \texttt{others} are positional, and allow us to pass the ``successors variables'' either in sequence (individually) or under the form of a list.
The next two parameters are optional (and must be named): \texttt{start\_index} gives the number used for indexing the first variable of the specified list (0, by default), and \texttt{size} indicates that the circuit must be of a given size (\nn{None} by default indicates that no specific size is required). 
The last parameter, \texttt{no\_self\_looping}, indicates, when \texttt{true}, that loops are not authorized (i.e., the successor of a variable/node cannot be itself); this parameter was introduced in Version 2.6. 

Hence, it is important to note that the circuit is not required to cover all nodes (the nodes that are not present in the circuit are then self-looping).
Hence \gb{circuit}, with loops being simply ignored, basically represents \gb{subcircuit} (e.g., in \mzinc).
If ever you need a full circuit (i.e., without any loop), you have three solutions:
\begin{itemize}
\item simply specify \texttt{no\_self\_looping=True}
\item initially define the variables without the self-looping values, 
\item post unary constraints.
\end{itemize}

\paragraph{Mario.\label{pb:mario}} \index{Problems!Mario} From Amaury Ollagnier and Jean-Guillaume Fages, in the context of the 2013 Minizinc Competition:
``This models a routing problem based on a little example of Mario's day. 
Mario is an Italian Plumber and his work is mainly to find gold in the plumbing of all the houses of the neighborhood. 
 Mario is moving in the city using his kart that has a specified amount of fuel. Mario starts his day of work from his house 
 and always ends at his friend Luigi's house to have the supper. The problem here is to plan the best path for
 Mario in order to earn the most money with the amount of fuel of his kart. 
 From a more general point of view, the problem is to find a path in a graph:
\begin{itemize}
\item path endpoints are given (from Mario's to Luigi's)
\item the sum of weights associated to arcs in the path is restricted (fuel consumption)
\item the sum of weights associated to nodes in the path has to be maximized (gold coins)''
\end{itemize}

An example of data is given by the following JSON file:
\begin{json}
{
  "marioHouse": 0,
  "luigiHouse": 1,
  "fuelLimit": 2000,
  "houses":[
    {
      "fuelConsumption": [0,221,274,80,13,677,670,921,93,969,13,18,217,86,322],
      "gold":0
    },
    {
      "fuelConsumption":[0,0,702,83,813,679,906,246,35,529,79,528,451,242,712],
      "gold":0
    },
    ...
  ]
}
\end{json}

\vspace{-0.4cm}
\begin{figure}[h!]
\begin{center}
  \includegraphics[scale=0.8]{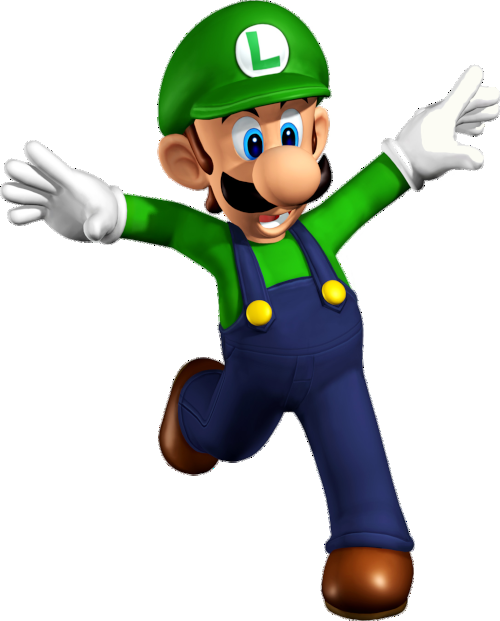}
\end{center}
\caption{Finding the Best Path for Mario. \tiny{(image from \href{https://pngimg.com/download/30515}{pngimg.com})} \label{fig:mario}}
\end{figure}

A \p3 model\footnote{This model is inspired by the one proposed by Ollagnier and Fages for the 2013 Minizinc Competition.} of this problem is given by the following file `Mario.py':

\begin{boxpy}\begin{python}
@\imp@

marioHouse, luigiHouse, fuelLimit, houses = data
fuels, golds = zip(*houses)  # using cp_array is not necessary since intern arrays
                             # have the right type (for the constraint Element)
nHouses = len(houses)

# s[i] is the house succeeding the ith house (itself if not part of the route)
s = VarArray(size=nHouses, dom=range(nHouses))

satisfy(
  # we cannot consume more than the available fuel
  Sum(fuels[i][s[i]] for i in range(nHouses)) <= fuelLimit,

  # Mario must make a tour (not necessarily complete)
  Circuit(s),
  
  # Mario's house succeeds Luigi's house
  s[luigiHouse] == marioHouse
  )

maximize(
  # maximizing collected gold
  Sum((s[i] != i) * golds[i] for i in range(nHouses) if golds[i] != 0)
)
\end{python}\end{boxpy}

When computing consumed fuel, note how some \gb{element} constraints are internally involved.
The lists \nn{fuels[i]} involved in these constraints can be directly indexed by variables (objects).
This is because the type of \nn{fuels[i]} is a \p3 subclass of 'list'; and this is automatically handled when loading the JSON file.
Suppose that we would have written instead: 
\begin{python}
fuels = [[v for v in house.fuelConsumption] for house in houses]
\end{python}
Here, \nn{fuels[i]} would be a simple 'list', and we would get an error when compiling.
In that case, to fix the problem, it is possible to call the \p3 function \nn{cp\_array}():
\begin{python}
fuels = [cp_array(v for v in house.fuelConsumption) for house in houses]
\end{python}
but of course, the code we have chosen for our model above is simpler.


\section{Meta-Constraint \gb{slide}} \label{sec:slide}

A general mechanism, or meta-constraint, that is useful to post constraints on sequences of variables is \gb{slide} \cite{BHHKW_slide}.
The scheme \gb{slide} ensures that a given constraint is enforced all along a sequence of variables.
To represent such sliding constraints in \x3, we simply build an element \xml{slide} containing a constraint template (for example, one for \xml{extension} or \xml{intension}) to indicate the abstract (parameterized) form of the constraint to be slided, preceded by an element \xml{list} that indicates the sequence of variables on which the constraint must slide.

For the semantics, we consider that $\gb{ctr}(\%0,\ldots,\%q-1)$ denotes the template of the constraint \gb{ctr} of arity $q$, and that $\gb{slide}^{circ}$ means the circular form of \gb{slide} 

\begin{boxse}
\begin{semantics}
$\gb{slide}(X,\gb{ctr}(\%0,\ldots,\%q-1))$, with $X=\langle x_0,x_1,\ldots \rangle$, iff  
  $\forall i : 0 \leq i \leq |X|-q, \gb{ctr}(x_i,x_{i+1},\ldots,x_{i+q-1})$ 
$\gb{slide}(X,\nm{os},\gb{ctr}(\%0,\ldots,\%q-1))$, with an offset $os$, iff  
  $\forall i : 0 \leq i \leq (|X|-q)/\nm{os}, \gb{ctr}(x_{i \times \nm{os}},x_{i \times \nm{os}+1},\ldots,x_{i \times \nm{os}+q-1})$ 
$\gb{slide}^{circ}(X,\gb{ctr}(\%0,\ldots,\%q-1))$ iff 
  $\forall i : 0 \leq i \leq |X|-q+1, \gb{ctr}(x_i,x_{i+1}\ldots,x_{(i+q-1)\%|X|})$ 
\end{semantics}
\end{boxse}

In \p3, to post a (meta-)constraint \gb{slide}, we must call the function \nn{Slide}() whose signature is:

\begin{python}
  def Slide(*args):
\end{python}

The specified arguments must correspond to a list (or a set, or even a generator) of sliding constraints.
The \p3 compiler will then attempt to build the \x3 sliding form.

It is important to note that \gb{slide} is interesting only if reasoning with the meta-constraint is stronger than reasoning with each constraint individually.
It is also interesting for generating compacter \x3 files (however, you can simply use the option \nn{-recognizeSlides}).
An illustration is given in Section \ref{sec:blackhole}.

\section{Constraint \gb{adhoc}} \label{sec:adhoc}

In some situations, you may want to introduce a particular constraint by arbitrarily defining its semantics (and arguments).
This may be a (new) global constraint that some user wants to try out (and implement in a solver).
In \p3, it is possible to handle such constraints, called \gb{adhoc}.
It suffices to choose a name (label) and create a dictionary with some expected arguments.
More precisely, in \p3, to post a constraint \gb{adhoc}, we must call the function \nn{Adhoc}() whose signature is:

\begin{python}
  def Adhoc(form, note=None, **d)
\end{python}

Here, \nn{form} is a label (string) indicating the form (name) of the adhoc constraint, \nn{note} is an optional argument, and \nn{d} is a dictionary containing all arguments of the adhoc constraint.

For illustrating \gb{adhoc} constraints, we propose to simulate two well-known constraints (of course, there is no real interest in doing so; this is just for simplicity).
Let us consider the following \p3 model: 

\begin{boxpy}\begin{python}
@\imp@

x = VarArray(size=10, dom=range(10))

satisfy(
  AllDifferent(x),

  x * range(1, 11) <= 165
)
\end{python}\end{boxpy}

When compiling this simple model, we obtain an \x3 file with the following content:

\begin{xcsp}
<instance format="XCSP3" type="CSP">
  <variables>
    <array id="x" size="[10]"> 0..9 </array>
  </variables>
  <constraints>
    <allDifferent> x[] </allDifferent>
    <sum>
      <list> x[] </list>
      <coeffs> 1 2 3 4 5 6 7 8 9 10 </coeffs>
      <condition> (le,165) </condition>
    </sum>
  </constraints>
</instance>
\end{xcsp}

The following \p3 model uses two constraints \gb{adhoc} for simulating \gb{AllDifferent} and \gb{Sum}:

\begin{boxpy}\begin{python}
@\imp@

x = VarArray(size=10, dom=range(10))

satisfy(
    Adhoc("myAllDiff", "this is my demo simulating alldifferent", list=x),

    Adhoc("mySum", list=x, coeffs=list(range(1, 11)), condition="(le,165)")
)
\end{python}\end{boxpy}

When compiling this revisited model, we obtain an \x3 file with the following content:

\begin{xcsp}
<instance format="XCSP3" type="CSP">
  <variables>
    <array id="x" size="[10]"> 0..9 </array>
  </variables>
  <constraints>
    <adhoc>
      <form> myAllDiff </form>
      <note> this is my demo simulating @alldifferent@ </note>
      <list> x[] </list>
    </adhoc>
    <adhoc>
      <form> mySum </form>
      <list> x[] </list>
      <coeffs> 1 2 3 4 5 6 7 8 9 10 </coeffs>
      <condition> (le,165) </condition>
    </adhoc>
  </constraints>
</instance>
\end{xcsp}

In the Java \x3 parser, the callback function is: 

\begin{json}
void buildCtrAdhoc(String id, String from, Map<String, Object> map);
\end{json}

In the constraint solver \ace, we can implement something like:

\begin{json}
public void buildCtrAdhoc(String id, String form, Map<String, Object> map) {
  if (form.equals("myAllDiff")) {
    XVarInteger[] list = (XVarInteger[]) map.get("list");
    problem.allDifferent(trVars(list));
  } else if (form.equals("mySum")) {
    XVarInteger[] list = (XVarInteger[]) map.get("list");
    int[] coeffs = (int[]) map.get("coeffs");
    Condition condition = (Condition) map.get("condition");
    problem.sum(trVars(list), coeffs, trVar(condition));
  }
}
\end{json}

It means that you can rather easily call the propagators (of possibly new constraints) you want by intercepting the  right forms of adhoc constraints.

\chapter{Logically Combining Constraints}\label{ch:logical}

When modeling, it happens that, for some problems, constraints must be logically combined.
For example, assuming that $x$ is a 1-dimensional array of variables, the statement: 
\begin{equation}\label{eq:meta1}
\gb{Sum}(x) > 10 \lor \gb{AllDifferent}(x)
\end{equation}
  enforces that the sum of values assigned to the variables of $x$ must be greater than 10, or the values assigned to $x$ variables must be all different.
  As another example, assuming that $y$ is an integer variable, the statement:
\begin{equation}\label{eq:meta2}
y \neq -1 \Rightarrow x[y] = 1
\end{equation}
enforces that when the value of $y$ is different from $-1$ then the value in the array $x$ at index $y$ must be equal to 1.

\bigskip
\noindent The question is: how can we deal with such situations?
The answer is multiple, as one can:
\begin{enumerate}
\item use the complex (control) structures \nn{If} and \nn{Match} (or their logical equivalent forms involving the Python operators '|', '\&' and '\textasciitilde'),
\item benefit from automatic reformulation mechanisms, 
\item exploit tabulation,
\item post meta-constraints (possible in \x3, but not in the perimeter of \x3-core),
\item use explicit reification (possible in \x3, but not in the perimeter of \x3-core).
\end{enumerate}

As indicated above, meta-constraints and explicit reification are very general mechanisms that are not in the perimeter of \x3-core.
Consequently, in all our illustrative examples, we avoid them.
The very good news is that the complex (control) structures \nn{If} and \nn{Match} (or their logical equivalent forms), together with automatic reformulation mechanisms are strong instruments that allow us to write models in a declarative manner,
while generating instances that stay within the limits of \x3-core (and mainly preserving the structure of the models).
Tabulation can also be relevant for simplifying some complex logical constrained expressions (and for making sometimes the solving process more efficient).
What is stated above is proven by the production of compact and highly readable models for more than 400 problems (available on our website) of various nature and origin.

\section{Using Complex Structures (\nn{If} and \nn{Match})}\label{sec:complex}

In \p3, one simple way of combining constraints is to build complex expressions/structures by means of the \p3 functions \nn{If} and \nn{Match} (introduced in Version 2.2).
First, do note that the first letter is capitalized, and so \nn{If} is different from the Python keyword \nn{if} and \nn{Match} is different from the Python keyword \nn{match}.
Let us start with the function \nn{If} whose signature is:
\begin{python}
   If(test, *testOthers, Then, Else=None):  
\end{python}

The first two parameters (test and testOthers) are positional, and allow us to indicate a (possibly singleton) sequence of expressions/conditions (typically, constraints).
Since Version 2.6, although it is not the usual case, it is also possible to use classical conditions (i.e., expressions which are not constraints), as those employed with the Python keyword \nn{if}.
Two named parameters follow the sequence of tests/conditions:
\begin{itemize}
\item \nn{Then}, which is required and must be named, is either a single constraint or a list (or tuple) of constraints,
\item \nn{Else}, which is optional (as \nn{None} is a default value), is either \nn{None}, a single constraint or a list (or tuple) of constraints.
\end{itemize}

\noindent The signature of the function \nn{Match} is:
\begin{python}
  Match(Expr, *, Cases):   
\end{python}

The first parameter is typically a variable of the model, or a constraining expression, i.e., an expression involving at least one variable of the model.
Since Version 2.6, it is also possible to use a classical expression (i.e., an expression which is not a constraint), as those employed with the Python keyword \nn{match}.
The second parameter is required, and must be named.
Its value must be a dictionary such that each key of this dictionary must be a value or an expression, and the associated value must be a constraint or a list (or tuple) of constraints.

We illustrate various uses of these two high-level functions by introducing a model for the following problem.

\paragraph{Arithmetic Target.\label{pb:arithmetic}} \index{Problems!Arithmetic Target} From Minizinc Challenge 2022.
The Arithmetic Problem is to determine the set of operations to be applied on a sequence of numbers (given as input) so as to get some output (target).

\begin{figure}[h!]
\begin{center}
  \includegraphics[scale=0.15]{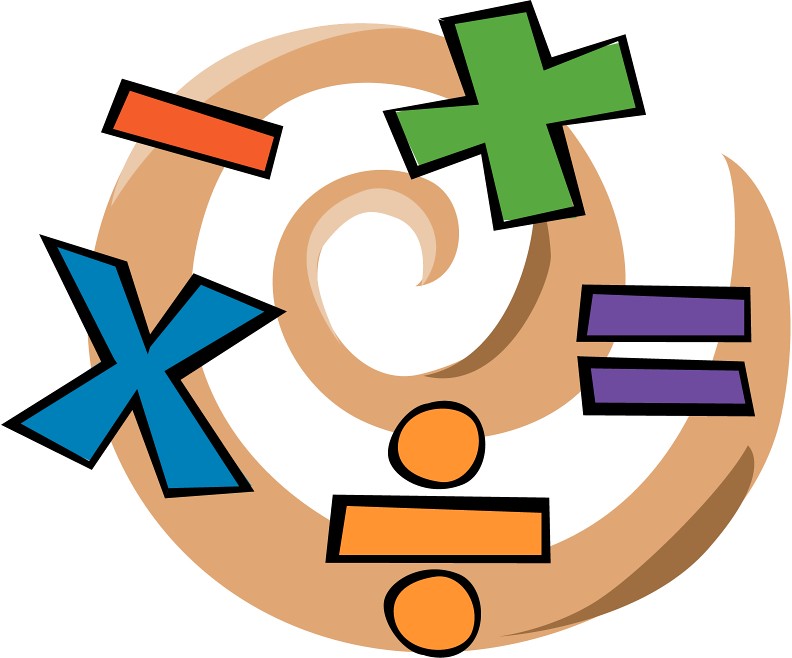} 
\end{center}
\caption{ \tiny{(image from \href{https://www.flickr.com/photos/85457190@N02/7832338798/in/photostream/}{flickr.com})} \label{fig:arithmetic}}
\end{figure}

An example of data is given by the following JSON file:
\begin{json}
{
  "numbers": [1, 2, 4, 6, 6, 7, 8, 9],
  "target": 814
}
\end{json}

A \p3 model (that can be seen as the close translation of the one submitted to the 2022 Minizinc challenge) is given by the following file `ArithmeticTarget.py':

\begin{boxpy}\begin{python}
@\imp@

numbers, target = data

n = len(numbers)
m = 2 * n  # - 1

M = range(1, m)
VAL, ADD, SUB, MUL, DIV, NO = Tokens = range(6)

# x[i] is the token associated with the ith node
x = VarArray(size=m, dom=lambda i: {-1} if i == 0 else Tokens)

# left[i] is the left child (or 0 if none) of the ith node
left = VarArray(size=m, dom=lambda i: {-1} if i == 0 else range(m))

# right[i] is the right child (or 0 if none) of the ith node
right = VarArray(size=m, dom=lambda i: {-1} if i == 0 else range(m))

# lowest[i] is the lowest descendant of the ith node
lowest = VarArray(size=m, dom=lambda i: {-1} if i == 0 else range(m))

# highest[i] is the highest descendant of the ith node
highest = VarArray(size=m, dom=lambda i: {-1} if i == 0 else range(m))

# index[i] is the index of the number associated with the ith node
index = VarArray(size=m, dom=lambda i: {-1} if i == 0 else range(n + 1))

# leaf[i] is 1 if the ith node is a leaf
leaf = VarArray(size=m, dom={0, 1})

# parent[i] is 1 if the ith node is a parent
parent = VarArray(size=m, dom={0, 1})

# unused[i] is the ith element is unused
unused = VarArray(size=m, dom={0, 1})

# the tree depth
depth = Var(dom=range(1, 2 * n))

# z1[i] is the value associated with the ith node
z1 = VarArray(size=m, dom=lambda i: {-1} if i == 0 else range(10 * target + 1))

# z2 is the number of used nodes
z2 = Var(dom=range(1, n + 1))

satisfy(
  # ensuring that the special value 0 appears n-1 times
  Count(index, value=0) == n - 1,

  # ensuring all indexes of numbers are different (except for the special value 0)
  AllDifferent(index, excepting=0),
   
  # ensuring that the tree has n leaves
  Count(x, value=VAL) == n,
   
  # computing the tree depth
  depth == highest[1],
   
  # computing the number of unused nodes
  2 * z2 - 1 == depth,
   
  # determining leaves
  [
    leaf[i] == conjunction(
      x[i] == VAL,
      left[i] == 0,
      right[i] == 0,
      highest[i] == i,
      lowest[i] == i,
      index[i] != 0
    ) for i in M
  ],

  # determining parents
  [
    parent[i] == conjunction(
      x[i] not in {VAL, NO},
      x[left[i]] != NO,
      x[right[i]] != NO,
      left[i] == i + 1,
      right[i] > left[i],
      right[i] == highest[left[i]] + 1,
      lowest[i] == i,
      highest[i] == highest[right[i]],
      index[i] == 0
    ) for i in M
  ],

  # determining unused elements
  [
    unused[i] == conjunction(
      x[i] in {NO, VAL},
      left[i] == 0,
      right[i] == 0,
      If(x[i] == VAL, Then=index[i] != 0),
      If(x[i] == NO, Then=index[i] == 0),
      lowest[i] == 0,
      highest[i] == 0
    ) for i in M
  ],

  # constraining leaves, parents and unused elements
  [
    If(
      i <= depth,
      Then=either(leaf[i], parent[i]),
      Else=unused[i]
    ) for i in M
  ],

  # computing values associated with all elements
  [
    Match(
      x[i],
      Cases={
        VAL: z1[i] == numbers[index[i]],
        ADD: z1[i] == z1[left[i]] + z1[right[i]],
        SUB: z1[i] == z1[left[i]] - z1[right[i]],
        MUL: z1[i] == z1[left[i]] * z1[right[i]],
        DIV: z1[i] * z1[right[i]] == z1[left[i]],
        NO: z1[i] == 0}
    ) for i in M
  ],

  # tag(symmetry-breaking)
  [
    # associativity
    [
      Match(
        x[i],
        Cases={
          ADD: x[left[i]] != ADD,
          MUL: x[left[i]] != MUL,
          SUB: x[left[i]] != SUB}
      ) for i in M
    ],

    # identity
    [
      [If(x[i] == ADD, Then=[z1[left[i]] != 0, z1[right[i]] != 0]) for i in M],
      [If(x[i] == MUL, Then=[z1[left[i]] != 1, z1[right[i]] != 1]) for i in M]
    ],
    
    # symmetry of addition and multiplication
    [If(x[i] in (ADD, MUL), Then=x[left[i]] <= x[right[i]]) for i in M],
    
    # distributivity of multiplication
    [
      If(
        x[i] in {ADD, SUB}, x[left[i]] == MUL, x[right[i]] == MUL,
        Then=[
          z1[left[left[i]]] != z1[left[right[i]]],
          z1[left[left[i]]] != z1[right[right[i]]],
          z1[right[left[i]]] != z1[left[right[i]]],
          z1[right[left[i]]] != z1[right[right[i]]]]
      ) for i in M
    ],
    
    # distributivity of division
    [
      If(
        x[i] in {ADD, SUB}, x[left[i]] == DIV, x[right[i]] == DIV,
        Then=z1[right[left[i]]] != z1[right[right[i]]]
      ) for i in M
    ],
    
    # conditions with respect to addition and multiplication
    [
     (
       If(x[i] == ADD, x[right[i]] == VAL, Then=x[left[i]] == VAL),
       If(x[i] == MUL, x[right[i]] == VAL, Then=x[left[i]] == VAL),
       If(
         x[i] == ADD, x[left[i]] == VAL, x[right[i]] == VAL,
         Then=index[left[i]] < index[right[i]]
       ),
       If(
         x[i] == MUL, x[left[i]] == VAL, x[right[i]] == VAL,
         Then=index[left[i]] < index[right[i]]
       ),
       If(
         x[i] == ADD, x[left[i]] == VAL, x[right[i]] == ADD, x[left[right[i]]]==VAL,
         Then=index[left[i]] < index[left[right[i]]]
       ),
       If(
         x[i] == MUL, x[left[i]] == VAL, x[right[i]] == MUL, x[left[right[i]]]==VAL,
         Then=index[left[i]] < index[left[right[i]]]
       )
     ) for i in M
    ],

    # all numbers with the same value should be assigned in sorted order
    [
      If(
        x[i] == VAL, x[j] == VAL, numbers[index[i]] == numbers[index[j]],
        Then=index[i] < index[j]
      ) for i, j in combinations(M, 2)
    ],
    
    # sorting nodes of equivalent value
    [If(z1[i] == z1[j], Then=x[i] >= x[j]) for i, j in combinations(M, 2)]
  ]
)

minimize(
   10 * abs(z1[1] - target) + z2
)
\end{python}\end{boxpy}

\smallskip
An illustration of how classical conditions can be used now (since Version 2.6) with the complex structure \nn{Match} is given by the model of the following problem.

\paragraph{FAPP (Frequency Assignment Problem with Polarization).} \index{Problems!Freq. Assign. Problem with Polarization}

The frequency assignment problem with polarization constraints (FAPP) is an optimization problem\footnote{This is an extended subject of the CALMA European project.} that was part of the ROADEF'01 challenge.
In this problem, there are constraints concerning frequencies and polarizations of radio links.
Progressive relaxation of these constraints is explored: the relaxation level is between 0 (no relaxation) and 10 (the maximum relaxation).  
For a complete description of the problem, the reader is invited to see \href{https://roadef.org/challenge/2001/files/fapp_roadef01_rev2_tex.pdf}{https://roadef.org/challenge/2001}.

As an illustration of data specifying an instance of this problem, we have:

\begin{json}
{
  "domains": [
    [1, 2, 3, ..., 99, 100],
    [25, 26, 27, 28, 29, 30, 31, 32, 33, 34, 35, 36, 37, 38, 39, 40],
    [55, 56, 57, 63, 64, 65]
  ],
  "frequencies": [1, 1, 0, 2, 0, 0, 2, 0, 0],
  "polarizations": [-1, 0, 0, 1, 0, 0, 0, 0, 0],
  "hards": [
    [1, 2, "F", "E", 36],
    [0, 1, "F", "E", 0],
    [6, 7, "F", "I", 0],
    [2, 3, "P", "E", 0],
    [1, 6, "P", "E", 0],
    [5, 6, "P", "I", 0]
  ],
  "softs": [
    [
      0,
      2,
      [46, 44, 42, 42, 40, 40, 35, 35, 35, 30, 20],
      [39, 37, 35, 35, 33, 33, 28, 28, 28, 23, 13]
    ],
    ...,
  ]
}
\end{json}

A \p3 model is given by the following file `FAPP.py':

\begin{boxpy}\begin{python}
@\imp@

domains, frequencies, polarizations, hard_constraints, soft_constraints = data 

frequencies = [domains[f] for f in frequencies]  # we skip the indirection

n, nSofts = len(frequencies), len(soft_constraints)

def soft_table(i, j, eqr, ner):  # building table for a soft constraint
   ... code for building T not included
   ... see https://github.com/xcsp3team/pycsp3-models/tree/main/realistic/FAPP
   return T

def domain_p(i):
   if polarizations[i] == 0:
      return {0,1}
   if polarizations[i] == 1:
      return {1}
   return {0}

# f[i] is the frequency of the ith radio-link
f = VarArray(size=n, dom=lambda i: frequencies[i])

# p[i] is the polarization of the ith radio-link
p = VarArray(size=n, dom=domain_p)

# k is the relaxation level to be optimized
k = Var(dom=range(12))

# v1[q] is 1 iff the qth constraint is violated when relaxing another level
v1 = VarArray(size=nSofts, dom={0, 1})

# v2[q] is the number of times the qth constraint is violated when relaxing 
v2 = VarArray(size=nSofts, dom=range(11))

satisfy(
   # imperative constraints
   [
      Match(
         (c3, c4),
         Cases={
            ("F", "E"): abs(f[i] - f[j]) == gap,
            ("P", "E"): abs(p[i] - p[j]) == gap,
            ("F", "I"): abs(f[i] - f[j]) != gap,
            ("P", "I"): abs(p[i] - p[j]) != gap
         }
      ) for (i, j, c3, c4, gap) in hard_constraints
   ],    

   # soft radio-electric compatibility constraints
   [
      Table(
         scope=(f[i], f[j], p[i], p[j], k, v1[q], v2[q]),
         supports=soft_table(i, j, eqr, ner)
      ) for q, (i, j, eqr, ner) in enumerate(soft_constraints)
   ]
)

minimize(
   k * (10 * nSofts ** 2) + Sum(v1) * (10 * nSofts) + Sum(v2)
)
\end{python}\end{boxpy}

The model, above, corresponds to the main variant that can be found at \href{https://github.com/xcsp3team/PyCSP3-models/blob/main/realistic/FAPP/FAPP.py}{github.com\-/xcsp3team/PyCSP3-models}. 
Another variant (``aux'') was used for the 2025 \x3 competition. 
Here, note how the structure \nn{Match} allows us to post hard constraints according to the values of specific pairs (c3,c4) of characters.

\section{Benefiting from Automatic Reformulation Mechanisms}\label{sec:reformulation}

In Section \ref{sec:meta}, we will see that meta-constraint operators can be applied by calling specific \p3 functions \nn{And()}, \nn{Or()}, $\dots$
However, this is not compatible with the perimeter of \x3-core.
So, it is better to use the complex structures \nn{If} and \nn{Match}, as well as the classical Python operators '|', '\&' and '\textasciitilde', because some reformulation mechanisms are automatically applied when compiling complex logic-based expressions (so as to stay within the limits of \x3-core).  
These operators, which are redefined in \p3, can be used to build \gb{intension} constraints, but also more complex forms obtained by logically combining (global) constraints. 
Let us try this with the following \p3 model:
\begin{boxpy}\begin{python}
@\imp@

x = VarArray(size=4, dom=range(4))
y = Var(dom=range(-1, 4))

satisfy(
  (Sum(x) > 10) | AllDifferent(x),

  If(y != -1, Then=x[y] == 1)
)
\end{python}\end{boxpy}

Note that for the first complex constraint, instead of using the operator '|', we could equivalently write \verb!either(Sum(x) > 10, AllDifferent(x))!, or \verb!If(Sum(x) <= 10, Then=AllDifferent(x))!, or even \verb!imply(Sum(x) <= 10, AllDifferent(x))!.
Also, we could equivalently write for the second constraint \verb!(y == -1) | (x[y] == 1)!, or \verb:imply(y != -1, x[y] == 1):, instead of using the function \nn{If()}. 

In any case, when compiling, we obtain the following \x3 instance:

\begin{xcsp}
<instance format="XCSP3" type="CSP">
  <variables>
    <array id="x" size="[4]"> 0..3 </array>
    <var id="y"> -1..3 </var>
    <array id="aux" note="auxiliary variables automatically introduced" size="[4]">
      <domain for="aux[0]"> 0..12 </domain>
      <domain for="aux[1]"> 1..4 </domain>
      <domain for="aux[2] aux[3]"> 0..3 </domain>
    </array>
  </variables>
  <constraints>
    <extension>
      <list> y aux[2] </list>
      <supports> (-1,*)(0,0)(1,1)(2,2)(3,3) </supports>
    </extension>
    <sum>
      <list> x[] </list>
      <condition> (eq,aux[0]) </condition>
    </sum>
    <nValues>
      <list> x[] </list>
      <condition> (eq,aux[1]) </condition>
    </nValues>
    <intension> @or@(gt(aux[0],10),eq(aux[1],4)) </intension>
    <intension> imp(ne(i,-1),eq(aux[3],1)) </intension>
    <element>
      <list> x[] </list>
      <index> aux[2] </index>
      <value> aux[3] </value>
    </element>
  </constraints>
</instance>
\end{xcsp}

One can observe that four auxiliary variables have been automatically introduced.
The generated \x3 instance has been the subject of some reformulation rules which, importantly, allow us to remain within the perimeter of \x3-core.
Actually, the main reformulation rule is the following: if a condition-based global constraint is involved in a complex formulation, it can be replaced by an auxiliary variable while ensuring apart that what is 'computed' by the constraint is equal to the value of the new introduced variable.
For example, \verb!Sum(x) > 10! becomes \verb!aux[0] > 10! while posting \verb!Sum(x) == aux[0]! apart (after having introduced the auxiliary variable \verb!aux[0]!).
By proceeding that way, we obtain classical (i.e., non complex) \gb{intension} constraints.

Many global constraints are {\em condition-based}, i.e., involve a condition in their statements.
This is the case for:
\begin{itemize}
 \item \gb{AllDifferent}, since $\gb{AllDifferent}(x)$ is equivalent to $\gb{NValues}(x) = |x|$
 \item \gb{AllEqual}, since $\gb{AllEqual}(x)$ is equivalent to $\gb{NValues}(x) = 1$
 \item \gb{Sum}
 \item \gb{Count}
 \item \gb{NValues}
 \item \gb{Minimum} and \gb{Maximum}
 \item \gb{Element}
 \item \gb{Cumulative}
 \item \gb{BinPacking} (first form)
 \item \gb{Knapsack}
\end{itemize}

\medskip\noindent In the rest of this section, three additional illustrations are given.

\paragraph{Stable Marriage.\label{pb:stable}}\index{Problems!Stable Marriage}
See \href{https://en.wikipedia.org/wiki/Stable_marriage_problem/}{Wikipedia}. 
Consider two groups of men and women who must marry.
Consider that each person has indicated a ranking for her/his possible spouses.
The problem is to find a matching between the two groups such that the marriages are stable.
A marriage between a man $m$ and a woman $w$ is stable iff:
\begin{itemize}
\item whenever $m$ prefers another woman $o$ to $w$, $o$ prefers her husband to $m$
\item whenever $w$ prefers another man $o$ to $m$, $o$ prefers his wife to $w$
\end{itemize}

In 1962, David Gale and Lloyd Shapley proved that, for any equal number $n$ of men and women,
it is always possible to make all marriages stable, with an algorithm running in $O(n^2)$.
Nevertheless, this problem remains interesting as it shows how a nice and compact model can be written.

\begin{figure}[h!]
\begin{center}
  \includegraphics[scale=0.08]{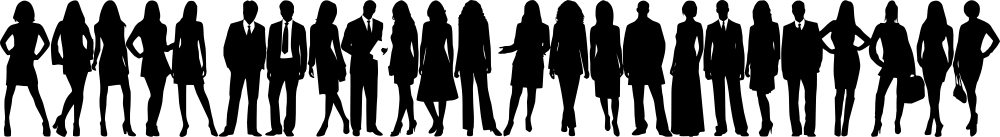}
\end{center}
\caption{Marrying People. \tiny{(image from \href{https://freesvg.org/group-of-men-and-women-silhouettes}{freesvg.org})} \label{fig:people}}
\end{figure}

An example of data is given by the following JSON file (here, $n=5$) :

\begin{json}
{
  "women_rankings": [[1,2,4,3,5],[3,5,1,2,4],[5,4,2,1,3],[1,3,5,4,2],[4,2,3,5,1]],
  "men_rankings": [[5,1,2,4,3],[4,1,3,2,5],[5,3,2,4,1],[1,5,4,3,2],[4,3,2,1,5]]
}
\end{json}

A \p3 model of this problem is given by the following file `StableMarriage.py':

\begin{boxpy}\begin{python}
@\imp@

wr, mr = data  # ranking by women and men
n = len(wr)
Men, Women = range(n), range(n)

# x[m] is the wife of the man m
x = VarArray(size=n, dom=Women)

# y[w] is the husband of the woman w
y = VarArray(size=n, dom=Men)

satisfy(
  # spouses must match
  Channel(x, y),

  # whenever m prefers another woman o to his wife, o prefers her husband to m
  [
    If(
      m_rankings[m][o] < m_rankings[m][x[m]],
      Then=w_rankings[o][y[o]] < w_rankings[o][m]
    ) for m in Men for o in Women
  ],
  
  # whenever w prefers another man o to her husband, o prefers his wife to w
  [
    If(
      w_rankings[w][o] < w_rankings[w][y[w]],
      Then=m_rankings[o][x[o]] < m_rankings[o][w]
    ) for w in Women for o in Men
  ]
)
\end{python}\end{boxpy}

Note how the last two lists (groups) of constraints combine \gb{element} constraints.
When compiling, auxiliary variables will then be introduced.
Note that a cache is used to avoid generating equivalent auxiliary variables.

\begin{figure}[h!]
\begin{center}
  \includegraphics[scale=0.15]{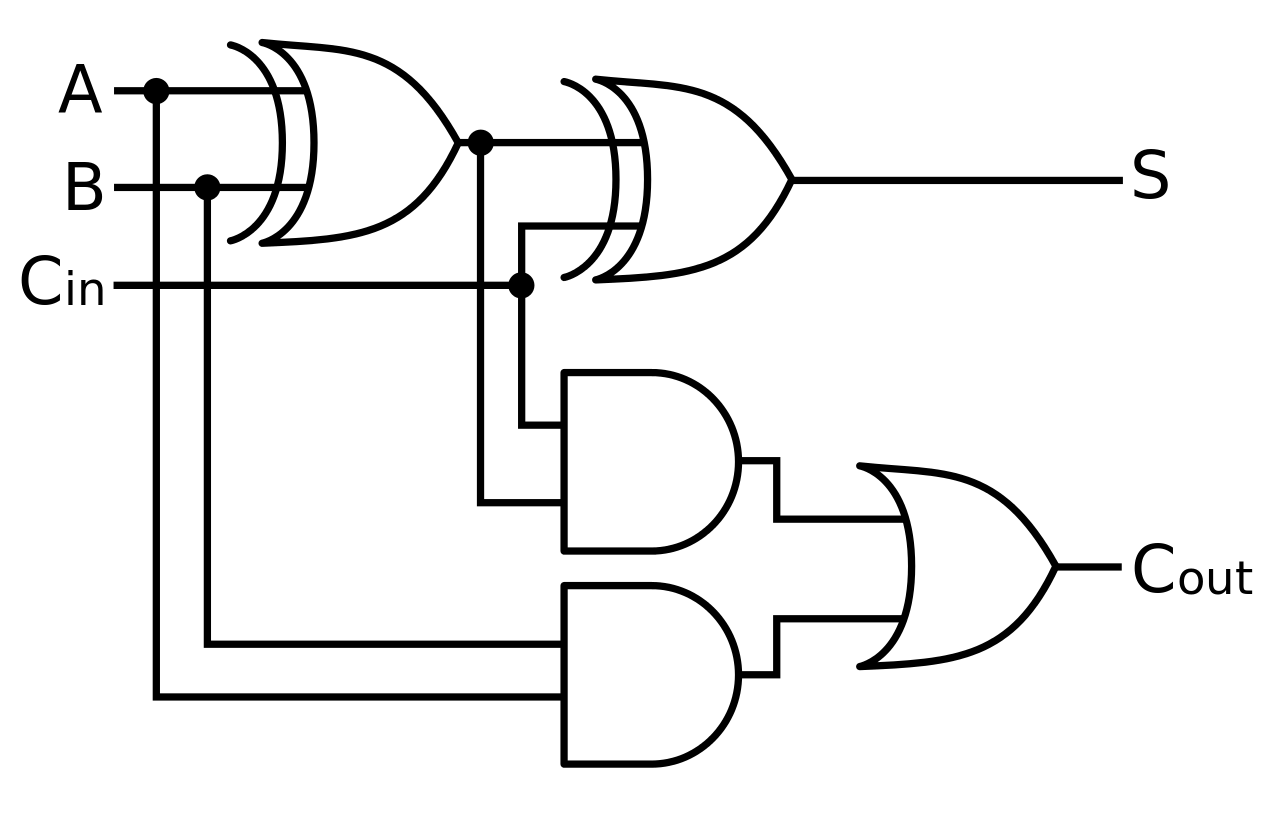}
\end{center}
\caption{Full Adder. \tiny{(image from \href{https://commons.wikimedia.org/wiki/File:Full-adder.svg}{commons.wikimedia})} \label{fig:adder}}
\end{figure}

\paragraph{Diagnosis.\label{pb:diagnosis}}\index{Problems!Diagnosis}
From \href{https://www.csplib.org/Problems/prob042/}{CSPLib}:
``
Model-based diagnosis can be seen as taking as input a partially parameterized structural description of a system and a set of observations about that system.
Its output is a set of assumptions which, together with the structural description, logically imply the observations, or that are consistent with the observations.
Diagnosis is usually applied to combinational digital circuits, seen as black-boxes where there is a set of controllable input bits but only a set of primary outputs is visible.
The problem is to find the set of all (minimal) internal faults that explain an incorrect output (different than the modelled, predicted, output), given some input vector.
The possible faults consider the usual stuck-at fault model, where faulty circuit gates can be either stuck-at-0 or stuck-at-1, respectively outputting value 0 or 1 independently of the input.
As an example, for the full-adder circuit displayed in Figure \ref{fig:adder}, if we assume that the input is $A=0, B=0, c_{in}=0$ and the observed output is $S=1, C_{out}=0$ (although it should be $S=0, C_{out}=0$),
the single faults that explain the incorrect output are the first XOR gate stuck-at-1 or the second XOR gate stuck-at-1.''

An example of data is given by the following JSON file:

\begin{json}
{
  "functions": [[[0,1],[1,1]], [[0,0],[0,1]],[[0,1],[1,0]]],
  "gates": [
    null,
    null,
    {"f": 2, "in1": 0, "in2": 0, "out": -1},
    {"f": 1, "in1": 0, "in2": 0, "out": -1},
    {"f": 2, "in1": 0, "in2": 2, "out": 1},
    {"f": 1, "in1": 0, "in2": 2, "out": -1},
    {"f": 0, "in1": 3, "in2": 5, "out": 0}
  ]
}
\end{json}

Logical functions are given under their matrix forms; here, we have the functions OR (index 0), AND (index 1), and XOR (index 2).
Each gate is given its logical function (index given by 'f'), its input (0 for False, 1 for True and the index of another gate otherwise), and its observed output (if any).
A \p3 model of this problem is given by the following file `Diagnosis.py':

\begin{boxpy}\begin{python}
@\imp@

# note that the first two gates are special
# they are inserted for reserving indexes 0 and 1 (for false and true)
funcs, gates = data  
nGates = len(gates)

def apply(gate):
  return functions[gate.f][y[gate.in1]][y[gate.in2]]

# x[i] is -1 if the ith gate is not faulty (otherwise 0 or 1 when stuck at 0 or 1)
x = VarArray(size=nGates, dom=lambda i: {-1} if i < 2 else {-1, 0, 1})

# y[i] is the (possibly faulty) output of the ith gate
y = VarArray(size=nGates, dom=lambda i: {i} if i < 2 else {0, 1})

satisfy(
  # ensuring that y is coherent with the observed output
  [y[i] == j for i in range(2, nGates) if (j := gates[i].out) != -1],

  # ensuring that each gate either meets expected outputs based on its function
  # or is broken (either stuck on or off)
  [
    If(
      y[i] != x[i],
      Then=[
        y[i] == apply(gates[i]),
        x[i] == -1
      ]
    ) for i in range(2, nGates)
  ]
)
\end{python}\end{boxpy}

Note how the last list (group) of constraints involves \gb{element} constraints under their matrix forms.
Once again, when compiling, auxiliary variables will be introduced, and the generated \x3 instances will be guaranteed to be within \x3-core.

\paragraph{Vellino's Problem.\label{pb:vellino}} \index{Problems!Vellino} From {\em Constraint Programming in OPL} by L. Michel, L. Perron, and J.-C. R\'egin, CP'99:
this problem involves putting components of different materials (glass, plastic, steel, wood, copper) into bins of various types (identified by red, blue, green colors), subject to capacity (each bin type has a maximum capacity) and compatibility constraints.
Every component must be placed into a bin and the total number of used bins must be minimized.
The compatibility constraints are:
\begin{itemize}
\item red bins cannot contain plastic or steel
\item blue bins cannot contain wood or plastic
\item green bins cannot contain steel or glass
\item red bins contain at most one wooden component
\item green bins contain at most two wooden components
\item wood requires plastic
\item glass excludes copper
\item copper excludes plastic
\end{itemize}
See also \href{https://www.csplib.org/Problems/prob116/}{CSPLib--Problem 116}.

\begin{figure}[h]
      \begin{center}
        \includegraphics[width=1.8cm]{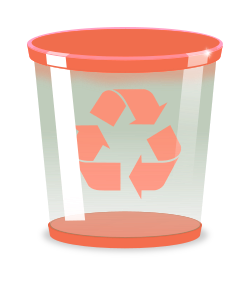} \includegraphics[width=1.8cm]{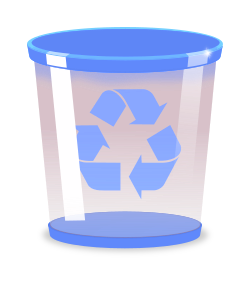} \includegraphics[width=1.8cm]{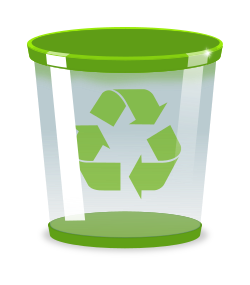}  
    
      \end{center}
        \caption{Red, Blue and Green Bins for Vellino's Problem. \tiny{(image from \href{https://freesvg.org/trash-bin}{freesvg.org})}} \label{fig:vellino} 
\end{figure}

\bigskip
An example of data is given by the following JSON file:

\begin{json}
{
  "capacities": [3,1,4],
  "demands": [1,2,1,3,2]
}
\end{json}

Capacities are orderly given for red, blue and green bins, and demands (numbers of components) are orderly given for glass, plastic, steel, wood, and copper materials.
A \p3 model of this problem is given by the following file `Vellino.py':

\begin{boxpy}\begin{python}
@\imp@

# 0 is a special color, 'Unusable', to be used for any empty bin
Unusable, Red, Blue, Green = BIN_COLORS = 0, 1, 2, 3  
Glass, Plastic, Steel, Wood, Copper = MATERIALS = 0, 1, 2, 3, 4
nColors, nMaterials = len(BIN_COLORS), len(MATERIALS)

capacities, demands = data
capacities.insert(0, 0)  # unusable bins have capacity 0
maxCapacity, nBins = max(capacities), sum(demands)

# c[i] is the color of the ith bin
c = VarArray(size=nBins, dom=range(nColors))

# p[i][j] is the number of components of the jth material put in the ith bin
p = VarArray(size=[nBins, nMaterials],
             dom=lambda i, j: range(min(maxCapacity, demands[j]) + 1))

satisfy(
  # every bin with a real colour must contain something, and vice versa
  [(c[i] == Unusable) == (Sum(p[i]) == 0) for i in range(nBins)],

  # all components of each material are spread across all bins
  [Sum(p[:, j]) == demands[j] for j in range(nMaterials)],
  
  # the capacity of each bin is not exceeded
  [Sum(p[i]) <= capacities[c[i]] for i in range(nBins)],
  
  # handling compatibility of materials
  [
    Match(
      c[i],
      Cases={
        Red: [p[i][Plastic] == 0, p[i][Steel] == 0, p[i][Wood] <= 1],
        Blue: [p[i][Wood] == 0, p[i][Plastic] == 0],
        Green: [p[i][Steel] == 0, p[i][Glass] == 0, p[i][Wood] <= 2]
      }
    ) for i in range(nBins)
  ],

  # wood requires plastic
  [If(p[i][Wood] > 0, Then=p[i][Plastic] > 0) for i in range(nBins)],
  
  # glass excludes copper
  [If(p[i][Glass] > 0, Then=p[i][Copper] == 0) for i in range(nBins)],
  
  # copper excludes plastic
  [If(p[i][Copper] > 0, Then=p[i][Plastic] == 0) for i in range(nBins)],
  
  # tag(symmetry-breaking)
  [LexIncreasing(p[i], p[i + 1]) for i in range(nBins - 1)]
)
  
minimize(
  # minimizing the number of used bins
  Sum(c[i] != Unusable for i in range(nBins))
)   
\end{python}\end{boxpy}

Note how the first list (group) of constraints involve \gb{sum} constraints in a more general expression.
Automatic reformulation at compilation time will then be applied.
Some other lists in the model also involve \gb{element} constraints that will be reformulated.

\section{Using Tabulation}\label{sec:tabling}

In this section, we show with two illustrations how modeling with tables can be relevant to logically combine involved constraints.

First, let us recall that table constraints are important in constraint programming because
\begin{enumerate*}[label=(\roman*)]
\item they are easily handled by end-users of constraint systems,
\item they can be perceived as a universal modeling mechanism since any constraint can theoretically be expressed in tabular form (although this may lead to time/space explosion),
\item sometimes, they happen to be simple and natural choices for dealing with tricky situations: this is the case when no adequate (global) constraint exists or when a logical combination of (small) constraints must be represented as a unique table constraint for efficiency reasons.
\end{enumerate*}
If ever needed, another argument showing the importance of universal structures like tables, and also diagrams, is the rising of (automatic) tabulation techniques, i.e., the process of converting sub-problems into tables, by hand, using heuristics \cite{AGJMNS_automatic} or by annotations \cite{DBCFM_auto}.

\paragraph{Amaze.\label{pb:amaze}} \index{Problems!Amaze} From Minizinc, Challenge 2012.
Given a grid containing $p$ pairs of numbers (ranging from 1 to $p$), connect the pairs (1 to 1, 2 to 2, $\dots$, $p$ to $p$) by drawing a line horizontally and vertically, but not diagonally.
The lines must never cross.

An example of data is given by the following JSON file:
\begin{json}
{
  "n": 5,
  "m": 5,
  "points": [
    [[3,4], [5,1]],
    [[2,2], [4,2]]
  ]
}
\end{json}
Here, we have a grid of size $5 \times 5$ with value 1 in cells at index $(3,4)$ and $(5,1)$, and value 2 in cells at index $(2,2)$ and $(4,2)$; here, $p=2$, and indexing is assumed to start at 1. 
For representing a solution, we can fill up the grid with either value 0 (empty cell) or a line number (value from 1 to $p$).  
For example, here is a solution corresponding to the data given above (with a border put all around the grid).  

\begin{footnotesize}\begin{verbatim}
    [
      [0, 0, 0, 0, 0, 0, 0],
      [0, 0, 0, 0, 0, 0, 0],
      [0, 0, 2, 0, 0, 0, 0],
      [0, 0, 2, 0, 1, 0, 0],
      [0, 0, 2, 0, 1, 0, 0],
      [0, 1, 1, 1, 1, 0, 0],
      [0, 0, 0, 0, 0, 0, 0]
    ]
\end{verbatim}
\end{footnotesize}

When analysing this problem, one can find that any non-empty cell (i.e., any cell with a value different from 0) is such that if it is not an end-point then it has exactly two 
 horizontal or vertical neighbours with the same value.
The piece of code in Minizinc to handle such constraints is:
 
\begin{footnotesize}
\begin{verbatim}
  % Return true if the given point is one of the end points of a path. 
  %
  test is_end_point(int: i, int: j) =
    exists (v in 1..N) (
      end_points_start_x[v] = i /\ end_points_start_y[v] = j \/
      end_points_end_x[v] = i /\ end_points_end_y[v] = j
  );

  % Constrain any non-empty cell that is not an end-point to have exactly two
  % horizontal or vertical neighbours of the same value.
  %
  constraint forall(i in 1..n, j in 1..m) (
    if is_end_point(i, j) then	
      true
    else
      x[i, j] != 0 -> count([x[i, j+1], x[i+1, j], x[i, j-1], x[i-1, j]], x[i, j], 2) 
    endif
  );
\end{verbatim}
\end{footnotesize}

As an alternative, table constraints can be posted, leading to the \p3 model given by the following file `Amaze.py':

\begin{boxpy}\begin{python}
@\imp@

n, m, points = data  # points[v] gives the pair of points for value v+1
nValues = len(points) + 1  # number of pairs of points + 1 (for 0)

free_cells = [(i, j)  for i in range(1, n + 1) for j in range(1, m + 1)
              if [i, j] not in [p for pair in points for p in pair]]

T = {(0, ANY, ANY, ANY, ANY)} |
    {tuple(ne(v) if k in (i, j) else v for k in range(5))
      for i, j in combinations(range(1, 5), 2) for v in range(1, nValues)}

def domain_x(i, j):
  return {0} if i in {0, n + 1} or j in {0, m + 1} else range(nValues)

# x[i][j] is the value at row i and column j (a boundary is put around the board).
x = VarArray(size=[n + 2, m + 2], dom=domain_x)

satisfy(
  # putting two occurrences of each value on the board
  [x[i][j] == v for v in range(1, nValues) for i, j in points[v - 1]],

  # each fixed cell has exactly one neighbour with the same value
  [ExactlyOne(x.beside(i, j), value=v) for v in range(1, nValues)
     for i, j in points[v - 1]],

  # each free cell either contains 0 or has exactly two neighbours with its value
  [(x[i][j], x[i - 1][j], x[i + 1][j], x[i][j - 1], x[i][j + 1]) in T
     for i, j in free_cells]
)

minimize(
  Sum(x)
)
\end{python}\end{boxpy}

Each table indicates the possible combinations of values for exactly 5 variables (forming a cross shape in the grid).
We use the function \nn{ne} to stand for any value 'not equal' to the specified parameter (in the near future, we shall let the user the opportunity to generate so-called hybrid (or smart) tables \cite{MDL_smart}).
For example, we obtain the following group of constraints with respect to the above data:

\begin{xcsp}
  <group>
    <extension>
      <list> %... </list>
      <supports>
        (0,*,*,*,*)(1,0,0,1,1)(1,0,1,0,1)(1,0,1,1,0)(1,0,1,1,2)(1,0,1,2,1)(1,0,2,1,1)
        (1,1,0,0,1)(1,1,0,1,0)(1,1,0,1,2)(1,1,0,2,1)(1,1,1,0,0)(1,1,1,0,2)(1,1,1,2,0)
        (1,1,1,2,2)(1,1,2,0,1)(1,1,2,1,0)(1,1,2,1,2)(1,1,2,2,1)(1,2,0,1,1)(1,2,1,0,1)
        (1,2,1,1,0)(1,2,1,1,2)(1,2,1,2,1)(1,2,2,1,1)(2,0,0,2,2)(2,0,1,2,2)(2,0,2,0,2)
        (2,0,2,1,2)(2,0,2,2,0)(2,0,2,2,1)(2,1,0,2,2)(2,1,1,2,2)(2,1,2,0,2)(2,1,2,1,2)
        (2,1,2,2,0)(2,1,2,2,1)(2,2,0,0,2)(2,2,0,1,2)(2,2,0,2,0)(2,2,0,2,1)(2,2,1,0,2)
        (2,2,1,1,2)(2,2,1,2,0)(2,2,1,2,1)(2,2,2,0,0)(2,2,2,0,1)(2,2,2,1,0)(2,2,2,1,1)
      </supports>
    </extension>
    <args> x[2][3] x[1][3] x[3][3] x[2][2] x[2][4] </args>
    <args> x[2][4] x[1][4] x[3][4] x[2][3] x[2][5] </args>
    <args> x[3][2] x[2][2] x[4][2] x[3][1] x[3][3] </args>
    <args> x[3][3] x[2][3] x[4][3] x[3][2] x[3][4] </args>
    <args> x[4][3] x[3][3] x[5][3] x[4][2] x[4][4] </args>
    <args> x[4][4] x[3][4] x[5][4] x[4][3] x[4][5] </args>
  </group>
\end{xcsp}


\paragraph{Layout Problem.\label{pb:layout}}\index{Problems!Layout}\label{sec:layout}
From {\em Exploiting symmetries within constraint satisfaction search} by P. Meseguer and C. Torras, Artificial Intelligence 129, 2001: 
given a grid, we want to place a number of pieces such that every piece is completely included in the grid and no overlapping occurs between pieces.
An example is given in Figure \ref{fig:layout}, where three pieces have to be placed inside the proposed grid.
See also \href{https://www.csplib.org/Problems/prob132}{CSPLib--Problem 132}.

\begin{figure}[h]
  \centering
  \begin{subfigure}{0.2\textwidth}
    \centering
    \begin{tikzpicture}[thick,scale=0.5]
      \draw[fill=black!20!white] (0,-1) grid (1,6) rectangle (0,-1) ;
      \draw[fill=black!20!white] (1,3) grid (2,6) rectangle (1,3) ;
      \draw[fill=black!20!white] (2,4) grid (4,6) rectangle (2,4);
    \end{tikzpicture}
    \caption{Grid\label{fig:grid}}
  \end{subfigure}%
  \begin{subfigure}{0.2\textwidth}
    \centering
    \begin{tikzpicture}[thick,scale=0.5] 
      \draw[dgreen] (0,1) grid (1,6) rectangle (0,1) ;
      \draw[dgreen] (1,5) grid (2,6) rectangle (1,5) ;
      \node[draw=none] at (0,-1) {};
    \end{tikzpicture}
    \caption{Piece 1\label{fig:piece1}}
  \end{subfigure}
  \begin{subfigure}{0.2\textwidth}
    \centering
    \begin{tikzpicture}[thick,scale=0.5] 
      \draw[dblue] (0,4) grid (2,6) rectangle (0,4) ;
      \node[draw=none] at (0,-1) {};
    \end{tikzpicture}
    \caption{Piece 2\label{fig:piece2}}
  \end{subfigure}
   \begin{subfigure}{0.2\textwidth}
    \centering
    \begin{tikzpicture}[thick,scale=0.5]
      \draw[dred] (0,4) grid (2,6) rectangle (0,4) ;
      \node[draw=none] at (0,-1) {};
    \end{tikzpicture}
    \caption{Piece 3\label{fig:piece3}}
  \end{subfigure}
  \caption{Layout Problem \label{fig:layout}}
\end{figure}

An example of data (corresponding to the problem instance of Figure \ref{fig:layout}) is given by the following JSON file:

\begin{json}
{
  "grid": [[1,1,1,1],[1,1,1,1],[1,1,0,0],[1,0,0,0] ,[1,0,0,0],[1,0,0,0],[1,0,0,0]],
  "shapes": [
    [[1,1], [1,0],[1,0],[1,0],[1,0]],
    [[1,1], [1,1]],
    [[1,1], [1,1]]
  ]
}
\end{json}

Note how the grid and the pieces are represented by two-dimensional matrices (0 being used to discard some cells).
A solution can be represented by storing in each cell of the grid either the index of a piece or -1.
For example, here is a solution corresponding to the data given above.

\begin{footnotesize}\begin{verbatim}
    [
      [1,  1,  2,  2],
      [1,  1,  2,  2],
      [0,  0, -1, -1],
      [0, -1, -1, -1],
      [0, -1, -1, -1],
      [0, -1, -1, -1],
      [0, -1, -1, -1]
    ]
\end{verbatim}
\end{footnotesize}

A model for this layout problem in language Essence is:

\begin{footnotesize}
\begin{verbatim}
  given n, m, nShapes : int(1..)
 
  letting Shape be domain int(1..nShapes),
        N be domain int(1..n),
        M be domain int(1..m),
        Cell be domain tuple (N,M)
 
  $ grid: the set of pairs of i and j coordinates that make up the grid shape
  $ form: the form of each shape, as a set of pairs of i and j coordinates
  given grid : set of Cell,
      form : function (total) Shape --> set of Cell
 
  $ x: a mapping from each cell in the grid to the shape id occupying it
  find x : function Cell --> Shape
 
  such that
  $ only cells in the grid are part of the layout
    forAll c in defined(x) . c in grid,
  $ the cells that map to a shape match the shape's form.
  $ this is long and complicated because we need the minimum i and j coordinates
  $ (min(sn) and min(sm)) that map to each shape, ... 
    forAll s : Shape . exists sn : set of N . exists sm : set of M .
        (forAll (i,j) : Cell . i in sn /\ j in sm <-> (i,j) in preImage(x,s)) /\
        forAll (i,j) in form(s) . x((min(sn) + i, min(sm) + j)) = s,
  $ a shape has exactly the right number of cells mapping to it
    forAll s : Shape . |form(s)| = |preImage(x,s)|
\end{verbatim}
\end{footnotesize}

This model is elegant (Essence handles rather high level mathematical objects),
but its compilation may possibly yield complex instances. 
Posting table constraints substantially simplifies this task.
Of course, this can be performed in Essence.
A \p3 model based on table constraints is given by the following file `Layout.py':

\begin{boxpy}\begin{python}
@\imp@

grid, shapes = data
n, m, nShapes = len(grid), len(grid[0]), len(shapes)

def domain_x(i, j):
  return {-1} if grid[i][j] == 0 else range(nShapes)

def domain_y(k):
  shape, height, width = shapes[k], len(shapes[k]), len(shapes[k][0]) 
  return [i * m + j for i in range(n - height + 1) for j in range(m - width + 1)
            if all(grid[i + gi][j + gj] == 1 or shape[gi][gj] == 0
              for gi in range(height) for gj in range(width))]

def table(k):
  shape, height, width = shapes[k], len(shapes[k]), len(shapes[k][0])
  tbl = []
  for v in domain_y(k):
    i, j = v // m, v % m
    t = [(i + gi) * m + (j + gj) for gi in range(height) for gj in range(width)
          if shape[gi][gj] == 1]
    tbl.append((v,) + tuple(k if w in t else ANY for w in range(n * m)))
  return tbl

# x[i][j] is the index of the shape occupying the cell at row i and column j, or -1
x = VarArray(size=[n, m], dom=domain_x) 

# y[k] is the base cell index in the grid where we start putting the kth shape
y = VarArray(size=nShapes, dom=domain_y)

satisfy(
  # putting shapes in the grid
  (y[k], x) in table(k) for k in range(nShapes)
)
\end{python}\end{boxpy}

As an illustration, the table constraints that are generated from the above data are:

\begin{xcsp}
  <block note="putting shapes in the grid">
    <extension>
      <list> y[0] x[][] </list>
      <supports> (0,0,0,*,*,0,*,*,*,0,*,*,*,0,*,*,*,0,*,*,*,*,*,*,*,*,*,*,*)(4,*,*,*,*,0,0,*,*,0,*,*,*,0,*,*,*,0,*,*,*,0,*,*,*,*,*,*,*)(8,*,*,*,*,*,*,*,*,0,0,*,*,0,*,*,*,0,*,*,*,0,*,*,*,0,*,*,*) </supports>
    </extension>
    <extension>
      <list> y[1] x[][] </list>
      <supports> (0,1,1,*,*,1,1,*,*,*,*,*,*,*,*,*,*,*,*,*,*,*,*,*,*,*,*,*,*)(1,*,1,1,*,*,1,1,*,*,*,*,*,*,*,*,*,*,*,*,*,*,*,*,*,*,*,*,*)(2,*,*,1,1,*,*,1,1,*,*,*,*,*,*,*,*,*,*,*,*,*,*,*,*,*,*,*,*)(4,*,*,*,*,1,1,*,*,1,1,*,*,*,*,*,*,*,*,*,*,*,*,*,*,*,*,*,*) </supports>
    </extension>
    <extension>
      <list> y[2] x[][] </list>
      <supports> (0,2,2,*,*,2,2,*,*,*,*,*,*,*,*,*,*,*,*,*,*,*,*,*,*,*,*,*,*)(1,*,2,2,*,*,2,2,*,*,*,*,*,*,*,*,*,*,*,*,*,*,*,*,*,*,*,*,*)(2,*,*,2,2,*,*,2,2,*,*,*,*,*,*,*,*,*,*,*,*,*,*,*,*,*,*,*,*)(4,*,*,*,*,2,2,*,*,2,2,*,*,*,*,*,*,*,*,*,*,*,*,*,*,*,*,*,*) </supports>
    </extension>
  </block>
\end{xcsp}

\section{Posting Meta-Constraints}\label{sec:meta}

In \p3, some functions have been specifically introduced to build meta-constraints: \nn{And()}, \nn{Or()}, \nn{Not()}, \nn{Xor()}, and \nn{Iff()}.
It is important to note that the first letter of these function names is uppercase.
If one also wants to generate a meta-constraint form with the function \nn{If()}, one has to specify the parameter \nn{meta} with value \nn{True}, because that function is usually employed for posting classical constraints (i.e., not meta-constraints).
As an illustration, here is a \p3 model showing how to capture statements of Equations \ref{eq:meta1} and \ref{eq:meta2}:

\begin{boxpy}\begin{python}
@\imp@

x = VarArray(size=4, dom=range(4))
y = Var(range(-1, 4))

satisfy(
  Or(Sum(x) > 10, AllDifferent(x)),
  If(y != -1, Then=x[y] == 1, meta=True)
)
\end{python}\end{boxpy}

When compiling, we obtain the following \x3 instance:

\begin{xcsp}
<instance format="XCSP3" type="CSP">
  <variables>
    <array id="x" size="[4]"> 0..3 </array>
    <var id="i"> -1..3 </var>
  </variables>
  <constraints>
    <or>
      <sum>
        <list> x[] </list>
        <condition> (gt,10) </condition>
      </sum>
      <allDifferent> x[] </allDifferent>
    </or>
    <ifThen>
      <intension> ne(i,-1) </intension>
      <element>
        <list> x[] </list>
        <index> i </index>
        <value> 1 </value>
      </element>
    </ifThen>
  </constraints>
</instance>
\end{xcsp}

As you can see, with meta-constraints, we can stay very close to the original (formal) formulation.
Unfortunately, there is a price to pay: the generated instances are no more in the perimeter of \x3-core (and consequently, it is not obvious to find an appropriate constraint solver to read such instances).
Of course, in the future, some additional tools could be developed to offer the user the possibility of reformulating \x3 instances (and possibly, the perimeter of \x3-core could be slightly extended).
Meanwhile, the solutions presented in Sections \ref{sec:complex}, \ref{sec:reformulation} and \ref{sec:tabling} should be chosen in priority.

\section{Using Explicit Reification}\label{sec:reification}

Reification is the fact of representing the satisfaction value of certain constraints by means of Boolean variables.
Reifying a constraint $c$ requires the introduction of an associated variable $b$ while considering the logical equivalence $b \Leftrightarrow c$. 
The two equations given earlier could be transformed by reifying three constraints, as follows:
\begin{quote}
  $b_1 \Leftrightarrow \gb{Sum}(x) > 10$\\
  $b_2 \Leftrightarrow \gb{AllDifferent}(x)$\\
  $b_1 \lor b_2$\\

  $b_3 \Leftrightarrow  x[y] = 1$\\
  $y \neq -1 \Rightarrow b_3$
\end{quote}

Currently, there is no \p3 function (or mechanism) to deal with explicit reification (i.e., explicitly associating a Boolean variable with a reified constraint), although this is possible in \x3.
The main reason is that when reification is involved, \x3 instances are no more in the perimeter of \x3-core (and consequently, it is not obvious to find an appropriate constraint solver to read such instances).
Actually, reification is outside the scope of \x3-core because it complexifies the task of constraint solvers. 
Even if this restriction could be relaxed in the future (e.g., half reification), for the moment, we are not aware of any situation (based on our experience of having modeled more than 400 problems) that cannot be (efficiently) handled with the solutions presented in Sections \ref{sec:complex}, \ref{sec:reformulation} and \ref{sec:tabling}.

\chapter{Interface of the Library}\label{ch:interface}

In this chapter, we are interested in the interface of the library \p3.
First, in Section \ref{sec:options}, we review all options that can be used on the command line.
Second, in Section \ref{sec:members}, we review all components (constants, variables and functions) that are available when importing the library (package) \p3.
Finally, we briefly discuss control of imports in Section \ref{sec:imports}, which is actually a classical Python issue.

\section{Command-Line Interface}\label{sec:options}

The following options, concerning data, are described in Section \ref{sec:specdata}.

\begin{itemize}
\item \verb!-data!
\item \verb!-parser! (or \verb!-dataparser!)
\item \verb!-export! (or \verb!-dataexport!)
\item \verb!-format! (or \verb!-dataformat!)
\end{itemize}

\bigskip
\noindent The following option allows us to indicate what must be the name of the generated filename (instead of the one that is automatically chosen).

\begin{itemize}
\item \verb!-output!
\end{itemize}

\noindent For example, the name of the generated \x3 file is `Queens-4.xml' when executing:

\begin{command}
python Queens.py -data=4 
\end{command}

whereas it is `myname.xml' when executing:
\begin{command}
python Queens.py -data=4 -output=myname
\end{command}

This option can be also given the name of a directory.
See examples given from Page \pageref{par:bibd} for Problem BIBD.

\bigskip
\noindent The following option allows us to choose between several possible variants of a model.

\begin{itemize}
\item \verb!-variant!
\end{itemize}

Actually, it is possible to reason with both a variant name and a subvariant name.
It is the case when the specified name contains the character '-' separating the variant name from the subvariant name.
In \p3, we then use the functions \nn{variant()} and \nn{subvariant()}.
Let us consider the following example (piece of code in a file called `TestVariant.py'):

\begin{boxpy}\begin{python}
@\imp@

x = Var(0,1)

if not variant():
    print("no variant")
elif variant("v1"):
    print("variant v1")
elif variant("v2"):
    print("variant v2")
    if not subvariant():
        print("no subvariant")
    elif subvariant("a"):
        print("subvariant a")
    elif subvariant("b"):
        print("subvariant b")
\end{python}\end{boxpy}

Here are the results we obtain for various command lines:
\begin{command}
python TestVariant.py                 // no variant
python TestVariant.py -variant=v1     // variant v1
python TestVariant.py -variant=v2     // variant v2  no subvariant
python testVariant.py -variant=v2-a   // variant v2  subvariant a 
python testVariant.py -variant=v2-b   // variant v2  subvariant b
\end{command}

\bigskip
\noindent The following options concern the solving process.

\begin{itemize}
\item \verb!-solve!
\item \verb!-solver!
\end{itemize}

When using \verb!-solve!, the default solver, \ace, is called. However, when using \verb!-solver!, one must indicate the name of the solver (ace or choco, case insensitive), and possibly other solver options, in which case, square brackets are required.
Among the solver options, one can use \verb!v! (for verbose) or \verb!vv! (for very verbose), and \verb!args! that must then be followed by the symbol '=' and a string corresponding to some specific solver options. 
Here are a few examples:

\begin{command}
python Queens.py -data=4 -solve
python Queens.py -data=4 -solver=choco
python Queens.py -data=4 -solver=ace
python Queens.py -data=4 -solver=[choco,v]
python Queens.py -data=4 -solver=[ace,vv]
python Queens.py -data=4 -solver=[ace,v,args="-s=2"]
\end{command}

To see which options can be used with ACE and Choco, see Chapter \ref{cha:options}. 

\bigskip
Finally, there are some other options, used as flags, i.e., requiring no argument.
By default, these flags have \nn{False} as value.
They are:

\begin{itemize}
\item \verb!-display! displays the XCSP3 instance in the system standard output, instead of generating an XCSP3 file (not compatible with \verb!-solve! and  \verb!-solver!)
\item \verb!-verbose! displays some additional information when compiling
\item \verb!-sober! does not include side notes in the \x3 file 
\item \verb!-ev! may display additional information when an error occurs
\item \verb!-lzma! compresses the \x3 file with lzma (requires lzma to be installed)
\item \verb!-safe! performs some operations (possibly based on parallelism) in a safer manner 
\item \verb!-keepHybrid! keeps compressed forms of hybrid tables
\item \verb!-restrictTablesWrtDomains! removes useless tuples (because invalid) in ordinary/starred tables
\item \verb!-dontRunCompactor! prevents from compacting the representation of constraints and objectives 
\item \verb!-dontCompactValues! prevents from compressing lists of integers (with character 'x' as in 0x20)
\item \verb!-groupSumCoeffs! gathers coefficients that are associated with the same variables (e.g., in a constraint \gb{sum}) 
\item \verb!-mini! attempts to generate instances in the perimeter of the mini-tracks of XCSP competitions
\item \verb!-uncurse! prevents from redefining some operators (with module 'forbiddenfruit')
\end{itemize}

\section{Main Module Interface}\label{sec:members}

In this section, we briefly review all components (constants, variables, functions) that are available from the main module of the library \p3.
This is what you get when executing:
\begin{quote}
\verb!from pycsp3 import *!
\end{quote}
To list all of them, one can simply execute:

\begin{quote}
\verb!import pycsp3!\\
\verb!dir(pycsp3)!
\end{quote}

In the next sub-sections, we introduce the different categories of such components.

\subsection{Building Models}

The main functions for building CSP and COP models are about:
\begin{itemize}
\item declaring stand-alone variables, and arrays of variables
  \begin{itemize}
  \item \nn{Var()}
  \item \nn{VarArray()}
  \end{itemize}
\item posting constraints
  \begin{itemize}
  \item \nn{satisfy()}
  \end{itemize}
\item specifying an objective
  \begin{itemize}
  \item \nn{minimize()}
  \item \nn{maximize()}
  \end{itemize}
\item managing several model variants
  \begin{itemize}
  \item \nn{variant()}
  \item \nn{subvariant()}
  \end{itemize}
\end{itemize}

How to declare variables is discussed in Section \ref{sec:variables}.
How to post constraints is made by calling \nn{satisfy()}, as recalled in the introduction of Chapter \ref{ch:twenty}.
How to specify an objective is discussed in Section \ref{sec:objectives}.
How to manage variants and subvariants is illustrated in Section \ref{sec:options}.

\subsection{Building Expressions}

When building expressions of intensional constraints, one can use constants, variables, and arithmetic, relational, and logic operators (which are redefined to this particular purpose).
In addition to the Python functions:
\begin{itemize}
\item \nn{abs()}
\item \nn{min()}
\item \nn{max()}
\end{itemize}
which are also extended (redefined), one can use the following specific functions: 

\begin{itemize}
\item \nn{xor()}
\item \nn{iff()}
\item \nn{imply()}
\item \nn{ift()}
\item \nn{expr()}
\item \nn{both()}
\item \nn{either()}
\item \nn{conjunction()}
\item \nn{disjunction()}
\end{itemize}

\medskip
For example, the 8 constraints of this demonstration model:

\begin{boxpy}\begin{python}
@\imp@

x = VarArray(size=6, dom=range(6))

satisfy(
  xor(x[0] == 0, x[1] == 1, x[2] == 2),
  iff(x[0] < 3, x[1] != 2),
  iff(x[i] != i for i in range(6)),
  imply(x[0] > 2, x[1] == 4),
  ift(x[0] == 1, x[1] == 2, x[2] == 3),
  expr("<", x[0], 4),
  both(x[0] > 0, x[1] < 3),
  either(x[0] > 0, x[1] < 3),
  conjunction(x[i] != i for i in range(6)),
  disjunction(x[i] != i for i in range(6)),
)
\end{python}\end{boxpy}

correspond to intensional constraints whose expressions in prefix notation are: 
\begin{command}
xor(eq(x[0],0),eq(x[1],1),eq(x[2],2))
iff(lt(x[0],3),ne(x[1],2))
iff(ne(x[0],0),ne(x[1],1),ne(x[2],2),ne(x[3],3),ne(x[4],4),ne(x[5],5))
imp(gt(x[0],2),eq(x[1],4))
if(eq(x[0],1),eq(x[1],2),eq(x[2],3))
lt(x[0],4)
and(gt(x[0],0),lt(x[1],3))
or(gt(x[0],0),lt(x[1],3))
and(ne(x[0],0),ne(x[1],1),ne(x[2],2),ne(x[3],3),ne(x[4],4),ne(x[5],5))
or(ne(x[0],0),ne(x[1],1),ne(x[2],2),ne(x[3],3),ne(x[4],4),ne(x[5],5))
\end{command}

A related function is \nn{protect} that allows us to execute some piece of code while all redefined operators are temporarily deactivated.
To be effective, one must chain the call to \nn{protect} with a call to \nn{execute} with the piece of code to be executed in protected mode.
As an illustration, if we execute: 

\begin{python}
x = Var(0,1)
y = Var(0,1)

print(x == y)
print(protect().execute(x == x))
print(protect().execute(x == y))
\end{python}

we obtain:

\begin{command}
eq(x,y)
True
False
\end{command}

\subsection{Building Global Constraints}

Some constraints can be built by simply using the (redefined) operators (and functions) of Python.
This is mainly the case for \nn{intension}, \nn{extension} and also \nn{element}.
For the other global constraints, here is the list of functions to be called:

\begin{itemize}
\item \nn{Automaton()} and \nn{MDD()}
\item \nn{AllDifferent()}, \nn{AllDifferentList()}, \nn{AllEqual()}
\item \nn{Increasing()}, \nn{Decreasing()}, \nn{LexIncreasing()}, \nn{LexDecreasing()}, \nn{Precedence()}
\item \nn{Sum()}, \nn{Count()}, \nn{NValues()}, \nn{Cardinality()}
\item \nn{Maximum()}, \nn{Minimum()}, \nn{MaximumArg()}, \nn{MinimumArg()}, \nn{Channel()}
\item \nn{NoOverlap()}, \nn{Cumulative()}, \nn{BinPacking()}, \nn{Knapsack()}
\item \nn{Circuit()}, \nn{Clause()}
\end{itemize}

Details about these functions can be found in the docstrings and in Chapter \ref{ch:twenty} of this document.

Notice that, as derived constraint forms, you can find \nn{Hamming()}, derived from \gb{Sum}, and introduced in Section \ref{sec:sum},
as well as \nn{Exist()}, \nn{NotExist()}, \nn{ExactlyOne()}, \nn{AtLeastOne()}, \nn{AtMostOne()} and \nn{AllHold()}, derived from \gb{Count}, and introduced in Section \ref{sec:count}.

\subsection{Loading (Default) JSON Data}

Two useful functions to load some JSON data by default, or independently of the main object \nn{data} are: 

\begin{itemize}
\item \nn{default\_data()}
\item \nn{loading\_json\_data()}
\end{itemize}

These functions are described in Section \ref{sec:specdata}.

\subsection{Handling Lists (Matrices)}

Rather often, we need to handle matrices (i.e., two-dimensional lists) of integers or variables.
The following functions can be helpful: 

\begin{itemize}
\item \nn{columns()}
\item \nn{diagonal\_down()}
\item \nn{diagonals\_down()}
\item \nn{diagonal\_up()}
\item \nn{diagonals\_up()}
\end{itemize}

The function \nn{columns} actually computes a transpose matrix.
If we execute:

\begin{python}
x = VarArray(size=[3,4], dom={0,1})

print(x)
print(columns(x))
\end{python}

we obtain:
\begin{command}
[
  [x[0][0], x[0][1], x[0][2], x[0][3]]
  [x[1][0], x[1][1], x[1][2], x[1][3]]
  [x[2][0], x[2][1], x[2][2], x[2][3]]
]
[
  [x[0][0], x[1][0], x[2][0]]
  [x[0][1], x[1][1], x[2][1]]
  [x[0][2], x[1][2], x[2][2]]
  [x[0][3], x[1][3], x[2][3]]
]
\end{command}

As an illustration of functions that are useful for extracting diagonals, if we execute:

\begin{python}
y = VarArray(size=[4,4], dom={0,1})

print(diagonal_down(y))
print(diagonals_down(y))
print(diagonals_down(y, broken=True))
\end{python}

we obtain:
\begin{command}
[y[0][0], y[1][1], y[2][2], y[3][3]]
[
  [y[2][0], y[3][1]]
  [y[1][0], y[2][1], y[3][2]]
  [y[0][0], y[1][1], y[2][2], y[3][3]]
  [y[0][1], y[1][2], y[2][3]]
  [y[0][2], y[1][3]]
]
[
  [y[0][0], y[1][1], y[2][2], y[3][3]]
  [y[0][3], y[1][0], y[2][1], y[3][2]]
  [y[0][2], y[1][3], y[2][0], y[3][1]]
  [y[0][1], y[1][2], y[2][3], y[3][0]]
]
\end{command}

Finally, the function \nn{cp\_array} allows us to transform any list (of any dimension) of integers into a more specific type called 'ListInt' that inherits from list.
Similarly, it allows us to transform any list (of any dimension) of variables into a more specific type called 'ListVar' that inherits from list.
It is important to have such specific types of lists when using the constraint \gb{element}.
Importantly, when the data are loaded from a file (the usual case), all lists of integers have the specific type of list returned by \nn{cp\_array}, and so, it is very rare to need to call this function explicitly.

As an illustration, if we execute:

\begin{python}
t = [3, 4, 5]
print(type(t))
t = cp_array(t)
print(type(t))
\end{python}

we obtain:

\begin{command}
<class 'list'>
<class 'pycsp3.tools.curser.ListInt'>
\end{command}

If we execute:

\begin{python}
x = VarArray(size=5, dom={0,1})

print(type(x))
y = [x[0], x[2], x[4]]
print(type(y))
y = cp_array(y)
print(type(y))
\end{python}

we obtain:

\begin{command}
<class 'pycsp3.tools.curser.ListVar'>
<class 'list'>
<class 'pycsp3.tools.curser.ListVar'>
\end{command}

When a list is from type 'ListVar' or 'ListInt', it can be used in the expression of a constraint \gb{element}.

\subsection{Handling Tuples}

From package itertools, the following functions are directly available:

\begin{itemize}
\item \nn{product()}
\item \nn{permutations()}
\item \nn{combinations()}
\end{itemize}

Note that the function \nn{combinations()} is slightly extended so as to permit the first argument to be an integer.
In that case, this value is converted into a range.
For example, if we execute: 

\begin{python}
print([tuple for tuple in combinations(5,2)])
\end{python}

we obtain:

\begin{command}
[(0, 1), (0, 2), (0, 3), (0, 4), (1, 2), (1, 3), (1, 4), (2, 3), (2, 4), (3, 4)]
\end{command}

\subsection{Utility Computations}

Some utility functions are:

\begin{itemize}
\item \nn{different\_values()}
\item \nn{flatten()}
\item \nn{alphabet\_positions()}
\item \nn{all\_primes()}
\item \nn{integer\_scaling()}
\end{itemize}

The function \nn{different\_values} just checks that all specified arguments are different.
The function \nn{flatten} builds a one-dimensional list with all elements that can be encountered when looking into the specified arguments (typically, this is a list of possibly any dimension).
None values are discarded except if the optional named parameter \nn{keep\_none} is set to True.
For example, if we execute:
\begin{python}
x = VarArray(size=[3,3], dom=lambda i,j: {0,1} if i >= j else None)
print("x: ", x)
print("x flattened: ", flatten(x))
\end{python}

we obtain:
\begin{command}
x: [
  [x[0][0], None, None]
  [x[1][0], x[1][1], None]
  [x[2][0], x[2][1], x[2][2]]
]
x flattened: [x[0][0], x[1][0], x[1][1], x[2][0], x[2][1], x[2][2]]
\end{command}

The function \nn{alphabet\_positions} returns a list with the indexes of the letters (with respect to the 26 letters of the Latin alphabet) of a specified string.
The function \nn{all\_primes} returns a list with all prime numbers that are strictly less than the specified limit.

The function \nn{integer\_scaling} returns a list with all specified values after possibly converting them (when decimal) into integers by means of automatic scaling.
For example, if we execute:
\begin{python}
t = [3, 2.11, 0.0141]
print("t scaled: ", integer_scaling(t))
\end{python}

we obtain:

\begin{command}
t scaled: [30000, 21100, 141]
\end{command}

\subsection{Building Hybrid Tables}\label{sec:hybrid}

On the one hand, it is rather easy to build starred tuples, which are tuples involving '*', denoted by the constant ANY in \p3.
Illustrations are given by the models of problems TTPV, Section \ref{sec:ttpv}, and Layout, Section \ref{sec:layout}.

On the other hand, when creating tables to be used with extensional constraints, one can use some auxiliary functions that capture some patterns (conditions)
that can be put at some places inside tuples. Tables are then said to be {\em hybrid}.
The interest is that it is usually easier (quicker) to build tables, which are in compressed forms (possibly requiring far less memory space).
Besides, we can decide or not to generate such compressed tables when compiling.

For example, assuming that the possible values to work with are $\{0,1,2,3,4\}$, the hybrid tuple $(0,lt(3),2)$ represents the set of tuples $\{(0,0,2),(0,1,2),(0,2,2)\}$
since $lt(3)$ means any value that is strictly less than 3.
As a more concrete illustration, let us consider the following demonstration model:

\begin{boxpy}\begin{python}
@\imp@
    
T = [(0, ANY, gt(1)), (ne(0),(2,3),complement(2,3))]

x = VarArray(size=3, dom=range(4))

satisfy(
  x in T
)
\end{python}\end{boxpy}

The constraint expresses the fact that $x[0]$ can be 0 if $x[2] > 1$, or different from 0 if $x[1] \in \{2,3\}$ and $x[2] \in \{0,1\}$ (the complement of $\{2,3\}$).
When looking at the outcome of default compilation (i.e., the \x3 file), one can see that a starred table has been generated. 
Indeed, by default, hybrid tables are automatically transformed into ordinary/starred tables when compiling.
To generate hybrid tables in \x3, one has to use the option \nn{-keephybrid} as shown in \href{https://pycsp.org/documentation/constraints/xtension/}{Case 4} of the Jupyter notebook of constraint \gb{extension}.

More specifically, when we build tables, we can use a compressed expression at any place inside a tuple by using one of the following structures or functions:
\begin{itemize}
\item $\{v_1, v_2, \dots, v_k\}$ or $(v_1, v_2, \dots, v_k)$ corresponds to any value in the specified set or tuple
\item range$(a, b)$ corresponds to any value in the specified range
\item complement$(v_1, v_2, \dots, v_k)$ corresponds to any unspecified value
\item complement(range$(a, b)$) corresponds to any value not present in the specified range
\item $op(v)$ with $op$ being a relational operator in $\{ne, lt, le, gt, ge\}$ so that:
\begin{itemize}
\item $ne(v)$ corresponds to any value 'not equal' to $v$
\item $lt(v)$ corresponds to any value strictly 'less than' $v$
\item $le(v)$ corresponds to any value 'less than or equal to' $v$
\item $gt(v)$ corresponds to any value strictly 'greater than' $v$
\item $ge(v)$ corresponds to any value 'greater than or equal to' $v$
\end{itemize}
\item $op(col(i))$ with op being a relational operator in $\{eq, ne, lt, le, gt, ge\}$ and $col(i)$ denoting the 'column' of the tuple at index $i$, so that:
\begin{itemize}
\item $eq(col(i))$ corresponds to the value at index $i$ in the tuple
\item $ne(col(i))$ corresponds to any value 'not equal' to the value at index $i$
\item $lt(col(i))$ corresponds to any value strictly 'less than' the value at index $i$
\item $le(col(i))$ corresponds to any value 'less than or equal to' the value at index $i$
\item $gt(col(i))$ corresponds to any value strictly 'greater than' the value at index $i$
\item $ge(col(i))$ corresponds to any value 'greater than or equal to' the value at index $i$
\end{itemize}
\item $op(col(i) + v)$, defined similarly as above, with $v$ added to the value at index $i$
\item $op(col(i) - v)$, defined similarly as above, with $v$ subtracted to the value at index $i$
\item $op(col(i) + col(j))$, defined similarly as above, with the values at index $i$ and $j$ being added
\end{itemize}

As an example, if the compilation option \nn{-keephybrid} is enabled, the following model:
\begin{boxpy}\begin{python}
@\imp@
    
x = VarArray(size=3, dom=range(10))

T = {
  (range(4, 7), gt(7), ANY),
  (lt(3), ANY, ge(6)),
  (9, ne(2), ANY),
  ((3, 8), ANY, (6, 8)),
  (7, complement(range(2, 8)), complement(1, 3, 5, 7, 9))
}

satisfy(
  x in T
)
\end{python}\end{boxpy}

gives the following \x3 instance (file):

\begin{xcsp}
<instance format="XCSP3" type="CSP">
  <variables>
    <array id="x" size="[3]"> 0..9 </array>
  </variables>
  <constraints>
    <extension type="hybrid-1">
      <list> x[] </list>
      <supports>
        (4..6,@$\geq 8$@,*)(@$\leq 2$@,*,@$\geq 6$@)(9,@$\neq 2$@,*)({3,8},*,{6,8})(7,@$\complement$@2..7,@$\complement$@{1,3,5,7,9})
      </supports>
    </extension>
  </constraints>
</instance>
\end{xcsp}

Although the transformation from hybrid tables to ordinary/starred tables is automatic when compiling, one may want, for some reasons, to apply explicitly the transformation with the function \nn{to\_ordinary\_table}.
This function converts a specified table that may contain hybrid restrictions and stars into an ordinary table (or a starred table).
The first argument of the function is a table that contains r-tuples.
For converting, the domain to be considered are any index i of these tuples is given by domains[i] where domains is the second argument of the function.
If domains[i] is an integer, it is automatically transformed into a range.
An optional named parameter \nn{starred} allows us to choose between an ordinary and a starred table.

For example, if we execute:
\begin{python}
table = [(0, ANY, gt(1)), (ne(0),(2,3),complement(2,3))]
print("Hybrid table: ", table)
print("Starred Table: ", sorted(to_ordinary_table(table,[4,4,4], possibly_starred=True)))
print("Ordinary Table: ", sorted(to_ordinary_table(table,[4,4,4])))
\end{python}

we obtain:

\begin{quote}
Hybrid table:  [(0, *, $\geq$ 2), ($\neq$ 0, (2, 3), $\complement \{2,3\}$)] \\  \\
Starred Table:  [(0, *, 2), (0, *, 3), (1, 2, 0), (1, 2, 1), (1, 3, 0), (1, 3, 1),\\
  (2, 2, 0), (2, 2, 1), (2, 3, 0), (2, 3, 1), (3, 2, 0), (3, 2, 1), (3, 3, 0), (3, 3, 1)] \\ \\
Ordinary Table:  [(0, 0, 2), (0, 0, 3), (0, 1, 2), (0, 1, 3), (0, 2, 2), (0, 2, 3), \\
  (0, 3, 2), (0, 3, 3), (1, 2, 0), (1, 2, 1), (1, 3, 0), (1, 3, 1), (2, 2, 0), (2, 2, 1), \\
  (2, 3, 0), (2, 3, 1), (3, 2, 0), (3, 2, 1), (3, 3, 0), (3, 3, 1)]
\end{quote}

\subsection{Building Meta-constraints}

It is possible to build meta-constraints by using the following functions:
\begin{itemize}
\item \nn{And()}
\item \nn{Or()}
\item \nn{Not()}
\item \nn{Xor()}
\item \nn{If()}, but requires to set the parameter \nn{meta} to \nn{True}
\item \nn{Iff()}
\end{itemize}

It is important to note that the first letter of these function names is uppercase.
Some illustrations and details are given in Section \ref{sec:meta}.
For the moment, note that meta-constraints should be avoided as they are not in the perimeter of \x3-core.

\subsection{Solving}

Some constants are available.
Some concern the result of a solving process, when \nn{solve()} is called.
\begin{itemize}
\item \nn{UNSAT}, unsatisfiable (means that no solution is found by the solver) 
\item \nn{SAT}, satisfiable (means that at least one solution is found by the solver)
\item \nn{OPTIMUM}, optimum (means that an optimal solution is found by the solver)
\item \nn{UNKNOWN}, unknown (means that the solver is unable to solve the problem instance) 
\item \nn{CORE}, core (means that an unsatisfiable core has been extracted by the solver)
\end{itemize}

\noindent Some concern the choice of a solver:
\begin{itemize}
\item \nn{ACE}, Solver ACE (AbsCon Essence)
\item \nn{CHOCO}, Solver Choco
\end{itemize}

\noindent A last constant is
\begin{itemize}
\item \nn{ALL}, meaning that all solutions must be sought, when used with the parameter \nn{sols} of \nn{solve()}.
\end{itemize}

\bigskip
\noindent The functions that directly concern the solving process are:
\begin{itemize}
\item \nn{solve()}: runs the solver on the current instance
\item \nn{solver()}: returns the current solver, when no argument is given, or sets the current solver with an argument set to the constant ACE or the constant CHOCO
\item \nn{status()}: returns the result of the last solving process (last call to \nn{solve()})
\item \nn{solution()}: returns an object with various information (fields) concerning the last found solution
\item \nn{value()}:  returns the value assigned to the variable specified as parameter
\item \nn{values()}: returns the list of values assigned to the (list of) variables specified as parameter 
\item \nn{n\_solutions()}: returns the number of found solutions
\item \nn{bound()}: returns the value of the objective function corresponding to the last found solution
\item \nn{core()}: returns the core identified by the last extraction operation
\end{itemize}
These functions are described and/or illustrated in Chapter \ref{ch:piloting}.

\bigskip
\noindent Finally, some functions allow us to display the posted constraints (or objective), to remove some posted constraints and to clear everything (variables, constraints, objective):
\begin{itemize}
\item \nn{posted()}: displays the posted constraints
\item \nn{objective()}: displays the current objective
\item \nn{unpost()}: removes the constraints posted by the last call to \nn{satisfy()}.
\item \nn{clear()}: clears everything (variables, constraints, objective) 
\end{itemize}
These functions are described and/or illustrated in Chapter \ref{ch:piloting}.

\section{Controlling Imports} \label{sec:imports}

The practice of importing everything (i.e., $*$) into the current namespace is sometimes discouraged because it notably provides the opportunity for namespace collisions.
Although we shall always use \verb!from pycsp3 import *! in this guide, we give below an illustration of specific import statements.
Note that it is a general Python technical issue. 

\paragraph{Cookie Monster.}\index{Problems!Cookie Monster}

The Cookie Monster Problem is from Richard Green:
``Suppose that we have a number of cookie jars, each one containing a certain number of cookies.
The Cookie Monster (CM) wants to eat all the cookies, but he is required to do so in a number of sequential moves.
At each move, the CM chooses a subset of the jars, and eats the same (nonzero) number of cookies from each selected jar.
The goal of the CM is to empty all the cookies from the jars in the smallest possible number of moves, and the Cookie Monster Problem is to determine this number for any given set of cookie jars.''

Concerning data, we need a list of quantities in jars as e.g., $[15, 13, 12, 4, 2, 1]$,
meaning that there are six jars, containing 15, 13, 12, 4, 2, 1 cookies each.

\begin{figure}[h]
\begin{center}
  \includegraphics[scale=0.5,angle=90]{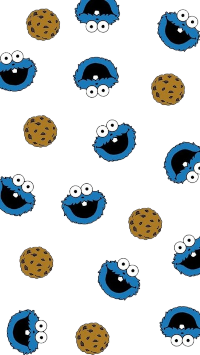}
\end{center}
\caption{Cookie Monsters. \tiny{(image from \href{https://pixabay.com/fr/illustrations/cookie-monster-les-cookies-5655016}{Pixabay})} \label{fig:cookie}} 
\end{figure}

A \p3 model (a variant can be found in OscaR) for this problem is given by the following file `CookieMonster.py':

\begin{boxpy}\begin{python}
from pycsp3 import @data@, Var, VarArray, @satisfy@, @minimize@

jars = data or [15, 13, 12, 4, 2, 1]
nJars, horizon = len(jars), len(jars) + 1

# x[t][i] is the quantity of cookies in the ith jar at time t
x = VarArray(size=[horizon, nJars], dom=range(max(jars) + 1))

# y[t] is the number of cookies eaten by the monster in selected jars at time t
y = VarArray(size=horizon, dom=range(max(jars) + 1))

# f is the first time where all jars are empty
f = Var(range(horizon))

satisfy(
  # initial state
  [x[0][i] == jars[i] for i in range(nJars)],

  # final state
  [x[-1][i] == 0 for i in range(nJars)],

  # handling the action of the cookie monster at time t (to t+1)
  [
    either(
      x[t + 1][i] == x[t][i],
      x[t + 1][i] == x[t][i] - y[t]
    ) for t in range(horizon - 1) for i in range(nJars)
  ],

  # ensuring no useless intermediate inaction
  [
    If(
      y[t] == 0,
      Then=y[t + 1] == 0
    ) for t in range(horizon - 1)
  ],
  
  # at time f, all jars are empty
  y[f] == 0
)
  
minimize(
  f
)
\end{python}\end{boxpy}

Note how the first line of the model avoids importing everything ($*$).

\bigskip
We can even go further, by only importing the package.
This way, no collision is possible; there is no risk of inadvertently redefining a \p3 function, for example.
However, one must prefix any \p3 member (constant, variable or function) with \verb!pycsp3.!
On our example, this gives:

\begin{boxpy}\begin{python}
import pycsp3

jars = pycsp3.data or [15, 13, 12, 4, 2, 1]
nJars, horizon = len(jars), len(jars) + 1

# x[t][i] is the quantity of cookies in the ith jar at time t
x = pycsp3.VarArray(size=[horizon, nJars], dom=range(max(jars) + 1))

# y[t] is the number of cookies eaten by the monster in selected jars at time t
y = pycsp3.VarArray(size=horizon, dom=range(max(jars) + 1))

# f is the first time where all jars are empty
f = pycsp3.Var(range(horizon))

pycsp3.satisfy(
  # initial state
  [x[0][i] == jars[i] for i in range(nJars)],

  # final state
  [x[-1][i] == 0 for i in range(nJars)],
  
  # handling the action of the cookie monster at time t (to t+1)
  [
    pycsp3.either(
      x[t + 1][i] == x[t][i],
      x[t + 1][i] == x[t][i] - y[t]
    ) for t in range(horizon - 1) for i in range(nJars)
  ],

  # ensuring no useless intermediate inaction
  [
    pycsp3.If(
      y[t] == 0,
      Then=y[t + 1] == 0
    ) for t in range(horizon - 1)
  ],
  
  # at time f, all jars are empty
  y[f] == 0
)
  
pycsp3.minimize(
  f
)
\end{python}\end{boxpy}

\chapter{Piloting the Solving Process}\label{ch:piloting}

In this chapter, we show how it is easy to pilot, in Python, the process of solving any problem instance by using the interface of \p3.
More specifically, we show how to run a solver, how to get several (possibly, all) solutions, how to conduct an incremental solving strategy, and how to extract an unsatisfiable core of constraints.   

\section{Running a Solver}

It is very simple to directly run a solver on a \p3 model. You just have to call the following function:

\begin{command}
solve()
\end{command}

This will start the solver \ace on the current problem instance.
The result of this command is the status of the solving operation, which is one of the following constants:
\begin{command}
  UNSAT
  SAT
  OPTIMUM
  UNKNOWN
\end{command}

More specifically, the result is:
\begin{itemize}
\item among UNSAT, SAT, and UNKNOWN for a CSP instance
\item among UNSAT, SAT, OPTIMUM and UNKNOWN for a COP instance
\end{itemize}

\bigskip
\noindent This function \nn{solve()} accepts several named parameters:
\begin{itemize}
\item \nn{solver}: name of the solver (ACE or CHOCO)
\item \nn{options}: specific options for the solver
\item \nn{filename}: the filename of the compiled problem instance
\item \nn{verbose}: verbosity level from -1 to 2
\item \nn{sols}: number of solutions to be found (ALL if no limit)
\item \nn{extraction}: True if an unsatisfiable core of constraints must be sought
\end{itemize}

\bigskip
As an illustration, let us consider the Warehouse Location Problem (WLP), introduced in Section \ref{sec:warehouse}.
In a first step, we consider the decision problem (i.e., the objective is not posted, so, we have a CSP instance), run the solver and print the solution if the problem instance is satisfiable (by default, only one solution is sought for a CSP instance).
Note that we can display the values assigned to the variables of a specified (possibly multi-dimensional) list by calling the function \nn{values()}.
The file `Warehouse.py' is:

\begin{boxpy}\begin{python}
@\imp@ 

fixed_cost, capacities, costs = data  
nWarehouses, nStores = len(capacities), len(costs)

# w[i] is the warehouse supplying the ith store
w = VarArray(size=nStores, dom=range(nWarehouses))

satisfy(
  # capacities of warehouses must not be exceeded
  Count(w, value=j) <= capacities[j] for j in range(nWarehouses)
)

if solve() is SAT:
    print(values(w))   
\end{python}\end{boxpy}

When we execute:
\begin{command}
python Warehouse.py -data=warehouse.json
\end{command}

we obtain:
\begin{command}
[0, 1, 1, 1, 1, 2, 2, 3, 4, 4]
\end{command}

The output is not very friendly/readable, but nothing prevents us from improving that aspect.
This is what we do now with a Python f-string, getting the value of individual variables with the function \nn{value()}.  
The new file `Warehouse.py' is:

\begin{boxpy}\begin{python}
@\imp@ 

fixed_cost, capacities, costs = data  
nWarehouses, nStores = len(capacities), len(costs)

# w[i] is the warehouse supplying the ith store
w = VarArray(size=nStores, dom=range(nWarehouses))

satisfy(
  # capacities of warehouses must not be exceeded
  Count(w, value=j) <= capacities[j] for j in range(nWarehouses)
)

if solve() is SAT:
    for i in range(nStores):
        print(f"Warehouse supplying the store {i} is {value(w[i])}
                with cost {costs[i][value(w[i])]}")
\end{python}\end{boxpy}

When we execute:
\begin{command}
python Warehouse.py -data=warehouse.json
\end{command}

we obtain:
\begin{command}
Warehouse supplying the store 0 is 0 with cost 100
Warehouse supplying the store 1 is 1 with cost 27
Warehouse supplying the store 2 is 1 with cost 97
Warehouse supplying the store 3 is 1 with cost 55
Warehouse supplying the store 4 is 1 with cost 96
Warehouse supplying the store 5 is 2 with cost 29
Warehouse supplying the store 6 is 2 with cost 73
Warehouse supplying the store 7 is 3 with cost 43
Warehouse supplying the store 8 is 4 with cost 46
Warehouse supplying the store 9 is 4 with cost 95
\end{command}

\bigskip
Now, we consider the objective function (and so, we have a COP instance).
This is the reason why we check if the status returned when calling \nn{solve()} is OPTIMUM.
Note that the function \nn{bound()} directly returns the value of the objective function corresponding to the found optimal solution.
The new file `Warehouse.py' is:

\begin{boxpy}\begin{python}
@\imp@ 

fixed_cost, capacities, costs = data  
nWarehouses, nStores = len(capacities), len(costs)

# w[i] is the warehouse supplying the ith store
w = VarArray(size=nStores, dom=range(nWarehouses))

satisfy(
  # capacities of warehouses must not be exceeded
  Count(w, value=j) <= capacities[j] for j in range(nWarehouses)
)

minimize(
  # minimizing the overall cost
  Sum(costs[i][w[i]] for i in range(nStores)) + NValues(w) * fixed_cost
)

if solve() is OPTIMUM:
    print(values(w))
    for i in range(nStores):
        print(f"Cost of supplying the store {i} is {costs[i][value(w[i])]}")
    print("Total supplying cost: ", bound())
\end{python}\end{boxpy}

When we execute:
\begin{command}
python Warehouse.py -data=warehouse.json
\end{command}

we obtain:
\begin{command}
[4, 1, 4, 0, 4, 1, 1, 2, 1, 2]
Cost of supplying the store 0 is 30
Cost of supplying the store 1 is 27
Cost of supplying the store 2 is 70
Cost of supplying the store 3 is 2
Cost of supplying the store 4 is 4
Cost of supplying the store 5 is 22
Cost of supplying the store 6 is 5
Cost of supplying the store 7 is 13
Cost of supplying the store 8 is 35
Cost of supplying the store 9 is 55
Total supplying cost:  383
\end{command}

\bigskip
One may be worried by the fact that the code mixes modeling and solving parts.
Interestingly, we can make a clear separation as described now. 
First, we write the model in the file `Warehouse.py':
\begin{boxpy}\begin{python}
@\imp@ 

fixed_cost, capacities, costs = data  
nWarehouses, nStores = len(capacities), len(costs)

# w[i] is the warehouse supplying the ith store
w = VarArray(size=nStores, dom=range(nWarehouses))

satisfy(
  # capacities of warehouses must not be exceeded
  Count(w, value=j) <= capacities[j] for j in range(nWarehouses)
)

minimize(
  # minimizing the overall cost
  Sum(costs[i][w[i]] for i in range(nStores)) + NValues(w) * fixed_cost
)
\end{python}\end{boxpy}

Then, we write the solving part in a file `WarehouseSolving.py':

\begin{boxpyno}\begin{python}
from Warehouse import *

if solve() is OPTIMUM:
    print(values(w))
    for i in range(nStores):
        print(f"Cost of supplying the store {i} is {costs[i][value(w[i])]}")
        print("Total supplying cost: ", bound())
\end{python}\end{boxpyno}

Then, we can execute:

\begin{command}
python WarehouseSolving.py -data=warehouse.json
\end{command}

If for some reasons, it is better to set data in the file containing the solving part, we can modify \verb!sys.argv!.
The file `WarehouseSolving.py' becomes: 

\begin{boxpyno}\begin{python}
import sys

sys.argv.append("-data=Warehouse_example.json")

from Warehouse import *

if solve() is OPTIMUM:
    print(values(w))
    for i in range(nStores):
        print(f"Cost of supplying the store {i} is {costs[i][value(w[i])]}")
    print("Total supplying cost: ", bound())
\end{python}\end{boxpyno}

Then, we can simply execute (do note that the option \verb!-data! is not used):

\begin{command}
python WarehouseSolving.py 
\end{command}

As another illustration, let us consider one of the two models, put in a file called `Queens.py', introduced (without variants) in Section \ref{sec:queens} for the Queens problem.
If we write this solving code in a file `QueensSolving.py':

\begin{boxpyno}\begin{python}
import sys
import chess.svg

sys.argv.append("-data=8")

from Queens import *

if solve() is SAT:
    solution = values(q)  # for example: [0, 4, 7, 5, 2, 6, 1, 3]
    board = chess.Board("/".join(("" if v == 0 else str(v)) + "q"
      + ("" if v == n - 1 else str(n - 1 - v)) for v in solution)
      + ' b KQkq - 0 1')
    with open('chess.svg', 'w') as f:
        f.write(chess.svg.board(board, size=350))
\end{python}\end{boxpyno}

Then, by means of the package \nn{chess.svg}, we can generate the rendering of the solution to the 8 queens problem in a SVG file:

\begin{center}
  \includegraphics[scale=0.5]{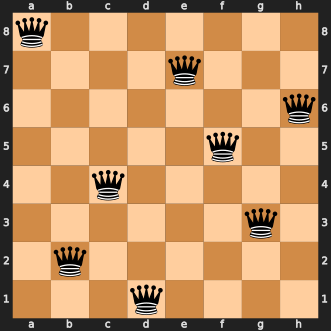}
\end{center}

\section{Finding One, Several or All Solutions}

The easiest and most efficient way of getting several (and even, all) solutions of a CSP instance is to ask the underlying solver to provide them.
We give an illustration with the Prime Looking Problem.

\paragraph{Prime Looking.}\index{Problems!Prime Looking}
This problem is from Martin Gardner: a number is said to be {\em prime-looking} if it is composite but not divisible by 2, 3 or 5.
We know that the three smallest prime-looking numbers are 49, 77 and 91. 
Can you find the prime-looking numbers less than 1000?

The model, which is rather simple, is written in a file `PrimeLooking.py':

\begin{boxpy}\begin{python}
@\imp@ 

# the number we look for
x = Var(range(1000))

# a first divider
d1 = Var(range(2, 1000))

# a second divider
d2 = Var(range(2, 1000))

satisfy(
  x == d1 * d2,
  x % 2 != 0,
  x % 3 != 0,
  x % 5 != 0,
  d1 <= d2
)
\end{python}\end{boxpy}

The solving part of the code is put in another file `PrimeLookingSolving.py':

\begin{boxpyno}\begin{python}
from PrimeLooking import *
    
instance = compile()
ace = solver(ACE)
result = ace.solve(instance)

print("Result:", result)
if result is SAT:
    print("The prime-looking number is: ", value(x))
\end{python}\end{boxpyno}

For the moment, we only get and display the first found solution.
Note how we can decide to compile, choose the solver and run the solver in separate statements.
By executing:

\begin{command}
python PrimeLookingSolving.py 
\end{command}

we obtain:
\begin{command}
Result: SAT
The prime-looking number is:  49
\end{command}

Of course, most of the time, we can prefer to use a simplified equivalent code.
This gives:

\begin{boxpyno}\begin{python}
from PrimeLooking import *

if solve() is SAT:
    print("The prime-looking number is: ", value(x))
\end{python}\end{boxpyno}

When executed, we obtain:

\begin{command}
The prime-looking number is:  49
\end{command}

Note that we can also call \nn{solution()} and get specialized information (field) as shown now: 

\begin{boxpyno}\begin{python}
from PrimeLooking import *

if solve() is SAT:
    solution = solution()
    print("Solution: ", solution)
    print("Solution Root: ", solution.root)
    print("Solution Variables: ", solution.variables)
    print("Solution Values: ", solution.values)
    print("Pretty Solution: ", solution.pretty_solution)
\end{python}\end{boxpyno}

When executed, we obtain:

\begin{command}
Solution:  <instantiation id="sol1" type="solution">
  <list> x d1 d2 </list>
  <values> 49 7 7 </values>
</instantiation>
Solution Root:  <Element instantiation at 0x7f061150d9b0>
Solution Variables:  [x, d1, d2]
Solution Values:  [49, 7, 7]
Pretty Solution:  <instantiation id="sol1" type="solution">
  <list> x d1 d2 </list>
  <values> 49 7 7 </values>
</instantiation>
\end{command}

Now, if we want to get and display all solutions, we need to set \nn{ALL} as value of the named parameter \nn{sols} of the function \nn{solve()}.
After solving, we can get the number of found solutions by calling \nn{n\_solutions()}, and, interestingly, we can use the named parameter \nn{sol} to indicate the index of a solution when calling the functions \nn{values()} and \nn{value()}.
The content of the file `PrimeLookingSolving.py' is now:

\begin{boxpyno}\begin{python}
from PrimeLooking import *

if solve(sols=ALL) is SAT:
    print("Number of solutions: ", n_solutions())
    print("Solutions: ", sorted([value(x, sol=i) for i in range(n_solutions())]))
\end{python}\end{boxpyno}

By executing:

\begin{command}
python PrimeLookingSolving.py 
\end{command}

we obtain (we use an ellipsis ... to avoid listing the 105 solutions):
\begin{command}
Number of solutions:  105
Solutions:  [49, 77, 91, 119, 121, 133, 143, 161, 169, 187, ...]
\end{command}

Actually, it is known that there are 100 prime-looking numbers less than 1000.
To check this, we can use a Python set to remove identical solutions: 

\begin{boxpyno}\begin{python}
from PrimeLooking import *

if solve(sols=ALL) is SAT:
    t = sorted(set([value(x, sol=i) for i in range(n_solutions())]))
    print("Number of prime looking numbers: ", len(t))
\end{python}\end{boxpyno}

When executed, we obtain:

\begin{command}
Number of prime looking numbers:  100
\end{command}

We can also choose to only find the first $k$ solutions.
We need $k$ to be a positive integer set as value of the named parameter \nn{sols} of the function \nn{solve()}.
For example, for $k=10$, we have:

\begin{boxpyno}\begin{python}
from PrimeLooking import *

if solve(sols=10) is SAT:
    print("Number of solutions: ", n_solutions())
    print("Solutions: ", [value(x, sol=i) for i in range(n_solutions())])
\end{python}\end{boxpyno}

When executed, we obtain:

\begin{command}
Number of solutions:  10
Solutions:  [49, 77, 91, 119, 133, 161, 203, 217, 259, 287]
\end{command}

\section{Incremental Solving}

Interestingly, one can really pilot the solving process by iteratively adding and/or removing constraints (and also adding/changing the objective), handling a form of incremental solving.
To add constraints, we already know that it suffices to call \nn{satisfy()}. To remove constraints, it suffices to call the function: 
\begin{command}
unpost()
\end{command}

When this function is called, the last {\em posting operation} is discarded: it corresponds to all constraints that were posted by the last call to \nn{satisfy()}.
It is also possible to give the index of the posting operation, and even a second parameter indicating the index of constraint(s) inside the specified posting operation.

In this section, we illustrate incremental solving by showing how to enumerate solutions by means of solution-blocking constraints, how to simulate an optimization procedure and how to compute diversified solutions.

\subsection{Enumerating Solutions with Solution-Blocking Constraints}

For a given CSP $P$, a {\em solution-blocking constraint} of $P$ is a constraint that forbids a solution of $P$ (i.e., forbids a complete instantiation of the variables of $P$ corresponding to a solution).
An original (but not necessarily efficient) way of enumerating the solutions of $P$ with a solver $S$ (that can, for example, only output a single solution) is to find solutions in sequence with $S$ while posting a new solution-blocking constraint every time a solution is found.

Let us consider the following toy model in a file called `ToyPb.py':
\begin{boxpy}\begin{python}
@\imp@

x = VarArray(size=4, dom=range(7))

satisfy(
  AllDifferent(x),
  Increasing(x),
  Sum(x) == 10
)
\end{python}\end{boxpy}

Enumerating the solutions of this model by successively posting solution-blocking constraints corresponds to the following piece of code, put in a file `ToyPbSolving.py':

\begin{boxpyno}\begin{python}
from ToyPb import *

cnt = 0
while solve() is SAT:
    cnt += 1
    print(f"Solution {cnt}: {values(x)}")
    satisfy(x != values(x))
\end{python}\end{boxpyno}

By writing \verb@satisfy(x != values(x))@, we post a constraint (technically, a table constraint with only one conflict) that will prevent us from finding the same solution again.
By executing:
\begin{command}
python ToyMaxSolving.py 
\end{command}

we display the 4 solutions of this problem instance:

\begin{command}
Solution 1: [0, 1, 3, 6]
Solution 2: [0, 1, 4, 5]
Solution 3: [0, 2, 3, 5]
Solution 4: [1, 2, 3, 4]
\end{command}

\subsection{Simulating an Optimization Procedure}

For a given CSP $P$, an independent integer cost function $f$ to be minimized, defined from (the Cartesian product of the domains of) a subset $X$ of variables of $P$ to $\mathbb{Z}$, and a solution $sol$ of $P$ whose cost computed by $f$ is $B$ , a {\em bound-improving constraint} of $P$ wrt $f$ and $sol$ is a constraint that forbids all solutions of $P$ with a bound greater than or equal to $B$: it can be written $f(X) < B$.
An original (but not necessarily efficient) way of finding an optimal solution of $P$ wrt $f$ with a CSP solver $S$ is to find solutions in sequence with $S$ while posting a new bound-improving constraint every time a solution is found.

Let us consider the Prime Looking problem introduced earlier, and let us consider that the cost function is simply the variable $x$ (to be maximized).
One way of ensuring that we get a better solution after finding a first solution is given by the following piece of code in a file `PrimeLookingSolving.py':

\begin{boxpyno}\begin{python}
from PrimeLooking import *

if solve() is SAT:
    print("The prime-looking number is: ", value(x))
    satisfy(x > x.value)
    if solve() is SAT:
        print("The prime-looking number is: ", value(x))
\end{python}\end{boxpyno}

By executing:

\begin{command}
python PrimeLookingSolving.py 
\end{command}

we obtain:

\begin{command}
The prime-looking number is:  49
The prime-looking number is:  77
\end{command}

If we want to find an optimal solution, we can write instead:

\begin{boxpyno}\begin{python}
from PrimeLooking import *
    
while True:
    if solve() is not SAT:
        break
    print("The prime-looking number is: ", value(x))
    satisfy(x > x.value)
\end{python}\end{boxpyno}

When executed, we obtain for example:

\begin{command}
The prime-looking number is:  49
The prime-looking number is:  77
The prime-looking number is:  121
...
The prime-looking number is:  899
The prime-looking number is:  961
The prime-looking number is:  989
\end{command}

In some cases, one may be worried of posting many bound-improving constraints, knowing that only the last one is relevant (since it is stronger than the other ones).
In our context, we can store the object (constraint) that was posted previously so as to be able to delete it afterwards.
This gives:

\begin{boxpyno}\begin{python}
from PrimeLooking import *

objective = None
while True:
    if solve() is not SAT:
        break
    print("The prime-looking number is: ", value(x))
    if objective is not None:
        objective.delete()
    objective = satisfy(x > x.value)
\end{python}\end{boxpyno}

As an alternative, it is possible to call the function \nn{unpost()} that discards the constraint(s) posted at the last call to \nn{satisfy()}.
This gives:

\begin{boxpyno}\begin{python}
from PrimeLooking import *

objective = False
while True:
    if solve() is not SAT:
        break
    print("The prime-looking number is: ", value(x))
    if objective:
      unpost()
    else:
       objective = True  
    satisfy(x > x.value)
\end{python}\end{boxpyno}

\subsection{Computing Diversified Solutions}

Instead of enumerating solutions in the order ``fixed'' by the solver, one may want to diversify computed solutions by exploiting some distances.
In other words, we may be interested in diverse solutions.
As a first illustration, let us consider the following toy model in a file called `ToyMax.py':
\begin{boxpy}\begin{python}
@\imp@
    
n = 8

x = VarArray(size=n, dom=range(5))

satisfy(
  Maximum(x) == 4
)
\end{python}\end{boxpy}

If we want to enumerate 5 solutions while maximizing the Hamming distance between found solutions, we can write this piece of code in a file `ToyMaxSolving.py':

\begin{boxpyno}\begin{python}
from ToyMax import *

solutions = []
while len(solutions) < 5 and solve() in {SAT, OPTIMUM}:
    print("Solution: ", values(x))
    solutions.append(values(x))
    maximize(
        Sum(x[i] != solution[i] for i in range(n) for solution in solutions)
    )
\end{python}\end{boxpyno}

Note that the problem instance is initially a CSP, and then becomes a COP because an objective is posted after the first turn of the loop (note also that any new objective overwrites the previous one, if any is present).
This is the reason why we check if the solving status is either SAT or OPTIMUM.

By executing:

\begin{command}
python ToyMaxSolving.py 
\end{command}

we obtain:

\begin{command}
Solution: [0, 0, 0, 0, 0, 0, 0, 4]
Solution: [1, 1, 1, 1, 1, 1, 4, 0]
Solution: [2, 2, 2, 2, 2, 4, 1, 1]
Solution: [3, 3, 3, 3, 4, 2, 2, 2]
Solution: [4, 4, 4, 4, 3, 3, 3, 3]
\end{command}

As a second illustration, let us consider the following model in a file called `ToySum.py':
\begin{boxpy}\begin{python}
@\imp@

n = 8

x = VarArray(size=n, dom=range(7))

satisfy(
  Sum(x) == 22
)
\end{python}\end{boxpy}

If we want to enumerate 5 solutions while maximizing the Euclidean distance between found solutions, we can write this piece of code in a file `ToySumSolving.py':

\begin{boxpyno}\begin{python}
from ToySum import *
   
solutions = []
while len(solutions) < 5 and solve() in {SAT,OPTIMUM}:
    print("Solution: ", values(x))
    solutions.append(values(x))
    maximize(
        Sum(abs(x[i] - solution[i]) for i in range(n) for solution in solutions)
    )
\end{python}\end{boxpyno}

By executing:

\begin{command}
python ToySumSolving.py 
\end{command}

we obtain:

\begin{command}
Solution: [0, 0, 0, 0, 4, 6, 6, 6]
Solution: [4, 6, 6, 6, 0, 0, 0, 0]
Solution: [6, 4, 0, 0, 6, 0, 0, 6]
Solution: [0, 0, 4, 6, 0, 6, 6, 0]
Solution: [0, 6, 6, 0, 6, 4, 0, 0]
\end{command}

\section{Extracting Unsatisfiable Cores}

In case a CSP instance is unsatisfiable, one may want to identify the cause of unsatisfiability.
Extracting a minimal unsatisfiable core (i.e. subset) of constraints may be relevant.
With \ace, this is possible by setting the value of the named parameter \nn{extraction}, of function \nn{solve()}, to True.
If a core is extracted by the solver, the constant CORE is returned.
In that case, one can call the function \nn{core()} to get the constraints of the identified core.

\bigskip
{\bf Important.} Currently, a string is returned by \nn{core()}. We shall revisit this simplistic way of getting the information in the next version of \p3.

Let us consider the following toy model in a file called `Core.py':
\begin{boxpy}\begin{python}
@\imp@

x = VarArray(size=10, dom=range(10))

satisfy(
  AllDifferent(x),
   x[0] > x[1],
   x[1] > x[2],
   x[2] > x[0]
)

if solve(extraction=True) is CORE:
    print(core())
\end{python}\end{boxpy}

By executing:

\begin{command}
python Core.py 
\end{command}

we obtain:

\begin{command}
{ c3(x[0],x[2]) c2(x[2],x[1]) c1(x[1],x[0]) }
\end{command}

\chapter{Frequently Asked Questions}


This chapter will contain frequently asked questions.
It needs to be extended.

\bigskip
{\bf Q.} Is it possible to post a constraint conditionally?

{\bf A.} Of course, it is always possible to put the condition (here, we check that the value of the variable $mode$ is strictly positive) outside the \p3 function \nn{satisfy}().
For example:

\begin{python}
if mode > 0:
   satisfy(
      AllDifferent(w, x, y, z)
   )   
\end{python}

but it is also possible to use the Python conditional operator 'if else' while returning 'None' if the condition does not hold.

\begin{python}
satisfy(
   AllDifferent(w, x, y, z) if mode > 0 else None
)   
\end{python}

\bigskip
    {\bf Q.} Is it possible to use the Python operators \nn{and}, \nn{or} and \nn{not} to combine (parts of) constraints?

    {\bf A.} No. These operators cannot be redefined. For a predicate (expression), you must use $|$, \& and \^{}; see Table \ref{tab:ops}.
    For posting two sets of constraints linked by \nn{and}, simply post two separate lists.



\chapter{Options for ACE and Choco}\label{cha:options}

In this chapter, the options that can be used with ACE and Choco are listed.

\section{Options for ACE}

Below, you will find some options that are available for ACE, as for example in:

\begin{command}
python Queens.py -data=8 -solver=[ace,v,args="-s=all -varh=PickOnDom"]
\end{command}

or
\begin{command}
java -jar ACE.jar Queens-8.xml  -s=all -varh=PickOnDom
\end{command}

\noindent ACE options are:

\begin{footnotesize}
\begin{verbatim}
To run ACE (AbsCon Essence), you must use a command like:
  java ace <xcsp3Instance> <options>
where:
  <xcsp3Instance> is the name of a file containing an XCSP3 instance
  
  <options> is the list of options  (separated by whitespace) used for solving the instance.
	Any such parameter is of the form -name=value (potentially -name for some flag parameters).
	The different parameters and their values are listed below.

General
  -General/solLimit -s
    The limit on the number of found solutions before stopping; for no limit, use -s=all or s=-1
    Default value is: -1
  -General/timeout -t
    The limit in milliseconds before stopping; seconds can be indicated as in -t=10s
    Default value is: 9223372036854775807
  -General/discardClasses -dc
    XCSP3 classes (tags) to be discarded (comma as separator)
    Default value is: "" (empty string)
  -General/trace -trace
    Displays a trace (with possible depth control as eg -trace=10-20
    Default value is: "" (empty string)
  -General/jsonLimit -jl
    The limit on the number of variables for displaying solutions in JSON
    Default value is: 10000
  -General/jsonAux -ja
    Take auxiliary variables when displaying solutions in JSON
    Default value is: false
  -General/jsonSave -js
    Save the first solution in a file whose name is this value
    Default value is: "" (empty string)
  -General/jsonQuotes -jq
    Surround keys with quotes when solutions are displayed on the standard output
    Default value is: false
  -General/jsonEachSolution -je
    During search, display all found solutions in JSON
    Default value is: false
  -General/xmlCompact -xc
    Compress values when displaying solutions in XML
    Default value is: true
  -General/xmlEachSolution -xe
    During search, display all found solutions in XML
    Default value is: false
  -General/noPrintColors -npc
    Don't use special color characters in the terminal
    Default value is: false
  -General/exceptionsVisible -ev
    Makes exceptions visible.
    Default value is: false
  -General/enableAnnotations -ea
    Enables annotations (currently, mainly concerns priority variables).
    Default value is: false
  -General/satisfactionLimit -satl
    Converting the objective into a constraint with this limit
    Default value is: 2147483647
  -General/seed -seed
    The seed that can be used for some random-based methods.
    Default value is: 0
  -General/verbose -v
    Verbosity level (value between -1 and 3)
    0 : only some global statistics are listed;
    1 : in addition, solutions are  shown;
    2 : in addition, additional information about the problem instance to be solved is given;
    3 : in addition, for each constraint, allowed or unallowed tuples are displayed.
    Default value is: 0

Problem
  -Problem/shareBits -shareBits
    Trying to save space by sharing bit vectors.
    Default value is: false
  -Problem/symmetryBreaking -sb
    Symmetry-breaking method (requires Saucy to be installed)
    Default value is: NO
    Possible values: NO SB_LE SB_LEX

Variables
  -Variables/selection -sel
    Allows us to give a list of variable that will form the subproblem to be solved.
    This list is composed of a sequence of tokens separated by commas (no whitespace).
    Each token is a variable id, a variable number or an interval of the form i..j with i and j integers.
    For example, -sel=2..10,31,-4 will denote the list 2 3 5 6 7 8 9 10 31.
    This is only valid for a XCSP instance.
    Default value is: "" (empty string)
  -Variables/instantiation -ins
    Allows us to give an instantiation (-ins) or refutation (-ref) of variables for the problem to be solved.
    This instantiation/refutation is given under the form vars:values.
    vars is a sequence of variable ids or numbers separated by commas (no whitespace).
    values is a sequence of integers (the values for variables) separated by commas (no whitespace).
    For example, -ins=x2,x4,x9:1,11,4 will denote the instantiation {x2=1,x4=11,x9=4} (or refutation).
    Default value is: "" (empty string)
  -Variables/refutation -ref
    Allows us to give an instantiation (-ins) or refutation (-ref) of variables for the problem to be solved.
    This instantiation/refutation is given under the form vars:values.
    vars is a sequence of variable ids or numbers separated by commas (no whitespace).
    values is a sequence of integers (the values for variables) separated by commas (no whitespace).
    For example, -ins=x2,x4,x9:1,11,4 will denote the instantiation {x2=1,x4=11,x9=4} (or refutation).
    Default value is: "" (empty string)
  -Variables/priority1 -pr1
    Allows us to give a list of variables that will become strict priority variables 
      (to be used by the variable ordering heuristic).
    This list is composed of a sequence of strings (ids of variables) or integers 
      (numbers of variables) separated by commas (no whitespace).
    For example, -pr1=2,8,1,10 will denote the list 2 8 1 10.
    Default value is: "" (empty string)
  -Variables/priority2 -pr2
    Allows us to give a list of variables that will become (non strict) priority variables.
    This list is composed of a sequence of tokens separated by commas (no whitespace).
    Each token is variable id, a variable number (integer) or an interval of the form i..j with i and j integers..
    For example, -pr2=2..10,31,-4 will denote the list 2 3 5 6 7 8 9 10 31.
    Default value is: "" (empty string)
  -Variables/priorityArrays -pra
    Index(es) or id(s) of the variable array(s) that must be assigned in priority
    Default value is: "" (empty string)
  -Variables/stayArrayFocus -saf
    Should we stay focused on arrays when assigning variables
    Default value is: false
  -Variables/reduceIsolated -riv
    Arbitrary keeping a single value in the domain of isolated variables
    Default value is: true

Constraints
  -Constraints/preserve1 -pc1
    Must we keep unary constraints (instead of filtering them straight)
    Default value is: true
  -Constraints/ignoreType -ignoreType
    Discard all constraints of this type
    Default value is: "" (empty string)
  -Constraints/ignoreArity -ignoreArity
    Discard all constraints of this arity
    Default value is: -1
  -Constraints/ignoreGroups -ig
    Index(es) of the group(s) of constraints that must be discarded
    Default value is: "" (empty string)

Optimization
  -Optimization/lb -lb
    Initial lower bound
    Default value is: -9223372036854775808
  -Optimization/ub -ub
    Initial upper bound
    Default value is: 9223372036854775807
  -Optimization/strategy -os
    Optimization strategy
    Default value is: DECREASING
    Possible values: INCREASING DECREASING DICHOTOMIC
  -Optimization/replaceObjVar -rov
    Must we replace the objective variable by an objective constraint, when possible?
    Default value is: true
  -Optimization/boundDescentCoeff -bdc
    Bound descent coefficient
    Default value is: 1

Extension
  -Extension/positive -positive
    Algorithm for (non-binary) positive table constraints
    Default value is: CT
    Possible values: V VA STR0 STR1 STR2 STR3 STR1N STR2N CT CMDDO CMDDS
  -Extension/negative -negative
    Algorithm for (non-binary) negative table constraint
    Default value is: V
    Possible values: V VA STR0 STR1 STR2 STR3 STR1N STR2N CT CMDDO CMDDS
  -Extension/generic2 -generic2
    Must we use a generic filtering scheme for binary table constraints?
    Default value is: true
  -Extension/structureClass2 -sc2
    Structures to be used for binary table constraints
    Default value is: Bits
    Possible values: TableHybrid Tries MDD Bits Matrix3D Matrix2D Table
  -Extension/structureClass3 -sc3
    Structures to be used for ternary table constraints
    Default value is: Matrix$Matrix3D
    Possible values: TableHybrid Tries MDD Bits Matrix3D Matrix2D Table
  -Extension/arityLimitToPositive -alp
    Limit on arity for converting negative table constraints to positive
    Default value is: -1
  -Extension/arityLimitToNegative -aln
    Limit on arity for converting positive table constraints to negative
    Default value is: -1
  -Extension/variant -extv
    Variant to be used for some algorithms (e.g., VA or CMDD)
    Default value is: 0
  -Extension/decremental -extd
    Must we use a decremental mode for some algorithms (e.g., STR2, CT or CMDD)
    Default value is: true
  -Extension/small -exts
    table size threshold for considering a special propagator
    Default value is: 10

Intension
  -Intension/decompose -di
    0: no decomposition, 1: conditional decomposition, 2: forced decomposition
    Default value is: 1
  -Intension/toExtension1 -ie1
    Must we convert unary intension constraints to extension?
    Default value is: true
  -Intension/arityLimitToExtension -ale
    Limit on arity for possibly converting to extension
    Default value is: 0
  -Intension/spaceLimitToExtension -sle
    Limit on space for possibly converting to extension
    Default value is: 20
  -Intension/recognizePrimitive2 -rp2
    Must we attempt to recognize binary primitives?
    Default value is: true
  -Intension/recognizePrimitive3 -rp3
    Must we attempt to recognize ternary primitives?
    Default value is: true
  -Intension/recognizeReifLogic -rlog
    Must we attempt to recognize logical reification forms?
    Default value is: true
  -Intension/recognizeExtremum -rext
    Must we attempt to recognize minimum/maximum constraints?
    Default value is: true
  -Intension/recognizeSum -rsum
    Must we attempt to recognize sum constraints?
    Default value is: true
  -Intension/recognizeIf -rif
    Must we recognize/decompose the ternary operator if
    Default value is: true
  -Intension/toHybrid -toh
    Must we convert toward hybrid tables, when possible?
    Default value is: false

Global
  -Global/allDifferent -g_ad
    Algorithm for AllDifferent
    Default value is: 0
  -Global/allDifferentExcept -g_ade
    Algorithm for AllDifferentExcept
    Default value is: 0
  -Global/gatherAllDifferent -g_gad
    Description is missing...
    Default value is: false
  -Global/distinctVectors -g_dv
    Algorithm for DistinctVectors
    Default value is: -1
  -Global/allEqual -g_ae
    Algorithm for AllEqual
    Default value is: 0
  -Global/notAllEqual -g_nae
    Algorithm for NotAllEqual
    Default value is: 0
  -Global/circuit -g_circ
    Algorithm for Circuit
    Default value is: 0
  -Global/noOverlap -g_no
    Algorithm for NoOverlap
    Default value is: 0
  -Global/redundNoOverlap -r_no
    Must we post redundant constraints for NoOverlap?
    Default value is: true
  -Global/binpacking -g_bp
    Algorithm for BinPacking
    Default value is: 0
  -Global/viewForSum -vs
    Must we use views for Sum constraints, when possible?
    Default value is: false
  -Global/eqDecForSum -eqs
    Must we post two constraints for Sum constraints, when the operator is EQ?
    Default value is: false
  -Global/permutation -permutation
    Must we use permutation constraints for AllDifferent if possible? (may be faster)
    Default value is: false
  -Global/allDifferentNb -adn
    Number of possibly automatically inferred AllDifferent
    Default value is: 10
  -Global/allDifferentSize -ads
    Limit on the size of possibly automatically inferred AllDifferent
    Default value is: 5
  -Global/starred -starred
    When true, some global constraints are encoded by starred tables
    Default value is: false
  -Global/hybrid -hybrid
    When true, some global constraints are encoded by hybrid/smart tables
    Default value is: false

Propagation
  -Propagation/clazz -p
    Class to be used for propagation (for example, FC, AC or SAC3)
    Default value is: AC
    Possible values: AP SAC SAC3 GIC2 FC GIC ESAC3 GIC3 GT BT AC DC1 SDC2 DC2 CDC1 CPC8 
      CPC2001 PC8 PC2001 CPC1 SAC3p MaxRPWC GIC4 TIC4
  -Propagation/variant -pv
    Propagation Variant (only used for some consistencies)
    Default value is: 0
  -Propagation/postponableConstraints -ppc
    Must we postpone filtering for expensive constraints?
    Default value is: false
  -Propagation/reviser -reviser
    Class to be used for performing revisions
    Default value is: Reviser$Reviser3
    Possible values: Reviser3 Reviser Reviser2
  -Propagation/residues -res
    Must we use residues (AC3rm)?
    Default value is: true
  -Propagation/bitResidues -bres
    Must we use bit residues (AC3bit+rm)?
    Default value is: true
  -Propagation/multidirectionality -mul
    Must we use multidirectionality
    Default value is: true
  -Propagation/arityLimit -al
    generic AC is systematically enforced if the arity is less than or equal to this value 
      (or this value is -1)
    Default value is: 2
  -Propagation/spaceLimit -sl
    generic AC is systematically enforced if the size of the Cartesian product of domains 
      is less than or equal to 2 to the power of this value (or this value is -1)
    Default value is: 20
  -Propagation/strongOnce -so
    Must we only apply the strong consistency (if chosen) before search?
    Default value is: false
  -Propagation/strongAC -sac
    Must we only apply the strong consistency (if chosen) when AC is effective?
    Default value is: false

Learning
  -Learning/nogood -ng
    Nogood recording technique (from restarts by default)
    Default value is: RST
    Possible values: NO RST RST_MIN RST_SYM
  -Learning/nogoodBaseLimit -ngbl
    The maximum number of nogoods that can be stored in the base
    Default value is: 200000
  -Learning/nogoodArityLimit -ngal
    The maximum arity of a nogood that can be recorded
    Default value is: 2147483647
  -Learning/nogoodDisplayLimit -ndl
    Size limit of the nogoods (from restarts) for being displayed
    Default value is: 0

Solving
  -Solving/clazz -s_class
    Class for solving (usually, Solver)
    Default value is: Solver
    Possible values: Solver
  -Solving/enablePrepro -prepro
    Must we perform preprocessing?
    Default value is: true
  -Solving/enableSearch -search
    Must we perform search?
    Default value is: true
  -Solving/branching -branching
    Branching scheme for search (binary or non-binary)
    Default value is: BIN
    Possible values: BIN NON

Restarts
  -Restarts/nRuns -r_n
    Maximal number of runs (restarts) to be performed
    Default value is: 2147483647
  -Restarts/cutoff -r_c
    Cutoff as a value of, e.g., the number of failed assignments before restarting
    Default value is: 10
  -Restarts/factor -r_f
    The geometric increasing factor when updating the cutoff
    Default value is: 1.1
  -Restarts/measure -r_m
    The metrics used for measuring and comparing with the cutoff
    Default value is: FAILED
    Possible values: FAILED WRONG BACKTRACK SOLUTION
  -Restarts/resetPeriod -r_rp
    Period, in term of number of restarts, for resetting restart data.
    Default value is: 90
  -Restarts/resetCoefficient -r_rc
    Coefficient used for increasing the cutoff, when restart data are reset
    Default value is: 2
  -Restarts/varhResetPeriod -r_vrp
    Description is missing...
    Default value is: 2147483647
  -Restarts/varhSolResetPeriod -r_vsrp
    Description is missing...
    Default value is: 30
  -Restarts/restartAfterSolution -ras
    Must we restart every time a solution is found?
    Default value is: false
  -Restarts/luby -luby
    Must we use a Luby series instead of a geometric one?
    Default value is: false

Revh
  -Revh/clazz -revh
    Class of the revision ordering heuristic
    Default value is: HeuristicRevisions$HeuristicRevisionsDynamic$Dom
    Possible values: WdegOnDom Last DdegOnDom Wdeg First Ddeg Dom Rand Lexico
  -Revh/anti -anti_revh
    Must we use the reverse of the natural heuristic order?
    Default value is: false

Varh
  -Varh/clazz -varh
    Class of the variable ordering heuristic
    Default value is: HeuristicVariablesDynamic$Wdeg
    Possible values: RunRobin Impact Dom Activity Wdeg Deg Memory DdegOnDom Ddeg CRBS PickOnDom 
      Rand Lexico WdegOnDom FrOnDom Srand ProcOnDom Regret FrbaOnDom DomThenDeg
  -Varh/anti -anti_varh
    Must we use the reverse of the natural heuristic order?
    Default value is: false
  -Varh/lc -lc
    Value for lc (last conflict); 0 if not activated
    Default value is: 2
  -Varh/weighting -wt
    How to manage weights for wdeg variants
    Default value is: CACD
    Possible values: VAR UNIT UNIT_EXP CACD CACD_EXP CHS
  -Varh/pickMode -pm
    How to manage incrementation of effective picked variables or constraints during propagation
    Default value is: 0
  -Varh/singleton -sing
    How to manage singleton variables during search
    Default value is: LAST
    Possible values: ANY FIRST LAST
  -Varh/connected -connected
    Must we select a variable necessarily connected to an already explicitly assigned one?
    Default value is: false
  -Varh/discardAux -da
    Must we not branch on auxiliary variables introduced by the solver?
    Default value is: false
  -Varh/arrayPriorityRunRobin -aprr
    Must we set priority to variable arrays in turn?
    Default value is: false

Valh
  -Valh/clazz -valh
    Class of the value ordering heuristic
    Default value is: HeuristicValuesDirect$First
    Possible values: Dist Vals OccsR Bivs3 Median AsgsFp Flrs Srand Rand Bivs Arbitrary First 
      InternDist Last Robin RunRobin Bivs2 AsgsFm Occs Asgs FlrsE Conflicts FlrsR AsgsE
  -Valh/anti -anti_valh
    Must we use the reverse of the natural heuristic order?
    Default value is: false
  -Valh/runProgressSaving -rps
    Must we use run progress saving?
    Default value is: false
  -Valh/solutionSaving -sos
    Solution saving (0: disabled, 1: enabled, otherwise deactivation period)
    Default value is: 1
  -Valh/warmStart -warm
    A starting instantiation (solution) to be used with solution saving
    Default value is: "" (empty string)
  -Valh/bivsFirst -bivs_f
    Must we stop BIVS at first found solution?
    Default value is: true
  -Valh/bivsOptimistic -bivs_o
    Must we use the optimistic BIVS mode?
    Default value is: true
  -Valh/bivsDistance -bivs_d
    0: only if in the objective constraint; 1: if at distance 0 or 1; 2: any variable
    Default value is: 2
  -Valh/bivsLimit -bivs_l
    BIVS applied only if the domain size is <= this value
    Default value is: 2147483647
\end{verbatim}
\end{footnotesize}

\section{Options for Choco}

Below, you will find some options that are available for Choco, as for example in:

\begin{command}
python Queens.py -data=8 -solver=[choco,v,args="-a -f -last"]
\end{command}

or
\begin{command}
java -jar choco.jar Queens-8.xml -a -f -last
\end{command}

\noindent Choco options are:

\begin{footnotesize}
\begin{verbatim}
ChocoXCSP [options...] file
 file                                   : File to parse.
 -a (--all)                             : Search for all solutions (default:
                                          false). (default: false)
 -ansi                                  : Enable ANSI colour codes (default:
                                          false). (default: false)
 -cos                                   : Tell the solver to use conflict
                                          ordering search. (default: false)
 -cs                                    : set to true to check solution with
                                          org.xcsp.checker.SolutionChecker
                                          (default: false)
 -csv (--print-csv)                     : Print statistics on exit (default:
                                          false). (default: false)
 -dfx                                   : Force default explanation algorithm.
                                          (default: false)
 -exp                                   : Plug explanation in (default: false).
                                          (default: false)
 -f (--free-search)                     : Ignore search strategy. (default:
                                          false)
 -flush N                               : Autoflush weights on black-box
                                          strategies (default: 32). (default:
                                          5000)
 -last                                  : Tell the solver to use use progress
                                          (or phase) saving. (default: false)
 -lc (--last-conflict) N                : Tell the solver to use last-conflict
                                          reasoning. (default: 1)
 -limit [String+]                       : Resolution limits (XXhYYmZZs,Nruns,Mso
                                          ls) where each is optional (no space
                                          allowed). (default: org.chocosolver.pa
                                          rser.ParserParameters$LimConf@e874448)
 -lvl (--log-level) [SILENT | COMPET |  : Define log level. (default: COMPET)
 RESANA | VERBOSE | JSON | IRACE |         
 INFO | FINE]                              
 -p (--nb-cores) N                      : Number of cores available for
                                          parallel search (default: 1).
                                          (default: 1)
 -pa (--parser) N                       : Parser to use.
                                          0: automatic
                                           1: FlatZinc (.fzn)
                                          2: XCSP3 (.xml or .lzma)
                                          3: DIMACS (.cnf),
                                          4: MPS (.mps) (default: 0)
 -restarts [String,int,double?,int]     : Define the restart heuristic to use.
                                          Expected format: (policy,cutoff,offset
                                          )  (no space allowed) (default:
                                          org.chocosolver.parser.ParserParameter
                                          s$ResConf@60285225)
 -seed N                                : Set the seed for random number
                                          generator.  (default: 0)
 -valh (--valHeuristic) [BEST | BMIN |  : Define the value heuristic to use.
 BLAST | DEFAULT | MAX | MED |            (default: DEFAULT)
 MIDFLOOR | MIDCEIL | MIN | RAND]          
 -varh (--varHeuristic) [ABS |          : Define the variable heuristic to use.
 ACTIVITY | CHS | DOM | DOMWDEG |         (default: DEFAULT)
 DOMWDEGR | DEFAULT | FRBA | FLBA |        
 IBS | IMPACT | INPUT | RAND |             
 MAB_CHS_DWDEG_STATIC | MAB_CHS_DWDEG_MOSS]        
\end{verbatim}
\end{footnotesize}

\chapter{Changelog}\label{cha:versioning}

\begin{itemize}
\item Version 2.6 (i) published on Septembre 2, 2026.
  Complex structures \nn{If} and \nn{Match} are slightly extended: now, classical expressions can be used in first position (parameter); see Section \ref{sec:complex}.
  A paragraph about correctly handling the indexing of constraints \gb{element} (when dealing with complex situations) is introduced Page \pageref{sec:indexElt}.
  Arrays of multiple variables (i.e., arrays of several fields representing arrays of named tuples of variables) have been introduced; see Section \ref{sec:multiple}.
  A new parameter, \texttt{no\_self\_looping}, has been added to the constraint \gb{Circuit}; see Section \ref{sec:circuit}. 
  It is now possible to use ranges (with some restrictions) when specifying the value of the parameter \texttt{size} of function \texttt{VarArray()}; see Section \ref{sec:varArray}.
\item Version 2.5. Some changes (published with Version 2.6).   
\item Version 2.4 published on August 28, 2024.
New forms are possible for posting \gb{extension} constraints; see Section \ref{sec:extension}.
New forms are possible for posting \gb{regular} constraints; see Section \ref{sec:regular}.
New forms are possible for posting \gb{lex} constraints; see Section \ref{sec:lex}.
Arbitrary constraints are now possible with \gb{adhoc} constraints; see Section \ref{sec:adhoc}.
\item Version 2.3. Minor changes. Unpublished. 
\item Version 2.2 published on December 5, 2023.
New (control) structures \nn{If} and \nn{Match} are introduced in Section \ref{sec:complex}.
The constraint form \gb{Hamming}, derived from \gb{Sum}, is introduced in Section \ref{sec:sum}.
The constraint forms \gb{Exist}, \gb{NotExist}, \gb{ExactlyOne}, \gb{AtLeastOne}, \gb{AtMostOne} and \gb{AllHold}, derived from \gb{Count}, are introduced in Section \ref{sec:count}.
The functions \nn{both} and \nn{either} have been introduced: they respectively correspond to a conjunction and a disjunction of two terms. 
An illustration of how to declare local arrays of variables (with Ghoulomb Problem) is introduced in Section \ref{sec:naming}.
Auto-adjustment of array indexing is introduced, and illustrated (with Traffic Lights Problem) in Section \ref{sec:extension}. 
The possibility of using a predefined named tuple \nn{Task}, when posting constraints like \gb{cumulative}, is illustrated with RCPSP Problem in Section \ref{sec:cumulative}.
Chapter \ref{ch:logical} has been reorganized, and includes the Arithmetic Target Problem (illustrating various use cases of \nn{If} and \nn{Match}).
Chapter \ref{cha:options} has been introduced for listing the options that can be used with solvers ACE and Choco.

\item Version 2.1 published on November 10, 2022.
New constraints are introduced: \gb{precedence}, \gb{knapsack}, \gb{binPacking}, \gb{maximumArg}, and \gb{minimumArg} in Sections \ref{sec:precedence}, \ref{sec:knapsack}, \ref{sec:binPacking}, \ref{sec:maximumArg}, and \ref{sec:minimumArg}, respectively.   
The constraints \gb{precedence}, \gb{knapsack}, and \gb{binPacking} belong now to \x3-core.
The construction of hybrid tables is documented; see Section \ref{sec:hybrid}. When declaring a stand-alone variable or array of variables, it is now possible to set its name with a parameter; see Section \ref{sec:naming}.

\item Version 2.0 published on December 15, 2021.
New functions allow us to pilot the solving process: this is described in the new chapter \ref{ch:piloting}.
Everything you need to know about the interface of the library is described in the new chapter \ref{ch:interface}.  
  How to format data in filenames, to use default data and to load independent JSON data files (possibly from URLs) is explained in Section \ref{sec:specdata}.   
\item Version 1.3 published on June 21, 2021.
It is now possible to load data from several files (see Section \ref{sec:specdata}).
How to avoid importing everything ($*$) is explained.
How to logically combine (global) constraints is explained in the new chapter \ref{ch:logical}.
\end{itemize}


\chapter*{Acknowledgments}

This work was supported by State funding managed by the French National Research Agency (ANR) under France 2030 and by the European Union through NextGenerationEU, reference 22-EXES-0009.

\begin{theindex}

  \item Problems
    \subitem All-Interval Series, \hyperpage{39}
    \subitem Allergy, \hyperpage{50}
    \subitem Amaze, \hyperpage{141}
    \subitem Arithmetic Target, \hyperpage{129}
    \subitem Balanced Academic Curriculum (BACP), \hyperpage{42}
    \subitem Balanced Incomplete Block Designs, \hyperpage{40}
    \subitem Blackhole, \hyperpage{32}
    \subitem Board Coloration, \hyperpage{20}
    \subitem CCMcP, \hyperpage{119}
    \subitem Community Detection, \hyperpage{86}
    \subitem Cookie Monster, \hyperpage{159}
    \subitem Costas Arrays, \hyperpage{76}
    \subitem Crossword Generation, \hyperpage{79}
    \subitem Crypto Puzzle, \hyperpage{89}
    \subitem Diagnosis, \hyperpage{137}
    \subitem Domino, \hyperpage{81}
    \subitem Flow Shop Scheduling, \hyperpage{113}
    \subitem Freq. Assign. Problem with Polarization, \hyperpage{133}
    \subitem Futoshiki, \hyperpage{77}
    \subitem Ghoulomb, \hyperpage{53}
    \subitem Golomb Ruler, \hyperpage{25}
    \subitem Hamming Vectors, \hyperpage{91}
    \subitem Labeled Dice, \hyperpage{97}
    \subitem Layout, \hyperpage{143}
    \subitem Magic Sequence, \hyperpage{23}
    \subitem Mario, \hyperpage{124}
    \subitem Open Stacks, \hyperpage{99}
    \subitem Optimized Knapsack, \hyperpage{122}
    \subitem Pizza Voucher, \hyperpage{93}
    \subitem Portal, \hyperpage{54}
    \subitem Prime Looking, \hyperpage{166}
    \subitem Progressive Party, \hyperpage{111}
    \subitem Quasigroup Existence, \hyperpage{104}
    \subitem Queens, \hyperpage{15}
    \subitem Rack Configuration, \hyperpage{35}
    \subitem Radio Link Frequency Assignment, \hyperpage{58}
    \subitem Rectangle Packing, \hyperpage{115}
    \subitem Resource Constrained Project Scheduling, \hyperpage{117}
    \subitem Riddle, \hyperpage{5}
    \subitem Sandwich, \hyperpage{102}
    \subitem Send-More-Money, \hyperpage{75}
    \subitem Social Golfers, \hyperpage{85}
    \subitem Sports Scheduling, \hyperpage{97}
    \subitem Stable Marriage, \hyperpage{136}
    \subitem Steel Mill Slab, \hyperpage{120}
    \subitem Steiner Triple Systems, \hyperpage{83}
    \subitem Subgraph Isomorphism, \hyperpage{69}
    \subitem Sudoku, \hyperpage{27}
    \subitem Template Design, \hyperpage{90}
    \subitem Traffic Lights, \hyperpage{66}
    \subitem Traveling Salesman Problem (TSP), \hyperpage{105}
    \subitem Traveling Tournament with Pred. Venues, \hyperpage{67}
    \subitem Vellino, \hyperpage{139}
    \subitem War or Peace, \hyperpage{95}
    \subitem Warehouse Location, \hyperpage{30}
    \subitem Word Design, \hyperpage{73}
    \subitem World Traveling, \hyperpage{12}
    \subitem Zebra, \hyperpage{63}

\end{theindex}

\bibliographystyle{plain}

\end{document}